\documentclass[a4paper,12pt,times,numbered,print,index]{Classes/PhDThesisPSnPDF}

\usepackage{algorithm}
\usepackage{algorithmic}
\usepackage{enumerate}

\usepackage{amsmath}
\DeclareMathOperator{\atantwo}{atan2}

\DeclareMathOperator*{\argmin}{arg\,min}
\usepackage[final]{listings}
\usepackage[utf8]{inputenc}
\usepackage{textcomp}
\usepackage[table]{xcolor}
\usepackage{gensymb}
\usepackage{float}
\usepackage{soul}
\usepackage{graphicx}

 \newcommand{\indep}{\perp\!\!\!\!\perp}

\usepackage[greek,english]{babel}

\input{Preamble/preamble}

\usepackage{graphicx}
\title{Big Earth Data and Machine Learning for Sustainable and Resilient Agriculture}
\author{Vasileios Sitokonstantinou}

\dept{School of Rural, Surveying \& 
Geoinformatics Engineering}

\university{National Technical University of Athens}
\degreetitle{Doctor of Philosophy}
\supervisor{Dr. V. Karathanassi, Professor of NTUA \newline Dr. C. Kontoes, Research Director of NOA \newline Dr. D. Argialas, Professor Emeritus of NTUA \newline Dr. K. Karantzalos, Associate Professor of NTUA \newline Dr. A. Voulodimos, Assistant Professor of NTUA \newline Dr. I. Manakos, Principal Researcher of CERTH  \newline Dr. I. Athanasiadis, Professor of Wageningen University and Research} 
\college{Remote Sensing Laboratory}
\subject{LaTeX} \keywords{{LaTeX} {PhD Thesis} {} {National Technical University of Athens}}
\ifdefineAbstract
\fi

\ifdefineChapter
\fi

\begin{document}

\frontmatter

\maketitle

\begin{figure}
\centering\includegraphics[scale=0.14]{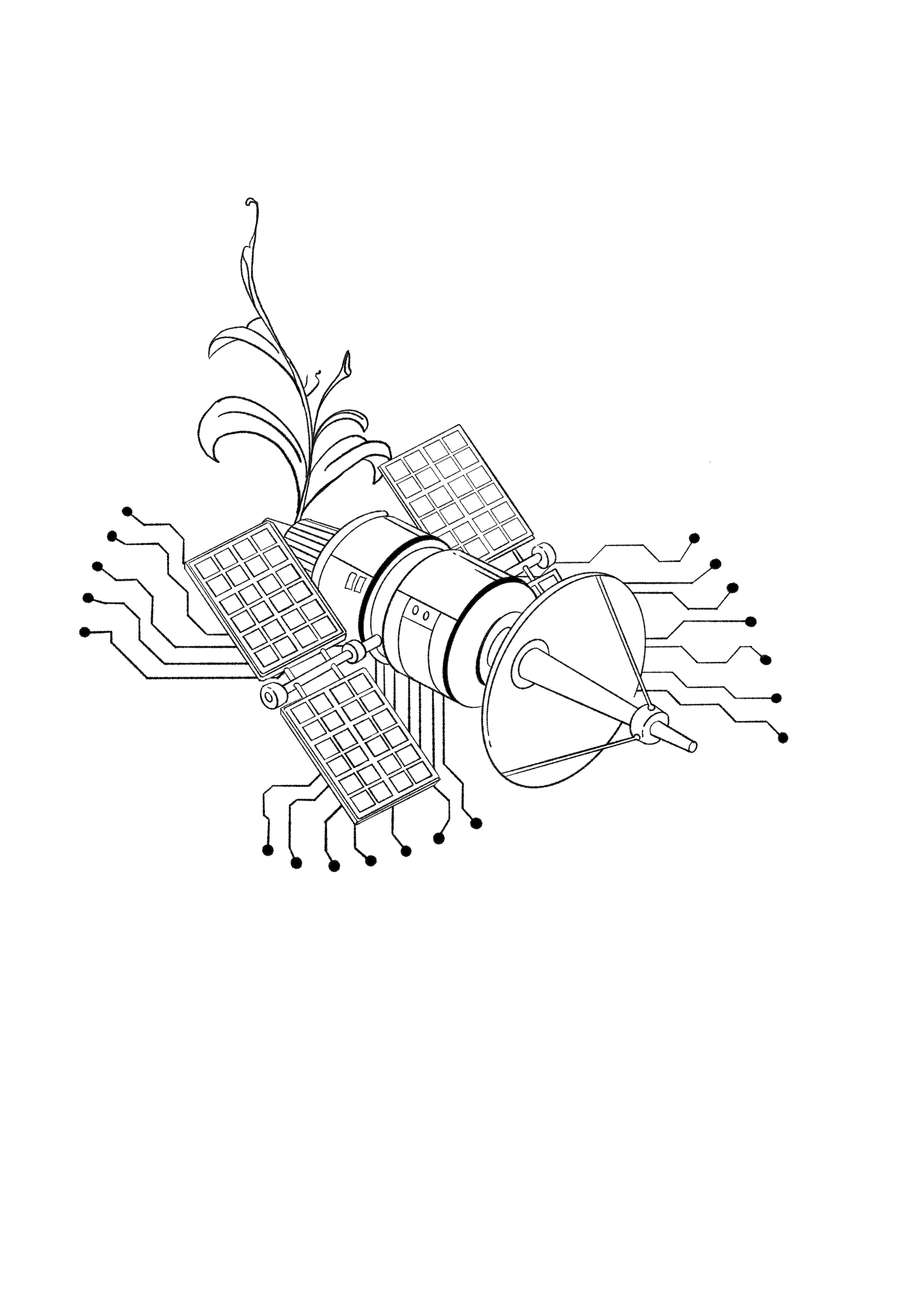}
\end{figure}

% ******************************* Thesis Dedidcation ********************************
\selectlanguage{english} 

\begin{dedication} 

\vspace{-3ex}%

I would like to dedicate this thesis to Suzie, Thanassis, Alkis and my family.

\end{dedication}
\selectlanguage{english} 

\begin{declaration}

I hereby declare that except where specific reference is made to the work of others, the contents of this dissertation are original and have not been submitted in whole or in part for consideration for any other degree or qualification in this, or any other university. This thesis contains material from 12 papers published in the following peer-reviewed journals and conference proceedings in which I am listed as an author. When I am not listed as a first author or I am listed as a first author with equal contribution with another individual, I explicitly mention my part in the work and acknowledge the effort of others. \\

\noindent Chapter 1 is published with material from all papers that are included in this thesis. \\

\noindent Chapter 2 is published with material from the following papers: \\

\noindent\textbf{(1)} Drivas, T.*, Sitokonstantinou, V.*, Tsardanidis, I., Koukos, A., Karathanassi, V. and
Kontoes, C. (2022). A Data Cube of Big Satellite Image Time-Series for Agriculture
Monitoring. In 2022 IEEE 14th Image, Video, and Multidimensional Signal Processing
Workshop, IVMSP. \\

\noindent The contributions of the co-authors are as follows:

\begin{itemize}
    \item I came up with the key idea, designed all experiments, implemented part of the source code and conducted most experiments. I prepared the manuscript drafts (in entirety).
    \item Part of the source code was implemented and part of the experiments were conducted by Mr. Thanassis Drivas, Mr. Alkiviadis Koukos and Mr. Iason Tsardanidis.
    \item The manuscripts were revised and edited mainly by Dr. Vassilia Karathanassi and Dr. Charalampos Kontoes.
\end{itemize}

\noindent\textbf{AND}\\

\noindent\textbf{(2)} Sitokonstantinou, V., Papoutsis, I., Kontoes, C., Lafarga Arnal, A., Armesto Andrés, A.
P., and Garraza Zurbano, J. A. (2018). Scalable parcel-based crop identification scheme
using Sentinel-2 data time-series for the monitoring of the common agricultural policy.
Remote Sensing, 10(6), 911.\\

\noindent The contributions of the co-authors are as follows:
\begin{itemize}
    \item I came up with the key idea, designed all experiments, implemented all of the source code and conducted all experiments. I prepared the manuscript drafts (in entirety).
    \item The manuscripts were revised and edited mainly by Dr. Ioannis Papoutsis and Dr. Charalampos Kontoes.
\end{itemize}

\noindent\textbf{AND}\\

\noindent\textbf{(3)} Rousi, M., Sitokonstantinou, V., Meditskos, G., Papoutsis, I., Gialampoukidis, I.,
Koukos, A., ... and Kompatsiaris, I. (2020). Semantically enriched crop type classification and linked earth observation data to support the common agricultural policy
monitoring. IEEE Journal of Selected Topics in Applied Earth Observations and Remote Sensing, 14, 529-552. \\

\noindent The contributions of the co-authors are as follows:
\begin{itemize}
    \item I came up with the key idea of the smart sampling algorithm, designed all experiments, implemented part of the source code (smart sampling algorithm and crop classification model) and conducted all experiments. Regarding the writing of the manuscript drafts, Ms. Maria Rousi and I contributed equally.
    \item Part of the source code was implemented by Mr. Alkiviadis Koukos (smart sampling algorithm) and Ms. Maria Rousi (geosparql code).
    \item The manuscripts were revised and edited mainly by Dr. Vassilia Karathanassi, Dr. Georgios Meditskos, Dr. Ilias Gialampoukidis and Dr. Charalampos Kontoes.
    
\end{itemize}

\noindent\textbf{AND}\\

\noindent\textbf{(4)} Choumos, G.*, Koukos, A.*, Sitokonstantinou, V. and Kontoes, C. (2022). Towards
space-to-ground data availability for agriculture monitoring.
In 2022 IEEE 14th Image, Video, and Multidimensional Signal Processing Workshop,
IVMSP. \\

\noindent The contributions of the co-authors are as follows:
\begin{itemize}
    \item Mr. George Choumos, Mr. Alkiviadis Koukos and I came up with the key idea and designed all experiments. I prepared part of the manuscript.
    \item The source code was implemented and the experiments conducted by Mr. George Choumos and Mr. Akiviadis Koukos.
    \item The manuscripts were revised and edited mainly by myself, Mr. Alkiviadis Koukos and Dr. Charalampos Kontoes.
\end{itemize}

\noindent\textbf{AND}\\

\noindent\textbf{(5)} Sitokonstantinou, V., Koukos, A., Drivas, T., Kontoes, C., and Karathanassi, V. (2022).
DataCAP: A Satellite Datacube and Crowdsourced Street-Level Images for the Monitoring of the Common Agricultural Policy. In International Conference on Multimedia Modeling (pp. 473-478). Springer. \\

\noindent The contributions of the co-authors are as follows:
\begin{itemize}
    \item I came up with the key idea, designed all experiments, implemented part of the source code and conducted all experiments. I prepared the manuscript drafts (in entirety).
    \item Part of the source code was implemented by Mr. Thanassis Drivas and Mr. Alkiviadis Koukos.
    \item The manuscripts were revised and edited mainly by Dr. Vassilia Karathanassi, Mr. Alkiviadis Koukos and Dr Charalampos Kontoes.
\end{itemize}

\noindent Chapter 3 is published with material from the following papers: \\

\noindent\textbf{(1)} Sitokonstantinou, V., Koutroumpas, A., Drivas, T., Koukos, A., Karathanassi, V., Kontoes, H., and Papoutsis, I. (2020). A Sentinel based agriculture monitoring scheme
for the control of the CAP and food security. In Eighth International Conference on Remote Sensing and Geoinformation of the Environment (RSCy2020) (Vol. 11524, pp.
48-59). SPIE \\

\noindent The contributions of the co-authors are as follows:

\begin{itemize}
    \item I came up with the key idea, designed part of the experiments, implemented part of the source code and conducted part of the experiments. I prepared the manuscript drafts.
    \item Part of the source code was implemented and part of the experiments were conducted by Mr. Antonis Koutroumpas.
    \item The manuscripts were revised and edited mainly by Dr. Vassilia Karathanassi and Dr. Charalampos Kontoes.
\end{itemize}

\noindent\textbf{AND}\\

\noindent\textbf{(2)} Sitokonstantinou, V., Drivas, T., Koukos, A., Papoutsis, I., Kontoes, C. and Karathanassi,
V. (2020). Scalable distributed random forest classification for paddy rice mapping. In
Proceedings of the Asian Remote Sensing Conference (ACRS 2019) (pp. 836-845). \\

\noindent The contributions of the co-authors are as follows:

\begin{itemize}
    \item I came up with the key idea, designed all experiments, implemented part of the source code and conducted most experiments. I prepared the manuscript drafts (in entirety).
    \item Part of the source code was implemented and part of the experiments were conducted by Mr. Alkiviadis Koukos and Mr. Thanassis Drivas.
    \item The manuscripts were revised and edited mainly by Dr. Ioannis Papoutsis, Dr. Vassilia Karathanassi and Dr. Charalampos Kontoes.
\end{itemize}

\noindent\textbf{AND}\\

\noindent\textbf{(3)} Sitokonstantinou, V., Koukos, A., Drivas, T., Kontoes, C., Papoutsis, I., and Karathanassi,
V. (2021). A Scalable Machine Learning Pipeline for Paddy Rice Classification Using
Multi-Temporal Sentinel Data. Remote Sensing, 13(9), 1769.\\

\noindent The contributions of the co-authors are as follows:

\begin{itemize}
    \item I came up with the key idea, designed all experiments, implemented part of the source code and conducted most experiments. I prepared the manuscript drafts (in entirety).
    \item Part of the source code was implemented and part of the experiments were conducted by Mr. Alkiviadis Koukos.
    \item The manuscripts were revised and edited mainly by Dr. Ioannis Papoutsis, Dr. Vassilia Karathanassi and Dr. Charalampos Kontoes.
\end{itemize}

\noindent Chapter 4 is published with material from the following papers: \\

\noindent\textbf{(1)} Sitokonstantinou, V., Koukos, A., Kontoes, C., Bartsotas, N. S., and Karathanassi, V.
(2021, July). Semi-Supervised Phenology Estimation in Cotton Parcels with Sentinel-2
Time-Series. In 2021 IEEE International Geoscience and Remote Sensing Symposium
IGARSS (pp. 8491-8494). IEEE\\

\noindent The contributions of the co-authors are as follows:

\begin{itemize}
    \item I came up with the key idea, designed part of the experiments, implemented part of the source code and conducted part of the experiments. I prepared the manuscript drafts.
    \item Part of the source code was implemented and part of the experiments were conducted by Mr. Alkviadis Koukos.
    \item The manuscripts were revised and edited mainly by Dr. Vassilia Karathanassi and Dr. Charalampos Kontoes.
\end{itemize}

\noindent\textbf{AND}\\

\noindent\textbf{(2)} Sitokonstantinou, V., Koukos, A., Tsoumas, I., Bartsotas, N., Kontoes, C. and Karathanassi,
V. (2022). Fuzzy clustering for the within-season estimation of cotton phenology. PLOS ONE (under review).\\

\noindent The contributions of the co-authors are as follows:

\begin{itemize}
    \item I came up with the key idea, designed all experiments, implemented part of the source code and conducted most experiments. I prepared the manuscript drafts (in entirety).
    \item Part of the source code was implemented and part of the experiments were conducted by Mr. Alkiviadis Koukos.
    \item The manuscripts were revised and edited mainly by Dr. Vassilia Karathanassi and Dr. Charalampos Kontoes.
\end{itemize}

\noindent Chapter 5 is published with material from the following papers: \\

\noindent\textbf{(1)} Giannarakis, G., Sitokonstantinou, V., Roxanne, L. and Kontoes, C. (2022). Towards
assessing agricultural land suitability using causal machine learning. In Proceedings
of the IEEE/CVF Conference on Computer Vision and Pattern Recognition Workshops (CVPRW).\\

\noindent The contributions of the co-authors are as follows:

\begin{itemize}
    \item Mr. Georgios Giannarakis and I came up with the key idea and designed all the experiments. Mr. Georgios Giannarakis and I prepared the manuscript drafts and Dr. Roxanne Suzette Lorilla contributed. 
    \item The source code was implemented and the experiments were conducted by Mr. Georgios Giannarakis.
    \item The manuscripts were revised and edited mainly by Dr. Roxanne Suzette Lorilla and Dr. Charalampos Kontoes.
\end{itemize}

\noindent\textbf{AND}\\

\noindent\textbf{(2)} Nanushi, O.*, Sitokonstantinou, V.*, Tsoumas, I. and Kontoes, C. (2022). Pest presence
prediction using interpretable machine learning. In 2022 IEEE 14th Image, Video, and
Multidimensional Signal Processing Workshop, IVMSP.\\

\noindent The contributions of the co-authors are as follows:

\begin{itemize}
    \item Ms. Ornela Nanushi and I contributed equally, i.e., came up with the key idea, designed most experiments, implemented the source code and prepared the manuscript drafts.
    \item Some experiments were designed by Mr. Ilias Tsoumas.
    \item The manuscript was revised and edited mainly by Dr. Charalampos Kontoes and Mr. Ilias Tsoumas.
\end{itemize}

% Author and date will be inserted automatically from thesis.tex \author \degreedate

\end{declaration}
% ************************** Thesis Acknowledgements **************************

\begin{acknowledgements}

I would like to acknowledge Prof. Vassilia Karathanassi for the supervision of my research work and her invaluable guidance throughout my PhD. I would also like to thank my mentor - professional and beyond - and supervisor Dr. Haris Kontoes, research director of the National Observatory of Athens (NOA). Throughout the course of my PhD, I have worked at the Operational Unit Beyond | IAASARS | NOA, led by Haris. The Beyond Unit is an inspiring research team that has embraced my scientific vision and helped to bring it to fruition. It was through my work at Beyond that I managed to secure the funding of my research but most importantly to grow as a person and scientist. \\

\noindent In detail, this work has been supported by the following projects of the Operational Unit Beyond | IAASARS | NOA that have received funding from the EU's Horizon 2020 research and innovation programme: RECAP No. 693171 (2016-2018), EOPEN No. 776019 (2018-2020), e-shape No. 820852 (2020-2023), ENVISION No. 869366 (2020-2023), CALLISTO No. 101004152 (2021-2024), EIFFEL No. 101003518 (2021-2024). \\

\noindent I would also like to acknowledge Dr. Ioannis Papoutsis and Dr. Dimitris Argialas for their guidance and supervision, and my beloved friends and colleagues Alkis Koukos and Thanassis Drivas. This thesis is the outcome of collaborative work and for this reason I would like to express my gratitude to all my co-authors: Ilias Tsoumas, Giorgos Giannarakis, Roxanne Suzette Lorilla, Iason Tsardanidis, Ornela Nanushi, George Choumos, Antonis Koutroumpas and Nikos Bartsotas. \\

\noindent I dedicate this thesis to Souzana and I thank her for her love and patience. And a big shout-out to my friends Vassilis, Yiannis, Lefteris, Tania, Nikos, Natalia (artwork creator), Leo, James, Dimitris D., Dimitris F., Eleni, Stella, Tamy and Michael. Last but not least, I would like to thank my family - Christina, Efthymis and Filippos - for their constant encouragement and unconditional love.

\end{acknowledgements}

% ************************** Thesis Abstract *****************************
% Use `abstract' as an option in the document class to print only the titlepage and the abstract.

\selectlanguage{english} 
\begin{abstract}
More than a tenth of the global population does not have access to sufficient quantities of affordable, nutritious food. At the same time, the food demand is expected to increase between 35-56\% over the period 2010-2050, which will require the intensification of agriculture. On the other hand, climate change calls for the employment of agricultural practices that will secure resilience and sustainability. There is, therefore, an urgent need for producing more, while changing our methods to account for future changes. In this regard, the large scale and frequent monitoring of agricultural land can provide significant insights for timely decision making based on evidence. \\

\noindent Big streams of Earth images from satellites or other platforms (e.g., drones and mobile phones) are becoming increasingly available at low or no cost and with enhanced spatial and temporal resolution. This thesis recognizes the unprecedented opportunities offered by the high quality and open access Earth observation data of our times and introduces novel machine learning and big data methods to properly exploit them towards developing applications for sustainable and resilient agriculture. The thesis addresses three distinct thematic areas, i.e., the monitoring of the Common Agricultural Policy (CAP), the monitoring of food security and applications for smart and resilient agriculture. The methodological innovations of the developments related to the three thematic areas address the following issues: i) the processing of big Earth Observation (EO) data, ii) the scarcity of annotated data for machine learning model training and iii) the gap between machine learning outputs and actionable advice.\\

\noindent The first contribution of this thesis is the Agriculture Monitoring Data Cube (ADC) that offers an automated, modular, end-to-end, cloud-based framework for handling big satellite data (Sentinel-1 and Sentinel-2) based on the Open Data Cube (ODC). ADC provides a set of powerful tools on top of the cube, including i) the generation of analysis-ready feature spaces of big satellite data to feed downstream machine learning tasks and ii) the support of Satellite Image Time-Series (SITS) analysis via services pertinent to the monitoring of the CAP, e.g., detecting trends and events, monitoring the growth status and more. \\

\noindent The second contribution of this thesis is a scalable and transferable machine learning method for multi-crop classification. The method uses the farmers’ declarations, as part of their subsidy applications in the CAP, in order to train the crop classification model. The method is extended by applying semantic enrichment on the crop type maps, increasing the value of knowledge extracted towards making decisions in operational scenarios of the paying agencies of the CAP. Specifically, a smart sampling method was developed to select parcels of potential wrong declaration (i.e., the farmers do not cultivate what they declared). This method provides actionable advice to the inspectors of the CAP paying agencies, early in the year, according to their operating model requirements. \\

\noindent Freely available satellite data, such as Sentinel-1 and Sentinel-2 data, used in the methods described above, cannot always provide confident crop type predictions in areas characterized by extended cloud coverage and/or small farm sizes. For this reason, ancillary sources of EO data are required. Towards this direction the space-to-ground paradigm is introduced, bringing together street-level and satellite images in an analysis-ready dataset, encouraging the community to experiment with fusion machine learning techniques for enhanced crop classification results. In this spirit, a preliminary late fusion approach was developed, where street-level image crop classification results were combined with the satellite based crop type maps.  \\

\noindent Another crop classification method is developed to classify paddy rice in South Korea. In this case, the focus is on weakly supervised learning as for South Korea there are not any available annotations. A pseudo-labeling approach is introduced using merely a confined number of labels that come from a small part of the country. Then the pseudo-labels are used to train a paddy rice classification model that generalizes to the entire country of South Korea. The nationwide classification required the processing of TBs of Sentinel images that would have been impossible in conventional machines. For this reason, a distributed implementation of the model was deployed using SPARK in a High Performance Data Analytics (HPDA) environment. \\

\noindent Finally, this thesis introduces two methods for phenology estimation in cotton fields. This is important work towards timely farm interventions that will secure the quality and volume of the yield and even increase it. Phenology ground observations are scarce in time and space, for this reason both methods focus on utilizing only few labels. In the first method, a semi-supervised approach is developed that uses a handful of labels to generate thousands of pseudo-labels that in turn train multiple supervised crop phenology classification models. The second method takes it a step further and develops a fuzzy clustering approach that not only estimates phenology in an unsupervised way but additionally predicts the transitional states between phenological stages using the membership score of fuzzy c-means. \\

\noindent The aforementioned contributions refer to detection methods, e.g., crop classification, phenology estimation. Although this work is important and addresses big data and machine learning issues associated with the large-scale and timely monitoring of agricultural land, there is still work to be done to reach actionable advice for the policy maker and the farmer. In this regard, causal and interpretable machine learning have been identified as key enablers to bridge this gap. Two methods have been developed to showcase preliminary results towards this direction. The first method uses interpretable machine learning to estimate the onset of pest harmfulness in cotton fields. The interpretability of the model allows for i) the rapid adoption of the application by the farmer and ii) the combination of the data-driven predictions with the empirical knowledge of the farmer, thus potentially increasing the value of the model outputs. The second method uses causal machine learning to assess agricultural land suitability for applying specific cultivation practices. In more detail, the heterogeneous impact of crop rotation and landscape crop diversity on Net Primary Productivity (NPP) was estimated, accounting for historical crop and environmental data. The results showed that the effect of crop rotation was insignificant, while landscape crop diversity had a small negative effect on NPP. Finally, considerable effect heterogeneity in space was observed for both practices. \\

\noindent All in all, this thesis showed that big EO data are a powerful tool for the large-scale and timely monitoring of agricultural land towards food security and climate resilience. In this context, it was demonstrated how big data technologies such as data cubes, distributed learning, linked open data and semantic enrichment can be used to exploit the data deluge and extract knowledge to address real user needs. Furthermore, this thesis argues for the importance of semi-supervised and unsupervised machine learning models that circumvent the ever-present challenge of scarce annotations and thus allow for model generalization in space and time. Specifically, it is shown how merely few ground truth data are needed to generate high quality crop type maps and crop phenology estimations. Finally, this thesis argues there is considerable distance in value between model inferences and decision making in real-world scenarios and thereby showcases the power of causal and interpretable machine learning in bridging this gap.\\


\end{abstract}
 
\begin{otherlanguage}{greek}  
% \begin{abstract}
\section*{\centering\textcolor{black}{Εκτενής Περίληψη}}

Οι απαιτήσεις για σίτιση θα αυξηθούν 35-56\% κατά την περίοδο 2010-2050 και ήδη 10\% του παγκόσμιου πληθυσμού δεν έχει πρόσβαση σε επαρκή τροφή. Ενώ η αύξηση της ζήτησης για τροφή απαιτεί την εντατικοποίηση της γεωργίας, η κλιματική αλλαγή απαιτεί την εφαρμογή καλλιεργητικών πρακτικών που θα εξασφαλίσουν την ανθεκτικότητα και τη βιωσιμότητα της. Ως εκ τούτου υπάρχει άμεση ανάγκη για μεγαλύτερη παραγωγή, αλλάζοντας παράλληλα τις μεθόδους μας, συνυπολογίζοντας μελλοντικές αλλαγές. Υπό αυτό το πρίσμα, η μεγάλης κλίμακας και τακτική παρακολούθηση της γεωργικής γης δύναται να προσφέρει σημαντικές πληροφορίες για την έγκαιρη λήψη αποφάσεων βάσει στοιχείων.\\

\noindent Μεγάλες ροές από εικόνες παρατήρησης της Γης είτε από δορυφόρους είτε από άλλες πλατφόρμες (για παράδειγμα \begin{otherlanguage}{english} drones \end{otherlanguage} και κινητά τηλέφωνα) γίνονται ολοένα και περισσότερο διαθέσιμες με χαμηλό ή και μηδενικό κόστος και σε καλύτερη χωρική και χρονική ανάλυση. Η παρούσα διατριβή αναγνωρίζει τις άνευ προηγουμένου ευκαιρίες που προσφέρουν τα υψηλής ανάλυσης και ανοικτής πρόσβασης  δεδομένα παρατήρησης της Γης της εποχής μας και υλοποιεί νέες μεθόδους μηχανικής μάθησης και επεξεργασίας μεγάλων δεδομένων για την κατάλληλη αξιοποίησή τους με σκοπό τη ανάπτυξη εφαρμογών για βιώσιμη και ανθεκτική γεωργία. Η διατριβή πραγματεύεται τρεις διακριτούς θεματικούς τομείς, δηλαδή την παρακολούθηση της Κοινής Αγροτικής Πολιτικής (ΚΑΠ), την παρακολούθηση της επισιτιστικής ασφάλειας και τις εφαρμογές ευφυούς γεωργίας. Οι μεθοδολογικές καινοτομίες, της παρούσας διατριβής, που σχετίζονται με τις τρεις θεματικές περιοχές αντιμετωπίζουν τα ακόλουθα ζητήματα: i) την επεξεργασία μεγάλων δεδομένων παρατήρησης της Γης, ii) την έλλειψη επισημειωμένων δεδομένων για εκπαίδευση μοντέλων μηχανικής μάθησης και iii) το χάσμα μεταξύ των αποτελεσμάτων της μηχανικής μάθησης και των πρακτικών συμβουλών. \\

\noindent Η πρώτη συνεισφορά της διατριβής αυτής είναι ο κύβος δεδομένων (\begin{otherlanguage}{english}data cube\end{otherlanguage}) για την παρακολούθησης γεωργίας (\begin{otherlanguage}{english}ADC\end{otherlanguage}), ο οποίος προσφέρει ένα πλήρως αυτοματοποιημένο πλαίσιο για την διαχείριση των μεγάλων δορυφορικών δεδομένων (\begin{otherlanguage}{english}Sentinel-1, Sentinel-2\end{otherlanguage}), βασισμένο στο \begin{otherlanguage}{english} Open Data Cube (ODC) \end{otherlanguage}. Ο \begin{otherlanguage}{english} ADC \end{otherlanguage} παρέχει ένα σύνολο ισχυρών εργαλείων που επιτρέπουν α) την δημιουργία χώρων χαρακτηριστικών μεγάλων δορυφορικών δεδομένων για την τροφοδοσία μοντέλων μηχανικής μάθησης και β) την υποστήριξη της ανάλυσης χρονοσειρών δορυφορικών εικόνων μέσω υπηρεσιών που σχετίζονται με την παρακολούθηση της ΚΑΠ (π.χ. ανίχνευση τάσεων και γεγονότων, παρακολούθηση της ανάπτυξης κτλ).\\

\noindent Η δεύτερη συνεισφορά της διατριβής είναι μια επεκτάσιμη και μεταφέρσιμη μέθοδος μηχανικής μάθησης για την ταξινόμηση πολλαπλών καλλιεργειών. Η μέθοδος αυτή κάνει χρήση των δηλώσεων των αγροτών, ως μέρος των αιτήσεων επιδότησης για την ΚΑΠ, προκειμένου να εκπαιδεύσει τα μοντέλα. Η μέθοδος επεκτείνεται με τον σημασιολογικό εμπλουτισμό στους χάρτες καλλιεργειών στοχεύοντας στην ενίσχυση της εξαγόμενης γνώσης και τη λήψη αποφάσεων σε επιχειρησιακά σενάρια των οργανισμών πληρωμών της ΚΑΠ. Συγκεκριμένα, αναπτύχθηκε μια έξυπνη μέθοδος δειγματοληψίας ώστε να επιλέγονται πιθανώς λανθασμένες δηλώσεις (δηλαδή οι παραγωγοί δεν καλλιεργούν αυτό που δηλώνουν). Αυτή η μέθοδος παρέχει πρακτικές συμβουλές στους επιθεωρητές των οργανισμών πληρωμών της ΚΑΠ νωρίς μέσα στο έτος βάσει των απαιτήσεων των επιχειρησιακών διαδικασιών.\\

\noindent Τα ελεύθερα διαθέσιμα δορυφορικά δεδομένα, όπως αυτά από τους δορυφόρους \begin{otherlanguage}{english} Sentinel \end{otherlanguage} που χρησιμοποιήθηκαν στις μεθόδους που περιγράφηκαν παραπάνω, δεν μπορούν να εγγυηθούν έγκυρες αποφάσεις σε περιοχές με υψηλή νεφοκάλυψη ή/και σε αγροτεμάχια με μικρό μέγεθος. Συνεπώς χρειαζόμαστε συμπληρωματικές πηγές δεδομένων παρατήρησης της Γης. Σε αυτή την κατεύθυνση αναπτύχθηκε ένα ανοιχτό σύνολο δεδομένων που συνδυάζει δορυφορικές εικόνες και \begin{otherlanguage}{english} street-level \end{otherlanguage} εικόνες, επιτρέποντας στην κοινότητα να πειραματιστεί με μοντέλα μηχανικής μάθησης για να ενισχύσει τα αποτελέσματα ταξινόμησης καλλιεργειών που χρησιμοποιούν μόνο δορυφορικές εικόνες. Σε αυτό το πνεύμα, αναπτύχθηκε μια προκαταρκτική προσέγγιση συνδυασμού των δορυφορικών και \begin{otherlanguage}{english} street-level \end{otherlanguage} εικόνων.\\

\noindent Επίσης αναπτύχθηκε μέθοδος ταξινόμησης των ορυζώνων στη Νότια Κορέα. Στην περίπτωση αυτή, εστιάσαμε στην μερικώς επιβλεπόμενη μάθηση, καθώς για τη Νότια Κορέα δεν υπάρχουν διαθέσιμα επισημειωμένα δεδομένα, όπως στην περίπτωση της ΚΑΠ. Έτσι, αναπτύχθηκε μία ημι-επιβλεπόμενη προσέγγιση που χρησιμοποιεί περιορισμένο αριθμό επισημειωμένων δεδομένων από ένα μικρό μόνο τμήμα της χώρας για να δημιουργήσει ψευδείς επισημειώσεις (προβλέψεις μοντέλου ομαδοποίησης) για την εκπαίδευση ενός μοντέλου επιβλεπόμενης ταξινόμησης ρυζιού που γενικεύεται σε ολόκληρη τη χώρα της Νότιας Κορέας. Η εθνικής κλίμακας ταξινόμηση απαιτούσε την επεξεργασία \begin{otherlanguage}{english} TB \end{otherlanguage} δεδομένων από εικόνες \begin{otherlanguage}{english} Sentinel \end{otherlanguage} που ήταν αδύνατη σε συμβατικά μηχανήματα. Ως εκ τούτου, εφαρμόστηκε μια κατανεμημένη υλοποίηση του μοντέλου (σε \begin{otherlanguage}{english} SPARK\end{otherlanguage}) σε περιβάλλον ανάλυσης δεδομένων υψηλής απόδοσης (\begin{otherlanguage}{english} HPDA\end{otherlanguage}).\\

\noindent Τέλος, η παρούσα διατριβή παρουσιάζει δύο νέες μεθόδους για την εκτίμηση της φαινολογίας σε καλλιέργειες βαμβακιού. Η δουλειά αυτή καθίσταται σημαντική για την έγκαιρη επέμβασης στο χωράφι που θα διασφαλίσει την ποιότητα και τον όγκο της παραγωγής.  Οι επίγειες παρατηρήσεις της φαινολογίας είναι ελλιπείς στο χρόνο και στον χώρο και για το λόγο αυτό και οι δύο μέθοδοι επικεντρώνονται στη χρήση μόνο λίγων επισημειωμένων δεδομένων. Στην πρώτη μέθοδο, αναπτύχθηκε ένα μοντέλο το οποίο χρησιμοποιεί ελάχιστα επισημειωμένα δεδομένα απο μία μόνο περιοχή για να δημιουργήσει χιλιάδες ψευδείς επισημειώσεις (προβλέψεις του μοντέλου στην γύρω περιοχή) που χρησιμοποιούνται για να εκπαιδεύσουν επιβλεπόμενα μοντέλα ταξινόμησης που μπορούν να γενικεύσουν στον χώρο. Η δεύτερη μέθοδος αναπτύσσει μια προσέγγιση ασαφούς ομαδοποίησης που όχι μόνο εκτιμά τη φαινολογία με μη επιβλεπόμενο τρόπο  αλλά προβλέπει επιπλέον τις μεταβατικές καταστάσεις μεταξύ φαινολογικών σταδίων χρησιμοποιώντας τη πιθανότητα μιας οντότητας να ανήκει σε μια συγκεκριμένη ομάδα, όπως αυτή υπολογίζεται από τον αλγόριθμο \begin{otherlanguage}{english} fuzzy c-means \end{otherlanguage}. \\

\noindent Οι προαναφερθείσες συνεισφορές αναφέρονται σε μεθόδους ανίχνευσης/εντοπισμού, π.χ. ταξινόμηση καλλιεργειών και εκτίμηση φαινολογίας. Παρόλο που η εργασία αυτή είναι σημαντική και δίνει απαντήσεις σε ζητήματα διαχείρισης μεγάλων δεδομένων και μηχανικής μάθησης σχετιζόμενα με την μεγάλης κλίμακας και έγκαιρη παρακολούθηση της γεωργικής γης, υπάρχει χώρος βελτίωσης ώστε να φτάσουμε σε πρακτικές και χρήσιμες συμβουλές τόσο για τον υπεύθυνο χάραξης πολιτικής όσο και για τον αγρότη. Από αυτή την άποψη, η αιτιώδης και η ερμηνεύσιμη μηχανική μάθηση (\begin{otherlanguage}{english}causal, intepretable machine learning\end{otherlanguage}) έχουν αναγνωριστεί ως κατάλληλες μέθοδοι για τη γεφύρωση αυτού του χάσματος. Προς αυτή την κατεύθυνση αναπτύχθηκαν δύο μεθοδολογίες για την επίδειξη προκαταρκτικών αποτελεσμάτων. Η πρώτη χρησιμοποιεί ερμηνεύσιμη μηχανική μάθηση στοχεύοντας στην εκτίμηση της έναρξης της βλαβερότητας του πράσινου σκουληκιού στο βαμβάκι. Η ερμηνεία του μοντέλου επιτρέπει την ταχεία δράση από τους αγρότες, καθώς εμπιστεύονται την εκτίμηση. Επίσης μπορούν να συνδυάσουν τα αποτελέσματα με την εμπειρική τους γνώση. Η δεύτερη μέθοδος χρησιμοποιεί αιτιώδη μηχανική μάθηση για να αξιολογήσει την καταλληλόλητα της γεωργικής γης για την εφαρμογή συγκεκριμένων καλλιεργητικών πρακτικών. Πιο συγκεκριμένα αξιολογείται ο αντίκτυπος της εναλλαγής των καλλιεργειών και της χωρικής διαφοροποίησης των καλλιεργειών στην ρύθμιση του κλίματος. Τα αποτελέσματα έδειξαν πως ο αντίκτυπος της εναλλαγής των καλλιεργειών δεν ήταν σημαντικός αλλά η χωρική διαφοροποίηση των καλλιεργειών είχε μία μικρή αρνητική επίδραση στην ρύθμιση του κλίματος.\\

\noindent  Αυτή η διατριβή έδειξε ότι τα μεγάλα τηλεπισκοπικά δεδομένα είναι ένα ισχυρό εργαλείο για την έγκαιρη και μεγάλης κλίμακας παρακολούθηση της γεωργικής γης. Συγκεκριμένα, η παρούσα εργασία επέδειξε πως τεχνολογίες επεξεργασίας μεγάλων δεδομένων, όπως κύβοι δεδομένων, κατανεμημένη μάθηση, συνδεδεμένα ανοιχτά δεδομένα και σημασιολογικός εμπλουτισμός, μπορούν να χρησιμοποιηθούν για να εξάγουν την απαραίτητη γνώση από μεγάλα δορυφορικά δεδομένα και να λύσουν πραγματικές ανάγκες χρηστών. Επιπροσθέτως, η διατριβή υποστηρίζει την σημασία των μοντέλων ημι-επιβλεπόμενης και μη επιβλεπόμενης μάθησης που ξεπερνούν το πανταχού παρόν πρόβλημα της έλλειψης επισημειωμένων δεδομένων. Επιδείχτηκε πως με ελάχιστα ή και χωρίς επισημειωμένα δεδομένα μπορούν να παραχθούν χάρτες καλλιεργειών και εκτιμήσεις φαινολογίας υψηλής ποιότητας. Τέλος, η διατριβή εντοπίζει πως υπάρχει απόσταση μεταξύ των αποτελεσμάτων των μοντέλων μηχανικής μάθησης και την λήψη αποφάσεων βάσει αυτών σε επιχειρησιακά σενάρια. Σε αυτή την κατεύθυνση, επιδεικνύεται η ισχύς της αιτιώδους και ερμηνεύσιμης μηχανικής μάθησης στην γεφύρωση του χάσματος μεταξύ εκτίμησης και πρακτικής συμβουλής.\\

\noindent \textbf{Λέξεις Κλειδιά}: μηχανική μάθηση, ημι-επιβλεμόμενη μάθηση, κατανεμημένη μάθηση, αιτιώδης μάθηση, ερμηνεύσιμη μάθηση, διαχείριση μεγάλων γεωχωρικών δεδομένων, σημασιολογικός εμπλουτισμός, βιώσιμη και ανθεκτική γεωργία, επισιτιστική ασφάλεια, κοινή αγροτική πολιτική, πρακτικές συμβουλές για λήψη αποφάσεων.

% \end{abstract}
\end{otherlanguage} 

% \include{Chapter1/chapter1}
% \include{Chapter2/chapter2}

% *********************** Adding TOC and List of Figures ***********************

\tableofcontents

\listoffigures

\listoftables

% \printnomenclature[space] space can be set as 2em between symbol and description
%\printnomenclature[3em]

% ******************************** Main Matter *********************************
\mainmatter

%!TEX root = ../thesis.tex
%*******************************************************************************
%*********************************** First Chapter *****************************
%*******************************************************************************

\chapter{Introduction}  %Title of the First Chapter

\ifpdf
    \graphicspath{{Chapter1/Figs/Raster/}{Chapter1/Figs/PDF/}{Chapter1/Figs/}}
\else
    \graphicspath{{Chapter1/Figs/Vector/}{Chapter1/Figs/}}
\fi

%********************************** %First Section  **************************************

\section{Contextual background}

As we move towards the end of the first quarter of the 21st century, we are continuously witnessing the effects of climate change and the growing global population. There are a multitude of ways in which such issues are impacting the sustainability of life on Earth, and this means that we are brought up against the need to adapt our activities to the new reality.\\

\noindent Climate change and overpopulation are not just random examples. They are, in fact, two of the main reasons for concern in various domains like agriculture, economy, weather, and infrastructure, which are greatly impacted from their manifestations both directly and indirectly. On the bright side, we are living in times where we are witnessing a blooming space applications sector and an exponential increase in our computing capacity. These two might initially seem to be disconnected, but they actually comprise a very powerful combination when it comes to addressing the above issues.\\

\noindent\textbf{Food security}\\

\noindent The increased needs for nutrition, in combination with the impact of climate change on food production, has affected and will continue to affect the food sector~\citep{fritz2013need}. The agricultural productivity needs to be strengthened in order to accommodate the needs of the growing population, while preserving environmentally-friendly and sustainable agricultural practices. In this context, there is demand for the timely, large-scale and accurate monitoring of agricultural production and the provision of the necessary knowledge for evidence-based decision-making on food security matters~\citep{yifang2015global}. \\

\noindent To address the large-scale monitoring of agricultural land, Earth Observation (EO) data are widely used and constitute a unique  basis for deriving the necessary information. For the purposes of food security monitoring, the Food and Agriculture Organization of the United Nations (FAO), the European Union (EU), but also global and regional initiatives such as the Group on Earth Observations Global Agricultural Monitoring (GEOGLAM) and Asian Rice Crop Estimation and Monitoring (Asia-RiCE), have invested greatly in the exploitation of EO to monitor the extent, health, growth and productivity of the agricultural land over very large areas~\citep{rashid2009global,becker2010monitoring,rembold2019asap,whitcraft2015framework}.\\

\noindent It is worth-noting that even in this digital age, the current operating models for decision-making, at the highest level, still use outdated methods. Most actors, e.g., agri-food companies, governments, policy makers etc., employ approaches that are based on costly and time-consuming field visits and the collection of field data at sampled points. In most cases, this point information is then spatially interpolated through statistical techniques in order to extract the required large-scale production assessments. In this regard, EO data and higher level knowledge extracted through the use of Artificial Intelligence (AI) methods, is a key enabler for facilitating detailed assessments over large areas. This is achieved by providing spatial exhaustiveness, timeliness and high thematic precision. \\

\noindent The blooming space sector can now offer a number of very capable Earth observing satellites, continuously generating massive amounts of multispectral and Synthetic Aperture Radar (SAR) images of high spatial resolution. The exponential increase in our computing capacity, on the other hand, has enabled the application of sophisticated AI methods on these massive data. Therefore, we have the satellite data, and we have a set of environmental and socio-economic problems to solve. However, to convert the available data into training datasets for Machine Learning (ML) pipelines, we also require the relevant annotations, i.e., ground truth labels. 
While satellite imagery is not in short supply, ground-level observations are hard to find and they lack consistency in terms of spatial and temporal availability. Apart from being difficult to acquire, such labels are also the most expensive data component to generate; not just in terms of monetary cost, but also in terms of time, workforce and the availability of expert knowledge for the annotation. \\

\noindent Thus, ground truth labels are usually the missing piece to convert the available data into datasets, which will then be used for the training of ML models. As mentioned, the collection of such ground reference data labels requires extensive effort and, it is common for them to be scarcely available, especially in remote and dangerous areas that would benefit the most from remote sensing applications. This narrows the application of ML-based techniques with satellite imagery down to specific parts of the world despite these images being available at a global scale. And apart from their scarcity, costly acquisition, and burdensome quality assurance, we also have to overcome the challenge of their discoverability and openness. The issue of getting access to reliable ground truth data becomes even more challenging when dealing with large scale applications that cover vast areas at national or continental level. \\

\noindent\textbf{Common Agricultural Policy}\\

\noindent Increased agricultural productivity under environmentally friendly practices is an increasingly interesting topic and a top priority for the EU, manifesting predominantly in the form of the Common Agricultural Policy (CAP)~\citep{schmedtmann2015reliable}. The CAP is transitioning to a new era, namely the CAP 2020+ reform, in which the current operating model is set to be simplified and improved significantly. This will be achieved by leveraging big satellite data and advanced Information and Communication Technologies (ICT) that will ultimately form the so-called Area Monitoring System (AMS). The AMS is expected to fully automate and optimize the administration and control system of the CAP~\citep{devos2018a}. Until now, paying agencies, the implementation bodies of the CAP, had had to inspect at least 5\% of farmers' declarations, performing field visits or visual inspections on very high resolution satellite images. However, these methods are non-exhaustive, time-consuming, complex, and reliant on the skills of the inspector~\citep{sitokonstantinou2018scalable}. Recently, several solutions have emerged towards the development of a comprehensive AMS, taking advantage of EO data and ML methods to systematically monitor the agricultural land over very large areas~\citep{sitokonstantinou2018scalable}. \\

\noindent The backbone of all these solutions is the Copernicus program, specifically its Sentinel-1 and Sentinel-2 satellite missions. The Sentinel-1 twin satellites are equipped with radar sensors, which capture useful information for the detection of mowing, grazing and harvest events. The Sentinel-2 twin satellites are equipped with optical sensors, which capture multispectral images of the optical and near-infrared parts of the spectrum. The exploitation of Sentinel-2 data enables crop type mapping, crop growth predictions and a plethora of other crop monitoring applications. The Sentinel data are freely available and characterized by high revisit frequency (6 days for Sentinel-1 and 5 days for Sentinel-2) and  high spatial resolution (10-60 m), which are key features for an AMS. \\

\noindent In order to extract knowledge from Satellite Image Time-Series (SITS), and thereby make decisions on CAP compliance, ML algorithms are employed. In an AMS, SITS are combined with the parcel geometries (vector data) from the Land Parcel Identification System (LPIS) that has attached, for each parcel, the declared crop type label~\citep{sitokonstantinou2018scalable}. Some of the most common ML tasks, found in literature, are crop type classification, phenology and yield estimation and grassland mowing detection~\citep{rousi2020semantically,sitokonstantinou2020sentinel,rs10081300, 9553456}. AMS can significantly improve the controls and reduce the administrative burden for paying agencies, however they are not equipped to manage big satellite data. For this reason, Analysis Ready Data (ARD) and multidimensional datacubes are key building blocks for such systems~\citep{picoli2020cbers}. EO datacubes, i.e., multi-dimensional arrays, organise the data in such a way that anyone can intuitively exploit them~\citep{appel}. \\

\noindent Nevertheless, EO-derived information is not panacea. Sentinels' spatial and temporal resolution limitations make it difficult to confidently decide on special scenarios, such as when we have small and narrow parcels, or broad crop type categories, or cloudy scenes. Therefore, EO data and EO-driven information needs to be accompanied by timely in-situ observations. Typical in-situ data collection methods are expensive, time-consuming and therefore cannot provide continuous data streams. However, crowdsourced street-level images or images at the edge constitute an excellent alternative source~\citep{rs10081300}. \\

\noindent Remote sensing systems usually produce a large volume of data due to the high spatial, spectral, radiometric and temporal resolutions needed for applications in agriculture monitoring~\citep{sishodia2020applications}. 
Looking at the current, rapidly advancing, state-of-the-art in ML for EO, data availability is growingly perceived to be of multi-dimensional nature, through the notion of data variability and data complementarity, on top of data volume. Thus, more and more heterogeneous data sources are collected and data fusion techniques are applied in an attempt to generate enhanced feature spaces from non-overlapping and complementary feature domains. In EO applications during the recent years, fusion of data in different altitude levels, such as Unmanned Aerial Vehicles (UAVs) and high resolution satellite images, is very common~\citep{zhao2019finer, zhou2021uav}. \\

\noindent CAP paying agencies are facing challenges in their attempt to monitor the compliance of the farmers with the rules. Through the implementation of the AMS, the decisions that will be made for the subsidy allocations of all parcels have to be supported by evidence, either as a result of the predictions of AI pipelines, or through photo-interpretation, which is a common process that the paying agencies undertake (in retrospect) for validation purposes and for cases of dispute resolution. Thus, the need for such a holistic approach regarding annotated data availability is becoming evident.\\

\noindent\textbf{Resilient and smart farming}\\

\noindent Crop phenology is key information for crop yield estimation and agricultural management and thereby actionable knowledge for the farmer, the agricultural consultant, the insurance company and the policy maker. In a few words, crop phenology is the physiological development of the plant from sowing to harvest. The precise and timely knowledge of the growth status of crops is crucial for estimating the yield early in the season, but also for taking prompt action on controlling the growth to i) maximize the production and ii) reduce the farming costs~\citep{gao2021mapping}. Phenology estimation is particularly important for the following reasons. \\

\noindent Crops' water needs are a function of the phenological stage. Using the example of cotton, which is the crop of interest in this study, there is higher water usage between the flowering and early boll opening stages than in the emergence and late boll opening stages~\citep{vellidis2016development}. In more detail, irrigation can be interrupted on the onset of boll opening to stop the continuous growth of cotton and allow the photosynthetic carbohydrates to start contributing to the development of bolls and not the development of leaves and flowers~\citep{Cotman}.  Therefore, we need crop phenology information in order to make irrigation recommendations towards fully utilizing the expensive and often scarce water, and at the same time reduce water stress and its potential adverse effects on the yield~\citep{anderson2016relationships}. Irrigation is one of the many examples of how phenology can benefit the agricultural practice management. Other examples include the precise fertilization, pest management and harvesting~\citep{gao2021mapping}. \\
 
% \noindent Farm profit and farmers' income is not certain, since crop yield is exposed to a number of risks~\citep{hardaker2004coping}. Adverse weather events pose a constant threat to farmers, as they can cause reduced yields, suboptimal product quality or even complete destruction of the production. For this reason, farmers usually insure their crops. Nonetheless, agricultural insurance and yield loss compensation is not always a fair nor transparent process. Yield losses caused by adverse weather events can differ greatly based on the phenological stage of the crop at the instance of the event. Indemnity-based agricultural insurance is characterized by asymmetries in information. In certain cases, weather index insurance has been used as an alternative, with the benefit of being contingent on specific events~\citep{turvey2001weather}. However, weather index insurance does not account for the actual farm losses. Here, phenology and yield estimation can play a key role and offer an event-specific assessment on the expected yield damage~\citep{sitokonstantinou2021scalable}. \\

\noindent Phenology information can also be used for taking action or schedule agricultural practices in view of upcoming weather events. Therefore, weather event predictions and phenology estimations make a powerful combination for meaningful decision making towards preventing yield losses and cutting costs. One example would be to intensify irrigation prior to a predicted heatwave or interrupt a scheduled irrigation when precipitation is expected. Another example would be the application of fertilizers depending on rainfall and phenology predictions. If rain is expected, the farmer can avoid the application of foliar fertilizer or fungicide, as they would rinse before they are absorbed. Moreover, cotton picking could be rushed prior to an anticipated hail event, if phenology estimations show a near complete boll opening status~\citep{sitokonstantinou2021scalable}. \\

% \noindent The recently published positioning papers of Gao and Zhang (2021), Lacueva-Perez et al. (2020) and Potgieter et al. (2021) identify the problem of remote phenology estimation as a fundamental one for the future of agriculture monitoring~\citep{gao2021mapping,lacueva2020multifactorial,potgieter2021evolution}. Particularly, the authors underline the importance and the expected impact of within-season estimations at high spatial resolutions. In this work we exploit Earth Observation (EO) data (Sentinel-2), together with numerical simulations of atmospheric and soil parameters, to address the within-season phenology estimation for cotton at the parcel level. Even more, since ground truth data is scarce and expensive to collect, we approach the problem in a fully unsupervised learning way, in order to be truly useful in real world scenarios. The detailed contributions of this study are listed in Section \ref{related} E. \\

\noindent For many years, phenology has been observed from the ground, through field visits and in-situ sensors. These approaches however are expensive, time-consuming and lack spatial variability. To this end, space-borne and aerial remote sensing Vegetation Index (VI) time-series have been used to systematically monitor crop phenology over large geographic regions; often termed land surface phenology~\citep{gao2021mapping}. The freely available Sentinel-2 data, but also the low-cost Planet data, offer optical imagery of unprecedented temporal and spatial characteristics that introduce new opportunities for the large-scale and near real-time monitoring of phenology~\citep{jianwu2016emerging,sitokonstantinou2020sentinel}. However, there are still a number of issues to be tackled. First of all, cloud coverage disrupts the continuity of the optical SITS, which is particularly damaging when monitoring fast progressing targets, like phenology. Additionally, the sensitivity of SITS on crop growth varies for different crop types and different growth stages. To this end, there have been multiple studies that additionally incorporate atmospheric and soil parameters that unlike VIs do not monitor the change in vegetation but the drivers of these changes. \\

\section{Objectives}
The main objectives of this thesis are outlined as follows:

\begin{itemize}
    \item Propose solutions to address the scarcity of annotated data for the EO-based ML tasks of crop type classification and crop phenology estimation.
    \item Propose solutions for handling the big data requirements associated with the large-scale monitoring of agriculture.
    \item Develop applications that address real-world problems in the thematic areas of i) monitoring of the CAP, ii) food security and iii) smart and resilient farming. The applications should satisfy the requirements of operational scenarios, i.e., a)  the ever-present lack of ground truth data for training and evaluating ML models, which leads to generalization limitations (w.r.t. area, time, crop type), b) enrich the ML outputs with high level knowledge to provide actionable advice. 
\end{itemize}

\section{Contributions}
This thesis is motivated by issues associated with three thematic domains of EO-enabled agriculture monitoring. These thematic domains define the thesis outline, as elaborated in the next subsection, i.e., i) the monitoring of the CAP (Chapter 2), ii) food security monitoring (Chapter 3) and iii) sustainable and resilient farming (Chapter 4). The main contributions of this thesis are summarized as follows.

\begin{itemize}
    \item The development of a pre-processing framework for the production of harmonized, analysis-ready, Sentinel-1 and Sentinel-2 time-series over large areas. \item The development of a multi-crop classification ML pipeline that exploits the farmers declarations for training. This is interesting from an engineering/scientific point of view for two reasons: i) these labels are widely available and thus resolve the ground-truth scarcity issue and ii) these labels are noisy, as farmers do not always declare the cultivated crop correctly.
    \item The development of a semantic enrichment method to enhance the knowledge extracted for the crop type maps towards addressing the operational needs of CAP inspectors. Using the crop type maps and the semantic enrichment method, a framework for the smart sampling of the on-the-spot checks was developed. Therefore, inspectors, instead of visiting randomly sampled parcels for checks, they visit the ones that are most probable wrong declarations.
    \item The development of a space-to-ground data availability framework. Satellite data cannot provide 100\% certainty for decision-making. Ancillary crowdsourced data, such as street-level images are an excellent alternative. A Sentinel and street-level analysis-ready dataset is provided, discussing the value of fusing such heterogenous sources and offering preliminary results. 
    \item The development of a semi-supervised rice classification scheme with two key considerations: i) use only a confined number of labels, ii) offer geographic generalization for large-scale applications. \item The development of a distributed data storage and processing framework for the large-scale classification of rice in a high performance data analytics environment. 
    \item The development of a semi-supervised crop phenology estimation method that uses merely a handful of labels. The method introduces a continuous phenology scale and addresses phenology estimation as a regression problem. This provides enhanced specification on the growth status of the crop, unlike the more common classification methods that are able to detect merely the principal phenological stages.
    \item The development of an unsupervised fuzzy clustering method for phenology estimation. The method does not use ground truth that makes it truly useful in real-world scenarios. Additionally, using the fuzziness scores one can identify the transitional periods between two phenological stages, increasing the specificity of the phenological stage prediction.
    \item The development of a causal ML method to assess the impact of crop rotation and landscape crop diversity on climate regulation, conditioning for crop and climatic factors. It is proposed to use the Conditional Average Treatment Effect (CATE) estimates as suitability scores for applying the practices of rotation and diversity.
    \item The development of an interpretable ML method for estimating the onset of harmfulness of pests in cotton fields. The glass-box nature of the model provides important insights on the drivers of pest presence and the inter-driver interactions.

\end{itemize}
\section{Scientific publications}

The developed methods and experimental results of this PhD thesis have been disseminated through a number of scientific publications as listed below.\\

\textbf{Journals}

\begin{enumerate}
    \item \textbf{Sitokonstantinou, V.}, Koukos, A., Tsoumas, I., Bartsotas, N., Kontoes, C. and Karathanassi, V. (2022). Fuzzy clustering for the within-season estimation of cotton phenology. PLOS ONE (under review).
    \item \textbf{Sitokonstantinou, V.}, Koukos, A., Drivas, T., Kontoes, C., Papoutsis, I., and Karathanassi, V. (2021). A Scalable Machine Learning Pipeline for Paddy Rice Classification Using Multi-Temporal Sentinel Data. Remote Sensing, 13(9), 1769.
    \item Rousi, M., \textbf{Sitokonstantinou, V.}, Meditskos, G., Papoutsis, I., Gialampoukidis, I., Koukos, A., ... and Kompatsiaris, I. (2020). Semantically enriched crop type classification and linked earth observation data to support the common agricultural policy monitoring. IEEE Journal of Selected Topics in Applied Earth Observations and Remote Sensing, 14, 529-552.
    \item \textbf{Sitokonstantinou, V.}, Papoutsis, I., Kontoes, C., Lafarga Arnal, A., Armesto Andrés, A. P., and Garraza Zurbano, J. A. (2018). Scalable parcel-based crop identification scheme using Sentinel-2 data time-series for the monitoring of the common agricultural policy. Remote Sensing, 10(6), 911.

\end{enumerate}

\textbf{Conferences}

\begin{enumerate}
    \item Giannarakis, G., \textbf{Sitokonstantinou, V.}, Roxanne, L. and Kontoes, C. (2022). Towards assessing agricultural land suitability using causal machine learning. In Proceedings of the IEEE/CVF Conference on Computer Vision and Pattern Recognition Workshops (CVPRW). 
    \item Drivas, T.*, \textbf{Sitokonstantinou, V.}*, Tsardanidis, I., Koukos, A., Karathanassi, V. and Kontoes, C. (2022). A Data Cube of Big Satellite Image Time-Series for Agriculture Monitoring. In 2022 IEEE 14th Image, Video, and Multidimensional Signal Processing Workshop, IVMSP.
    \item Nanushi, O.*, \textbf{Sitokonstantinou, V.}*, Tsoumas, I. and Kontoes, C. (2022). Pest presence prediction using interpretable machine learning. In 2022 IEEE 14th Image, Video, and Multidimensional Signal Processing Workshop, IVMSP.
    \item Choumos, G.*, Koukos, A.*, \textbf{Sitokonstantinou, V.} and Kontoes, C. (2022). Towards space-to-ground data availability for the monitoring of the common agricultural policy. In 2022 IEEE 14th Image, Video, and Multidimensional Signal Processing Workshop, IVMSP.
    \item \textbf{Sitokonstantinou, V.}, Koukos, A., Drivas, T., Kontoes, C., and Karathanassi, V. (2022). DataCAP: A Satellite Datacube and Crowdsourced Street-Level Images for the Monitoring of the Common Agricultural Policy. In International Conference on Multimedia Modeling (pp. 473-478). Springer.
    \item \textbf{Sitokonstantinou, V.}, Koukos, A., Kontoes, C., Bartsotas, N. S., and Karathanassi, V. (2021, July). Semi-Supervised Phenology Estimation in Cotton Parcels with Sentinel-2 Time-Series. In 2021 IEEE International Geoscience and Remote Sensing Symposium IGARSS (pp. 8491-8494). IEEE.
    \item \textbf{Sitokonstantinou, V.}, Drivas, T., Koukos, A., Papoutsis, I., Kontoes, C. and Karathanassi, V. (2020). Scalable distributed random forest classification for paddy rice mapping. In Proceedings of the Asian Remote Sensing Conference (ACRS 2019) (pp. 836-845).
    \item \textbf{Sitokonstantinou, V.}, Koutroumpas, A., Drivas, T., Koukos, A., Karathanassi, V., Kontoes, H., and Papoutsis, I. (2020). A Sentinel based agriculture monitoring scheme for the control of the CAP and food security. In Eighth International Conference on Remote Sensing and Geoinformation of the Environment (RSCy2020) (Vol. 11524, pp. 48-59). SPIE.
    
\end{enumerate}
\section{Thesis outline}

The contents of this thesis are divided into five parts. The first part (Chapter 2) includes work that is related to the monitoring of the CAP, from the retrieval and processing of big satellite data (data cubes) to crop type classification to the provision of actionable advice for the stakeholders (smart sampling of CAP checks). The second part (Chapter 3) focuses on the implications associated with large-scale applications for food security monitoring. The third part (Chapter 4) includes the development of methods for the timely and accurate estimation of crop phenology towards enabling sustainable and resilient farming. The fourth part (Chapter 5) sets the scene for future work, focusing on how we can close the gap between ML outputs and decision and/or policy making through causal and interpretable ML. Chapter 6 provides the conclusions and future work. A brief 
description of each chapter is presented as follows:\\

\noindent \textbf{Chapter 2}: \textit{Monitoring of the CAP}\\

\noindent This chapter presents a data cube that allows the effortless and transparent handling of big SITS. SITS processing functionalities are built on top of the data cube that enable the feature space creation for ML tasks, the animation of satellite data for decision making and the querying of complex questions related to the monitoring of the CAP. This chapter also includes a crop type classification method and a smart sampling algorithm that allows CAP inspectors to target their On-The-Spot-Checks (OTSCs) based on the crop map results. Finally, this chapter introduces the notion of space-to-ground data availability, combining satellite data and street-level images for crop classification. The aforementioned work culminates to DataCAP, an application that allows CAP inspectors to visually verify the compliance of farmers to their obligations. \\

\noindent \textbf{Chapter 3}: \textit{Food Security Monitoring}\\

\noindent This chapter presents work related to the monitoring of food security and particularly the big data considerations associated with the large-scale monitoring of agriculture. In this chapter, the focus is on paddy rice in South Korea, monitoring the growth and extracting biomass and yield indicators. Additionally, a methodology for the large-scale mapping of paddy rice is presented, focusing on i) using only few ground truth labels and ii) the geographic generalization of the model. \\

\noindent \textbf{Chapter 4}: \textit{Resilient \& Smart Farming}\\

\noindent This chapter presents two cotton phenology estimation methods towards resilient farming. Crop phenology is crucial for sustainable, resilient and cost-effective farm management. In the first section of this chapter phenology estimation is performed using a confined number of labels under a semi-supervised approach. In the second section, phenology estimation is done using a fuzzy clustering approach in an attempt to address the ever-present issue of scarce ground-truth. This chapter also includes the description of the development of unique dataset of crop phenology ground observations retrieved through extensive field campaigns using a new collection protocol.  \\

\noindent \textbf{Chapter 5}: \textit{From Knowledge to Wisdom}\\

\noindent This chapter briefly summarizes the conclusions of the previous chapters and discusses future perspectives. The core of this thesis are detection tasks that make use ML models and EO data to offer data products and services for the benefit of the agriculture sector. In this last chapter, it is discussed how we can go beyond mere detection (knowledge extraction), i.e. crop classification and phenology estimation, and move towards the understanding of the drivers of change in agro-ecosystems (wisdom extraction). In this direction, this chapter offers two preliminary works as such examples: i) assessment of agricultural land suitability using causal ML and ii) pest presence prediction using interpretable ML.

\nomenclature[z-DEM]{DEM}{Discrete Element Method}
\nomenclature[z-FEM]{FEM}{Finite Element Method}
\nomenclature[z-PFEM]{PFEM}{Particle Finite Element Method}
\nomenclature[z-FVM]{FVM}{Finite Volume Method}
\nomenclature[z-BEM]{BEM}{Boundary Element Method}
\nomenclature[z-MPM]{MPM}{Material Point Method}
\nomenclature[z-LBM]{LBM}{Lattice Boltzmann Method}
\nomenclature[z-MRT]{MRT}{Multi-Relaxation 
Time}
\nomenclature[z-RVE]{RVE}{Representative Elemental Volume}
\nomenclature[z-GPU]{GPU}{Graphics Processing Unit}
\nomenclature[z-SH]{SH}{Savage Hutter}
\nomenclature[z-CFD]{CFD}{Computational Fluid Dynamics}
\nomenclature[z-LES]{LES}{Large Eddy Simulation}
\nomenclature[z-FLOP]{FLOP}{Floating Point Operations}
\nomenclature[z-ALU]{ALU}{Arithmetic Logic Unit}
\nomenclature[z-FPU]{FPU}{Floating Point Unit}
\nomenclature[z-SM]{SM}{Streaming Multiprocessors}
\nomenclature[z-PCI]{PCI}{Peripheral Component Interconnect}
\nomenclature[z-CK]{CK}{Carman - Kozeny}
\nomenclature[z-CD]{CD}{Contact Dynamics}
\nomenclature[z-DNS]{DNS}{Direct Numerical Simulation}
\nomenclature[z-EFG]{EFG}{Element-Free Galerkin}
\nomenclature[z-PIC]{PIC}{Particle-in-cell}
\nomenclature[z-USF]{USF}{Update Stress First}
\nomenclature[z-USL]{USL}{Update Stress Last}
\nomenclature[s-crit]{crit}{Critical state}
\nomenclature[z-DKT]{DKT}{Draft Kiss Tumble}
\nomenclature[z-PPC]{PPC}{Particles per cell}
\chapter{Monitoring of the CAP}

% %changed 20/04
\ifpdf
    \graphicspath{{Chapter2/Figs/Raster/}{Chapter2/Figs/PDF/}{Figs/Chapter 2/}}
\else
    \graphicspath{{Figs/Chapter2/Vector/}{Figs/Chapter2/}}
\fi

\begin{figure}
\centering\includegraphics[scale=0.14]{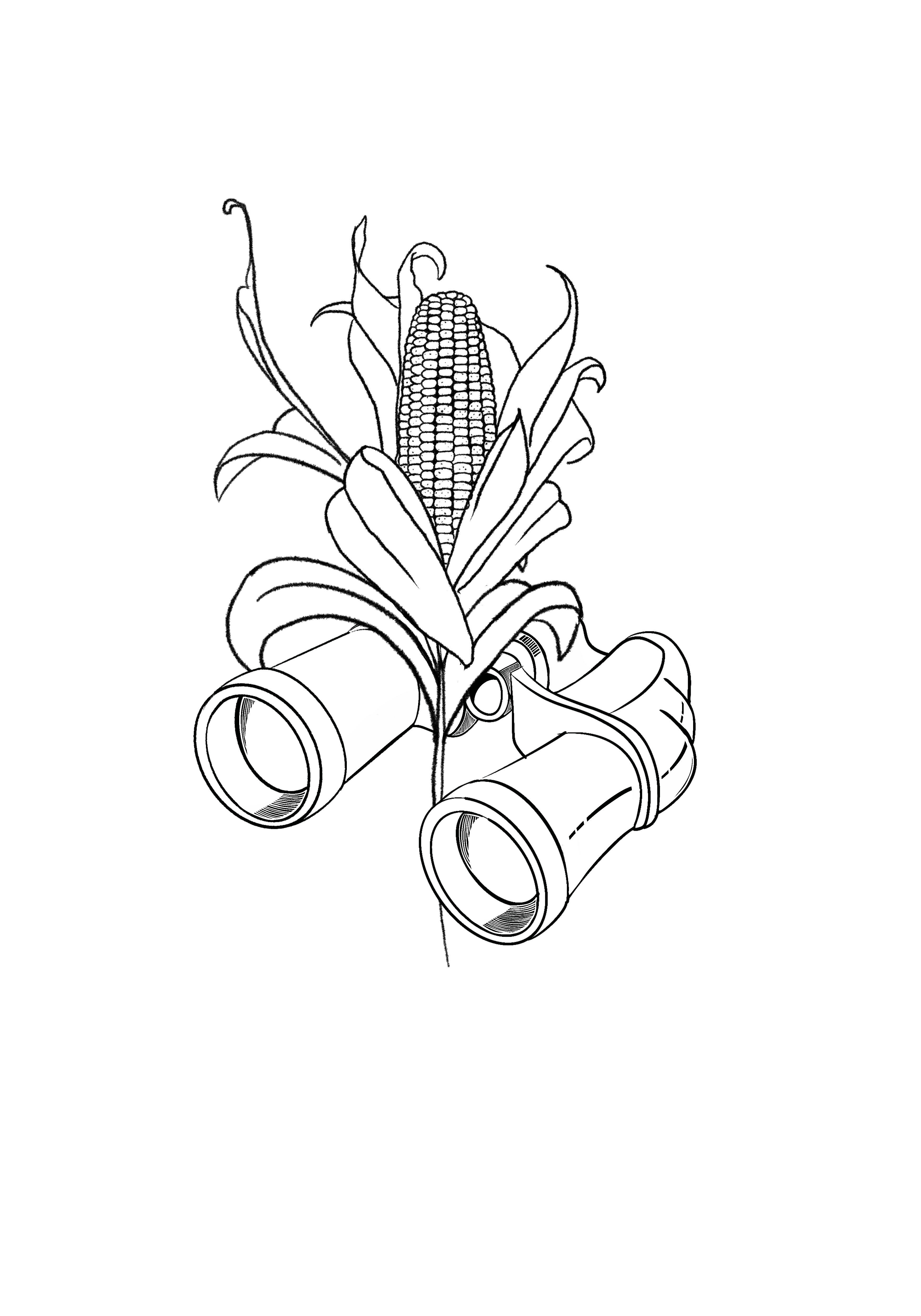}
\end{figure}

% \chapter{Monitoring of the CAP}

% %changed 20/04
% \ifpdf
%     \graphicspath{{Chapter2/Figs/Raster/}{Chapter2/Figs/PDF/}{Figs/Chapter 2/}}
% \else
%     \graphicspath{{Figs/Chapter2/Vector/}{Figs/Chapter2/}}
% \fi

\section{Introduction}

\noindent The modernization of the CAP requires the large scale and frequent monitoring of agricultural land. Towards this direction, the free and open Copernicus data (i.e., Sentinel missions) have been extensively used as the sources for the required high spatial and temporal resolution Earth observations. Nevertheless, monitoring the CAP at large scales constitutes a big data problem and puts a strain on CAP paying agencies that need to adapt fast in terms of infrastructure and know-how. Hence, there is a need for efficient and easy-to-use tools for the acquisition, storage, processing and exploitation of big satellite data. Section \ref{datacube} presents the Agriculture monitoring Data Cube (ADC), which is an automated, modular, end-to-end framework for discovering, pre-processing and indexing optical and SAR images into a multidimensional cube. ADC also offers a set of powerful tools on top of the cube, including i) the generation of analysis-ready feature spaces of big satellite data to feed downstream ML tasks and ii) the support of SITS analysis via services pertinent to the monitoring of the CAP (e.g., detecting trends and events, monitoring the growth status etc.). The knowledge extracted from the SITS analyses and the ML tasks returns to the data cube, building scalable country-specific knowledge bases that can efficiently answer complex and multi-faceted geospatial queries.\\

\noindent Furthermore, in Section \ref{croptype} is prestend a Sentinel-2 based crop classification method for the monitoring of the CAP. Specifically, a parcel-based supervised crop classification method was implemented and evaluated in Navarra, Spain. The method makes use of supervised classifiers Support Vector Machines (SVMs) and Random Forest (RF) to discriminate among the various crop types, based on a large feature space of Sentinel-2 images and Vegetation Index (VI) time-series. The classifiers are separately applied at three different levels of crop nomenclature hierarchy, comparing their performance with respect to accuracy and execution time. SVM provides optimal performance and proves significantly superior to RF for the lowest level of nomenclature, resulting in 0.87 Cohen’s kappa coefficient. Experiments were carried out to assess the importance of input variables, where top contributors are the Near-Infrared (NIR), vegetation red-edge, and Short-Wave Infrared (SWIR) multispectral bands, and the Normalized Difference Vegetation (NDVI) and Plant Senescence Reflectance (PSRI) indices, sensed during advanced crop phenology stages. The method is finally applied to a Landsat-8 OLI based equivalent variable space, offering 0.70 Cohen’s kappa coefficient for the SVM classification, highlighting the superior performance of Sentinel-2 for this type of application. This is credited to Sentinel-2’s spatial, spectral, and temporal characteristics.\\

\noindent Satellite data can be enriched with semantic annotations for the creation of value-added products. One challenge is to extract knowledge from the raw Sentinel-2 data in an automated way and to effectively manage the extracted information in a semantic way, to allow fast and accurate decisions of spatio-temporal nature in a real operational scenarios. Section \ref{semantics} presents a framework that combines supervised learning for crop type classification on SITS with Semantic Web and Linked Data technologies to assist in the implementation of the rules set by the CAP. The framework collects georeferenced data that are available online and satellite images from the Sentinel-2 mission. We analyze image time-series that cover the entire cultivation period and link each parcel with a specific crop. On top of that, we introduce a semantic layer to facilitate a knowledge-driven management of the available information, capitalizing on ontologies for knowledge representation and semantic rules, to identify possible farmers non-compliance according to the Greening 1 (Crop Diversification) and SMR 1 rule (protection of waters against pollution caused by nitrates) rules of the CAP. Experiments show the effectiveness of the proposed integrated approach in three different scenarios for crop type monitoring and consistency-checking for non-compliance to the CAP rules: i) the smart sampling of on-the-spot-checks, ii) the automatic detection of CAP's Greening 1 rule and iii) the automatic detection of susceptible parcels according to the CAP’s SMR 1 rule.\\

\noindent  Advanced remote sensing systems have been developed towards the large-scale evidence-based monitoring of the CAP. Nevertheless, the spatial resolution of satellite images is not always adequate to make accurate decisions for all fields. Section \ref{street} presents the notion of space-to-ground data availability, i.e., from the satellite to the field, in an attempt to make the best out of the complementary characteristics of the different sources. We present a space-to-ground dataset that contains Sentinel-1 radar and Sentinel-2 optical image time-series, as well as street-level images from the crowdsourcing platform Mapillary, for grassland fields in the area of Utrecht for 2017. The multifaceted utility of the dataset is showcased through the downstream task of grassland classification. We train traditional ML and DL models on these different data domains and highlight the potential of fusion techniques towards increasing the reliability of decisions. \\

\noindent In Section \ref{datacap}, we demonstrate DataCAP that combines the ADC, with ML pipelines and crowdsourced street-level images to assist in the monitoring of the CAP. DataCAP is an application that offers a suit of processing tools to simply and intuitively  search,  store and analyse radar and optical satellite images, along with visualisation tools that combine satellite and street-level imagery for the visual verification of model decisions.\\

\section{Data Cube}\label{datacube}

\subsection{Literature review}

\noindent The CAP has set out to implement radical changes towards fairer, greener and more performance-based policies \cite{newcap}. In this context, and inspired by the advent of free and open satellite data and the recent advancements in data science, the CAP aims at the country-wide evidence-based  monitoring of the farmers' compliance with the agricultural policies. Towards this effort, the Sentinel satellite missions, advanced ICT technologies and AI have been identified as key enablers \cite{datacap}. \\

\noindent The Sentinels provide frequent optical and SAR images of high spatial resolution and have been extensively used for the monitoring agriculture and specifically for the purposes of the CAP \cite{cap2, semantics}. Most Sentinel-based CAP monitoring systems utilize the parcel boundaries from the LPIS. LPIS is a database that connects the crop type label, as declared by the farmer, to each parcel object. The LPIS objects and crop labels are then combined with the SITS to feed AI models for CAP monitoring applications \cite{datacap}.\\

\noindent CAP monitoring systems need to be able to process and visualize  large amounts of satellite data, which is not possible using traditional local storage and processing workflows, and for this reason big EO management technologies are necessitated \cite{cubes_lewis}. One such example are the EO Data Cubes (EODCs) that can handle large volumes and provide a solid solution for accessing and managing ARD~\cite{cubes_giuliani}. Currently, several EODC technologies have been developed, e.g., Google Earth Engine \cite{gee_official}, Sentinel-hub \cite{sentinelhub}, gdalcube \cite{gdalcube}, Rasdaman \cite{RasDaMan_official}, ODC, OpenEO \cite{openeo}, and Earth System \cite{earth_system}.\\

\noindent In this work, we demonstrate an end-to-end workflow for building and exploiting a scalable agriculture monitoring data cube based on ODC, which is hosted on CreoDIAS, one of the five Data and Information Access Services (DIAS) cloud platforms. ODC is open and infrastructure-independent, so it can be installed in diverse environments, from personal computers to supercomputers \cite{odc_countries}. It enables effortless data management and simplifies data querying, using a Python Application Programming Interface (API) \cite{cubes_catalan}. In ODC-based systems, data are indexed or ingested into data cubes and can be then loaded into xarrays, which is a  powerful multidimensional data structure. The simplification of processing and exploitation of big spatio-temporal data, using data cubes, allows for unlocking their full potential and strengthens the connection between data and users \cite{cubes_kopp}. The ODC technology has been used for the implementation of a number of national data cubes, i.e., Australia \cite{odc_australia}, Africa \cite{odc_africa}, Switzerland \cite{odc_swiss}, Colombia \cite{odc_columbia}, Brazil \cite{odc_brazil} and Taiwan \cite{odc_taiwan}. \\

\noindent Our solution comes with several notable advantages when compared with the related work, as it includes both satellite data (SAR and optical) and auxiliary geospatial data, such as the farmers' declarations (crop type labels) and parcel boundaries from the LPIS, opening the door to fast and accurate operations between them. This is crucial for operational scenarios of CAP monitoring applications. In addition, we build several tools on top of the ADC. Specifically, we built functionalities that enable the effortless generation of corrected, cleaned and smoothed SITS that are formed into feature spaces that feed AI pipelines with ARD. We also offer functionalities for computing multidimensional statistics and in turn smart geospatial queries, enabling the recognition of patterns and trends and the detection of events on agricultural land. The outputs of the AI models (e.g., crop classification and grassland mowing detection) fed by ADC, along with the multidimensional information extracted from the monitoring of an area throughout the years, populate the cubes through a feedback loop. This way, we generate unique country-specific EO knowledge bases for agriculture monitoring. These knowledge bases allow for the evidence-based decision-making by fully exploiting big EO data products and results of AI models. \\

%%%%%%%%%%%%%%%%%%%%%%%%%%%%%%%%%%%%%%%%%%%%%%%%%%%%%%%%%%%%%%%%%%%%%%%%%%%%%%%%%%%%%%%%%%%%%%%%%%%%%%%%%%%%%%%%%%%%%%%%%%%%%%%%%%%%%%%%%%%%%%%%%%%%%%%%%%%%%%%%%%%%%%%%%%%%%%

\subsection{Agriculture monitoring data cube}\label{sec:ii}
% \medskip
\noindent The architecture of the ADC is based on two layers. The first layer is related to data discovery and acquisition, and the second layer is for the production of ARD. The ARD generation layer includes a number of pre-processing tools for SAR and optical images that make use of open-source libraries. Instructions and code for setting up a data cube like ADC can be found in https://github.com/Agri-Hub/datacap. 

\subsubsection{Access to Sentinel data}

\noindent Copernicus Sentinel missions and specifically Sentinel-1 (SAR) and Sentinel-2 (optical) have contributed to the monitoring of agriculture by providing high temporal and high spatial resolution images at global scale. Sentinel-2 offers images in the visible and near-infrared parts of the spectrum, making them ideal for vegetation monitoring. Apart from the surface reflectances of Sentinel-2, vegetation indices, e.g., NDVI, can also be extracted from optical images to enhance certain vegetation characteristics (e.g., water content, physiologocal stress etc.). SAR images (Sentinel-1) have also been used in related work, either as stand-alone or complementary to optical images, as they are not affected by clouds and allow for constructing dense SITS that are essential in agriculture monitoring \cite{s1cap2}. \\

\noindent To generate ARD we need to consider both data acquisition and data storage. To minimize the effort of accessing data, ADC has been developed within the CreoDIAS cloud platform. This allows us to access data directly via an offered catalogue (eodata), which according to the latest statistics includes more than 27 PB of Sentinel-1 and Sentinel-2 data \cite{creodias}. The CreoDIAS object storage repository ensures good performance and eliminates the need for repeated downloading of raw data locally. It is worth mentioning that CreoDIAS hosts the complete archive of Sentinel data, and thus we do not have to search across multiple data hubs. This is often a complicated and strenuous process due to the large number of sources and their varying performances (i.e., download speed, latency etc.). \\

\noindent Our ADC includes both Sentinel-1 and Sentinel-2 products that cover Cyprus and Lithuania over the span of three years (2019-2021). This results to a total number of approximately 15 TB of data, which are pre-processed before indexed in the ADC. The metadata of the products, as retrieved from the CreoDIAS API, are stored to a database. Thus, spatial operations are allowed giving the potential for statistics extraction. In addition, flags per product are created in every step of the pre-processing chain aiming at the monitoring of possible failures and the re-execution of the problematic task, if needed. \\

\subsubsection{Generation of Analysis Ready Data}
% \smallskip
\noindent In order to convert the raw satellite data to ARD, certain processing steps are required, i.e., i) atmospheric corrections for optical images (Sentinel-2) using the Sen2Cor software (Level 2A products) and ii) the generation of backscatter coefficients ($\sigma_{0}$) and interferometric coherence for the SAR data (Sentinel-1) using the python library snappy. Then all the products were converted to Cloud Optimized TIFFs that are more efficient in terms of storage and loading speed and thus fast enough for web applications. In addition, as ADC handles data from two sensors, it is important for all the pixels to have the same spatial resolution and be spatially aligned. For this, Sentinel-1 and Sentinel-2 data are resampled to 10 m and matched pixel-to-pixel, thus enabling efficient, sensor-agnostic time-series data analysis. The final step of the pre-processing chain is the masking of clouds and shadows on the optical images; the detection of which is done using the Sentinel-2 scene classification product of the Sen2Cor algorithm.\\
 
\begin{figure}[!ht]
\centering
\includegraphics[scale=1.8]{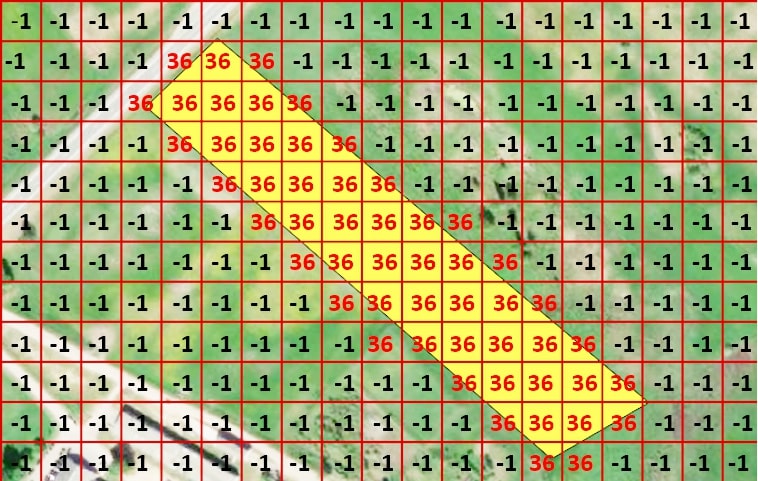}
\caption{Example of how parcels are indexed in the data cube as rasters. With $-1$ are the pixels that do not belong to parcel with $id=36$.}
\label{fig:geom_index}
\end{figure}

\noindent ADC is able to scale efficiently and cover large areas, which possibly include millions of objects (parcels), for which we need to compute statistics and execute geospatial queries. Besides serially querying the data cube for each parcel, using its boundaries in vector format, we can also rasterize them as shown in Fig.\ref{fig:geom_index}. Then we index the layer of rasterized parcels and load them into the data cube. This enables the fast and parallel computation of zonal statistics per parcel using the powerful xarrays data structure. We use the \textit{groupby} function, which allows for the grouping of the xarray dimensions based on the respective IDs. Table \ref{tab:processing_time} shows the processing time for generating zonal statistics for a varying number of parcels using the serial querying method and the groupby method. It is observed that for a large number of parcels, groupby is far more computationally economic. \\ 

\begin{table}[!ht]
\caption{Execution time for generating monthly averages for one year and one Sentinel-2 band over one tile.}
\label{tab:processing_time}
\centering
\scalebox{1}{
\begin{tabular}{|c|c|c|}
\hline
\textbf{\# parcels}& 
\textbf{groupby} & \textbf{serial querying}\\ \hline
% $10^3$ & 68.89 sec & 252.38 sec\\ \hline
% $10^4$ & 71.36 sec & 2401.83 sec\\ \hline
% $10^5$ & 151.84 sec & $24243.34$ sec\\ \hline
% $10^5$ & 151.84 sec & $>2\times10^4$ sec\\ \hline
$1$ k & $69$ sec & $250$ sec\\ \hline
$10$ k & $71$ sec & $40$ min\\ \hline
$100$ k  & $150$ sec & $400$ min\\ \hline
\end{tabular}}
\end{table}

\subsection{Satellite image time-series analysis}
% \medskip
\noindent Long and dense SITS are essential for agriculture monitoring since crops change dynamically with time. We need to capture these changes to i) detect trends and events (e.g., grassland mowing detection), ii) monitoring the growth status and health, iii) classify crops to verify the validity of declarations and many more applications pertinent to the monitoring of the CAP. We enable the effortless, rapid and large scale analysis of long and dense SITS, by exploiting the capabilities of xarrays. In this section, practical scenarios are described to showcase in detail the functionalities that are built on top of the ADC and specifically how we can reduce noise, detect trends, speed up work and improve reliability of decision-making. The code and data are available at \url{https://github.com/Agri-Hub/ADC}. 
\subsubsection{Spatial buffers}
\begin{figure}[!ht]
\centering
\includegraphics[scale=0.4]{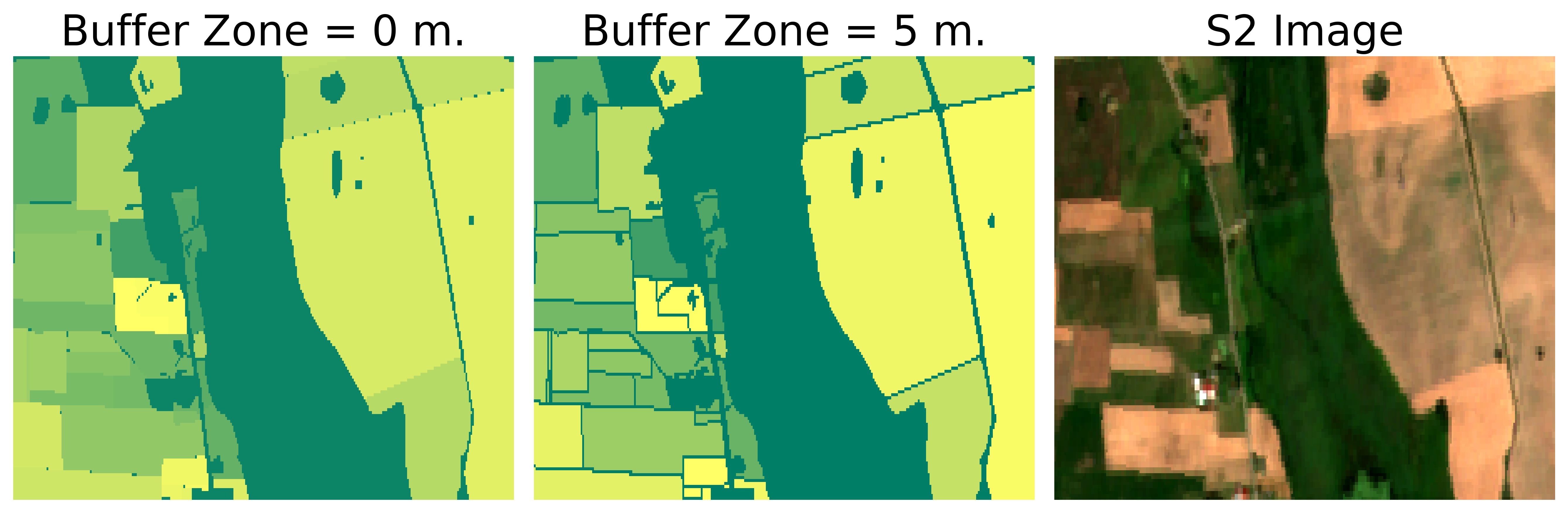}
\caption{Inward buffer to parcel boundaries to avoid using mixed pixels.}
\label{fig:parcels_buffering}
\end{figure}

% \smallskip
\noindent One of the main advantages of ADC is the rasterized information of the parcels, described in Sec. \ref{sec:ii}. Apart from the real parcel geometries, one can use our inward buffer functionality, as shown in Fig.\ref{fig:parcels_buffering}, which can alleviate the adverse consequences of mixed pixels. This way, we can extract only the representative information from the rest of the pixels encompassed inside the parcels' buffered geometries.\\

\begin{figure}[ht!]
\centering
\includegraphics[scale=0.4]{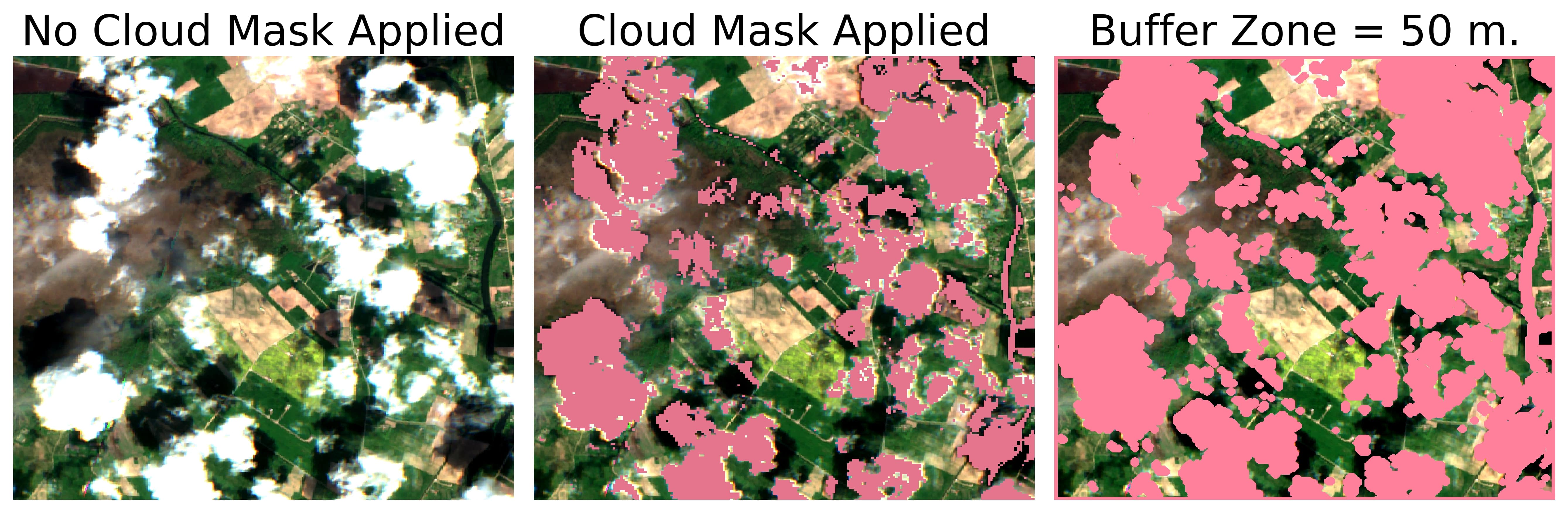}
\caption{Applying an outward buffer on cloud and shadow objects.}
\label{fig:cloud_masking}
\end{figure}

\noindent We additionally provide an adjustable outward buffer that can be applied on the less-than-perfect cloud and shadow mask products in order to reduce some of the noise. The Sen2Cor scene classification product has suboptimal recall for the cloud and shadow classes. However, for many agriculture monitoring applications it is important to have only clear pixels involved in the analyses. To tackle this issue, one can apply an outward buffer zone around the masked cloud and shadow objects. As a result, the pixels adjacent to clouds are now classified as cloudy, providing a trade-off between better cloud masking and fewer clear pixels for analysis. The original cloud mask and a cloud mask with a buffer are illustrated in Fig. \ref{fig:cloud_masking}.

\subsubsection{SITS preparation}
% \smallskip

\begin{figure}[!ht]
\centering
\includegraphics[scale=0.4]{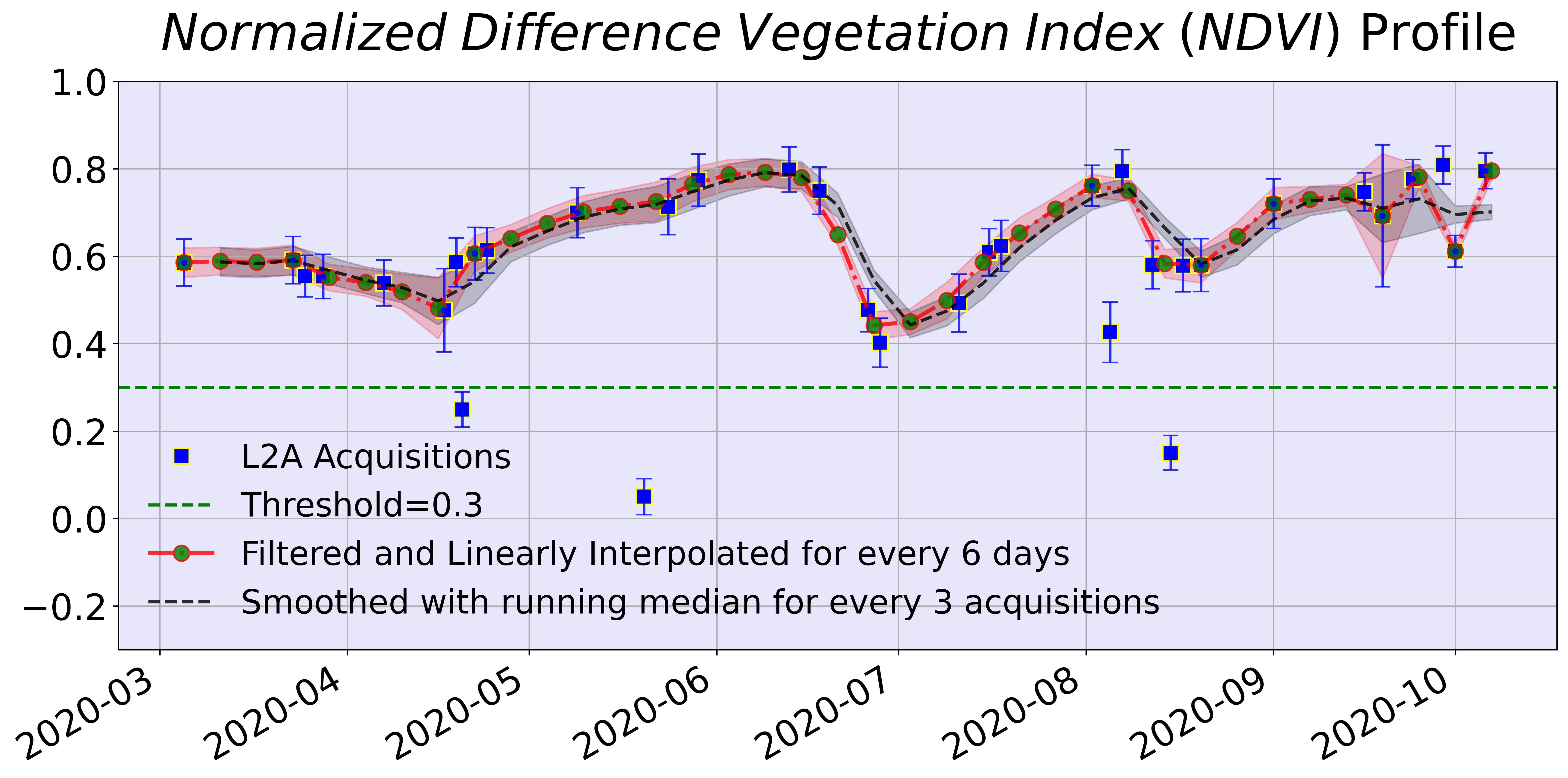}
\caption{An example of SITS preparation functionalities (filtering, interpolation, resampling and smoothing) applied to all pixels of a sampled parcel. }.
\label{fig:ts_preprocessing}
\end{figure}

\noindent Our ADC solution offers a number of SITS processing techniques to filter, interpolate, resample and smooth the pixel or parcel time-series. For instance, by applying specific thresholds we can filter out outliers, as seen in Fig. \ref{fig:ts_preprocessing}, and fill the missing values from clouds and shadows using a number of off-the-shelf interpolation techniques that we offer (e.g., Linear, Bicubic etc.). Interpolated SITS can then be sampled at any desired temporal resolution. This is useful when constructing feature spaces for machine learning tasks that require homogeneous input of fixed elements. Additionally, our solution provides smoothing functionalities (e.g., rolling median) to eliminate the noise caused by temporal fluctuations on the SITS. Thus, patterns can be more clear and reveal trends throughout the year(s). By refining the SITS, one can enhance the performance of AI models that are fed with these data, as well as improve photo-interpretation tasks. For example, crop classification tasks perform better using interpolated time-series and grassland mowing event detection is significantly enhanced through filtering and smoothing that removes outliers and abrupt changes that could be mistaken for real events. 

\subsubsection{Feature space generation}\label{sec:fs_generation}
% \smallskip
\noindent Using the ADC, one can go beyond single-date image features and combine spectral bands and vegetation indices that come from multiple temporal instances and thereby extract phenological features via computing the integrals and derivatives of the time curves (i.e., biomass indicator, yield indicator, end-of-season, peak-of-season, start-of-season) \cite{sitokonstantinou2020sentinel}. Such features have been shown to enhance the performance of crop classification AI tasks \cite{phenology_ids_crop_classification_Htitiou_1}. The capacity of the ADC to navigate in the spatial, temporal and spectral/variable dimensions, using the grid format of multidimensional xarrays, enables the effortless generation of SITS feature spaces. One can select to generate a feature space at the pixel or parcel level to match the downstream application requirements for spatial resolution. Additionally, one can segment the satellite images to chunks and generate patch-based datasets to feed deep learning models for computer vision tasks.

\subsubsection{Smart multidimensional queries}
% \smallskip

\noindent Time-series of EO hold important information on the evolution of crops and should be analyzed to extract knowledge on the vegetation development and to identify potential trends. By using the ADC, one can easily perform historical statistical analysis over an area at different time units, i.e., day, month, season or year. These statistics can be provided in the form of aggregated values for a parcel or any user-defined area. Apart from the coordinates of the area, a user can decide on additional parameters, such as the maximum cloud coverage percentage or the minimum number of cultivated crop fields over this area etc. \\

\begin{figure}[!ht]
\centering
\includegraphics[scale=0.4]{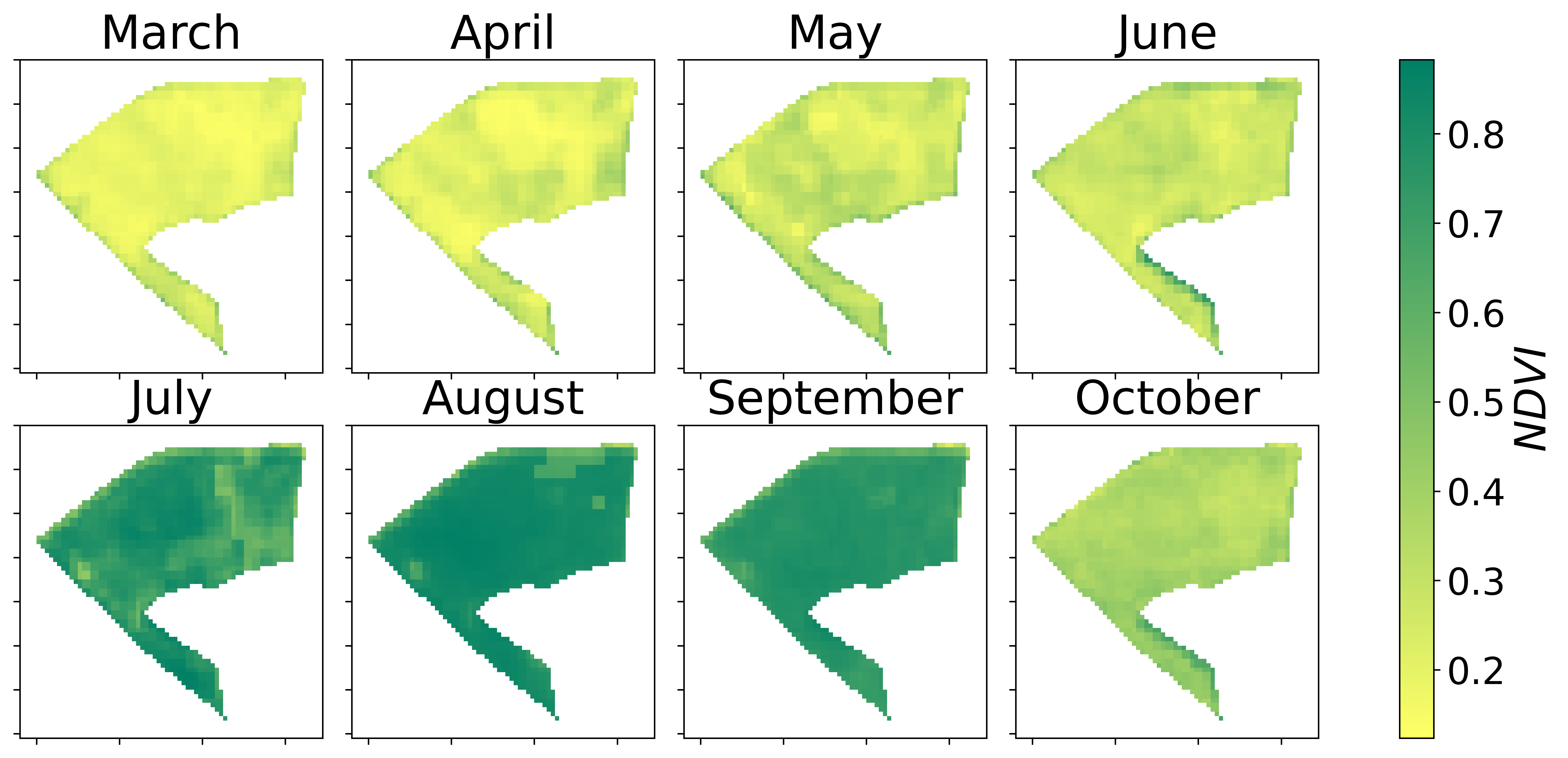}
\caption{An example of NDVI time-series animation for a maize field with monthly averages.}
\label{fig:analytic1}
\end{figure}

\noindent Using the ADC, we can generate animations of the evolution of a selected Sentinel variable (e.g., NDVI). Fig. \ref{fig:analytic1} shows a useful example on how temporal statistics can be used for validating if a CAP obligation is met or not. Specifically, the illustrated field was declared as \textit{spring triticale}, which is a particular type of \textit{spring cereal}, and it was predicted, by our AI crop classifier \cite{semantics} to be \textit{maize}, which is a summer cultivation. The animation in Fig. \ref{fig:analytic1} can be used to verify that indeed the prediction is correct and the declaration is erroneous, since the field has substantial vegetation during the summer months and spring triticale would have been harvested. This functionality is particularly useful for paying agency inspectors of the CAP that are not EO experts. This way, they can easily fetch a parcel-focused Sentinel time-series for the field of inspection and decide on the validity of the farmer's declaration.

\noindent Similarly to the temporal statistics described above, one can use ADC to compute pertinent statistics along the space and spectral/variable dimensions. All these statistics can be used in a synergistic manner to formulate complex geospatial queries. This is a game changer for the inspectors of the CAP paying agencies that are for the first time enabled to ask combinatory and multidimensional questions to make their inspection process faster, targeted and precise. As already mentioned, the ADC supports several services (e.g., crop classification, grassland mowing detection, SITS multidimensional statistics etc.) by providing gridded ARD. The results of these services are used to update a PostgreSQL/PostGIS database. The database contains aggregated results per parcel, which can then be easily accessed, enabling a back-and-forth communication with the ADC. \\

\noindent Thus, we have data cubes that include and keep on being dynamically populated by Sentinel-1 and Sentinel-2 products.  We also have auxiliary geospatial data (e.g., LPIS) that are used to enable the provision high level data products (e.g., crop classification) that in turn populate the cubes. This way, we end up with country specific knowledge bases for CAP monitoring. Meanwhile, useful operations such as the computation of distance between two geometries, the calculation of an area, buffer analysis and geospatial queries, can take place exploiting the power of the PostGIS extension. PostGIS and ADC queries can be combined to address the most sophisticated of questions inspectors might have. Below are listed examples of smart queries supported by our ADC-based framework.
\\

\noindent \textbf{Query 1.} \textit{Generate feature space}. Return monthly averages of Sentinel-1 coherence and Sentinel-2 NDVI at the parcel-level for Lithuania from 2017 to 2021 and Paphos, Cyprus for 2020. Apply inward buffer 5 m to avoid mixed pixels and outward buffer 50 m to reduce noise from clouds. Feed data to grassland mowing detection and crop classification algorithms.\\

\noindent \textbf{Query 2.} \textit{Select potential wrong declarations}. Using the output of query 1, return the fields in Paphos in 2020 that were declared to cultivate maize but have been predicted in a different crop type class. \textit{Visual verification.} Return animation of NDVI time-series, with a 10-day step, from June to October.\\

\noindent \textbf{Query 3.} \textit{Quantify grassland use intensity}. Using the output of query 1, return the number of mowing events in Lithunia per year from 2017 to 2021. Limit the results to grassland fields with average NDVI of less than $0.4$. Identify hotspots of  low grassland intensity with an average mowing event of less than 1 per year, over the years of inspection. Ultimately, this query will enable the decision makers to suggest spatially tailored mitigation or adaptation measures for the hotspots.

%%%%%%%%%%%%%%%%%%%%%%%%%%%%%%%%%%%%%%%%%%%%%%%%%%%%%%%%%%%%%%%%%%%%%%%%%%%%%%%%%%%%%%%%%%%%%%%%%%%%%%%%%%%%%%%%%%%%%%%%%%%%%%%%%%%%%%%%%%%%%%%%%%%%%%%%%%%%%%%%%%%%%%%%%%%%%%

\subsection{Conclusions}
% \medskip
\noindent Data cubes enable the transformation of EO data into i) analysis-ready information, ii) high-level knowledge and iii) intuitive visualizations to support timely and effective decision making. Our cloud-based approach allows for the efficient and automated discovery, pre-processing, data cube indexing and analysis of big satellite data. Currently, ADC is populated with Sentinel-1 and Sentinel-2 images that cover wall-to-wall Cyprus and Lithuania for three years. It also includes the parcel boundaries, crop type maps (LPIS) and ancillary data that enable the development of downstream applications for the monitoring of CAP rules. We indicatively used the outputs of an in-house crop classification model and Sen4CAP's$^1$\footnote{\url{https://github.com/Sen4CAP}}
grassland mowing detection model. Furthermore, a suite of tools has been built on top of the ADC. The users of our framework can straightforwardly generate spatial buffers, multidimensional statistics, animations, time-series plots and feature spaces, and execute complex multidimensional geospatial queries. We address the challenge of CAP controls by bringing together EO products, geospatial services and extracted knowledge from validated models. This solution is a stepping stone towards the modernization of the CAP and the seamless integration of big EO data in the operating models of non-expert users. 
\section{Crop Type Mapping} \label{croptype}

% Uncomment this line, when you have siunitx package loaded.
%The SI Units for dynamic viscosity is \si{\newton\second\per\metre\squared}.
% I'm going to randomly include a picture Figure~\ref{fig:minion}.

% If you have trouble viewing this document contact Krishna at: \href{mailto:kks32@cam.ac.uk}{kks32@cam.ac.uk} or raise an issue at \url{https://github.com/kks32/phd-thesis-template/}

% \begin{figure}[htbp!] 
% \centering    
% \includegraphics[width=1.0\textwidth]{minion}
% \caption[Minion]{This is just a long figure caption for the minion in Despicable Me from Pixar}
% \label{fig:minion}
% \end{figure}

\subsection{Literature review}

\noindent The administration and control of subsidy payments, associated with the relevant agrarian policies of the CAP require systematic and timely knowledge of the agricultural land cover and land use \citep{alganci2013parcel}. Paying agencies are the member state authorities responsible for the fair allocation of the CAP funds, and consequently for the realization of inspections to decide on the farmers’ compliance to their CAP obligations. The farmers submit their subsidy applications to their paying agency, declaring the location and the cultivated crop type for all owned parcels \citep{ee2018regulation}. In turn, paying agencies are required to inspect at least 5\% of declarations, via means of field visits and photointerpretation of Very High Resolution (VHR) and High Resolution (HR) satellite imagery (i.e., SPOT, Worldview, IKONOS) \citep{ee2018regulation,eyraud2011community}. However, these methods are time-consuming, complex, and reliant on the skills of the inspector. \\

\noindent Automated remote sensing methods, suitably designed with respect to computational efficiency, large-scale coverage, and spatial resolution, can offer a monitoring alternative with reliable solutions \citep{schmedtmann2015reliable}. In the context of the expected CAP 2020+ reform, there is a clear intention for simplification and a reduction of the administrative burden for both farmers and paying agencies. This study showcases how timely and accurate crop identification can assist the paying agency controls, suggesting an effective and efficient tool towards the required shift from sample inspections to large scale monitoring.\\

\noindent The freely available data and global coverage of Landsat satellites have established them, over the years, as one of the primary data sources for operational agriculture monitoring applications. Landsat data have been exploited in crop type mapping from as early as the 1980s, with Odenweller and Johnson \citep{odenweller1984crop} using VI time-series under threshold-based approaches, while in more recent studies their usage has evolved into non-parametric supervised classification schemes \citep{schmedtmann2015reliable,odenweller1984crop}. However, low (e.g., MODIS 250 m) and moderate (e.g., Landsat 30 m) spatial resolution data, with pixel sizes comparable to the parcel area, provide suboptimal thematic accuracy \citep{lebourgeois2017combined}. VHR (e.g., QuickBird and IKONOS) and HR (e.g., SPOT 5) satellite imagery has also been extensively used in crop classification studies \citep{yang2007using,turker2011field}. VHR offers an excellent target to pixel ratio and allows for the appropriate exploitation of image processing features, such as texture, context, and tone \citep{pena2011object}. Single-source missions that provide VHR imagery are often associated with high cost and long revisit times. However, the recently introduced small-satellite constellations, such as Planet, can offer global coverage at high revisit frequency and unprecedentedly low costs. Nonetheless, the inexpensive, off-the-shelf sensor components used imply suboptimal noise resilience, radiometric performance, and cross-sensor uniformity \citep{houborg2016high}.\\

\noindent Time-series of multispectral imagery that covers the agricultural cycle, from seeding to harvest, is necessitated for accurate crop identification \citep{vyas2005multi}. Mature crops are more distinct and carry varying textures, water absorbency, and colors, having greater chances of being detected with remote sensing. With respect to image processing, literature clearly points to Object-Based Image Analysis (OBIA) methods, assuming VHR and HR imagery is in use. Alternative pixel-based analyses often lead to misclassifications due to the canopy’s spectral variability, bare soil background reflectance, and mixed pixels. Grouping pixels into spectrally consistent objects can alleviate such issues \citep{pena2011object}.\\

\noindent Non-parametric supervised classifiers such as SVM, decision tree ensembles and artificial neural networks have been successfully employed in crop identification schemes \citep{foody2004supervised,duro2012comparison}. In the last ten years, SVM and RF classifiers have become increasingly popular in land cover mapping publications, doubling their appearances since 2012 \citep{thanh2017comparison}. Their success is attributed to effectively describing the nonlinear relationships between crops’ spectral characteristics and their physical condition \citep{pena2014object}. Both classifiers are resilient to noise and over-fitting and are therefore able to effectively cope with unbalanced data \citep{thanh2017comparison}. SVMs can ably handle small training datasets \citep{mountrakis2011support}, while ensemble classifiers like RF are characterized by high computational efficiency \citep{breiman2001random}. These constitute ideal classification features for accurate and efficient crop type mapping and specifically for the purposes of the CAP monitoring, where data are big and ground truth information is scarce.\\

\noindent The Sentinel-2 mission introduces a paradigm shift in the quality and quantity of open access EO data, opening a new era for operational terrestrial monitoring systems, a fortiori in the agriculture sector. The mission offers unprecedented 10 m and 20 m spatial resolution data at a 5-day revisit time with twin satellites. This, however, introduces the notion of big data and the challenge of their efficient handling. EO data are freely and systematically received over very large areas, which range beyond the limits of one region or a country, as they are nowadays required by the on-going and planned large scale initiatives and funded projects (e.g., GEOGLAM, GEO-CRADLE). In this regard, significant attention has been raised to Sentinels’ potential for agriculture monitoring and particularly in the context of the CAP control scheme \citep{valero2016production,matton2015automated,inglada2016improved,immitzer2016first}. One indicative example would be the European Space Agency’s (ESA) Sentinel-2 Agriculture project, aiming at the simplification of crop management. The system achieves an overall mapping accuracy of 85\% for five major crop types, making up 75\% of the regional agricultural zone \citep{inglada2015assessment}. Lebourgeois et al. \citep{lebourgeois2017combined} have been among the first to use a scheme of mixed OBIA and RF approach for smallholding crop type classification, achieving 64.4\% accuracy for 14 crop classes. In both publications, the authors tested their Sentinel specific methods, by simulating Sentinel-2 data using SPOT 5 and SPOT 4 (Take 5 experiment) as proxies. Lebourgeois et al. (2017) additionally incorporated Landsat-8 and VHR imagery, with the latter proving inconsequential in the classification process \citep{lebourgeois2017combined}. \\

\noindent This study attempts to capitalize on the most promising research paradigms discussed above, with a clear operationalization potential. In this context, we seek to suggest, implement, and validate a crop mapping scheme of methods, enabling the identification of nine different crop types and thus provide information on farmer’s compliance, with respect to their CAP requirements. A parcel-based Sentinel-2 MSI time-series approach is used for the construction of the feature space. The utilized imagery captures the growing stages of the various crop types in order to successfully discriminate among them. The local LPIS, which provides the geospatial input for crop delineation and the farmers’ declarations, as part of their CAP subsidy applications, are employed for the object partitioning of the images and the supervised classifiers’ training, respectively.\\

\noindent Thematic maps are produced for three different levels of crop nomenclature (crop type, crop family, and season of cultivation), employing separately a quadratic kernel SVM and a RF classifier. SVM and RF are two of the most popular non parametric classifiers in land cover mapping and their comparison has been an increasingly interesting topic in recent publications \citep{thanh2017comparison}. However, there is limited literature on their performance comparison under a Sentinel-2 based scheme, and even more under computational efficiency considerations that the Sentinel-induced big data shift demands. In this thesis, differences between the two classifiers are evaluated in terms of execution time, classification accuracy, and relevance to the crop identification issue. Further, the size of the variable space and the importance of individual variables are quantified and accordingly evaluated. The proposed methodology is finally applied to Landsat-8 OLI and down-sampled Sentinel-2 MSI equivalent variable spaces, for the comparison of spatial, temporal, and spectral characteristics between the two sensors. We also performed a compliance check analysis for the CAP Crop Diversification requirement to showcase the scheme’s capacity for effective decision making within the context of the control of CAP subsidies.

\subsection{Data and Study Area}

\begin{figure}[htbp!] 
\centering    
\includegraphics[width=1.0\textwidth]{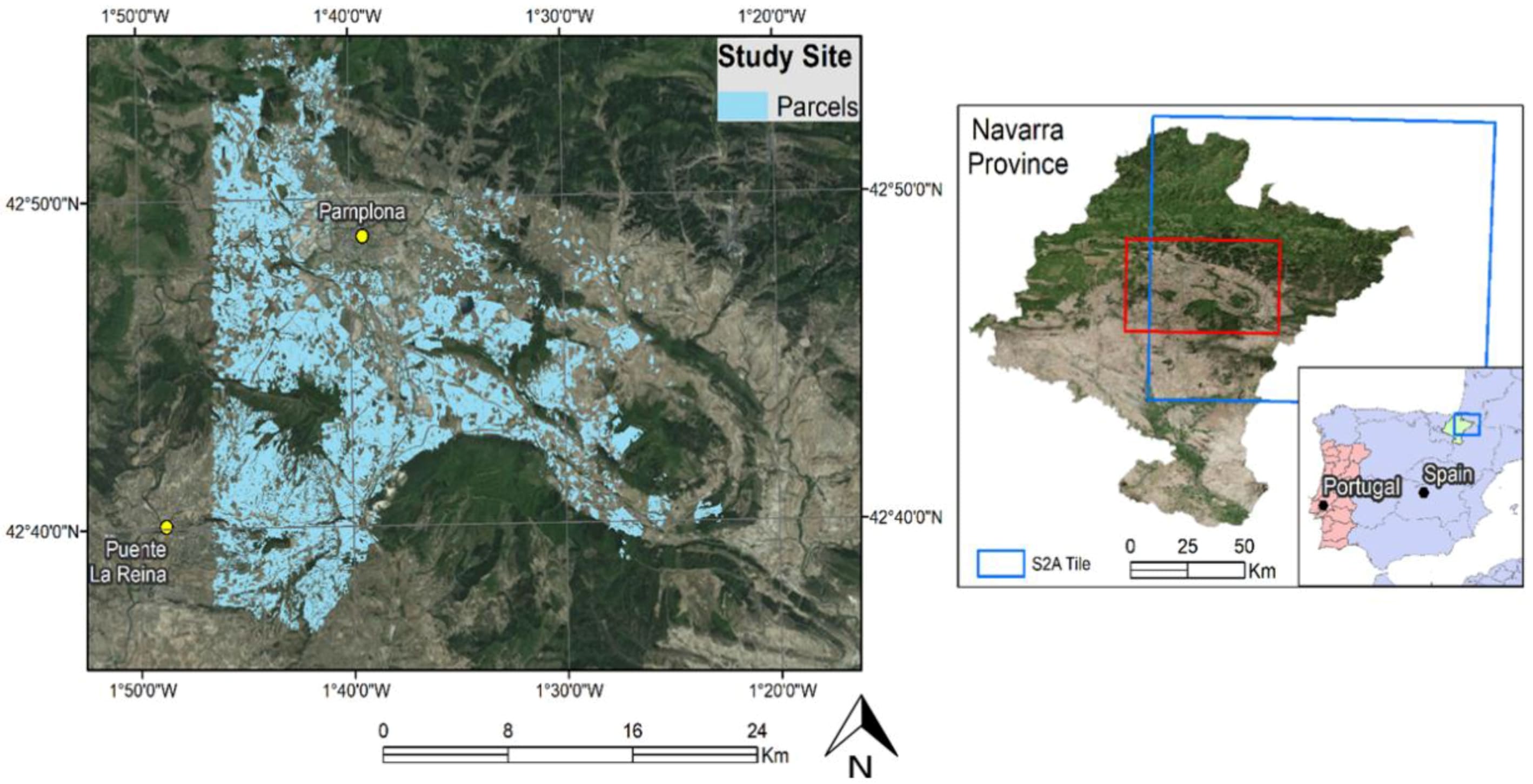}
\caption{Study area located in the Spanish province of Navarra and the corresponding Sentinel-2A tile (S2A tile) covering it. The site refers to the agricultural zone surrounding the city of Pamplona.}
\label{fig:navarra}
\end{figure}

\begin{figure}[htbp!] 
\centering    
\includegraphics[width=1.0\textwidth]{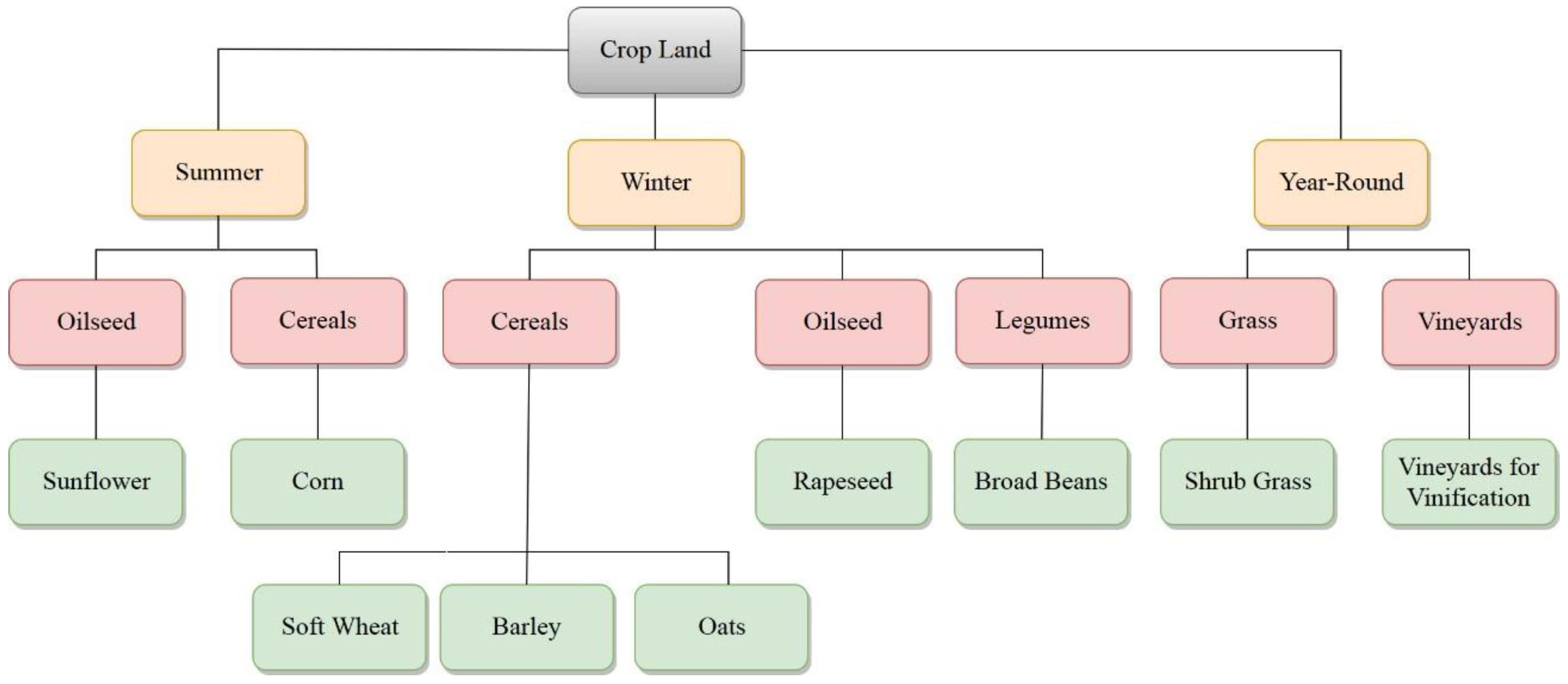}
\caption{Crop type nomenclature adopted for Navarra, covering the nine main crop types that amount to almost 90\% of parcels of the local agricultural zone. Crops are separated based on season, family, and type of cultivation.}
\label{fig:taxonomy}
\end{figure}

\noindent The study area is located in the province of Navarra in northeastern Spain, as shown in Figure \ref{fig:navarra}. It comprises of 9052 crop parcels and occupies approximately 215 km \textsuperscript{2}. The Pyrenees Mountains extend over the northern half of the province, as they stretch southward from France. Navarra’s landscape is characterized by a mixed relief of watered valleys and forested mountains. Its agricultural yield is comparatively low compared to the rest of the country due to farm fragmentation; however, the fertilizer usage in the community is far greater than the national average \citep{rodriguez2012assessment}. In this study, crop type maps are produced for the nine main crop types found in the available dataset, and this entails soft wheat (9217 ha), barley (4835 ha), oats (1554 ha), corn (255 ha), sunflower (602 ha), vineyards (250 ha), broad beans (826 ha), and rapeseed (1000 ha). These refer to the lowest level of the nomenclature hierarchy, as shown in Figure \ref{fig:taxonomy} , and explain nearly 90\% of the total number of parcels (9052/10,274 parcels) that is found in the regional agricultural zone. \\

% \begin{figure}[htbp!] 
% \centering    
% \includegraphics[width=1.0\textwidth]{Figs/Chapter2/scalable_fig2_taxonomy.png}
% \caption{Crop type nomenclature adopted for Navarra, covering the nine main crop types that amount to almost 90 of parcels of the local agricultural zone. Crops are separated based on season, family, and type of cultivation.}
% \label{fig:taxonomy}
% \end{figure}

\noindent The average parcel area in the study site is approximately 2 ha, which according to Lowder et al. (2016) belongs to the smallholding category \citep{lowder2016number}. This can be interpreted as 200 Sentinel-2 pixels and less than 25 Landsat-8 pixels per parcel. The small average parcel size forms a challenging test site to showcase the capacity of methods and proves ideal for comparisons among sensors of different spatial resolution.\\

\noindent The LPIS data used for the object partitioning of the time-series, along with the declared cultivated crop types, were made available by INTIA. INTIA is a public company, attached to the Department of Rural Development, Environment and Local Administration of Spain, which offers advanced services in the agri-food sector in Navarra. The provided dataset includes farmers’ crop type declarations for 2016, family- and season-based grouping of the crop types, and the parcels’ geospatial information in vector format. This unique dataset is provided by one of the direct end-users to the proposed scheme’s application, which adds to the transparency of the process. INTIA additionally offered a timeline of the phenology stages for key Navarra crop types, which was instrumental in the appropriate selection of the image acquisition dates, but also for the logical assessment and interpretation of the results.

\begin{figure}[htbp!] 
\centering    
\includegraphics[width=1.0\textwidth]{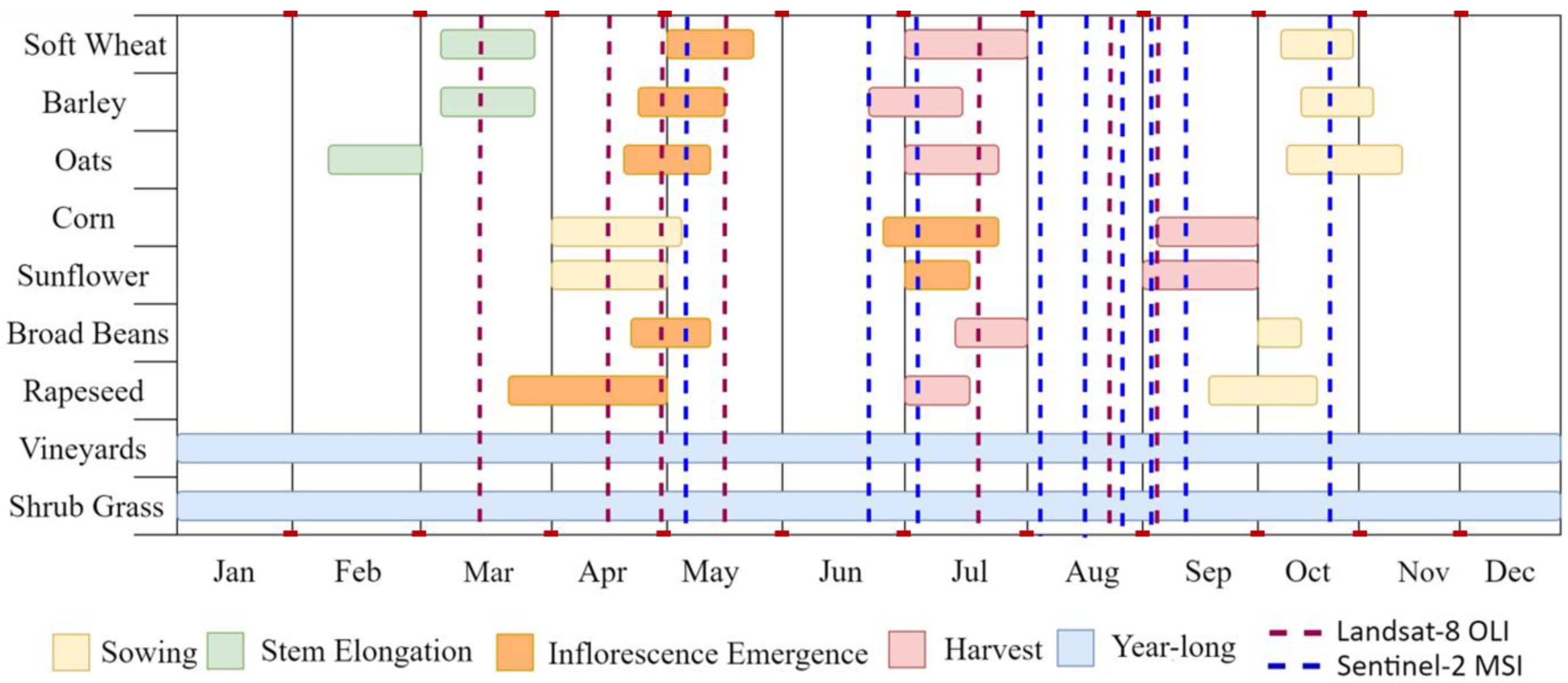}
\caption{Timeline of phenology stages for key crops in Navarra, overlaid with the acquisition dates of Sentinel-2 images for 2016.}
\label{fig:phenology_stages_navarra}
\end{figure}

\noindent An analysis was performed (\ref{appendix1}) on the impact and feasibility of utilizing a dense time-series of Sentinel-2 imagery, regardless of the cloud coverage. The analysis showed that the usage of solely cloud-free imagery (less than 3\% cloud coverage over the study site) that covers critical phenological stages was responsible for a near maximum crop type classification accuracy. On the other hand, the additional incorporation of cloudy imagery offered only a marginal increase in accuracy for a disproportionally larger cost in processing time. In this study, it was decided to use only cloud-free and nearly cloud-free imagery (<20\% cloud coverage), which guaranteed an even temporal distribution of acquisitions and at the same time filtered out potentially redundant information. This way the data management and computational efficiency considerations, integral to the design of the proposed methodology, are appropriately showcased. This further supports the ultimate objective of the present study to be viewed in the context of an operational monitoring system for the control of farmers’ compliance to their CAP obligations.
Sentinel-2 MSI’s 10 m and 20 m resolution bands (B02–B08A \& B11–B12) and Landsat-8 OLI’s 30 m resolution bands (B02–B08), in the visible, NIR, and SWIR parts of the spectrum, formed the Sentinel-based and Landsat-based feature spaces, respectively. Figure \ref{fig:phenology_stages_navarra}  depicts the nine selected cloud-free Sentinel-2A acquisitions (May to October) and the eight equivalent Landsat-8 scenes (March to September) for 2016.

\subsection{Methods}

\noindent Several machine learning algorithms of the supervised classification families of decisions trees, discriminant analysis, SVM, nearest neighbors, and tree ensembles were tested in a preliminary analysis. The analysis showed that the quadratic kernel SVM was the most accurate classifier, while RF came second best but with considerably improved computational efficiency. Hence, the implementation and comparison of these two classifiers was considered of great interest.\\

\begin{figure}[htbp!] 
\centering    
\includegraphics[width=1.0\textwidth]{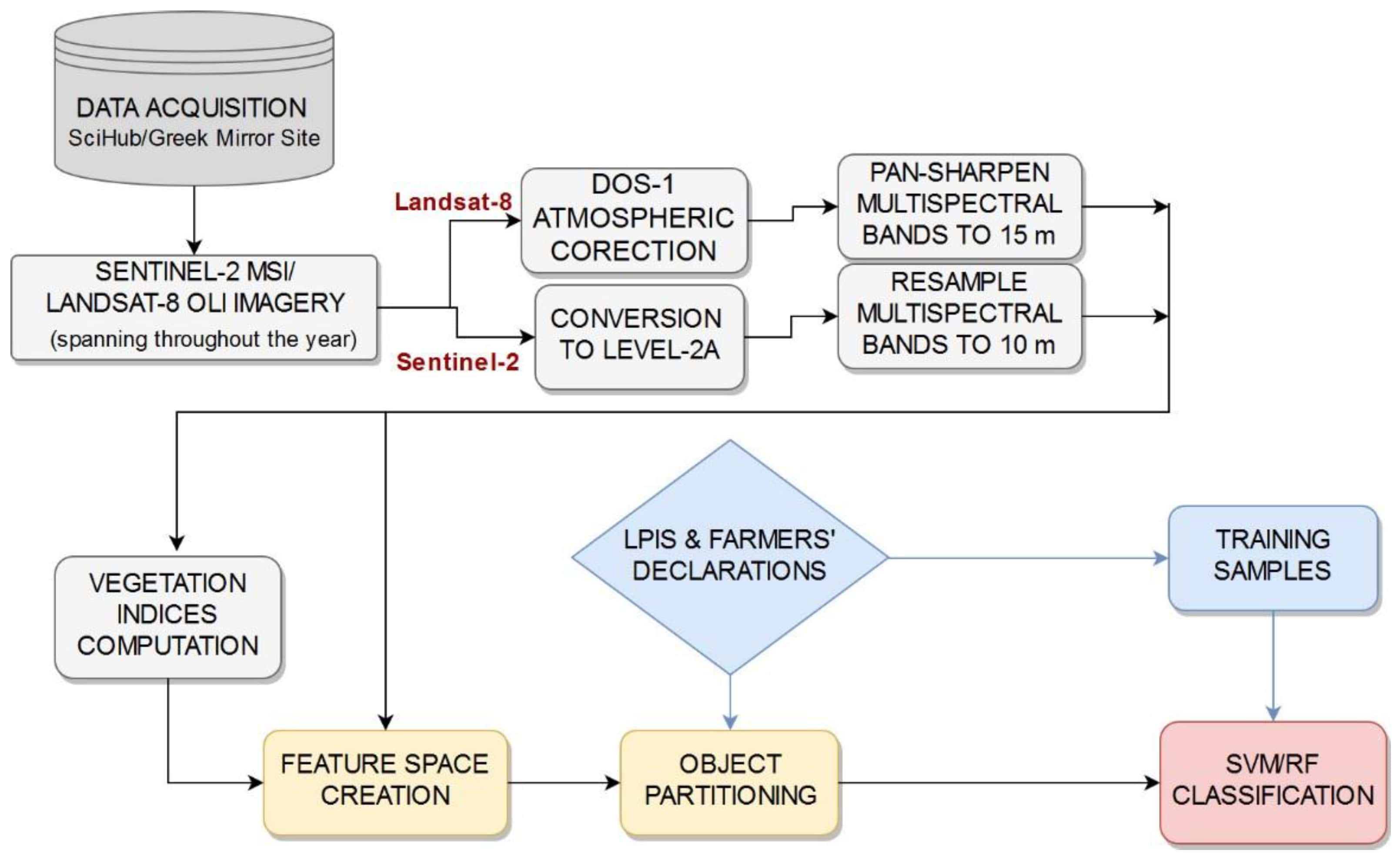}
\caption{Flowchart of overall methodology: imagery acquisition, data pre-processing, feature space creation, crop type classification.}
\label{fig:methodology}
\end{figure}

\noindent Crop type maps were produced via applying separately the SVM and RF classifiers to the Sentinel-2A MSI and Landsat-8 OLI imagery, under a parcel-based approach. An overview of the processing chain is shown in Figure \ref{fig:methodology}, with a brief description of the steps in the following subsections.

\subsubsection{Pre-Processing}

\noindent Sentinel-2A tiles were downloaded at Level 1C, which refers to 100 × 100 km2 ortho-images projected to cartographic geometry based on a Digital Elevation Model (DEM). Level 1C products are then transformed to Level-2A Bottom of Atmosphere (BOA) reflectances, using the Sen2Cor tool. Sentinel-2 bands of 20 m spatial resolution are resampled to 10 m spatial resolution, using the Nearest Neighbor (NN) resampling algorithm. Landsat-8 OLI imagery was acquired in GeoTiff format at Level 1TP, having been radiometrically calibrated and ortho-rectified. Image pixels have then been converted from digital numbers (DN) to radiances, to Top of Atmosphere (TOA) reflectances, using the Semi-Automatic Classification plugin in the geographic information system application QGIS \citep{congedo2016semi}. DOS-1 atmospheric correction has been additionally applied, converting data to surface reflectances.
Pan-sharpening was then performed by merging the moderate resolution (30 m) multispectral data with the higher resolution panchrormatic (15 m), providing multispectral imagery of higher resolution features. The employed technique implements a Brovey Transform, where the pan-sharpened values of the multispectral bands are calculated as in Equation (1) \citep{congedo2016semi}. 

\begin{equation}
  MSpan = MS \times \frac{PAN}{I}
\end{equation}

\noindent where I is the intensity, while MS is the multispectral, and PAN is the respective panchromatic pixel values. Intensity weights are a function of the multispectral data and are defined as in Equation (2) \citep{congedo2016semi}.

\begin{equation}
  I = \frac{0.42 \times Blue + 0.98 \times Green + 0.6 \times Red}{2}
\end{equation}

\subsubsection{Vegetation Indices}

\noindent VIs attempt to accentuate the vegetation signal, while diminishing soil background and solar irradiance contributions \citep{hatfield2010value}. NDVI and Normalized Difference Water Index (NDWI) have been widely used in crop monitoring and crop mapping applications \citep{lebourgeois,xiao2005mapping,huang2014analysis}. In this study, NDWI is used as defined by Gao (1996), utilizing the SWIR and NIR parts of the electromagnetic spectrum to correlate to plants’ water content \citep{gao1996ndwi}. SWIR is sensitive to dry matter content, leaf structure, and water content, while NIR only accounts for the first two; hence the water content property is set apart. VIs can be fully exploited in multi-temporal schemes, since crop types’ unique phenology calendars enable accurate crop discrimination. In this regard the Plant Senescence Reflectance Index (PSRI), which is defined as (Red-Green)/NIR, can form dissimilar spectral signatures for the different crop types, being particularly sensitive to their senescence phase \citep{hatfield2010value}.

\subsubsection{Image Partioning to Parcel Objects and Feature Space Creation}

\noindent Image segmentation that makes use of the spatial, temporal, and spectral features of satellite imagery is a complex and computationally demanding process, requiring fine-tuning that depends on the region and involved crop types \citep{inglada2015assessment}. Consequently, the adoption of image segmentation for partitioning the image time-series into parcels would not have supported the considerations of operationalization, transferability, and scalability, which the proposed methodology aspires to respect. The study assumes the viewpoint of a paying agency that is in charge of the management and control of CAP payments. The LPIS is one essential computerized database that paying agencies use within their operations and are mandated to frequently update. It should be noted that the reliability of the LPIS data is dependent on the update frequency, given that the evolution of the field limits can be significant over the years. However, in an attempt to exploit the benefits of an object-based approach, while at the same time preserving the scheme’s simplicity, it was decided that following a LPIS-based object partitioning approach is an acceptable trade off.\\

\noindent Table \ref{tab:moniroting_cap_table1} lists the variables that make up the Landsat- and Sentinel-based datasets. The Sentinel-2 feature space, initially, comprises of all multispectral bands, for each individual scene employed (90 variables). In turn, VIs (NDVI, NDWI, and PSRI) for all temporal instances (27 variables) are calculated at pixel level and thereafter incorporated into the feature space (117 variables). In the same manner the Landsat-8 feature space includes a total of 63 variables, including the pan-sharpened multispectral data (42 variables) and the equivalent VIs (21 variables). Then the time-series of imagery is partitioned into parcels via utilizing the geometry of the LPIS vector data. Specifically, the pixel values that fall within the boundaries of the LPIS polygons are averaged, giving a single parcel value for every variable.

\begin{table}[!ht]
\caption{Bands and Vegetation Indices of Sentinel-2 and Landsat-8}
\label{tab:moniroting_cap_table1}
\centering
\scalebox{0.85}{
\begin{tabular}{p{0.2\linewidth}|p{0.35\linewidth}|p{0.35\linewidth}}
 \hline
 \textbf{Type} & \textbf{Sentinel-2 MSI} & \textbf{Landsat-8 OLI} \\ 
 \hline
 Reflectances & Bands: B02, B03, B04 (VIS), B08, B8A (NIR) of 10 m spatial resolution. B05, B06, B07 (Red-Edge) and B11, B12 (SWIR) of 20 m spatial resolution  Mean values (per parcel) of 10 multispectral bands × 9 image acquisitions (90 variables) & Bands: B02, B03, B04 (VIS), B05 (NIR), B06, B07 (SWIR) of 30 m spatial resolution. B08 (panchromatic) of 15 m spatial resolution Mean values (per parcel) of 6 pan-sharpened multispectral bands × 7 image acquisitions (42 variables)
  \\
 Vegetation Indices	& Mean values (per parcel) of NDVI, PSRI, NDWI × 9 image acquisitions (27 variables) & Mean values (per parcel) of NDVI, PSRI, NDWI × 7 image acquisitions (21 variables)
  \\
 \hline
\end{tabular}}
\end{table}

\subsubsection{Supervised Classification}

\noindent We trained the classifiers based on a subset of the farmers’ declarations. Validated ground truth data were available but limited in number, as such information can only be acquired via field inspections. Therefore, we have assumed that the farmers’ declarations for 2016, as we received them, are valid in their majority. This is a fair assumption that is supported by INTIA inspectors, the workforce in charge of the field controls of the regional paying agency. In fact, cross-checking the validated data with the declarations showed that 99\% of the farmers stated truthfully the cultivated crop type, out of 464 validated parcels.\\

\noindent The dataset was separated into randomly selected training and test sets. Multiple such splits were used in order to showcase the sensitivity of the models to the training data. Accordingly, the classification estimations were evaluated against the respective test subsets of the farmers’ declarations. The number of training samples taken for each individual crop class was finalized at 20\% of the total crop parcels of that class. When experimenting with samples higher than 20\% for training, we encountered only a marginal increase in accuracy.\\

% \textbf{{Support Vector Machines}

\noindent\textbf{Support Vector Machines.} Binary SVMs set a hyperplane by exploiting the feature information of each entity in the training set. This hyperplane would be the decision boundary in the classification process. Optimally, it is defined to maximize the distance between itself and the nearest training entity of any class (functional margin), with larger margins relating to lower generalization errors [18]. In this case, multiple binary classifiers for all different 36 class pairs (1-to-1 mapping) are combined to construct a single multi-class classifier. \\

\noindent Since the classification problem in this study is not linearly solvable, a quadratic kernel is used to transform the original feature space onto higher dimensions where the crop classes become linearly separable. However, the data are not perfectly separable and the algorithm allows for some misclassification in the training set, which is applied by the box constraint parameter (C). Specifically, higher box constraint values suggest stricter data separation. The model was built using the MATLAB function fitcecoc, where the kernel scale parameter was set to ‘auto’ mode for optimization. The box constraint parameter was selected via hyperparameter optimization, based on the minimization of the 10-fold cross-validation loss, as shown in Table \ref{tab:moniroting_cap_table2}. Thirty different objective evaluations were performed, of which an indicative subset is presented. The final selection was made under both cross-validation loss minimization and processing time considerations.

\begin{table}[!ht]
\caption{Box constraint (C) parameter optimization using 10-fold cross-validation loss minimization.}
\label{tab:moniroting_cap_table2}
\centering
\scalebox{0.85}{
\begin{tabular}{|c|c|c|c|}
\hline
\textbf{\# Evaluation }& 
\textbf{Box Constrain} & \textbf{10-Fold Loss(\%) querying} & \textbf{Iteration Time (s)} \\ 
\hline
1	& 1.048 &	5.97 &	35.29  \\
2	& 736.8	& 8.54	& 286.29 \\
3	& 46.36	& 7.29	& 177.32  \\
4	& 220.22	& 8.32 &	689.06 \\
5	& 998.40 &	8.34 &	450.62 \\ 
6	& 0.0052 &	22.92 &	18.38  \\
7	& 0.17	& 8.82 &	17.86 \\
\hline
\end{tabular}}
\end{table}

\noindent It is evident that a box constraint of approximately 1, as in the 1st Evaluation, provides the optimal 10-fold cross-validation loss, while it also proves superior in terms of processing efficiency, as compared to the evaluations of comparable cross-validation loss (i.e., Evaluations 2–5).\\

\noindent\textbf{Random Forest}
 Ensemble classifiers, such as RF, increasingly gain popularity in EO applications using high spatial resolution imagery, as they can effectively manage large volumes of data \citep{lebourgeois2017combined}. RF starts with forming an ensemble of standard decision trees, also known as weak learners. In a simple decision tree, the input entity is entered at the top and as it moves towards the bottom it gets grouped into progressively smaller subsets. The labels are then assigned to parcels based on the maximum number of votes among the ensemble of weak learners \citep{breiman2001random,lawrence2006mapping}. The RF model was built using the MATLAB function fitensemble, under the “Bag” method. Bagging refers to the repeated selection of random samples from the entirety of the training set, on which the weak learners train separately. Even though the estimates of a single decision tree would be noisy, the mean estimates over multiple decision trees are both unbiased and resilient to over-fitting \citep{breiman2001random}. In the same manner as for the SVM classification, the minimization of 10-fold cross-validation loss is used for the optimal hypermeter tuning. Once again, thirty different objective evaluations were performed, of which an indicative subset is presented in Table \ref{tab:moniroting_cap_table3}.

\begin{table}[!ht]
\caption{Number of weak learners (trees) and maximum number of splits optimization based on 10-fold cross-validation loss minimization.
}
\label{tab:moniroting_cap_table3}
\centering
\scalebox{0.85}{
\begin{tabular}{|c|c|c|c|c|}
\hline
\textbf{\# Evaluation }& 
\textbf{\# of Weak Learners} & \textbf{Max \# of Splits} & \textbf{10-Fold Loss} & \textbf{Iteration Time (s)} \\ 
\hline
1	& 479 &	1182 &	10.69 &	206.31  \\
2	& 141 &	783 &	19.86 &	34.19  \\
3	& 30 &	8925 &	11.12 & 	14.70  \\
4	& 781 &	2 &	33.87 & 11.81 \\
5	& 11 &	3083 &	16.98 &	3.258  \\
\hline
\end{tabular}}
\end{table}

\noindent The hyperparameters “number of weak learners” and “maximum number of splits” were varied and evaluated against the minimization of the objective function. The third evaluation of 30 decision trees was selected as the optimal set of parameters. Nonetheless, it is observed that the first evaluation that uses 479 weak learners gives a lower 10-fold cross-validation loss, but this marginal increase in accuracy cannot excuse the additional processing effort required.

\subsubsection{Accuracy Assessment}
\noindent Classification performance was assessed according to producer’s accuracy (PA), user’s accuracy (UA), and Cohen’s kappa coefficient (Kc), computed as shown in Equations (3)–(5). PA is the ratio of correctly classified parcels over the total number of parcels for a ground truth class. Alternatively, UA is the ratio of correctly classified parcels for a given class to the total number of parcels predicted to belong to that class \citep{story1986accuracy}. Statistical metric kappa is used to describe the overall classification accuracy and it is generally preferred over simple accuracy metrics, as it accounts for random agreement between truth and estimation \citep{cohen1960coefficient}. Hence, Kc better showcases the capacity of both input data and classification techniques \citep{liu2007comparative}.

\begin{equation}
  PA = \frac{n_{ii}}{n_{irow}}
\end{equation}

\begin{equation}
  UA = \frac{n_{ii}}{n_{icol}}
\end{equation}

\begin{equation}
  Kc = N \sum_{i=0}^{r}n_{ii} - \sum_{i=0}^{r} \frac{n_{irow}n_{icol}}{{N^2}} - \sum_{i=0}^{r}n_{icol}n_{irow}
\end{equation}

\noindent where n\textsubscript{ii} is the number of parcels correctly classified in a particular crop type class; N is the total number of parcels in the confusion matrix; r is the number of rows;     n\textsubscript{icol} and n\textsubscript{irow} are the column (predicted class labels) and row (ground truth) total, respectively \citep{petropoulos2012land}.

\subsection{Results}
\subsubsection{Accuracy Performance}
\noindent PA and UA results presented in Table \ref{tab:moniroting_cap_table4} refer to the quadratic kernel SVM and RF classifications at the crop type nomenclature level. Since the dataset is strongly dominated by soft wheat parcels, the overall accuracy would be respectively affected. Therefore, PA and UA better depict the classifiers performance for all individual crop classes. The values in Table \ref{tab:moniroting_cap_table4} have been averaged over 20 iterations of random training sample splits. \\

\begin{table}[!ht]
\caption{Producer’s accuracy and user’s accuracy for both Random Forest (RF) and Support Vector Machine (SVM) classifications for the crop type nomenclature level. The standard deviation of the mean accuracy values is also shown in parentheses.
}
\label{tab:moniroting_cap_table4}
\centering
\scalebox{0.8}{
\begin{tabular}{|c|c|c|c|c|}
\hline
\textbf{Crop Type}& 
\textbf{RF PA (\%)} & \textbf{RF UA (\%)} & \textbf{SVM PA (\%)} & \textbf{SVM UA (\%)} \\ 
\hline 
Soft Wheat	& 94.89 (0.6) &	82.88 (0.9) &	95.60 (0.2) &	91.49 (1.0)  \\
Corn &	96.35 (2.3) &	93.29 (4.0) &	92.56 (3.2) &	96.48 (3.6)  \\
Barley &	81.81 (1.6)	& 87.78 (1.0) &	90.10 (0.9) &	90.43 (0.9)  \\
Oats &	44.38 (2.3) &	90.36 (2.4) &	78.57 (3.0) &	89.59 (1.9)  \\
Sunflower &	85.69 (3.5) &	92.03 (2.7)	& 85.36 (3.6)	& 95.45 (2.6)  \\
Rapeseed &	92.30 (1.7)	& 95.03 (1.4) &	90.19 (1.8)	& 96.12 (0.7) \\
Broad Beans &	85.41 (1.8) &	92.41 (1.5) &	86.83 (1.7) &	95.08 (1.1)  \\
Shrub Grass &	77.93 (4.3) &	76.90 (3.8) &	84.07 (3.8) &	83.48 (2.3)  \\
Vineyards & 86.15 (4.0) &	83.72 (3.7) &	87.52 (4.0) & 83.48 (3.5)  \\
\hline
\end{tabular}}
\end{table}

\begin{figure}[!ht]
\centering
\includegraphics{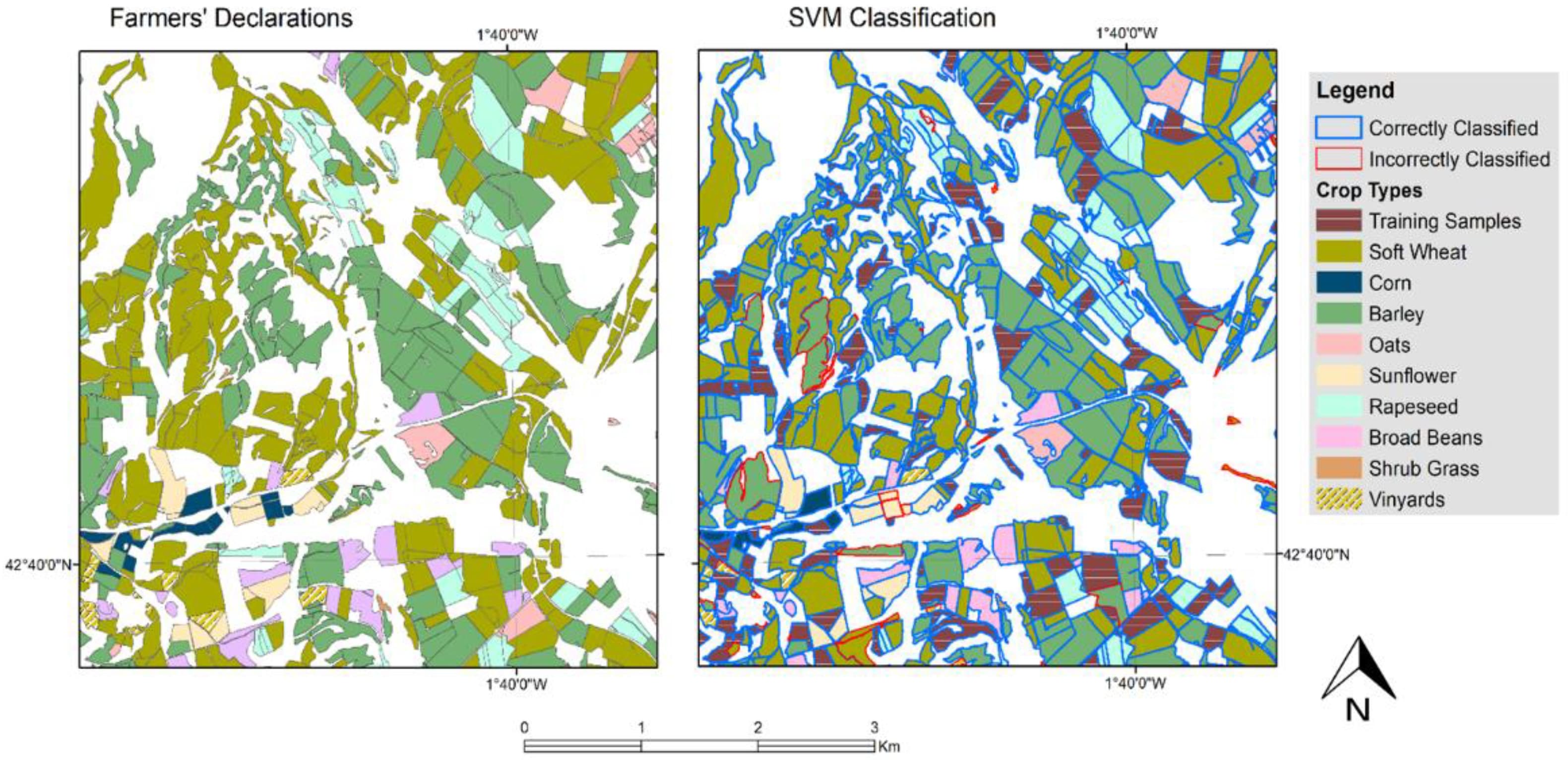}
\caption{Farmers’ declarations and quadratic SVM classification at the lowest nomenclature level (crop type). Only a clipped part of the study site is shown, with correctly and incorrectly classified parcels also identified.}
\label{fig:scalable_fig5}
\end{figure}

\noindent Both classifiers provide excellent results with an overall accuracy 85.59\% for RF and 91.30\% for SVM. Nonetheless, according to the PAs, SVM considerably outperforms RF for the barley and oats ground truth classes. These are two of the most indicative classes as their proneness to misclassification can be physically interpreted and thus better describe SVM’s superiority. Barley, oats, and soft wheat are crops of the cereal family with strong physical and spectral resemblance. Even more, their similar cultivation calendars (Figure \ref{fig:phenology_stages_navarra}) minimize the discriminating qualities of the multi-temporal approach. The same holds true for shrub grass and vineyards, which are year-round crops, and thus their spectral profiles would not bear any solid temporal particularities. Shrub grass is also vegetation of a diverse nature, providing generic spectral signatures, and hence misclassifications are evident for both classifiers, particularly for RF. Finally, UA values identify SVM as the more reliable classifier, with very high accuracies for all crop types apart from shrub grass and vineyards. A clip of the SVM crop type classification map is shown in Figure \ref{fig:scalable_fig5}, identifying both correctly and incorrectly crop type estimations.\\

\noindent Performance differences between the two classifiers become even more apparent using Kc, as RF classification is subject to substantial random agreement. Maximum kappa coefficients of 0.87 and 0.78 for SVM and RF, respectively, showcase more reliably the dominance of the first classifier over the latter (Table \ref{tab:moniroting_cap_table5}). The listed accuracy values have been averaged over 20 classification iterations of randomly split training sets. The standard deviation of the accuracies for each classification scenario is also given. This is done to showcase the very low sensitivity of the overall accuracy results to the training set, with standard deviations varying in the range of 0.3–1\%.

\begin{table}[!ht]
\caption{Maximum Kappa coefficient values for all nomenclature levels; values are averaged over 20 classification iterations, with the standard deviation of the mean accuracy values (st.d.) also listed.}
\label{tab:moniroting_cap_table5}
\centering
\scalebox{0.9}{
\begin{tabular}{|c|c|c|c|c|}
\hline
\textbf{Nomenclature Level} &	\textbf{Kappa RF}&	\textbf{st.d. RF} & \textbf{Kappa SVM} & \textbf{st.d. SVM} \\ 
\hline 
Type	& 0.7834 &	0.0070	& 0.8721 &	0.0054 \\
Family	& 0.8902 &	0.0030	& 0.9118 &	0.0053 \\
Season	& 0.8856 &	0.0092 &	0.8729 &	0.0115 \\ 
\hline
\end{tabular}}
\end{table}

\subsubsection{Feature Importance}
\begin{figure}[!ht]
\centering
\includegraphics{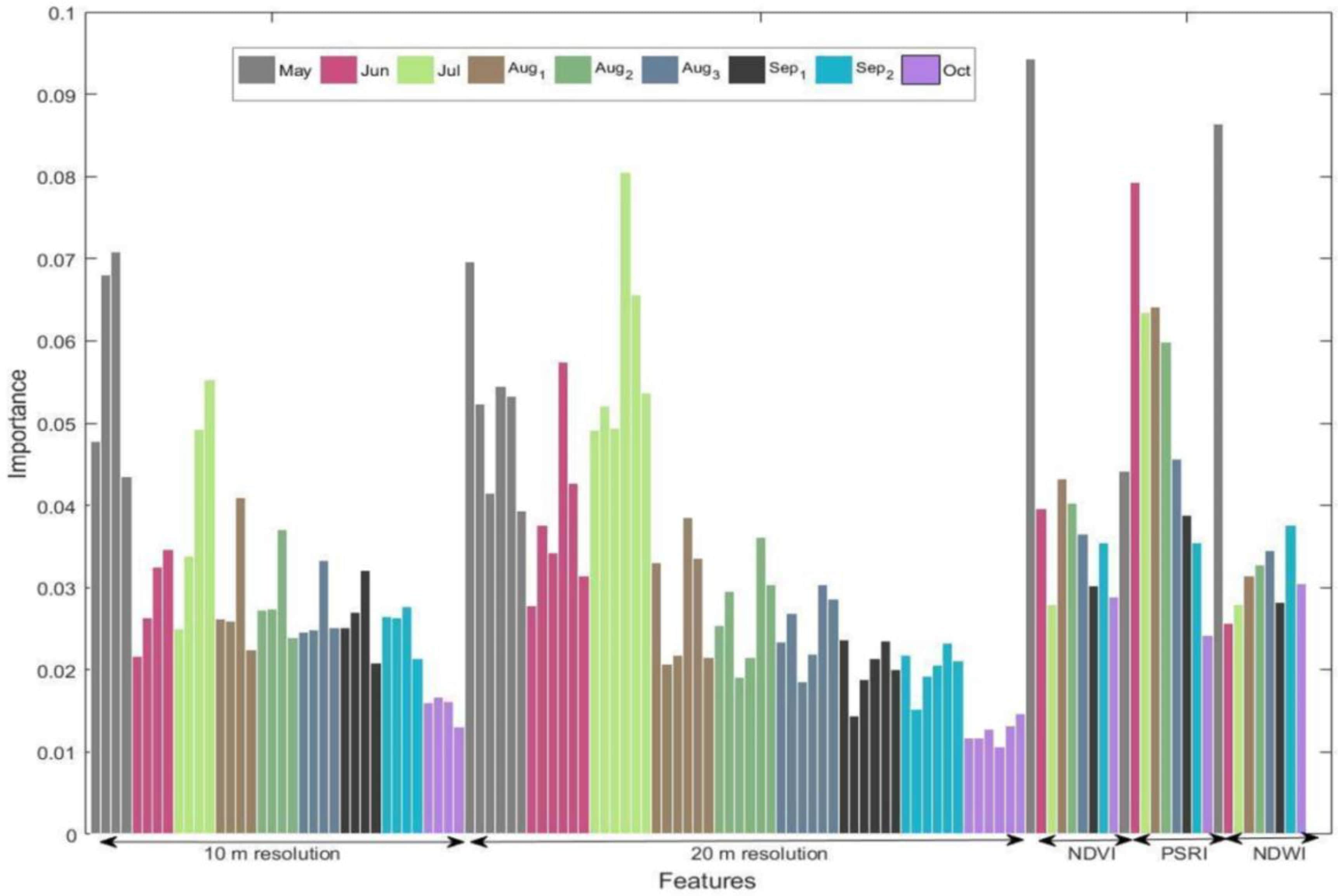}
\caption{Feature importance bar graph of Sentinel-2 features, for the crop type classification. Features are grouped with respect to their spectral characteristics by position, and to their acquisition dates by color. The order for the 10 m resolution bands is B02-B03-B04-B08 and for the 20 m resolution bands is B05-B06-B07-B08A-B11-B12.}
\label{fig:scalable_fig6}
\end{figure}

\noindent Quantifying the importance of individual features of a large variable space enables the better understanding of the dataset itself. This would in turn provide the means for the physical interpretation of results and the proper evaluation of the input data. Importance weights for the entirety of the Sentinel-2 variable space are shown in Figure \ref{fig:scalable_fig6}, with features being grouped into thematic divisions with respect to spectral characteristics and acquisition dates.

\noindent Feature importance, for the Sentinel-2 feature space, is calculated using the MATLAB function ReliefF. It is an extension of the original Relief algorithm suggested by Kira and Rendell (1992), which estimates the quality of attributes with respect to the distinguishability they offer among classification entities near each other \citep{kira1992practical}. Further, the ReliefF extension enables feature importance estimation even in multi-class scenarios, whilst being more resilient to incomplete and noisy data \citep{kononenko1994estimating}. The algorithm randomly selects entities (Ei) and thereafter searches for the k nearest neighbors coming from the same class, known as nearest hits (Hj), and the k nearest neighbors from different classes, known as nearest misses (Mj). Feature weights are estimated based on the Ei, Hj, and Mj values. The contribution of all hits and misses is averaged, with a contribution weight for each class of the misses being dependent on the prior probability of that class \citep{robnik2003theoretical}. The function was set to search for 80 nearest neighbors, as trialing with larger values did not produce better data modeling. \\

\begin{figure}[!ht]
\centering
\includegraphics{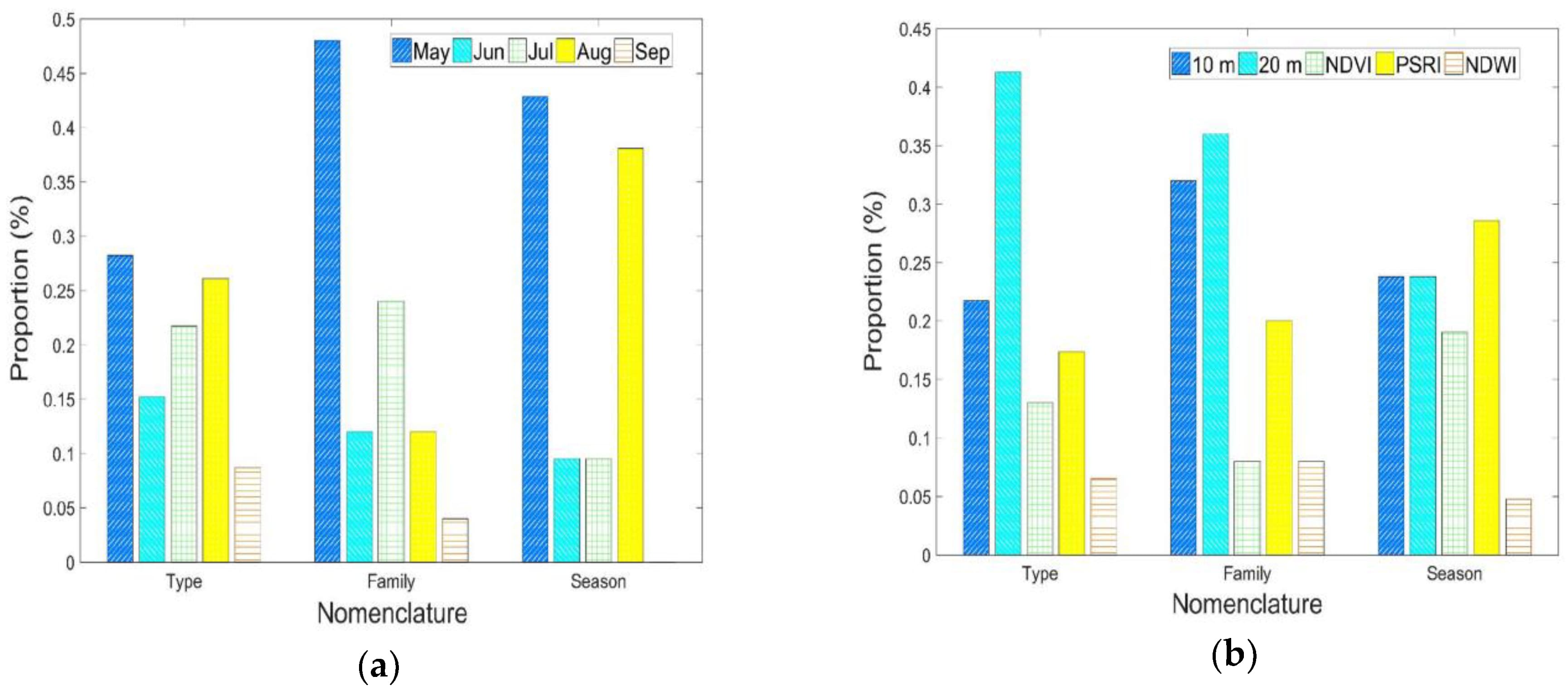}
\caption{Proportion of feature contribution bar graphs for the Sentinel-2 scenario, calculated based on the feature importance values. The two figures are plotted for two thematic categories (a) month of image acquisition and (b) spectral information.}
\label{fig:scalable_fig7}
\end{figure}

\noindent May and July appear to attain the highest weights of importance. May coincides with the inflorescence emergence period for the winter crops, while July matches the winter crop senescence and summer crop leaf development (refer to Figure \ref{fig:phenology_stages_navarra}). PSRI is the most consistent of the VIs, having high weights of importance for nearly all scenes. In Figure \ref{fig:scalable_fig7} the highest ranked features, based on the feature importance estimations, are grouped into two thematic categories that relate to (a) month of image acquisition and (b) their spectral characteristics. Figure \ref{fig:scalable_fig7}a quantifies the proportion of the top ranked features that stem from each month of image acquisition. In the same fashion, Figure \ref{fig:scalable_fig7}b quantifies the proportion of top ranked features coming from (a) the visible and NIR bands of 10 m spatial resolution, (b) the vegetation red-edge and SWIR bands of 20 m resolution, and (c) the three VIs. The number of total top ranked features, for which the proportion values are calculated against, differ for each level of the nomenclature hierarchy. This refers to the least number of top ranked features that offers the first near-maximum classification accuracy; and amounts to 46, 25, and 21 features for the “type”, “family” and “season” classifications, respectively.

\noindent For the crop type level in Figure \ref{fig:scalable_fig7}a, the most important features are distributed among May, June, July, and August. These months cover winter crops’ flowering to senescence stages and summer crops’ leaf development to flowering stages. Time-series of satellite imagery that covers the advanced phenological phases of both summer and winter crops is characterized by high discriminatory qualities. Furthermore, the image sensed during May proves of great significance in all nomenclature levels, explained by the dataset’s abundance in soft wheat and barley and thus in cereals and winter crops. In volume-wise skewed datasets, the dependence of variable importance in classes’ prior probability would affect the weights correspondingly. With respect to the spectral characteristics, it is the VIs and the bands of 20 m spatial resolution that contribute the most, despite the small volume of the first and the lower spatial resolution of the latter. More specifically, what the red-edge, narrow-NIR, and SWIR data lack in spatial resolution, they gain in spectral resolution (narrower spectral windows) and reflectance profile distinctiveness among the different crop types.\\

\noindent Inspecting the higher levels of nomenclature in Figure \ref{fig:scalable_fig7}b, crop season classification exposes considerable differences, with May and August being the main contributors in the discrimination process. Summer crops in May are germinating, while winter crops are already flowering or even ripening. On the other hand, summer crops in August are fully grown, whereas winter crops have already been harvested. Hence, the two months describe periods of maximal vegetation density difference between the two classes. This in turn explains the significant contribution of VIs in crop season classification.

\subsubsection{Cohen’s Kappa Evolution with Increasing Number of Features}
\begin{figure}[!ht]
\centering
\includegraphics[scale=0.85]{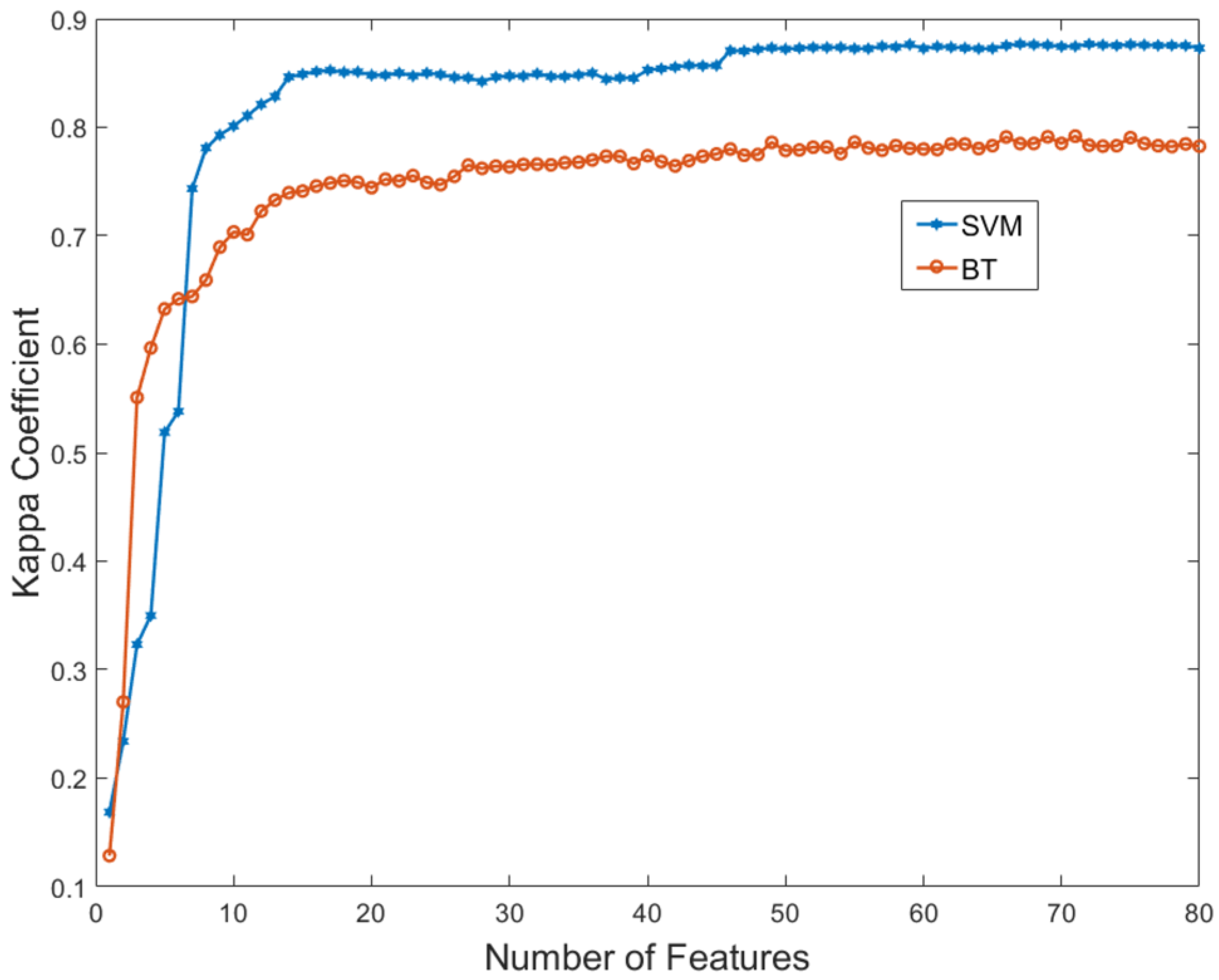}
\caption{Kappa coefficient evolution for an increasing number of features for both SVM and RF crop type classifications.}
\label{fig:scalable_fig8}
\end{figure}

\noindent Based on the feature importance computation (Figure \ref{fig:scalable_fig6}), the two classification algorithms are trained and tested for progressively larger feature spaces, as shown in Figure \ref{fig:scalable_fig8}. The first point on the x-axis refers to a feature space comprising solely of the most important feature. For the second point onwards, features keep on populating the feature space, one by one, according to their importance ranking. It is observed that the total number of variables is large, with certain variables being strongly correlated and others being redundant by default.

\begin{figure}[!ht]
\centering
\includegraphics[scale=0.85]{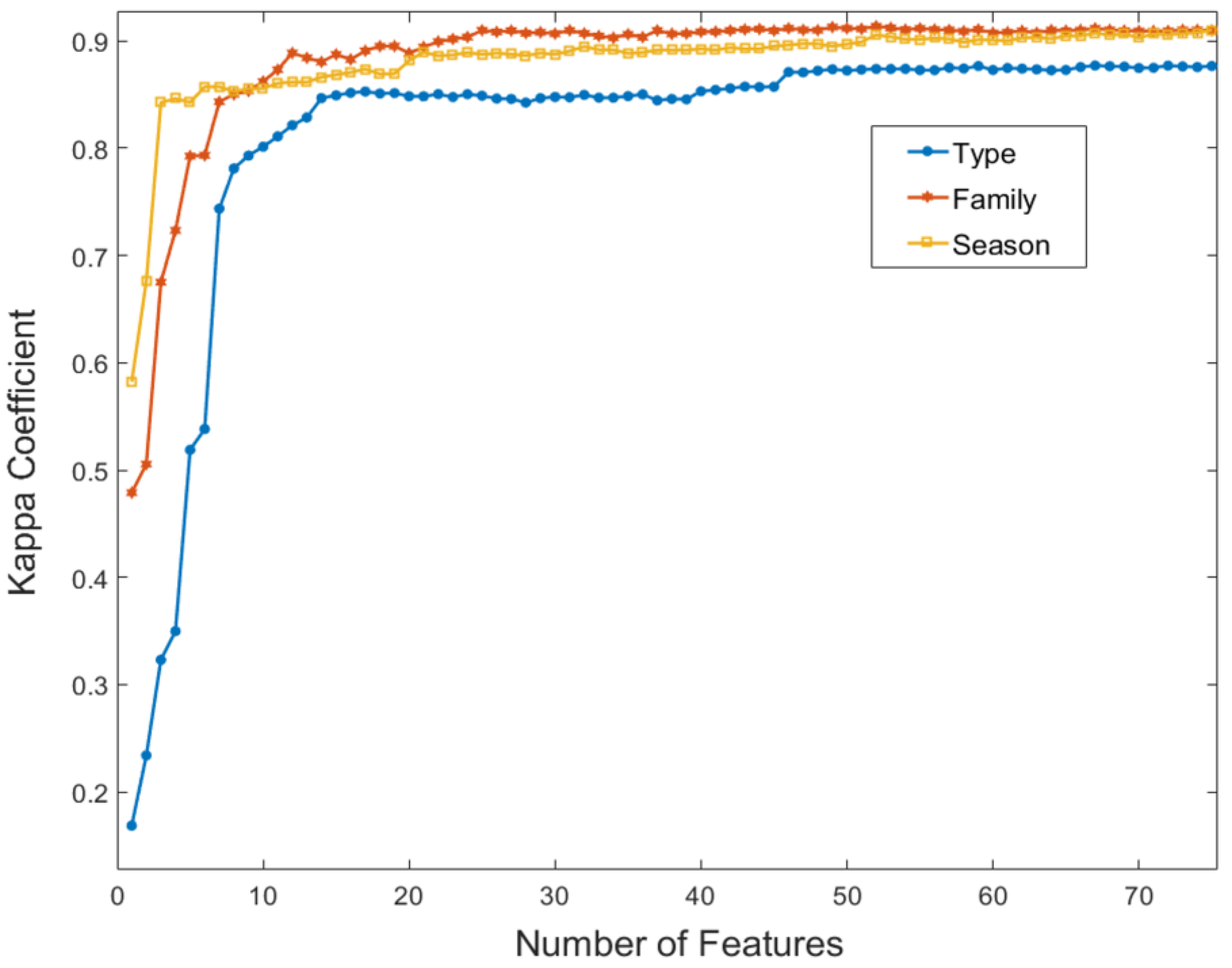}
\caption{Kappa coefficient evolution for all levels of nomenclature using the SVM classifier.}
\label{fig:scalable_fig9}
\end{figure}

\noindent In Figure \ref{fig:scalable_fig9} the SVM kappa coefficient evolution for all three levels of the nomenclature is depicted. All curves are almost monotonically increasing up until they reach a plateau, hence demonstrating the successful estimation of feature importance. Crop family and season of cultivation classifications reach their first optimal solution for spaces of fewer than 25 features, as indicated by the onset of the curves’ flat region. On the other hand, crop type Kc evolution demands more than 45 to stabilize.\\

\begin{figure}[!ht]
\centering
\includegraphics[scale=0.85]{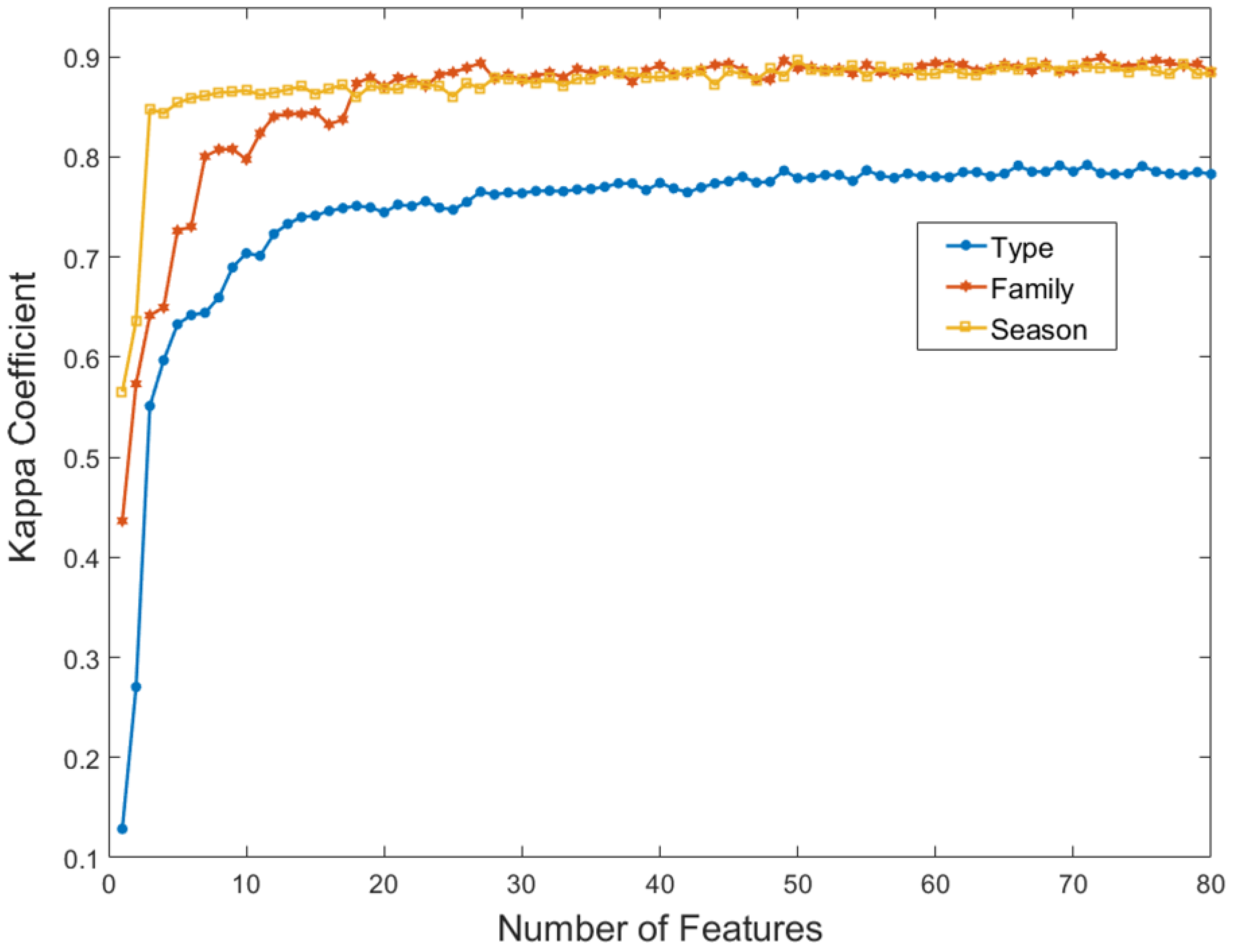}
\caption{Kappa coefficient evolution for all levels of nomenclature using the RF classifier.}
\label{fig:scalable_fig10}
\end{figure}

\noindent Inspecting Figure \ref{fig:scalable_fig10}, it is evident that RF performs exceptionally well for the crop family and season of cultivation classifications, with maximum Kc values comparable to the ones of SVM. However, crop type classification provides suboptimal performance. Introducing a larger number of spectrally similar classes, such as the constituent types of the cereal family, presents a challenge to the RF classification.

\subsubsection{Execution Time Considerations}
\begin{figure}[!ht]
\centering
\includegraphics{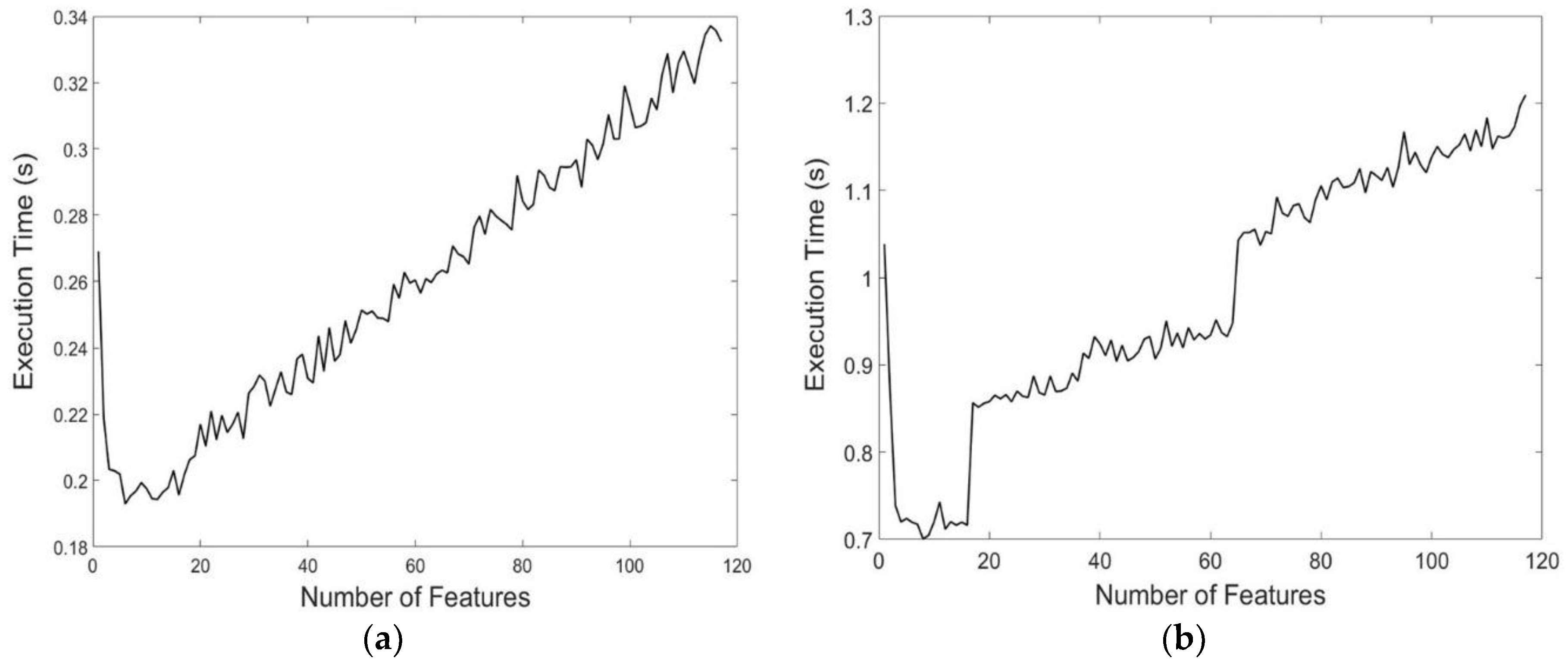}
\caption{Execution time evolution with increasing number of features for the Sentinel-2 based RF crop type classification, showing in (a) the training time evolution and (b) the implementation time evolution.}
\label{fig:scalable_fig11}
\end{figure}

\begin{figure}[!ht]
\centering
\includegraphics{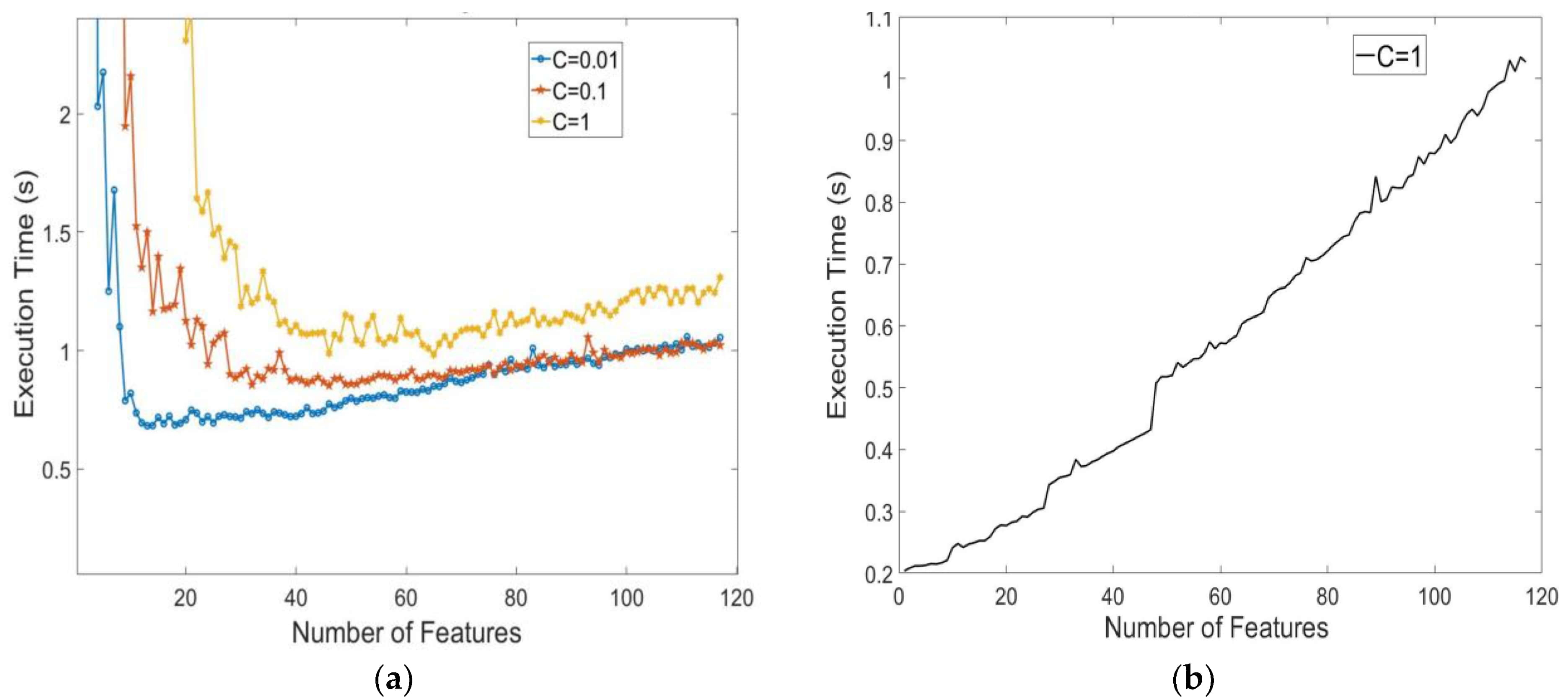}
\caption{Execution time evolution with increasing number of features for the Sentinel-2 quadratic SVM crop type classification, showing in (a) the training time for three different C values and in (b) the implementation time for C = 1.}
\label{fig:scalable_fig12}
\end{figure}

\noindent Sentinel’s freely received high temporal and spatial resolution imagery demands big data handling, establishing computational efficiency as a key consideration. In this regard, RF and SVM classifiers are assessed in terms of both their training and implementation times. The values in Figure \ref{fig:scalable_fig11} and Figure  \ref{fig:scalable_fig12} have been averaged for five separate iterations.

\noindent RF training time and both SVM and RF implementation times exhibit a linear relationship with an increasing number of features. The two distinct steps on the RF implementation curve are attributed to pre-processing overhead, evident in the MATLAB implementation of the algorithm. SVM training time curve acquires linearity only for feature spaces comprising more than 45 features. Experimenting with lower box constraint values (0.01–0.1), thus forcing the optimizer to find larger margin hyperplanes, showed that training time is both low and linearly increasing, for even smaller feature spaces (Figure \ref{fig:scalable_fig12}a). However, training accuracy is compromised as the misclassification cost constraint is relaxed. Inspecting Figure \ref{fig:scalable_fig8}, the threshold of 45 features coincides with the onset of the classifiers accuracy evolution plateau. Since the constraint is strict, the optimizer takes longer to find a satisfying hyperplane for spaces of fewer than 45 features. All in all, higher execution times are evident for quadratic kernel SVM classifications, with differences strengthening as the number of variables increases (Table \ref{tab:moniroting_cap_table6}).

\begin{table}[!ht]
\caption{Total RF and SVM (C = 1) execution times, including both training and implementation, as number of features increases (Sentinel-2 feature space).}
\label{tab:moniroting_cap_table6}
\centering
\scalebox{0.9}{
\begin{tabular}{|c|c|c|c|c|}
\hline
\textbf{Number of Features} &	\textbf{Total Time RF (s)}&	\textbf{Total Time SVM (s)} & \textbf{Difference (\%)} \\
\hline 
65	& 1.32 &	1.46 &	10.6 \\
90	& 1.42 &	1.80 &	26.7 \\
117	& 1.55 &	2.19 &	41.2 \\
\hline
\end{tabular}}
\end{table}

\subsection{Discussion}

\subsubsection{Sentinel-2 MSI and Landsat-8 OLI}

\noindent We tested the performance of our processing workflow using Landsat data as input, as well. The aim is to highlight the performance improvement, if any, achieved with Sentinel-2 data, based on the unique spatial and temporal resolution, and the spectral characteristics in utilizing parts of the spectrum of key significance in discriminating between types of vegetation. The values in Table \ref{tab:kc_svm} have been averaged over 20 iterations of random training sample splits.\\

\begin{table}[!ht]
\caption{Maximum Kc values for SVM crop type classifications based on the Sentinel-2 MSI, pan-sharpened Landsat-8 OLI and combined feature spaces.}
\label{tab:kc_svm}
\centering
\scalebox{1}{
\begin{tabular}{|c|c|c|}
\hline
\textbf{Feature Space} &  \textbf{Kappa RF} & \textbf{Kappa SVM} \\ \hline
Landsat-8 & 0.5991 & 0.7011 \\ \hline
Sentinel-2 & 0.8055 & 0.8830 \\ \hline
\end{tabular}}
\end{table}

\noindent Landsat crop-type classification proves to perform significantly worse than the Sentinel equivalent (\ref{tab:McNemar}). A combination of Landsat and Sentinel feature spaces marginally increases the overall accuracy and thus does not justify the higher complexity introduced. This is expected as the two feature spaces are highly correlated.\\

\noindent We used McNemar’s test to evaluate the statistical significance in the classification accuracy between (i) the two classifiers in the Sentinel scenario and (ii) Sentinel-Landsat pairs in the SVM scenario. This is an interpretable statistical test that quantifies the superiority between two thematic maps \citep{foody2004supervised}. McNemar’s test is essentially a standardized normal chi-square statistic, computed from a two by two matrix based on correctly and incorrectly classified parcels in both classifications, as shown in Equation \ref{eq:chi_square}. \\

\begin{equation}\label{eq:chi_square}
    x^2 = \frac{(n\textsubscript{ab} - n\textsubscript{ba})^2}{n\textsubscript{ab}+ n\textsubscript{ba}} 
\end{equation}

where n\textsubscript{ab} is the number of parcels correctly classified by classifier one but incorrectly classified by classifier two; n\textsubscript{ba} is the number of parcels correctly classified by classifier two but incorrectly classified by classifier one \citep{petropoulos2012land}.\\

\noindent The difference between all pairs, shown in Table \ref{tab:McNemar}, proves to be statistically significant at a 99.99\% confidence level. In this table, higher x\textsuperscript{2} values indicate better accuracy performance for the first item in the “pairs” column. These results further support the argument of SVM’s dominance over RF (pair 5) and similarly MSI’s dominance over OLI (pairs 1, 2 and 4). Relative x\textsuperscript{2} differences among the pairs can in ways determine the importance of parameters such as spatial resolution, spectral characteristics, and choice of classifier\\

\begin{table}[!ht]
\caption{McNemar’s test results for all different classification pairs (spatial resolution is defined in parentheses).}
\label{tab:McNemar}
\centering
 \begin{tabular}{||c c c c c c c c||} 
 \hline
 \multicolumn{3}{|c|}{Pair} & n\textsubscript{ab} & n\textsubscript{ba} & x\textsuperscript{2} & p-Value & h \\ \hline
 
1 & Sentinel (10 m) & Landsat (15 m) & 205 & 1073 & 589.53 & <0.0001 & True \\ \hline
2 & Sentinel (10 m) & Landsat (30 m) & 207 & 1204 & 704.47 & <0.0001 & True \\ \hline
3 & Landsat (15 m) & Landsat (30 m) & 371 & 500 & 19.11 & <0.0001 & True \\ \hline
4 & Sentinel (30 m) & Landsat (30 m) & 322 & 988 & 338.54 & <0.0001 & True \\ \hline
5 & Sentinel SVM & Sentinel RF & 114 & 577 & 310.23 & <0.0001 & True \\ \hline
 
 \end{tabular}
 \end{table}

\noindent OLI’s data pan-sharpening and MSI’s data down-sampling, in pairs (1) and (4) respectively, attempt to reduce the effect of the spatial resolution difference and thus isolate and quantify the importance of other sensor attributes. The difference in accuracy is significant for both scenarios, which can be attributed to Sentinel-2’s superiority in spectral characteristics, specifically having four vegetation red-edge bands (B05, B06 and B07). The difference for pair (3) is statistically significant but, nonetheless, implies a marginal increase in accuracy. In other words, pan-sharpening proves indeed useful for Landsat-8 but cannot be compared with the higher spatial resolution, in which Sentinel-2 multispectral bands sense directly.\\

\noindent Moreover, pair (2) depicts the maximum difference between the two sensors, accounting for the impact of both spatial resolution and spectral characteristics. Sentinel’s superiority in temporal resolution could not be adequately exhibited for the present case study, as the cloud free images for the two sensors are comparable in both number and temporal span. Nevertheless, the sensors’ difference in temporal resolution is significant and even larger performance differences are expected in most of relevant scenarios. \\

\subsubsection{Relevance of Methods}

\noindent The overall performance of the proposed scheme is in accordance with the requirements of an operational agriculture monitoring system. Discrimination and characterization of the cultivated crop types, is of paramount importance in the development, conservation, and management of natural resources at both macro- and micro-scales [41]. The identification of crops, along with their
distribution, management practices, and annual rotation schemes provide essential information for the implementation, control, and monitoring of agricultural policies and environmental measures imposed by the CAP \citep{alganci2013parcel}.\\

\noindent Open access to the unmatched features of the Sentinel mission, in temporal and spatial resolution, and their efficient exploitation, signal a new era in the field. A revisit time of five days ensures the proper construction of imagery time-series, able for the consistent and timely monitoring of the agricultural landscape. Additionally, the alternative Landsat data, featuring a three-fold reduction in temporal resolution, can offer inadequate number of quality images, particularly in heavily clouded regions. \\

\noindent On the other hand, Sentinel-2’s high resolution multispectral data enable the successful employment of a parcel-based approach and the generation of parcel-specific thematic information. Besides, alternative VHR imagery is unrealistically costly for operational, large-scale, and consistent monitoring. The proposed methods, which effectively take advantage of the Sentinel-2 attributes, could ably function as the backbone for the EO-assisted compliance validation of CAP requirements, especially to what concerns the enhancement of transparency and the simplification of subsidy administration. The scheme was designed to directly assist the CAP paying agencies in their compliance inspections, by utilizing the data used (LPIS and farmers’ declarations) in their existing operations, to offer a monitoring alternative to their inefficient sample-based controls. The scheme is finally characterized by robustness, in the sense that it is largely independent of manual fine-tuning or case-specific optimizations and is thereby reproducible.\\

\subsubsection{Relevance in Operational Scenarios}

\noindent The notion of transferability, although addressed in the design process and the selection of input data, remains to be validated by applying the scheme to other agricultural landscapes in the European Union (EU). An evident issue of transferability is the absence of standardization and the dissimilarity of the LPIS data and farmers’ declaration among the different EU countries. There are significant differences in the definition and description of cultivation practices and crop types. Dealing with this variability requires manual pre-processing and polishing of the input data and entails the appropriate definition of the crop types to be classified, merging or breaking them down to classes that describe spectrally coherent vegetation types. \\

\noindent In this study, it was shown that a time-series of multispectral imagery is required for accurate crop classification, as capturing the growing of crops exposes the most significant differences in their spectral signatures and thus enables their successful discrimination. Nonetheless, it was also exhibited that only a handful of images, appropriately distributed in time, is required to offer high thematic accuracy. The twin Sentinel-2 satellites offer a combined revisit time of five days, amounting to more than 70 images throughout the year. It is expected that in most cases this is a sufficient number of images to produce several quality cloud-free images within the year. However, in northern countries, where cloud-free imagery is scarce, Landsat-8 OLI data can be employed to enrich the feature space. Alternatively, fusion techniques with weather independent Sentinel-1 SAR data could also be explored.

\noindent When applying the scheme over large areas, the corresponding Sentinel-2 tiles can cover multiple adjacent satellite tracks, thus having different sensing dates. To overcome this issue, all tracks should be resampled with respect to a common temporal frame of reference, as proposed in \citep{inglada2015assessment}. Starting from the first image acquisition, a sampling step of five days, namely the Sentinel-2 revisit time, is set to create a grid of virtual sensing dates \citep{inglada2015assessment,inglada2017operational}. Then all images from all tracks are linearly interpolated to these predefined sample instances. The temporal interpolation was shown to marginally affect the classification accuracy \citep{inglada2015assessment}.\\

\noindent Since the methodology introduced was designed to be fully transferable by making use of predominantly open access data and data provided by the targeted end-users, the concept of scalability in regional or national scales is of great interest. Scaling up comes against certain trade-offs, mainly regarding the computational complexity, overall accuracy performance, and geographic scale—which in ways can be addressed based on this study’s results. Processing should be performed in adequately small regional extents, such as the one exhibited in the present study, in order to avoid the geographic variability of the crop type spectral signatures. Thereby, the difference in processing time between SVM and RF application, as presented in Section 4.4, can be thought of as linearly increasing in scaling up scenarios. SVM usage would indeed compromise computational economy, but not significantly. In that respect, accuracy is favored over computational complexity, as the relevant differences are less important. Nevertheless, in cases where the lowest level of nomenclature is not required, RF functions as an excellent alternative. It performs exceptionally for crop family and season of cultivation classifications, where it deals with fewer classes, of more distinct spectral profiles. All in all, it could be argued that both classifiers provide excellent results for a large number of classes, under an overall computationally efficient scheme.\\

\noindent Finally, the dependability of the methodology relies on the assumption that farmers’ declarations are truthful. In this study, it was shown that declarations were correct in their vast majority (99\%), which might not be the case for every relevant scenario. In \citep{pelletier2017effect} the authors have analyzed the effect of training class label noise for SVM (linear and radial basis function kernels) and RF classifiers that were applied on simulated vegetation profiles for 10 crop type classes. The results showed that SVM and particularly RF are robust for low noise levels (up to 20\%), with a marginal decrease in the OA. The robustness of the RF can be attributed to it excellent generalization ability. Multiple uncorrelated weak learners are trained with a random subset of the training set and then decisions are made based on the majority vote of the ensemble, making the classifier resistant to overfitting \citep{pelletier2017effect}. All in all, it can be argued that the original assumption is fair and that it can ultimately shape a robust scheme, even for cases where incorrect declarations amount to a considerable percentage of the training set. \\

\subsubsection{Greening 1: Crop Diversification}

\noindent Monitoring of compliance to CAP’s Greening 1: Crop Diversification requirement is one example of the proposed scheme’s direct application. Crop diversification entails the growing of different crop types based on farmers’ total land area, aiming to improve biodiversity and reduce soil erosion. Farmers owning land of less than 10 ha are automatically exempted from the rule. If however, their land extends between 10 and 30 ha, at least two different crop types must be cultivated, with the main one not exceeding 75\% of the total land. Similarly, arable land larger than 30 ha should involve at least three different crop types, with the main one occupying up to 75\% and the main two occupying less than 95\% of total land \citep{greening2018online}. Based on the SVM crop type classification, 42.6\% of farmers were exempted by having total arable land smaller than 10 ha, while only 2.2\% were found to not comply with the requirement The Greening 1 requirement considers the total arable land owned by the farmer in order to decide on their compliance. Since our knowledge was limited to this particular dataset, we assumed that every farmer’s total land is in fact encompassed within the borders of the study area. Therefore, the results are not representative of local farmers’ compliance to Greening 1 but merely exhibit the ease of decision making based on the crop identification product.\\

\subsection{Conclusions}

\noindent Sentinel-2 data time series using a parcel-based quadratic SVM classification provided a successful crop identification product, reaching an overall accuracy higher than 0.87 Cohen’s kappa, for the lowest level of the crop nomenclature hierarchy. RF classification provided comparable results for the family and season of cultivation nomenclature levels, while it underperformed for the crop type level. Also, Landsat-8 OLI imagery, of lower spatial resolution, resulted in inferior performance, validating the argument of Sentinel 2’s dominance in smallholding agriculture mapping. Accurate crop classification enables the monitoring of the CAP and allows for effective and efficient decision making on farmer compliance. The proposed methods were designed appropriately to be geographically independent and potentially scalable from the extent of a small region to national or even continental scales. In the same fashion, input data are deliberately kept least and freely available in order to achieve maximum transferability and ease of data retrieval.

\section{Semantic Enrichment} \label{semantics}

\subsection{Literature review}
\label{section:intro}

\noindent In recent years, a massive quantity of georeferenced data is generated %which are stored and distributed via the web. Data come 
from many different sources like human activity and EO, in-situ sensors, satellite missions (e.g. Copernicus) and mobile phones. %The Copernicus EU programme, is believed to be a game changer for both science and industry. 
The semantic enrichment and linking of these free and open data of this scale, frequency, and quality constitute a fundamental challenge for interoperability and automation in decision-making. %paradigm change in EO. Today, Copernicus is producing 15 terabytes of data every day, while every product is downloaded 10 times on average. %However, the availability of the sheer volume of Copernicus data outstrips our capacity to extract meaningful and actionable information. The EO community needs technology enablers from the ICT to propel the development of entirely new applications at scale.
% \subsection{The Linked Data paradigm}
%However, 
EO data become useful only when analyzed together with other sources of data (e.g. geospatial data, in-situ data) and turned into actionable information and knowledge for decision making. In this context, linked data\footnote{https://www.w3.org/standards/semanticweb/data} %\hl{[ref]}
is a data paradigm that studies how one can make Resource Description Framework (RDF) \cite{mcbride2004resource,kolbe2019linked} %\hl{[ref]} 
data available on the web and interconnect it with other data with the aim to increase its value. In the last few years, linked geospatial data has received attention as researchers have started tapping the wealth of geospatial information available on the web using semantic web technologies ~\cite{koubarakis2016managing, zhu2017multidimensional}. 
Nevertheless, there are only a handful of applications that showcase the semantic integration of linked EO and non-EO products. The scalability to accommodate big linked EO data also remains an open issue \cite{koubarakis2017big}.\\

% \hl{need refs to support this}.

% \subsection{Post 2020 CAP: Context and identified gaps}
\noindent One of the domains that is already heavily dependent on the effective and efficient knowledge extraction from EO data is the control of the Common  Agricultural Policy (CAP) \cite{pe2019greener}. The European Union (EU), through the CAP, aims at increasing the European agricultural productivity under sustainable practices, while at the same time making sure that the farmers maintain a decent standard of living\footnote{http://esa-sen4cap.org/}. It is the EU’s aim to reinforce the competitiveness of European agriculture, whilst maintaining and strengthening its sustainability. This manifests as a major priority, with CAP’s annual budget amounting to approximately 59 billion Euros. The Integrated Administration and Control System (IACS) of the CAP, consumes the majority of its annual budget. The IACS functions as the management system for the CAP payments; and is implemented by the national paying agency of each EU member state\footnote{https://ec.europa.eu/info/food-farming-fisheries/key-policies/common-agricultural-policy/financing-cap/controls-and-transparency/managing-payments\_en}. The CAP legal framework is transitioning to its new form, the post-2020 CAP reform, which aims to modernize and simplify the current operating model\footnote{https://www.consilium.europa.eu/el/policies/cap-future-2020/}. Based on the post-2020 CAP ambitions and towards the direction of the so-called monitoring approach for the implementation of IACS, Earth Observation has been identified as a key enabler.\\

\noindent Multiple EC funded projects have employed Copernicus data, using advanced ICT and A technologies, to address the monitoring of the CAP. The RECAP project\footnote{https://www.recap-h2020.eu/} has been one of the first to develop Copernicus based machine learning pipelines to assist the Paying Agencies in reducing the costs and increasing the efficiency of the control of CAP’s Cross-Compliance. Additionally, the Sen4CAP project\footnote{http://esa-sen4cap.org/}, building on the legacy of RECAP, has focused on reducing the costs of IACS towards the post-2020 CAP objectives; exploring the applicability of an evidence-based monitoring approach. To the best of our knowledge, none of the existing approaches are able to support the following CAP scenarios to deliver true business value. The selected scenarios require significant human resources, with On-The-Spot Checks (OTSCs) on only a small sample of the farmers' applications.\\

\paragraph{Smart Sampling of on-the-spot-checks (OTSCs)}.
Farmers declare the cultivated crop type for their arable land each year, around the months of May and June. Paying agencies are responsible for validating the declarations to then grant the requested subsidy to the farmer. The MSs randomly sample and inspect 1-5\% of the total number of declarations. It is necessary to automatically monitor the farmers' declarations, using and linking additional and online available data, in order to create an targeted, instead of random, sample set that requires OTSC.

\paragraph{Automatic detection of CAP’s Greening 1 rule}.
This rule seeks to improve biodiversity and reduce soil erosion by imposing limits in the size and number of the different cultivations in a farm. Specifically, the farmers that own 10 to 30 ha of arable farm should grow at least two different crop types, while farmers that own more than 30 ha should grow at least three different crop types.  In the first case, the main crop should not cover more than 75\% of the land, while for the latter case the two main crops should additionally not exceed 95\% of the total land
%  , as explained in Figure \ref{fig:crop_diversification}%
\footnote{https://ec.europa.eu/info/food-farming-fisheries/key-policies/common-agricultural-policy/income-support/greening\_en}. 
This rule requires accurate and semantically enriched geospatial data, so as to both detect correctly the crop types and perform semantic reasoning to infer the consistency between the farmers' declaration and the Greening 1 rule.
% \begin{figure}[h]
%     \centering
%     \includegraphics[width=8cm]{images/greening_130420_dpi300.png}
%     \caption{Greening 1: Crop diversification conditions for direct payment}
%     \label{fig:crop_diversification}
% \end{figure}

% Using i) the accurate crop classification of the cultivated crop type to each parcel, ii) the area of the parcel computed through geosparql and iii) the Farmer ID that defines the parcels owned by a specific farmer, the user can automatically query the IDs that do not comply with the Greening 1 requirement. 

\paragraph{Automatic detection of susceptible parcels according to CAP’s SMR 1 rule}.
This particular CAP requirement expects from the farmers, among other things, to perform a risk assessment on the susceptibility of their parcel to contribute nitrate-rich soil to nearby surface waters. The farmer should account for the slope of land, the ground cover, the proximity to surface water, weather conditions, soil type and conditions and the presence of land drains\footnote{GCCE 2017 v1.0, "The guide to cross compliance in England 2017”, produced by the Department for Environment, Food and Rural Affairs}. The SMR 1 requirement defines buffer zones, which shall be respected in terms of fertilizer spreading. Specifically, i) manufactured fertilizer spreading should be at least 2 m from surface water and ii) organic manure spreading should be at least 10 m from surface water\footnote{https://www.gov.uk/guidance/using-nitrogen-fertilisers-in-nitrate-vulnerable-zones}. Therefore, measuring the proximity of the parcel boundaries to the nearest surface water is of great significance both for the inspections of the paying agency, but also for the farmer who wishes to better comply with the requirement. To that end, the semantic data fusion of Sentinel images and other linked open geospatial data, would lead to a more efficient monitoring of SMR 1.\\

% \subsection{Big Data Life Cycle}

\noindent In this work we adopt and extend the linked open EO data life cycle paradigm %proposed by Koubarakis et al.
~\cite{koubarakis2016managing}, and provide for the first time an end-to-end implementation to address operational needs of the CAP. Motivated by the above mentioned use cases and their inherent requirements to effectively and intelligently combine and interlink various geospatial data sources (e.g. LPIS), hydrographic network, Natura2000 zones etc.), we instantiate the linked open EO data life cycle in the domain of CAP. To this end, we propose a hybrid data- and knowledge-driven framework, developing concrete CAP-related scenarios and demonstrate automatic pipelines for satellite data processing, content extraction, semantic annotation and transformation to RDF, interlinking layer, validation and querying of Linked Open spatio-temporal data.\\

% \subsection{Contributions of this paper}
%More specifically, 
\noindent Our contributions are summarized as follows: 
\begin{itemize}
    \item  We propose the use of semantic reasoning in the context of CAP monitoring, to check compliance with Greening-1 requirements, taking into account satellite derived products and ancillary geospatial data.
    \item We demonstrate the use of spatial relationships in LOD (GeoSPARQL) towards assessing vulnerable parcels according to CAP SMR-1 specifications.
    \item We propose the Smart Sampling Scheme, i.e. the use of spatiotemporal queries to define a new, educated sampling of parcels that need to be checked for compliance with CAP rules with in-field visits.
    \item We evaluate our framework under the light of national scale application, in line with post-2020 CAP monitoring needs. Therefore, we discuss scalability implications for both the knowledge extraction from satellite imagery module and the semantic reasoning framework.
\end{itemize}

\noindent This section is structured as follows. Section \ref{related_work} presents related technologies considering both the EO-based CAP monitoring and semantic web technologies. The technologies are grouped under a common framework, aiming to cover the whole life-cycle of the linked EO data for the control of the CAP. Section \ref{methodology} describes our proposed semantically enriched crop type classification model for checking the compliance of farmers' declarations to the CAP regulations. % Our proposed framework's layers correspond to the stages of the life-cycle that have been presented in Section \ref{related_work}. 
Section \ref{area_of_interest} presents experiments and results regarding our proposed methodology, along with the area of interest, the considered CAP scenarios, their implementations and results regarding the effectiveness and efficiency of our pipeline. Section \ref{sec:conclusion_semantics} concludes this work.\\

\subsection{Life-cycle of linked EO data for the control of the CAP} 
\label{related_work}

%\todo[inline]{
%I suggest the following structure:

%\begin{enumerate}
%    \item \textbf{CAP monitoring}: what is about, aims, objectives, problems, challenges, relevant technologies, data used, existing solutions / frameworks / papers / etc.
    
%    \item \textbf{Semantic Web technologies}: ontology languages and standards (OWL/RDF, SPARQL, GeoSPARQL), reasoning solutions (DL, rules), Linked Data, LOD, what they offer, benefits, etc.
    
%    \item \textbf{Linked Data and EO}: motivation, benefits, etc., presenting also the general concepts of [8]:
%    \begin{enumerate}
%        \item Knowledge Extraction
%        \item Semantic Annotation
%        \item Transformation into RDF
%        \item Storage and querying
%        \item Interlinking
%    \end{enumerate}
%    However, apart from generic concepts on the 5 pillars, we need also to present related work relevant to the CAP / Agri domain, as we do for Knowledge Extraction which is adapted to CAP. For example, Semantic Annotation needs to describe solutions / ontologies used in Agriculture or CAP.
%\end{enumerate}
%}

\begin{figure}[h]
    \centering
    \includegraphics[width=\linewidth]{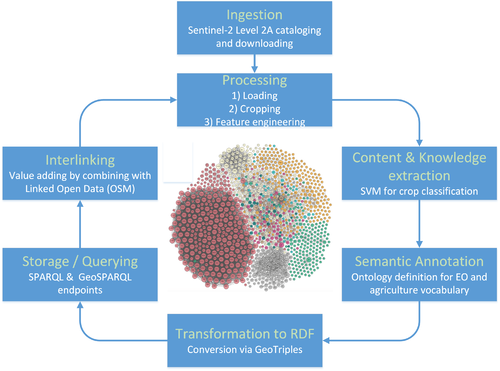}
    \caption{The main stages of the complete life-cycle of linked open EO data for the monitoring of the CAP}
    \label{fig:stages}
\end{figure}

\noindent Existing works for the monitoring of CAP mainly focus on knowledge extraction technologies, while the semantic technologies cover more generic agricultural needs. However, our main objective is to cover the complete life cycle of linked open EO data paradigm, as originally discussed in ~\cite{koubarakis2016managing}, following a multi-disciplinary approach. The stages of the life-cycle of linked EO data for the control of the CAP are presented in Figure \ref{fig:stages}. The first step covers the content extraction machine learning methodologies so as to get new information layers out of the large streams of raw satellite data. The second step involves the standardized data representation and ontological modeling for semantic annotation. The semantic annotation is based on semantic Web technologies which are being adapted and developed under the EO and Agricultural domains. The next step of the life-cycle regards the transformation of the extracted content into RDF, allowing the population of the knowledge base (triplestore) to perform semantic queries that offer a better knowledge of the data (Storage/Querying). Applying useful interconnections in the semantic data using external datasets can additionally enrich the content and extract hidden knowledge (Interlinking).

%As stated in ~\cite{koubarakis2016managing} creating a methodology that covers the complete life cycle of linked open EO data has not yet been implemented and is a vision to the Semantic Web community. A system that applies the paradigm shift that has been proposed in ~\cite{koubarakis2016managing} to support the various stages can be extremely useful in applications development. 

%Data published following the linked data paradigm can be further analyzed using machine learning techniques (Content and Knowledge extraction). Ontologies need to be defined to meet the needs of exploiting the data (Semantic Annotation). Results need to be semantically represented (Transformation into RDF) and saved into semantic triplestore to perform semantic queries that offer a better knowledge of the data (Storage/Querying). Applying useful interconnections in the semantic data using external datasets can additionally enrich the content and extract hidden knowledge (Interlinking).

%https://www.overleaf.com/project/5e8c2ff104ded00001b65b3e

%In the next sections we present the related work based on the stages defined on this section.

\subsubsection{Content and Knowledge Extraction: Crop Classification}
\noindent Over the last decades there have been multiple studies that have utilized EO data to extract high level thematic knowledge for the agricultural land. Recently and since the introduction of the Sentinel missions, there have been a plethora of scientific publications that have exploited either Sentinel-1 or Sentinel-2 imagery, or in certain cases both, to classify crop types. The high spatial and temporal resolution of the Sentinel missions makes them ideal for constructing dense image time-series of high quality that capture all the phenological stages of the different crop types and thus allowing for their accurate discrimination. \\

\noindent The state-of-the-art in EO-based and specifically the Sentinel-based crop classification has advanced significantly over the last years, with the majority of publications reaching optimal accuracy levels ($>$85\%) for multi-class problems. The relative differences in the published approaches are based on the nature and level of specificity of the investigated crop classes, the computational complexity restrictions, the scale of application and the ground truth information that is available for training and validation. \\

\noindent In order to reduce the computational complexity of crop classification and develop scalable solutions, multiple studies have followed Object-based Image Analysis (OBIA) approaches. For instance, in \cite{lebourgeois2017combined} the authors have segmented their image stack into objects using spectral segmentation techniques on Very High Resolution imagery, while in \cite{sitokonstantinou2018scalable} the authors made use of the LPIS to partition their feature space into parcel objects. \\

\noindent Synthetic Aperture Radar (SAR) and optical imagery, retrieved from Sentinel-1 and Sentinel-2 missions respectively, have been used either individually or combined. In \cite{brinkhoff2020land} a combination of both Sentinel-1 and Sentinel-2 is used in order to create very dense time-series, thus alleviating the cloud coverage limitations. In \cite{arias2018crop}, the authors employ solely Sentinel-1 data, suggesting a weather independent crop classification scheme for the monitoring of the CAP, hence accounting for northern European countries that suffer from year round cloud coverage. Other studies focus on generating multiple diverse features from Sentinel imagery, beyond the most common spectral bands and VIs. In \cite{feng2019crop} and \cite{akbari2020crop}, the authors create deep feature spaces, additionally including variations of VIs, texture and phenology parameters. Such methods are shown to be particularly useful in classifying spectrally heterogeneous crop classes, i.e. vegetables.\\ 

\noindent With respect to the classification methods employed, both supervised and semi-supervised learning approaches can be found in literature. In \cite{solano2019semi}, for example, the authors have combined a hierarchical correlation clustering with an artificial neural network. The vast majority of studies, however, make use of supervised learners, such as Support Vector Machines (SVM) and Random Forest (RF) (\cite{sitokonstantinou2018scalable}, \cite{lebourgeois2017combined}, \cite{feng2019crop} , \cite{akbari2020crop}, \cite{brinkhoff2020land}, \cite{ray2019exploring}, \cite{singh2019assessment}). Their effectiveness stems from their ability to accurately describe the nonlinear relationships between crops’ physical condition and their spectral characteristics, while being particularly insensitive to noise and overfitting. Finally, there are important studies that have used Convolutional Neural Networks (CNN) or Recurrent Neural Networks (RNN) or a combination of both \cite{mazzia2020improvement}, which allow the learning of time and space correlation over the Sentinel time-series, thus reducing manual feature engineering.\\
% Especially, SVM is very popular for land cover mapping among other classification methods. The success of SVM is related to the identification of the non linear relationships between crops’ spectral characteristics and their physical conditions. Moreover, SVM is resilient to noise and over-fitting and in the same time is able to be effective on small training datasets.

% \noindent In \cite{sitokonstantinou2018scalable}, the authors have developed a scalable crop identification scheme, employing a 2nd order polynomial SVM on a time-series of Sentinel-2 data. The authors of \cite{sitokonstantinou2018scalable} have additionally performed an extensive comparison between SVM and RF, them being the most widely used classifiers for crop mapping problems. The results showcased the superiority of SVM over RF for the classification of multiple and spectrally similar classes. This conclusion is additionally supported by \cite{zhang2019cropclass}.\\

\noindent This study builds on the methods and results that were presented in Section \ref{croptype}, following the state-of-the-art in crop classification as described earlier. The crop classification method of this study, however, has been applied to three different areas of interest, of diverse characteristics, thus proving its transferability. Finally, we perform multiple crop classifications, starting from very early in the year; therefore with truncated feature spaces. Nonetheless, the results, even early in the year, are satisfactory for the purposes of smart sampling the CAP OTSCs.\\

\subsubsection{Semantic Web Technologies}
\noindent Web Ontology Language (OWL) ~\cite{mcguinness2004owl} is an ontology language that provides classes, properties and individuals under the semantic web aspect. Ontologies offer the taxonomy of semantic objects and the relationship between them. RDF ~\cite{patel2002yin} ~\cite{bizer2011linked} is the W3C recommendation standard that offers data representation under subject–predicate–object standard, which is known as triples. Each subject is a resource and each object can be either a resource, a value or an empty node. Predicates or properties express the relationship between a specific subject and object. Data expressed in this format are saved into RDF triplestores, named Knowledge Graphs.\\

\noindent SPARQL Protocol and RDF Query Language (SPARQL) ~\cite{perez2009semantics} is the most popular querying language for the retrieval and manipulation of data in RDF format. SPARQL offers a wide range of query forms and operators to access and retrieve the data. stSPARQL ~\cite{koubarakis2010modeling}~\cite{kyzirakos2018geotriples} is a SPARQL version that applies semantic queries into data in stRDF %~\cite{koubarakis2010modeling}~\cite{kyzirakos2018geotriples} 
format. Such formats offer representation and querying of thematic and spatial data which contain a temporal dimension. GeoSPARQL ~\cite{kyzirakos2018geotriples} focuses more on geospatial data querying, by providing a wide list of functions to support semantic queries execution over geometry and feature objects. Topological relationships are also taken into account.\\ 

\noindent Reasoning ~\cite{maarala2016semantic} is the procedure of infering logical consequences based on asserted facts or axioms. In RDF graphs, reasoning takes advantage of data triples using in many cases different data sources to specify the rules that can lead to useful knowledge extraction. Reasoning with rules is usually based on Description Logics (DLs). Description Logics~\cite{baader2008description} constitute a family of logic-based representation formalisms and are usually used to represent well-structured knowledge over the application domain. Its name comes from a combination of descriptions, which are the expressions namely predicates, and the fact that they support logic-based semantics. DLs are strongly associated with structuring ontology languages such as OWL, but are also widely used in application domain.\\

\noindent The vision of linked data is associated with the transformation of data into RDF formats. Data in this format can be published on the web and linked with other existing data that come from different sources ~\cite{bizer2011linked}. Linked data are easily accessible using semantic queries. %The result is an easily accessible data model with a great amount of interconnected information.
The main advantage of semantics is that they have the means to create intelligent interconnections over objects that come from heterogeneous sources as they support better information management, complexity limitation and useful inferences extraction ~\cite{dam2011knowledge}. %The dynamic of semantics is extremely useful in decision making systems. 
In this work we provide a list of functions and semantic queries, using semantic web technologies, to support three CAP-related scenarios in real operational problems, that require content extraction and semantic linking of data for compliance checking.

%\h

\subsubsection{Semantic Annotation under the Earth Observation and Agriculture domains}
\noindent Building appropriate ontologies to describe the different aspects of EO and agriculture are presented in this section. EO ontologies focus more on the environmental monitoring domain.\\ 

\noindent The ontology presented in ~\cite{wang2017building} deals with hydrological monitoring issues and captures the main components of hydrological monitoring which are the events, the sensors and the observations. Sensors and observations are divided into many subcategories such as physical and meteorological, while events are associated with any hydrological cycle change. Modular Environmental Monitoring Ontology (MEMOn) ~\cite{masmoudi2018predicat} extends the abovementioned ontology as, except from sensor and observation data, it provides a structure to model a plethora of different aspects that are identified on an emergency situation under the environmental monitoring domain. The ontology provides the structures to represent environmental features (procedure and material), physical conditions (disaster) and spatiotemporal information (geolocation and time).\\

\noindent An agricultural ontology representation method is described in ~\cite{song2012study}. The suggested model contains information such as cultivation and processing practices, storage, pests control, genetic attributes etc. OntoCrop ontology ~\cite{maliappis2009applying} offers knowledge representation for common cultivation practices, pests control and in general the crops physiology. Each plant is characterized by properties such as name, growth stage, type of infection, infected part, information about disorders. Agriculture ontology for the purpose of agriculture internet of things (AgOnt) ~\cite{hu2010agont} presents a more product-oriented view of agricultural products containing information related to the product, the seeding procedures, the physical conditions, the phase, the location and temporal dimensions. The main purpose of this ontology is supporting healthy food management. The ontology presented in ~\cite{su2012agricultural} offers a uniform representation of text classification and concepts extraction results. The ontology matches specific concepts into ontology classes which include many different types of products such as agricultural, planting, livestock, fishery and agricultural material.\\

\noindent Most works that have been previously mentioned describe either EO or agricultural data. What is actually missing is a combination of both describing information from EO with information from the agricultural domain. In this work we reuse and extend the ontology found in ~\cite{sitokonstantinou2018scalable} that describes the relationship among crop types, families and season, and create a combination with GeoSPARQL vocabulary that represents geospatial-related data such as points in polygon geometry.

\subsubsection{Transformation into RDF}
\noindent Another widely investigated issue is combining semantics with EO data to discover hidden knowledge. This section describes some frameworks that deal with data transformation into semantic format, integration and searching.\\

\noindent Intelligent Interactive Image Knowledge Retrieval ($I^3KR$) ~\cite{durbha2005semantics} is a framework that utilizes EO data archives and applies image segmentation (PCA kernel approach) and classification techniques (SVM learning method). From the semantic aspect, the system achieves high-level query processing into context information from distributed data archives. Domain-specific ontologies provide the appropriate structures to integrate heterogeneous data sources in order to support complex semantic queries. A hybrid ontology approach has been used to integrate data coming from different ontologies. Semantic restrictions have been applied using DL reasoning to determine the conditions under which an instance will belong to a class.\\

\noindent GeoTriples ~\cite{kyzirakos2018geotriples} is a tool that deals with the geospatial data transformation into semantic RDF format. The system gets as an input a file in various formats and creates a mapping which is based on GeoSPARQL vocabulary, using RML and R2RML rules (mapping generator). Users have the opportunity to define the rules if needed. Initial data are transformed into RDF graph format using the RML rules defined in previous phase (mapping processor). Various RDF syntax formats are supported. Querying is also available in a relational database using R2RML mapping (stSPARQL/GeoSPARQL evaluator).\\

\noindent In this study, we use the GeoTriples tool as a basis in order to transform shapefile data into RDF format under the GeoSPARQL standard for semantic representation.

\subsubsection{Storage and querying}
\noindent We choose to handle the three different CAP scenarios using semantic technologies. The problem could have been solved using relational databases, though this selection would be accompanied with an inflexible data schema and higher execution times \cite{jaiswal2013comparative}. Additionally, the relationships between the entities handled in this work are quite complex to be represented using SQL keys\footnote{https://www.sqlshack.com/understanding-benefits-of-graph-databases-over-relational-databases-through-self-joins-in-sql-server/}. Information is coming from three layers and, with the usage of semantics, are combined in the most effective way, while OWL 2 RL rules are used to enrich the data \cite{meditskos2009rule}.\\

\noindent RDF triplestores are semantic databases that offer data saving in semantic graph format. Strabon~\cite{kyzirakos2012strabon} is a geospatial–oriented RDF triplestore that offers a broad amount of querying functions over georeferenced information, supporting both stSPARQL and GeoSPARQL. GraphDB\footnote{http://graphdb.ontotext.com/}~\cite{guting1994graphdb} is also a popular triplestore that supports saving and querying over georeferenced and non-georeferenced semantic data, supporting native OWL 2 reasoning. It is considered as one of the best triplestores available in terms of storage, supported functionalities, performance and execution time~\cite{bellini2018performance}. Other RDF triple stores that provide geospatial support include RDF4J\footnote{https://rdf4j.org/}, Virtuoso\footnote{https://virtuoso.openlinksw.com/}~\cite{erling2009rdf}, OntopSpatial\footnote{http://ontop-spatial.di.uoa.gr/}~\cite{bereta2019ontop}, Oracle spatial and Graph\footnote{https://www.oracle.com/database/technologies/spatialandgraph.html}, AllegroGraph\footnote{https://allegrograph.com/}, Stardog\footnote{https://www.stardog.com/}, uSeekM\footnote{https://www.openhub.net/p/useekm} and Parliament\footnote{https://github.com/SemWebCentral/parliament}.  \\
%Many semantic querying languages have been developed which apply semantic queries into geospatial data. stSPARQL ~\cite{koubarakis2010modeling}~\cite{kyzirakos2018geotriples} is a SPARQL version that applies semantic queries into data in stRDF ~\cite{koubarakis2010modeling}~\cite{kyzirakos2018geotriples} format. Such formats offer representation and querying of thematic and spatial data which contain a temporal dimension. GeoSPARQL ~\cite{kyzirakos2018geotriples} focuses more on geospatial data querying, by providing a wide list of functions to support semantic queries execution over geometry and feature objects. Topological relationships are also taken into account.%

\noindent The storage and query capabilities of our framework capitalise on an existing RDF triple store, on top of which SPARQL and GeoSPARQL standards are used to form the queries that support the rules of agriculture policies. The current implementation uses the GraphDB semantic graph database, taking full advantage of the provided dashboard to explore and manage the RDF repositories. It also supports different reasoning profiles, such as OWL 2 Rule (RL) reasoning, allowing us to use off-the-shelf reasoning on top of our domain ontology. It is worth mentioning, however, that since our framework capitalises on existing, well-known standards (RDF, OWL, SPARQL, etc.), it is interoperable and it does not depend on specific implementations. For example, it requires minor updates to migrate to different triple stores, according to the application requirements, such as Strabon and AllegroGraph, or to use different SPARQL query engines.\\

% user-friendly dashboard  for exploring data, while it requires the installation of additional frameworks, such as a Java Servlet Container, and a spatial DBMS to store the data. Strabon has been related with some limitations at the time we started working on this paper as some OGC GeoSPARQL functionalities were not supported ~\cite{nishanbaev2019survey}. Finally, there is no native OWL reasoning engine in Strabon to support basic reasoning services, such as subsumption hierarchies and property paths \cite{bellini2018performance}. On the other hand, GraphDB provides a 

\subsubsection{Interlinking}
\noindent To exploit the wealth of data, there comes the need of generating intelligent interconnections between different datasets. In literature, many systems have been implemented dealing with this issue but due to the vast heterogeneity of data, using existing systems into new datasets does not work in most cases. Interlinking is achieved based on geospatial data characteristics in some cases \cite{zhu2017multidimensional}, while in others specific mechanisms have been developed to meet the needs of the data ~\cite{yang2016earth}. \\

\noindent The system that is presented in \cite{zhu2017multidimensional} receives data from heterogeneous sources such as meteorological, health and EO. It creates an appropriate RDF representation and associations between specific characteristics. Data interlinking is achieved by calculating the similarity between different datasets. In ~\cite{yang2016earth} a system that integrates EO data to support data management is introduced. The system receives data from different data sources and has two different functionalities. For data that are related with China multiple components have been developed to adapt in different interfaces, while for international data the GEO DAB agent is used. PREDICAT \cite{masmoudi2018predicat} is a system that focuses on natural catastrophes prediction. PREDICAT uses different ontologies to semantically represent the data that are pertinent to the system (semantic layer). The system overcomes data heterogeneity and provides a common structure of interconnected objects containing spatiotemporal information (data integration layer), while a reasoner and a decision maker are implemented to provide the appropriate responses to the user (data processing layer). In CANDELA project\footnote{http://candela-h2020.eu/} semantic search is supported on EO images and other associated metadata, using existing technologies, namely GeoSPARQL, OWL-Time, SOSA, DCAT and PROV-O. These technologies support a monitoring use case in agriculture, where the impact of a storm on vineyards is measured by first extracting knowledge from Sentinel images and then semantically fusing them with weather reports in the same area of interest\footnote{http://candela-h2020.eu/content/semantic-search-v2}. However, the semantic search targets mainly to offer a mature solution for insurance companies, while it is not straightforward to monitor CAP-related regulations that require to be checked in terms of the farmers' compliance to CAP rules.\\

\subsubsection{The complete life-cycle for CAP monitoring using semantic technologies and linked open EO data}

\noindent Despite the fact that a lot of progress has been achieved in different aspects of the  individual components mentioned above (e.g. Semantic annotation has been implemented under the environmental monitoring domain \cite{wang2017building} ~\cite{masmoudi2018predicat} \cite{kontopoulos2018ontology}, data integration has been widely investigated (Interlinking) ~\cite{durbha2005semantics} \cite{audebert2017joint} ~\cite{audebert2017fusion} \cite{yao2017semantic} ~\cite{masmoudi2018predicat} \cite{hilbringharmonizing}), not much effort has been achieved in implementing a system that supports the \textbf{complete life-cycle}. 
%According to literature, 
The major challenge, which we address in this study, is mostly related with interlinking phase and dealing with the heterogeneity of data (e.g. sensors \cite{hilbringharmonizing}, aerial and satellite imagery \cite{audebert2017joint}, OpenStreetMap data \cite{audebert2017joint} \cite{yao2017semantic}, Google Earth imagery \cite{yao2017semantic}) and defining the ways to exploit these data and enhance knowledge discovery \cite{durbha2005semantics}.\\

\noindent Contrary to the presented approaches, in this work we focus on the linked open EO data life cycle paradigm proposed in  \cite{koubarakis2016managing}, aiming to support impactful use cases in CAP. To the best of our knowledge, this is the first attempt to reuse and adapt the proposed architecture to the domain of CAP monitoring. The proposed framework implements a hybrid scheme of data analysis and annotation: the results of a data-driven crop classification framework are semantically annotated and interlinked in order to foster advanced interpretation, such as improving classification accuracy through domain knowledge, and querying solutions. We demonstrate the added value and feasibility of our approach in a number of challenging use cases in CAP monitoring. \\

% knowledge-driven framework for satellite data classification, semantic annotation and transformation to RDF, interlinking and querying over linked geospatial data. An intelligent interlinking mechanism has been developed, to couple data from different sources (pre-classified data, data from OpenStreetMap, crop ontology) using the geolocation as the main interconnection point. 

% % covers an end-to-end example of the non-complying farmers detection based on the CAP requirements.
\subsection{Methodology}
\label{methodology}

\begin{figure}[!ht]
    \centering
    \includegraphics[scale=0.4]{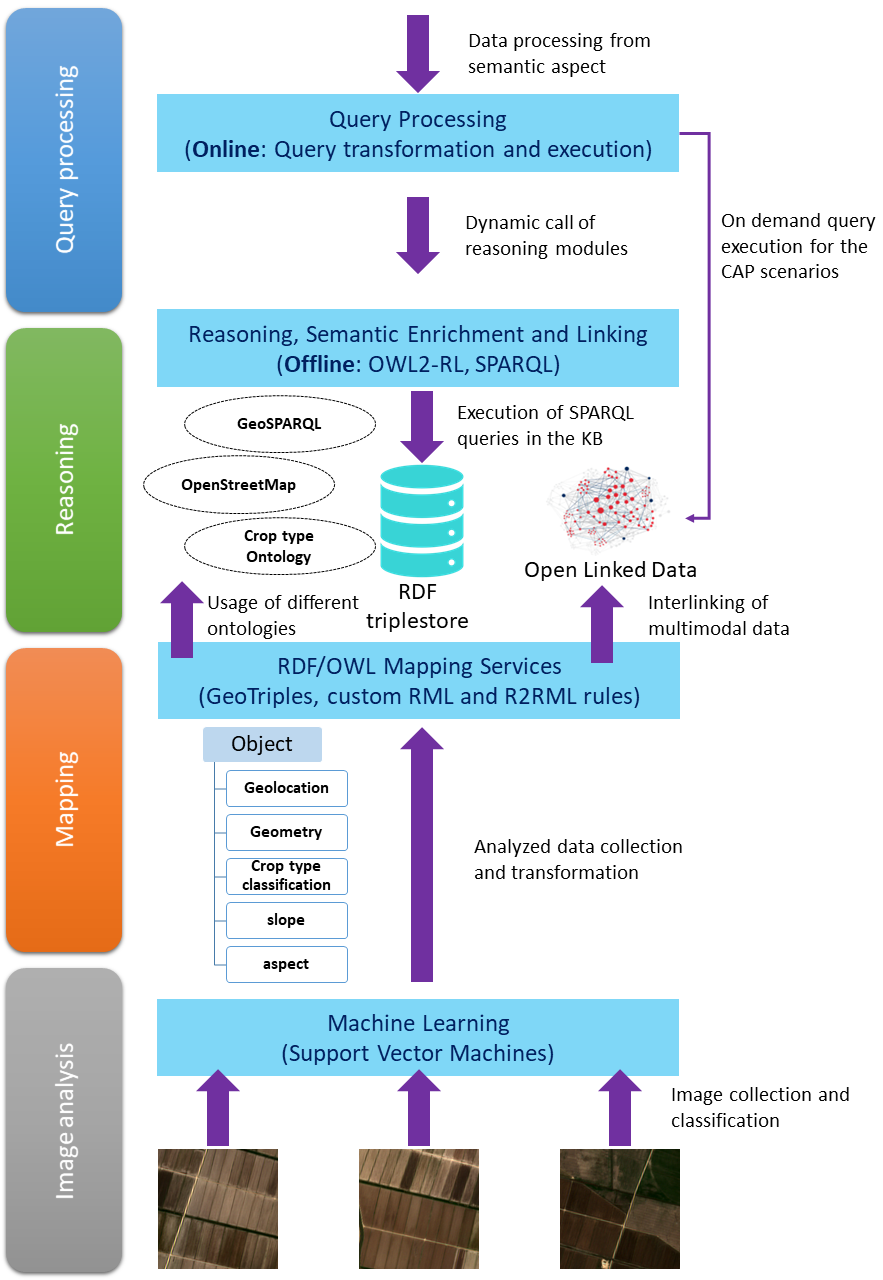}
    \caption{System architecture overview}
    \label{fig:methodology_arch}
\end{figure}

\noindent The overall framework of our proposed methodology is presented in Figure \ref{fig:methodology_arch}. The layers are the image analysis layer, the mapping layer, the data ingestion and reasoning layer and the query processing layer. The knowledge extraction phase of the life-cycle, which was presented in Section \ref{related_work}, consists of the image analysis layer using machine learning techniques for the content extraction. The semantic web technologies in the context of EO and Agriculture domains involve also the semantic annotation and transformation into RDF of the Mapping layer. The data ingestion and reasoning layer populates the knowledge base with the extracted knowledge, for storage and querying in a standard data representation model. Finally, the interlinking is done as part of the query processing layer that allows for checking the compliance of the farmers' declarations to the CAP regulations under the reasoning mechanism of the previous layers, using the open linked data paradigm. These layers are presented in detail in the following sections.

\subsubsection{Satellite image analysis for the monitoring of the CAP}
%\subsubsection{Crop classification}

%\textbf{Background and Legacy from Related Work} 

%This study is an extension of the work published by ~\cite{sitokonstantinou2018scalable}. In the aforementioned work, the authors have developed a Sentinel based crop classification system for the monitoring of the CAP. A similar implementation is followed in this study.  The input data that comprise the feature space for crop classification are listed in the previous section. A 2nd order polynomial Support Vector Machines (SVM) classifier is trained using a randomly selected 30\% of the total number of parcels. The farmer declarations, as part of the annual subsidy application for the CAP, have been used for labeling the dataset and training the supervised classifier. 

\noindent Paying agencies of EU member states, usually receive the annual subsidy applications in May or June. The paying agency inspectors require the information of the cultivated crop type, even as early as May. This way, inspectors can select and organize their OTSCs, which follow in the coming months. Additionally, crop classification results, received prior to the annual farmer declarations, can assist as an alerting mechanism during the application process. The image analysis layer is the first layer of our proposed pipeline (Figure \ref{fig:methodology_arch}), where a Sentinel-based crop classification system for the monitoring of the CAP is developed. \\

\noindent The Area of Interest (AOI) is located in northeastern Spain and specifically the district of Navarra. The AOI covers the agricultural land surrounding the city of Pamplona, capital of Navarra. In detail, the dataset includes 9,052 parcels and amounts to approximately 215 km\textsuperscript{2} of total land area. The Northern part of Navarra is surrounded by the Pyrenees Mountains, as they stretch southward from France. The landscape of the district is a mixture of forested mountains and watered valleys, while the agricultural land is characterized by substantial fragmentation ~\cite{rodriguez2012assessment}. This study builds upon the crop classification results, as described in Section \ref{croptype} and  ~\cite{sitokonstantinou2018scalable}. SVM based crop maps are produced, including the crop types of soft wheat (50\%), barley (26\%), oats (8.4\%), maize (1.4\%), sunflower (3.2\%), vineyards (1.3\%), broad beans (4.5\%), rapeseed (5.4\%) and cherry trees (0.2\%). The aforementioned crop types are the lowest level of ontology, as shown in Figure \ref{fig:croptype_ontology}. \\

% \begin{figure}[!ht]
% \centering
% \includegraphics[scale=0.4]{Figs/Chapter2/geom_buffer.png}
% \caption{Inward buffer to parcel boundaries to avoid using mixed pixels.}
% \label{fig:parcels_buffering}
% \end{figure}

% \begin{figure}[!ht]
\begin{figure}[!ht]
    \centering
    \includegraphics[scale=0.8]{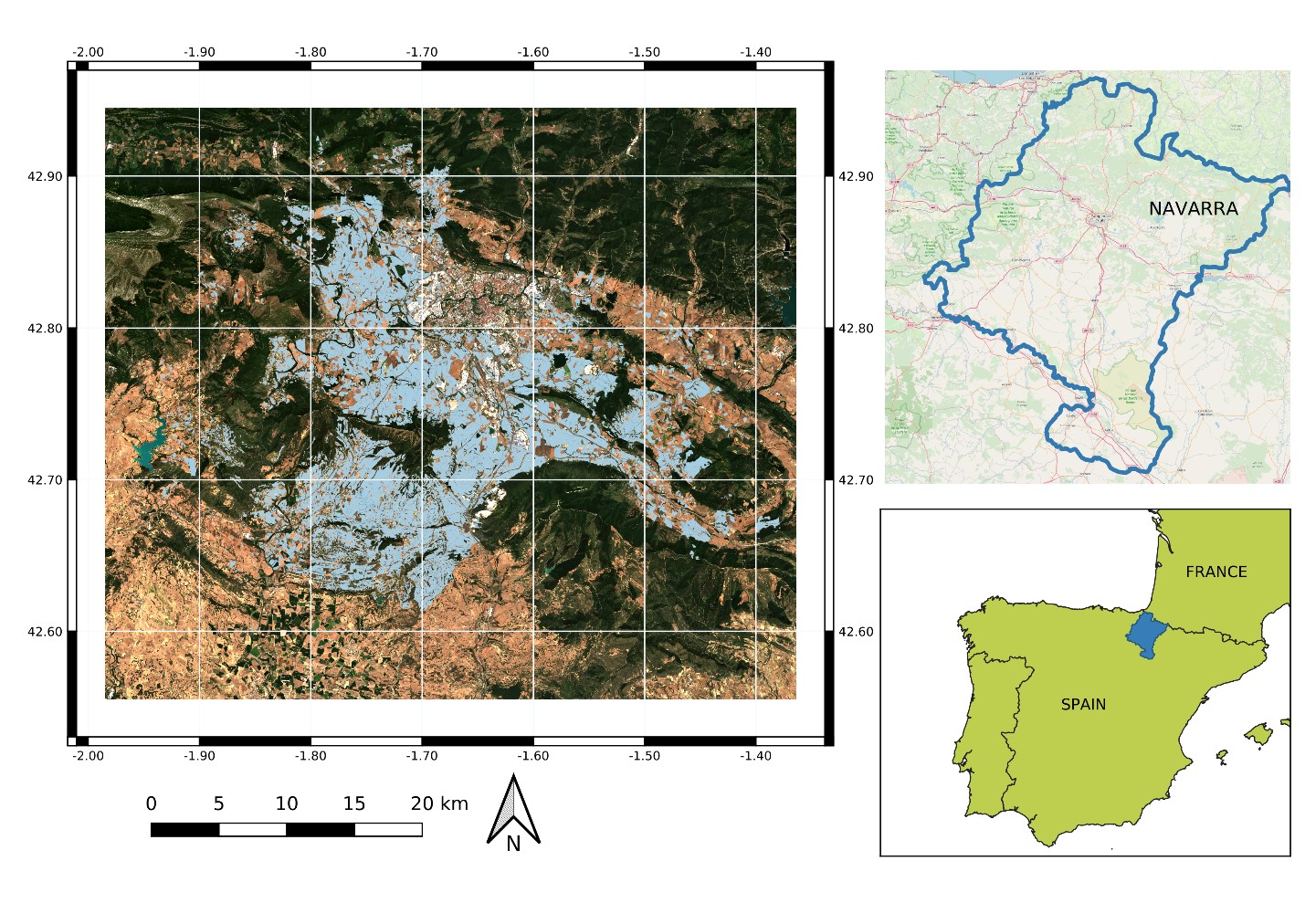}
    \caption{Study area located in northeastern Spain and specifically in the Navarra district.The parcels of interest are shown in light blue color.}
    \label{fig:area_of_interest}
\end{figure}

\begin{figure}[!ht]
    \centering
    \includegraphics[width=\linewidth]{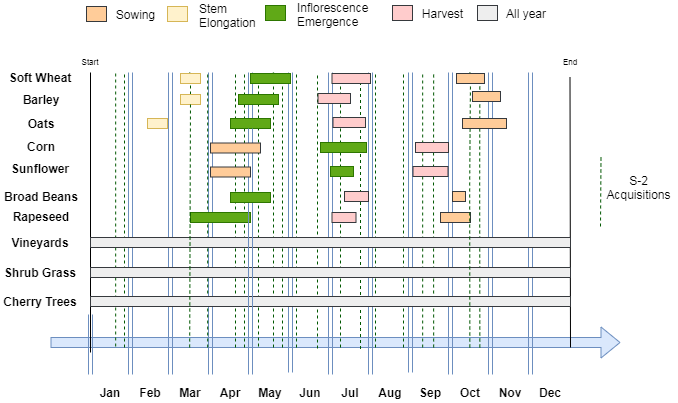}
    \caption{Timeline of the growth cycle of major crops in Navarra, reworked from ~\cite{sitokonstantinou2018scalable}, together with the acquisition dates of the Sentinel-2 images.}
    \label{fig:growth_cycle_timeline}
\end{figure}

%\subsection{Data Ingestion and Pre-processing}
\noindent The dataset for training the SVM classifier is based on the LPIS, which includes the parcel polygons in vector format and the associated farmer declaration for the 2018 CAP subsidy applications. The parcel polygons are used for segmenting the stack of Sentinel imagery to objects. The LPIS was provided by INTIA\footnote{https://www.intiasa.es/en/}, a public company, part of the Department of Rural Development, Environment and Local Administration of Spain. INTIA serves the role of paying agency for the district of Navarra, performing all CAP compliance inspections for the area. INTIA has additionally provided the timeline of growth for the major crops of the area. Figure \ref{fig:growth_cycle_timeline} illustrates the acquisitions of Sentinel-2 images, spanning over the entirety of the various crop cycles. \\

\noindent The feature space used for the crop classification includes the Sentinel-2 images for the acquisitions depicted in Figure \ref{fig:growth_cycle_timeline}. The acquisitions have been selected to have minimal cloud coverage over the AOI. All spectral bands, except B09 and B10, were used, along with the VIs NDVI, NDWI and PSRI. In this study, NDWI is used as defined by Gao (1996) \cite{gao1996ndwi}. Sentinel-2 images are atmospherically corrected to Bottom Of Atmosphere (BOA) reflectances using the Sen2Cor tool, and all bands are resampled to 10 m spatial resolution.The feature space comprises of parcel entities described by the mean value, for all features, of the pixels that fall within their LPIS boundaries.\\

\begin{algorithm}
\caption{Smart Sampling}
\label{alg:alg1}
\textbf{Input:} $X_{train}=\{(x_i,d_i),x_i \in R^l, d_i\in \{1,2,\ldots, m\}, i=1, \ldots N\}$, $X_{test}=\{(x_i,d_i),x_i \in R^l, d_i\in \{1,2,\ldots, m\}, i=1, \ldots, M\}$, 
$acqDate = \{acqDate_k, k = 1,\ldots, \lambda, \ldots, A \}$, number of iterations $t = 0 $, the acquisition number that bounds that starting feature space $\lambda$,
the date the algorithm is executed $currentDate$, persistence threshold for each iteration $Pt=0$, number of misclassifications $mis = $\{$mis_i, i = 1,...M\}$\vspace*{2mm}\\ 
\textbf{Output:} Alarms for potential breaches of compliance\\ 
\begin{algorithmic}
\WHILE {$acqDate_t \leq currentDate$}
\STATE $X_{train}(t)=\{(x_i,d_i),x_i \in R^{(acqDate_{(t+\lambda)} \cdot f)}, d_i\in \{1,2,..m\}, i=1, ... N\}$\\
\STATE $X_{test}(t)=\{(x_i,d_i),x_i \in R^{(acqDate_{(t+\lambda)} \cdot f)}, d_i\in \{1,2,..m\}, i=1, ... M\}$\\
  \vspace*{2mm}\STATE $h(\overrightarrow{x}) = \sum \alpha_i y_i (\overrightarrow{x_i} \cdot \overrightarrow{x} + b )^2 + b$ \textit{(Train SVM)}
   \vspace*{2mm}\STATE $\pi_d$ = $P(d | X_{test(\tau)}) = \frac{1}{(1 + {\rm e}^(A * 
   f(X_{test(t)}) + B) }$ \textit{(Calculate SVM Scores)}
 
% \vspace*{2mm}\STATE $score = $max($P(d | X_{test(t)}))-$max($P(d | X_{test(t)})- $max($P(d | X_{test(t)})))$
 \vspace*{2mm}\STATE $score =  \max{\pi_d} - \max(\pi_d - \max{\pi_d)} $

 \vspace*{2mm}\IF{$t$ mod 2 = 1 }
    \STATE $P_t = P_t + 1$
 \ENDIF

 \STATE $alarms = $ \{\}
 
 \FOR{$i = $1 to $M$}
   
    \IF{not $score \geq threshold$} 
    \STATE continue \hspace*{7mm} (Bypass unreliable decisions)
    \ENDIF
        \IF{parcel is misclassfied}
            \STATE $mis_i = mis_i$ + 1
             
        \ENDIF
        \IF {$mis_i \geq P_t $ \AND $mis_i > 0 $}
              \STATE $alarms = alarms \cup i$
        \ENDIF
 \ENDFOR
 \STATE $t = t + 1$
  \ENDWHILE
 
\end{algorithmic}
\end{algorithm}
\noindent The proposed methodology is based on a traffic light system approach. Specifically, each parcel is categorized into four groups, each offering different levels of confidence. These categories comprise of the green, yellow, red and unreliable classes, indicating high to low levels of confidence, in that order. The categorization of each parcel is based on the  difference  between the two highest SVM scores. This study focuses predominantly on the green category, namely the decision of highest confidence. It is on those most confident samples that we then record the mismatches of model predictions and the farmer declarations. An alarm mechanism is then introduced, identifying the green parcels that have been systematically misclassified (mismatch of declaration and prediction) during the cultivating season. \\

\noindent In the proposed algorithm (Algorithm \ref{alg:alg1}), the alarms of potential false declarations are detected for any time instance throughout the year, with variable accuracy considering the satellite imagery available to date. The $X_{train}$ and $X_{test}$ are the training and test feature spaces, respectively. The feature spaces are dynamically populated with all new acquisitions. Therefore, when the algorithm is executed ($currentDate$), it uses the up to date feature spaces as input; containing imagery until the latest available acquisition within the cultivation season (Equation \ref{eq1_2}).
\begin{equation}\label{eq1_2}
    acqDate = acqDate_k, k = 1, \ldots, \lambda, \ldots, A  
\end{equation}
where $A$ is the index to the latest acquisition prior to $currentDate$ and $\lambda$ is the index to the acquisition that defines the starting feature space, early in the year. \\

\noindent The algorithm iterates $A$-$\lambda$ times, each time recording the misclassifications ($mis$). Misclassifications, in this context and as previously stated, refer to the mismatch between the SVM model's prediction and the farmers' declaration. For each iteration, a 2\textsuperscript{nd} order polynomial SVM model is trained based on $X_{train}(t)$ (Equation \ref{eq21}) and the SVM scores are computed after applying the model to $X_{test}(t)$ (Equation \ref{eq3}). 
$X_{train}(t)$ and $X_{test}(t)$ are the training and test data for each iteration, and $f$ is the number of individual features for each acquisition. The farmer declarations, as part of the annual subsidy application for the CAP, are used  for labeling the parcels and thus training the model. A stratified random split was performed to split the samples into 30\% and 70\% subsets for $X_{train}(t)$ and $X_{test}(t)$ respectively. This amounts to 2,716 parcels, which have been used for training. All classification metrics that are presented in later sections have been averaged for 20 random splits of different seeds.The percentage of training samples was ultimately set to 30\% after experimenting with larger datasets, which have provided only a marginal increase in performance. \\

\begin{align}
\label{eq21}
X_{train}(t)&=\{(x_i,d_i),x_i \in R^{(acqDate_{(t+\lambda)} \cdot f)}, d_i \in \{1,2,..m\}, \\ \nonumber
& i=1, ... N\}
\end{align}

\begin{align}
\label{eq3}
X_{test}(t)&=\{(x_i,d_i),x_i \in R^{(acqDate_{(t+\lambda)} \cdot f)}, d_i\in \{1,2,..m\}, \\ \nonumber
&i=1, ... M\}
\end{align}
where $x_{i}$ is the feature representation of the i-th out of N parcels, belonging to $R^{(acqDate_{(t+\lambda)} \cdot f)}$. The superscript represents the dimensionality of the feature space. In each iteration, starting with $t=0$, the feature space comprises of the starting feature space, i.e. the one including all features, $f$, of all acquisitions until $acqDate_{\lambda}$, plus all features, $f$, of acquisitions $acqDate_{\lambda+t}$ ; $d_{i}$ is the label for each parcel, ranging from 1 to $m$ (=10), representing the different crop types. \\

\noindent The difference between the two highest per class scores $P(d | X_{test(t)})$ for each sample is recorded as the overall score value (Equation \ref{eq42}) for the selection of the most confident decisions against a $threshold$. These parcels constitute the $green$ labels in the aforementioned defined traffic light system. We denote by $\pi_d$ the difference scores, i.e.: $\pi_d = P(d | X_{test(t)})$, so the overall score is given by:
\begin{equation}
\label{eq42}
score = \max{\pi_d} - \max(\pi_d - \max{\pi_d)}
\end{equation}

\noindent The algorithm returns the misclassifications of the last iteration, namely the confident decisions of mismatch between the prediction and the declaration, which are classified as $alarms$. $alarms$ are the samples that have been found misclassified at least $P_t$ times, in all previous iterations. $P_t$ is varying based on the time within the year the algorithm is executed (Equation \ref{eq52}). 
\begin{equation}
\label{eq52}
 P_t = \sum_{t=1}^{A-\lambda}t\mod 2 
\end{equation}

\noindent Early classifications are characterized by limited reliability, as the imagery included the training datasets does not cover the entirety of crops’ growth cycle. For this reason Algorithm \ref{alg:alg2} can be optionally used to further refine the selected alarms.

\begin{algorithm}
\caption{Smart Sampling Fine Tuning}\label{alg:alg2}
\textbf{Input:} \textit{alarms} from Algorithm 1, $Y_{test}=\{(y_i \in {1,2,..m\}}$ the actual estimations of the classifier, $D_{test}=\{(d_i \in {1,2,..m\}}$ the declared labels\\
\textbf{Output:} Updated alarms for potential breaches of compliance\\ 
\begin{algorithmic}
\STATE $updatedAlarms$ = \{\} 
 \STATE $i = 1$
 \STATE $n$ =  size($alarms$)
    \WHILE{$i \leq n$}
        \IF{season of $y_i \neq$ season of $d_i$ } 
            \STATE $updatedAlarms = updatedAlarms \cup i$
        \ENDIF
    \STATE $i = i + 1$
\ENDWHILE
 
\end{algorithmic}
\end{algorithm}

\noindent Algorithm \ref{alg:alg2} uses the alarms of Algorithm \ref{alg:alg1} as input and returns an updated set of alarms. In order to increase the reliability of the smart sampling algorithm, we select alarms for which the crops are classified to a type of a completely different crop season class that then one of the type declared.

%\subsection{Semantic Annotation and Transformation into RDF}
%\subsubsection{Mapping and Integration}
\subsubsection{Mapping layer for semantic representation}
\label{mapping_integration}
\noindent An important aspect of the framework is the representation of the available information, e.g. crop classification results, as well as capturing of domain knowledge needed to further correlated results. For the former, we use GeoTriples to transform data into the RDF format, while for the later we developed a domain-specific ontology. The metadata and analysis results are integrated using geotagged information and analyzed using semantic queries.\\ 

\noindent As far as the domain ontology is concerned, it consists of three layers:

\begin{enumerate}[I]
\item \label{itm:first}	Data, which describe crop land taxonomy. Figure \ref{fig:croptype_ontology} shows the relationship between period, crop family, crop type and crop code classes. More specifically, each crop type (crop type classes) has a unique crop code (crop code property) and belongs to a specific crop family (family classes). Crop families are connected with the season that each crop thrives (period classes). Each crop thrives in a different season, while some crops seem to thrive in year-round basis. Same crop families may belong to different seasons when they represent different crop types.
\begin{figure*}[!ht]
    \centering
    \includegraphics[width=\textwidth]{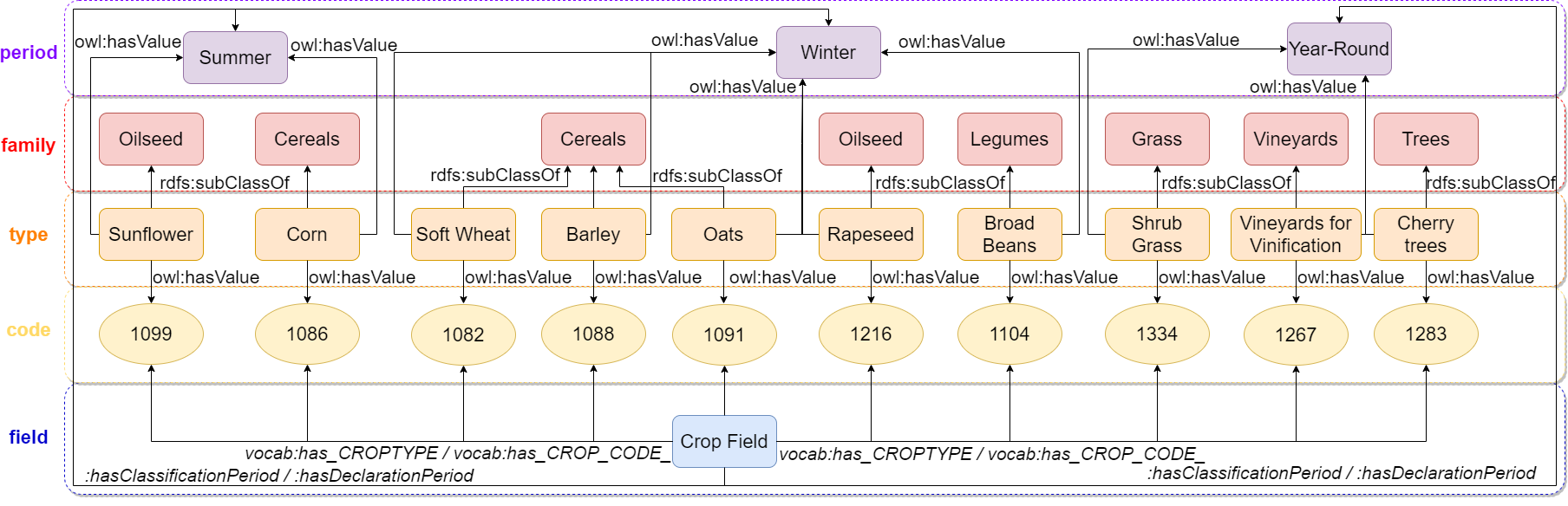}
    \caption{Crop type ontology}
    \label{fig:croptype_ontology}
\end{figure*}
\item \label{itm:second} The crop type classification data that contain a linking of metadata which describe field information like parcel identifier, geometry, slope, aspect and classification scores for all different crop types.
\item Data collected from OpenStreetMap, containing water and waterways information in geospatial format i.e. the geometries of hydrographic network objects.
\end{enumerate}

\noindent Figure \ref{fig:croptype_ontology} depicts the relationship between crop fields, periods, families, types and codes. Crop field is a class that corresponds to the crop fields that have been identified in a classification run. Each crop type is a subclass of a specific family and related to a specific crop code. All periods, families and types are different classes in the ontology. Crop codes are the main interconnection point between the crop fields and the characteristics of each crop type. OWL property restrictions have been identified to automatically detect the period of the crop code declaration or classification using the value of the crop code. These relationships are useful for extending the dataset described in layer \ref{itm:second} of the ontology (crop type classification data). For instance, in the example presented in Listing \ref{mapping_example} the crop code that has been identified by classification is 1082 and the crop field is associated with the classification period winter. The crop code that the farmer declared is 1334, whose classification period is year-round. OWL2 RL is used to make useful inferences and apply reasoning rules into crop type classification data. \\

\begin{lstlisting}[caption={Example OWL2 RL},captionpos=b,label={owl2rl_example},basicstyle=\scriptsize,linewidth=\columnwidth]
:SoftWeat
  a owl:Class ;
  rdfs:subClassOf :Cereals ;
  rdfs:subClassOf [
      a owl:Restriction ;
      owl:hasValue :winter ;
      owl:onProperty :hasClassificationPeriod ;
    ] ;
  owl:equivalentClass [
      a owl:Restriction ;
      owl:hasValue "1082" ;
      owl:onProperty vocab:has_CROPTYPE ;
    ] ;
.
\end{lstlisting}

\noindent In the example presented in Listing \ref{owl2rl_example}, the classification period and hyperclass have been automatically assigned to the crop field, taking advantage of the dynamics of OWL2 RL. T-Box reasoning (in the form of OWL 2 RL entailment rules supported by the GraphDB implementation) is applied to infer that soft wheat is a cereal, crop type “1082” corresponds to soft wheat and classification period “winter”, etc. More specifically, the rdfs:subClassOf has been used to assign the crop family values, while owl:Restrictions have been used to assign the value “winter” on the property hasClassificationPeriod  when the value of the property vocab:has\_CROPTYPE is “1082”. \\

\begin{lstlisting}[caption={Example of crop field information as described in RDF combining the crop type ontology (\ref{itm:first}) and the results of crop type classification (\ref{itm:second})},captionpos=b,label={mapping_example},basicstyle=\tiny,linewidth=\columnwidth]
@prefix :      <http://mklab.iti.gr/ontologies/croptypes/> .
@prefix owl:   <http://www.w3.org/2002/07/owl#> .
@prefix xsd:   <http://www.w3.org/2001/XMLSchema#> .
@prefix vocab: <http://example.com/ontology#> .
@prefix rdfs:  <http://www.w3.org/2000/01/rdf-schema#> .
@prefix fa:    <http://example.com/farmer_ontology#> .
@prefix map:   <http://example.com/#> .
@prefix geo: <http://www.opengis.net/ont/geosparql#> .

<http://example.com/parcels_classification_wscores_v3/Geometry/14356>
        a       geo:Geometry ;
        geo:asWKT
                "<http://www.opengis.net/def/crs/EPSG/0/4326> 
                MULTIPOLYGON (((-1.6927107344688275 42.649935712372795, 
                ..., -1.692387508984902  42.65003808978244,
                -1.692421288894873 42.65001269177615, 
                -1.6925231315390605 42.649985701208124,
                -1.6927107344688275 42.649935712372795))
                )"^^geo:wktLiteral .

<http://example.com/parcels_classification_wscores_v3/id/14356>
        a                         vocab:parcels_classification_wscores_v3 ;
        fa:hasOwner               fa:farmid5 ;
        vocab:has_ASPECT          5.33798E1 ;
        vocab:has_CROPTYPE        "1082" ;
        vocab:has_CROP_CODE_      1334 ;
        vocab:has_ID              773280 ;
        vocab:has_SLOPE           5.0291E0 ;
        vocab:has_scores_t_1      7.511E-3 ;
        vocab:has_scores_t_2      4.33932E-1 ;
        vocab:has_scores_t_3      5.4892E-2 ;
        vocab:has_scores_t_4      1.41E-2 ;
        vocab:has_scores_t_5      4.9191E-2 ;
        vocab:has_scores_t_6      3.6199E-2 ;
        vocab:has_scores_t_7      2.17041E-1 ;
        vocab:has_scores_t_8      2.802E-2 ;
        vocab:has_scores_t_9      3.8657E-2 ;
        vocab:has_scores_typ      1.20457E-1 ;
        :hasClassificationPeriod  :winter ;
        :hasDeclarationPeriod     :year-round ;
        geo:hasGeometry
                <http://example.com/parcels_classification_wscores_v3/Geometry/14356> .
\end{lstlisting}

\noindent The  transformation of crop type classification data and data collected from OpenStreetMap into the RDF model has been supported by GeoTriples. The tool accepts data in shapefile format and automatically produces an RDF Mapping Language (RML) file containing the rules that the RDF file should satisfy. Then, it produces the RDF serialisation that satisfies the RML rules. The Well-Known Text (WKT) representation of coordinate reference systems standard has been reused to capture location-related information.\\ %The reason behind this selection is that GraphDB offers a plugin for GeoSPARQL queries execution, while it offers high ease of use.%

\noindent For the classification results, GeoTriples tool is used multiple times to convert the results of each run into RDF format. Each file has a unique name and because of that, new instances are created in the Knowledge Base. In the end of this procedure, the Knowledge Base contains many different instances of the same parcel having in common the parcel identifier. In such way we keep crop type classification data of past runs in the Knowledge Base, which can be aligned with new classification data.\\

% Additional analysis is performed in the metadata, to integrate the data that come from different sources. Geolocation information is the base of the integration between them. In order to exploit this data need to be represented under the same geographic system representation standards. Other analysis is achieved in terms of data preprocessing to easily tag the polygons that represent hydrographic network objects and polygons formulation. 

%Additional analysis is performed in the metadata, to randomly assign fields to farmers. Each farmer may own from 1 to 8 fields.

%\subsubsection{Semantic enrichment} 
\subsubsection{Data ingestion and querying layer for storage and semantic enrichment}
\label{sem_enrichment}
\noindent Semantic enrichment aims at interconnecting and further enriching the contents of the generated knowledge graphs, applying semantic rules. The focus is given on improving the smart sampling methodology, executing a set of queries (rules) to improve the selection process of OTSCs. In the following, we describe the specifics of the approach, presenting the defined rules. With every new image acquisition a new crop classification is performed and the  classification results are dynamically populating the knowledge base.\\
%Each time a new dataset containing pre-classified results is inserted into the knowledge base, a semantic query retrieves all parcel entities that exist in this dataset.% %The following queries are running per parcel, given the parcel instance identifier that comes from the previous query in the queries that is needed.%

\noindent Procedural code is used to run semantic queries in sequence and pass the needed values from past queries (parcel, value1, value2) into next ones (Listings \ref{retrieve_parcels} to \ref{enrichment5}). The reason for this decision is that SPARQL lacks in terms of arguments saving or passing into next queries and these calculations are better expressed using many SPARQL queries, improving the execution time. The queries that are presented in this section are running for each parcel instance. The parcel instances are retrieved by the query depicted in Listing \ref{retrieve_parcels}. A parcel instance (described as {\footnotesize$<parcel>$}) may be for example: http://example.com/parcels\_classification\_wscores\_v8/id/1109. The first step (Listing \ref{enrichment1}) is to retrieve the two highest classification scores among the 10 different crop types, as these are defined in the ontology (Figure \ref{fig:croptype_ontology}). This is done in order to compute the $score$ value in Algorithm \ref{alg:alg1}. This process takes place for each of the different classification instances in the knowledge base. In this query, typical SPARQL functions are used, such as BIND, to group different type results under the same variable, and ORDER BY to arrange the results in descending order. 
%\begin{figure}
%\scriptsize{
%\begin{verbatim}
%PREFIX vocab: <http://example.com/ontology#>
%SELECT * WHERE { 
%    ?parcel a vocab:parcels_classification_wscores_v8.
%} 
%\end{verbatim}}
%\caption{Semantic query to retrieve all parcel instances for the latest classification run}
%Retrieve all parcel instances%}
%\label{retrieve_parcels}
%\end{figure}

\begin{lstlisting}[caption={Semantic query to retrieve all parcel instances for the latest classification run},captionpos=b,label={retrieve_parcels},basicstyle=\scriptsize]
PREFIX vocab: <http://example.com/ontology#>
SELECT * WHERE { 
  ?parcel a vocab:parcels_classification_wscores_v8.
} 
\end{lstlisting}

%\begin{figure}
%\scriptsize{
%\begin{verbatim}
%PREFIX vocab: <http://example.com/ontology#>
%select ?max_types where { 
 %   <parcel> vocab:has_scores_t_1 ?type_1. 
 %   <parcel> vocab:has_scores_t_2 ?type_2. 
 %   <parcel> vocab:has_scores_t_3 ?type_3. 
 %   <parcel> vocab:has_scores_t_4 ?type_4. 
 %   <parcel> vocab:has_scores_t_5 ?type_5. 
 %   <parcel> vocab:has_scores_t_6 ?type_6. 
 %   <parcel> vocab:has_scores_t_7 ?type_7. 
 %   <parcel> vocab:has_scores_t_8 ?type_8. 
 %   <parcel> vocab:has_scores_t_9 ?type_9. 
 %   <parcel> vocab:has_scores_typ ?type_10.
 %   
 %   {BIND(?type_1 as ?max_types)}
 %   union
 %   {BIND(?type_2 as ?max_types)}
 %   union
 %   {BIND(?type_3 as ?max_types)}
 %   union
 %   {BIND(?type_4 as ?max_types)}
 %   union
 %   {BIND(?type_5 as ?max_types)}
 %   union
 %   {BIND(?type_6 as ?max_types)}
 %   union
 %   {BIND(?type_7 as ?max_types)}
 %   union
 %   {BIND(?type_8 as ?max_types)}
 %   union
 %   {BIND(?type_9 as ?max_types)}
 %   union
 %   {BIND(?type_10 as ?max_types)}
%} 
%ORDER BY DESC (?max_types)
%LIMIT 2
%\end{verbatim}}
%\caption{Semantic query to retrieve the two highest classification score values per parcel}
%Retrieve the two highest  score values per parcel}
%\label{enrichment1}
%\end{figure}

\begin{lstlisting}[caption={Semantic query to retrieve the two highest classification score values per parcel},captionpos=b,label={enrichment1},basicstyle=\scriptsize]
PREFIX vocab: <http://example.com/ontology#>
select ?max_types where { 
    <parcel> vocab:has_scores_t_1 ?type_1. 
    <parcel> vocab:has_scores_t_2 ?type_2. 
    <parcel> vocab:has_scores_t_3 ?type_3. 
    <parcel> vocab:has_scores_t_4 ?type_4. 
    <parcel> vocab:has_scores_t_5 ?type_5. 
    <parcel> vocab:has_scores_t_6 ?type_6. 
    <parcel> vocab:has_scores_t_7 ?type_7. 
    <parcel> vocab:has_scores_t_8 ?type_8. 
    <parcel> vocab:has_scores_t_9 ?type_9. 
    <parcel> vocab:has_scores_typ ?type_10.
    {BIND(?type_1 as ?max_types)}
    union
    {BIND(?type_2 as ?max_types)}
    union
    {BIND(?type_3 as ?max_types)}
    union
    {BIND(?type_4 as ?max_types)}
    union
    {BIND(?type_5 as ?max_types)}
    union
    {BIND(?type_6 as ?max_types)}
    union
    {BIND(?type_7 as ?max_types)}
    union
    {BIND(?type_8 as ?max_types)}
    union
    {BIND(?type_9 as ?max_types)}
    union
    {BIND(?type_10 as ?max_types)}
} 
ORDER BY DESC (?max_types)
LIMIT 2
\end{lstlisting}

\noindent The difference between the two largest values (Listing \ref{enrichment2}) is calculated using the results of the previous query (value1, value2). If the difference is bigger than a specific threshold, the query in Listing \ref{enrichment3} is used to mark the parcel as ``green'', i.e. parcels with a high confidence that the prediction is correct.\\

%\begin{figure}
%\scriptsize{
%\begin{verbatim}
%PREFIX xsd: <http://www.w3.org/2001/XMLSchema#>
%SELECT * where { 
%    BIND (xsd:double(value1)-xsd:double(value2) 
 %       AS ?result)
 %   FILTER (?result>"0.5"^^xsd:double)    		    			
%}
%\end{verbatim}}
%\caption{Semantic query to detect the parcels that the difference from the two highest score values is above threshold}
%Checking if result is above threshold}
%\label{enrichment2}
%\end{figure}

\begin{lstlisting}[caption={Semantic query to detect the parcels that the difference from the two highest score values is above threshold},captionpos=b,label={enrichment2},basicstyle=\scriptsize]
PREFIX xsd: <http://www.w3.org/2001/XMLSchema#>
SELECT * where { 
    BIND (xsd:double(value1)-xsd:double(value2) 
        AS ?result)
    FILTER (?result>"0.5"^^xsd:double)    		    		
}
\end{lstlisting}

%\captionof{lstlisting}{Semantic query to detect the parcels that the difference from the two highest score values is above threshold}

%\begin{figure}
%\scriptsize{
%\begin{verbatim}
%PREFIX vocab: <http://example.com/ontology#>
%PREFIX xsd: <http://www.w3.org/2001/XMLSchema#>
%PREFIX map: <http://example.com/#>
%INSERT {
%    <parcel> map:risk "green parcel". 
%} WHERE { 
%    <parcel> ?p ?o. 	    	    		    			
%}
%\end{verbatim}}
%\caption{Semantic query to mark green parcels that satisfy the conditions of the two previous queries}
%Marking green parcels}
%\label{enrichment3}
%\end{figure}

\begin{lstlisting}[caption={Semantic query to mark green parcels that satisfy the conditions of the two previous queries},captionpos=b,label={enrichment3},basicstyle=\scriptsize]
PREFIX vocab: <http://example.com/ontology#>
PREFIX xsd: <http://www.w3.org/2001/XMLSchema#>
PREFIX map: <http://example.com/#>
INSERT {
    <parcel> map:risk "green parcel". 
} WHERE { 
    <parcel> ?p ?o. 	    	    		    			
}
\end{lstlisting}

\noindent In addition, each time a green parcel is identified, the query in Listing \ref{enrichment4} marks the parcels that have been incorrectly classified. In the end, the query in Listing \ref{enrichment5} marks the parcels that the season of the declaration does not agree with the season of the classification, based on the domain ontology (Figure \ref{fig:croptype_ontology}). SPARQL INSERT function is used to enrich the parcels with ``misclassification'', ``green parcel'' and ``same season'' information. All in all, semantic enrichment tries to enrich smart sampling, taking into account the provided classification results and domain knowledge about crop types, so as to detect parcels that have a  high probability to have a false crop type declaration by the farmers.\\

%\begin{figure}
%\scriptsize{
%\begin{myverb}
%PREFIX vocab: <http://example.com/ontology#>
%PREFIX xsd: <http://www.w3.org/2001/XMLSchema#>
%PREFIX map: <http://example.com/#> 
%INSERT { 
%    <parcel> map:misclassification "misclassification". 
%} WHERE { 
%    <parcel> map:risk "green parcel".    
%    <parcel> vocab:has_CROP_CODE_ ?decl . 
%    <parcel> vocab:has_CROPTYPE ?class . 
%	    	    	    		    	 		
%    FILTER(?decl!=xsd:integer(?class))    		    	        
%}
%\end{myverb}}
  
%\caption{Semantic query to mark parcels as misclassified when parcels have been marked as green and declaration does not agree with classification}
%Marking misclassified parcels}
%\label{enrichment4}
%\end{figure}
\begin{lstlisting}[caption={Semantic query to mark parcels as misclassified when parcels have been marked as green and declaration does not agree with classification},captionpos=b,label={enrichment4},basicstyle=\scriptsize]
PREFIX vocab: <http://example.com/ontology#>
PREFIX xsd: <http://www.w3.org/2001/XMLSchema#>
PREFIX map: <http://example.com/#> 
INSERT { 
    <parcel> map:misclassification "misclassification". 
} WHERE { 
    <parcel> map:risk "green parcel".    
    <parcel> vocab:has_CROP_CODE_ ?decl . 
    <parcel> vocab:has_CROPTYPE ?class . 
    FILTER(?decl!=xsd:integer(?class))    		    	        
}
\end{lstlisting}

%

%\begin{figure}
%\footnotesize {
%\begin{verbatim}
%PREFIX vocab: <http://example.com/ontology#>
%PREFIX xsd: <http://www.w3.org/2001/XMLSchema#>
%PREFIX map: <http://example.com/#>
%PREFIX owl: <http://www.w3.org/2002/07/owl#>
%PREFIX rdfs: <http://www.w3.org/2000/01/rdf-schema#>
%PREFIX : <http://mklab.iti.gr/ontologies/croptypes/>
%INSERT {
%    <parcel> map:same_season "false". 
%} WHERE { 	
%    <parcel> vocab:has_ID ?id.
%    <parcel> vocab:has_CROP_CODE_ ?decl .
%    <parcel> vocab:has_CROPTYPE ?class .

%    ?type a owl:Class.
%    ?type rdfs:subClassOf ?object1 .
%    ?object1 owl:hasValue ?code.
%    ?type rdfs:subClassOf ?object2 .
%    ?object2 owl:hasValue ?period1.
%    ?period1 a :Period.
%    FILTER regex (str(?code),str(?decl))
    
%    ?type2 a owl:Class.
%    ?type2 rdfs:subClassOf ?object3 .
%    ?object3 owl:hasValue ?code2.
%    ?type2 rdfs:subClassOf ?object4 .
%    ?object4 owl:hasValue ?period2.
%    ?period2 a :Period.
%    FILTER regex (str(?code2),?class)
    
%    FILTER (?period1!=?period2)
%}
%\end{verbatim}}
%\caption{Marking parcels where declaration and classification belong to different seasons}
%\label{enrichment5}
%\end{figure}%

\begin{lstlisting}[caption={Marking parcels where declaration and classification belong to different seasons},captionpos=b,label={enrichment5},basicstyle=\scriptsize]
PREFIX vocab: <http://example.com/ontology#>
PREFIX xsd: <http://www.w3.org/2001/XMLSchema#>
PREFIX map: <http://example.com/#>
PREFIX owl: <http://www.w3.org/2002/07/owl#>
PREFIX rdfs: <http://www.w3.org/2000/01/rdf-schema#>
PREFIX : <http://mklab.iti.gr/ontologies/croptypes/>
INSERT {
    <parcel> map:same_season "false". 
} WHERE { 	
    <parcel> vocab:has_ID ?id.
    <parcel> vocab:has_CROP_CODE_ ?decl .
    <parcel> vocab:has_CROPTYPE ?class .

    ?type a owl:Class.
    ?type rdfs:subClassOf ?object1 .
    ?object1 owl:hasValue ?code.
    ?type rdfs:subClassOf ?object2 .
    ?object2 owl:hasValue ?period1.
    ?period1 a :Period.
    FILTER regex (str(?code),str(?decl))
    
    ?type2 a owl:Class.
    ?type2 rdfs:subClassOf ?object3 .
    ?object3 owl:hasValue ?code2.
    ?type2 rdfs:subClassOf ?object4 .
    ?object4 owl:hasValue ?period2.
    ?period2 a :Period.
    FILTER regex (str(?code2),?class)
    FILTER (?period1!=?period2)
}
\end{lstlisting}

%\subsection{Querying} 
\subsubsection{Query processing layer for interlinking spatial data}
\label{querying}
\noindent By capturing data in the RDF model space, we enable  spatial relationships-based querying and easy integration with other data sources, such as Linked Data. We demonstrate the query answering capabilities of the framework, as well as the ability to integrate  external datasets, by defining queries to detect possible non-compliance of the farmers according to the specified rules.

\subsubsection{Greening 1 requirement}

\noindent The query in Listing \ref{rule1} calculates the number of different crop types that a farmer cultivates and the total area of their farm. %As described in Section \ref{section:intro}, if the total area of the farm is more than \hl{30 hectares} and there are less than three different crop types cultivated, then there is a breach of compliance. Additional queries are applied to also check, for instance, the farmers that have less than two different crop types cultivated in a total farm area between 10 and 30 hectares using % \\
\noindent More specifically, the query detects a breach of compliance when the farmers own a total farm area between 10 and 30 hectares and cultivate at least two different crop types, but the dominant one is more than 75\% of the total farm area. The crop type as it is calculated in the SVM classification is expressed in the mapping via the vocab:has\_CROPTYPE property. In the RDF space, we use count(distinct ?ctype) as ?count to detect the number of the different crop type values that are detected for each farmer (GROUP BY ?owner).
% cultivate less than two different crop types and the area that a crop type covers is more than 75\% of the total area farm area that the specific farmer owns.%%
Additional queries are applied to also check, for instance, the farmers that have less than three different crop types cultivated in a total farm area of more than 30 hectares if:
%$HAVING \quad (?sum > 100000 \quad \&\& \quad ?sum <= 300000 \quad \&\& \quad ?count<2) $. %More information about this scenario are described in section \ref{greening1desc}. 
%$HAVING \quad \{ ?sum > 100000 \} \land \{?sum \leq 300000 \} \land \{ ?count<2 \} $
$ \{ sum > 300000 \} \land \{ count<3 \} \land \{ max_{croptype} [1]+max_{croptype} [2]>0.95*sum \}  $
as it has been described in Section \ref{section:intro}. The query (Listing \ref{rule1b}) further supports the greening 1 requirement by detecting the farmers that grow less than 3 different crop types in a total area of more than 30 hectares and farmers that grow less than 2 different crop types in a total area of 10 to 30 hectares. \\

%change text?%
%In the query that we present in this section we are calculating the number of different crop types that a farmer has into his possession and the total area that his fields cover. Such information are very important for checking the compliance of a farmer in the Crop Diversification scenario of Greening 1 requirement.

%The query in Figure \ref{rule1} implements such calculations. If the total area and the different crop types count are among specific values, defined in section \ref{greening1desc} the fields need to be checked because a possible non-compliance of the farmer is detected. To support different values checking an additional query is applied, by transforming the query of figure 10 using 
%$HAVING \quad (?sum > 100000 \quad \&\& \quad ?sum <= 300000 \quad \&\& \quad ?count<2) $.
%$HAVING \quad \{ ?sum > 100000 \} \land \{?sum \leq 300000 \} \land \{ ?count<2 \} $

% The area of the fields is calculated using GeoSPARQL ext:area function, taking advantage of the polygon points coordinates. The number of different crop types is calculated using the result of the SVM classification prediction of crop type on the field, while the sum of the fields area using the results of GeoSPARQL area calculations. All operations are implemented using SPARQL functions such as count, distinct and sum.

%

\begin{lstlisting}[caption={Semantic query to extract possible non-compliance in the Greening 1 requirement for the farmers owning a total farm area between 10 and 30 hectares and cultivating at least two different crop types, where the dominant one is more than 75\% of the total farm area},captionpos=b,label={rule1},basicstyle=\scriptsize]
PREFIX geo: <http://www.opengis.net/ont/geosparql#>
PREFIX ext: <http://rdf.useekm.com/ext#>
PREFIX fa: <http://example.com/farmer_ontology#>
PREFIX vocab:   <http://example.com/ontology#>
PREFIX xsd: <http://www.w3.org/2001/XMLSchema#>
select * where {
{
  select ?owner (sum(?area) as ?max) ?sum where { 
  	{ select ?owner (count(distinct ?ctype) as ?count) 
      	(sum(?area) as ?sum) where { 
      		?field fa:hasOwner ?owner .
             ?field vocab:has_CROPTYPE ?ctype .
             ?field geo:hasGeometry ?geo.
             ?geo geo:asWKT ?polygon.
             BIND(ext:area(?polygon) as ?area).
             ?owner a fa:Farmer . 
      	}
        GROUP BY ?owner
        HAVING (?sum > 100000 && ?sum <= 300000 && ?count>=2)
     }
     ?field fa:hasOwner ?owner .
     ?field vocab:has_CROPTYPE ?ctype .
     ?field geo:hasGeometry ?geo.
     ?geo geo:asWKT ?polygon.
     BIND(ext:area(?polygon) as ?area).
     ?owner a fa:Farmer .  
     }   
     GROUP BY ?owner ?ctype ?sum
     ORDER BY DESC (?max)    
    }
    FILTER (?max>0.75*?sum)
}
\end{lstlisting}

\begin{lstlisting}[caption={Semantic query to extract possible non-compliance in the Greening 1 requirement for the farmers that grow less than 3 different crop types in a total area of more than 30 hectares and farmers that grow less than 2 different crop types in a total area of 10 to 30 hectares},captionpos=b,label={rule1b},basicstyle=\scriptsize]
PREFIX geo: <http://www.opengis.net/ont/geosparql#>
PREFIX ext: <http://rdf.useekm.com/ext#>
PREFIX fa: <http://example.com/farmer_ontology#>
PREFIX vocab:   <http://example.com/ontology#>
PREFIX xsd: <http://www.w3.org/2001/XMLSchema#>
select ?owner (count(distinct ?ctype) as ?count)
	(sum(?area) as ?sum) where { 
    ?field fa:hasOwner ?owner .
    ?field vocab:has_CROPTYPE ?ctype .
    ?field geo:hasGeometry ?geo.
    ?geo geo:asWKT ?polygon.
    BIND(ext:area(?polygon) as ?area).
    ?owner a fa:Farmer . 
}
GROUP BY ?owner	
HAVING ((?sum > 300000 && ?count<3) || 
    (?sum > 100000 && ?sum <= 300000 && ?count<2))
\end{lstlisting}

\noindent The area of the fields is calculated using GeoSPARQL ext:area function, taking advantage of the polygon points coordinates. The number of different crop types is calculated using the result of the SVM classification prediction of crop type on the field, while the sum of the fields area using the results of GeoSPARQL area calculations. All operations are implemented using SPARQL functions such as count, distinct and sum.\\
%\begin{figure}
%\scriptsize{
%\begin{verbatim}
%PREFIX geo: <http://www.opengis.net/ont/geosparql#>
%PREFIX ext: <http://rdf.useekm.com/ext#>
%PREFIX fa: <http://example.com/farmer_ontology#>
%PREFIX vocab:   <http://example.com/ontology#>
%select ?owner (count(distinct ?ctype) as ?count) 
%    (sum(?area) as ?sum) where { 
    
%  ?field fa:hasOwner ?owner .
%  ?field vocab:has_CROPTYPE ?ctype .
%  ?field geo:hasGeometry ?geo.
%  ?geo geo:asWKT ?polygon.
%  BIND(ext:area(?polygon) as ?area).
%  ?owner a fa:Farmer .          
%}   
%GROUP BY ?owner
%HAVING (?sum > 300000 && ?count<3)
%\end{verbatim}}
%\caption{Semantic query to extract possible non-compliance in the Greening 1 requirement}
%\label{rule1}
%\end{figure}

\subsubsection{SMR 1 requirement}
\noindent Another important information for the end users is the distance of the parcels from the hydrographic network objects. This is in accordance with the requirements of SMR1, as described in Section \ref{section:intro}. In order to effectively identify the parcels susceptible to contribute nitrate-rich soil to nearby surface water, a filtering mechanism takes place, accounting for the slope and aspect of the parcel. Results are important for both farmers and paying agencies that want to check the compliance according to SMR 1 requirement.\\

\noindent Listing \ref{rule2} presents the query that lists the parcel instances where the distance from surface waters is lower than 10 meters, which is the buffer for organic manure application. Since both parcel (LPIS) and hydrographic network data\footnote{https://download.geofabrik.de/europe/spain.html} (data from OpenStreetMap) contain geospatial information (e.g. multipolygon, polygon etc.), the distance is calculated using geof:distance function of GeoSPARQL. The function accepts two geometries and calculates the shortest distance between any two points of the specified geometries. A filtering mechanism selects the fields according to their distance from the hydrographic network objects, slope and aspect as described in section \ref{impl_scen3}. The GeoSPARQL function ext:closestPoint is used to compute the closest points of each geometry compared to the other geometry. A string replacement pattern is used in order to retrieve the coordinates of the two points, which are utilized to compute the angle of the two points in degrees. GraphDB math functions are used to achieve such computations.\\

\begin{lstlisting}[caption={Semantic query to detect susceptible parcels according to the SMR 1 requirement taking into account the slope, the aspect, the angle and the distance of the parcel from hydrographic network objects},captionpos=b,label={rule2},basicstyle=\tiny]
PREFIX geof: <http://www.opengis.net/def/function/geosparql/>
PREFIX geo: <http://www.opengis.net/ont/geosparql#>
PREFIX uom: <http://www.opengis.net/def/uom/OGC/1.0/>
PREFIX map: <http://example.com/#>
PREFIX ogc: <http://www.opengis.net/ont/geosparql#>
PREFIX ofn: <http://www.ontotext.com/sparql/functions/>
PREFIX xsd: <http://www.w3.org/2001/XMLSchema#>
PREFIX ext: <http://rdf.useekm.com/ext#>
PREFIX vocab: <http://example.com/ontology#>
select ?parcel ?distance where { 
  BIND(<parcel> AS ?parcel).
  ?parcel a vocab:parcels_classification_wscores_v8.
  ?a vocab:has_ID ?id.
  ?parcel vocab:has_SLOPE ?slope. 
  FILTER(?slope>12)    
  ?parcel ogc:hasGeometry ?s.
   
  ?s geo:asWKT ?o .
  ?fGeom geo:asWKT ?fWKT .
  ?fGeom map:containsWater ?b. 
  FILTER (?fGeom != ?s).
  FILTER NOT EXISTS {
    ?s map:containsWater ?wa.      
  }
  BIND(geof:distance(?o, ?fWKT) as ?distance).
   
  FILTER(?distance<=0.1)
  BIND (ext:closestPoint(?o, ?fWKT) as ?clpoint1)
  BIND (ext:closestPoint(?fWKT, ?o) as ?clpoint2)
   
  BIND( replace( str(?clpoint1), "^[^0-9\\.-]*([-]?[0-9\\.]+)
    .*$", "$1" ) as ?long )
  BIND( replace( str(?clpoint1), "^.*
    ([-]?[0-9\\.]+)[^0-9\\.]*$", "$1" ) AS ?lat )
    
  BIND( replace( str(?clpoint2), "^[^0-9\\.-]*([-]?[0-9\\.]+)
    .*$", "$1" ) AS ?long2 )
  BIND( replace( str(?clpoint2), "^.*
    ([-]?[0-9\\.]+)[^0-9\\.]*$", "$1" ) AS ?lat2 )
  BIND (xsd:double(?lat)-xsd:double(?lat2) AS ?x)
  BIND (xsd:double(?long)-xsd:double(?long2) AS ?y)
    
  BIND (ofn:atan2(?x, ?y) AS ?atan2)
  ?parcel vocab:has_ASPECT ?aspect.
  BIND(IF(ofn:toDegrees(?atan2)>"0"^^xsd:double,
    ofn:toDegrees(?atan2),ofn:toDegrees(?atan2)+
    "360"^^xsd:double) AS ?toDegrees )
  BIND(IF(?aspect-45>0, ?aspect-45, ?aspect+360-45) AS ?min)
  BIND(IF(?aspect+45>360, ?aspect-360+45, ?aspect+45) AS ?max)
     
  FILTER(?min<?toDegrees && ?toDegrees<?max)
}
LIMIT 1
\end{lstlisting}
\subsubsection{Smart sampling}

%The query in Figure \ref{rule3} supports the retrieval of the results of section \ref{sem_enrichment} by extending the functionality to a temporal dimension to achieve a higher confidence. More specifically, the query takes advantage of older pre-classified data to ensure that the classification results for the specific parcel agree that there is a misclassification. The older pre-classified data cover the crop type classification results over many historical periods per parcel, which are used as input in the semantic reasoning layer to infer the optimal threshold. The threshold for this query is defined in Figure \ref{rule_thres} and is strongly related with the number of classification runs. The query detects the parcels that a misclassification has been detected in high confidence level using the past measurements. If for a parcel the results are higher than the season threshold, an alarm is set and the misclassification for this parcel is described as “persistent”. The query (Figure \ref{rule3}) contains an optional part, which is mandatory in months before July as described in Algorithm \ref{alg:alg2}. 

\noindent The query in Listing \ref{rule3} supports the retrieval of the results of section \ref{sem_enrichment}. When querying for the parcels to be inspected through OTSC, at any given time in the year, all past classification instances until that point are used. More specifically, the query takes advantage of all past classification decisions for each parcel to ensure that the prediction is indeed a misclassification. The threshold, above which a parcel is considered to be persistently misclassified, is defined in Listing \ref{rule_thres} and is dynamically updated given the different parcel classes, that correspond to different times of the year, which exist in the triple store. The different classification runs are saved as instances of vocab:parcels\_classification\_wscores\_v1 for the first run, vocab:parcels\_classification\_wscores\_v2 for the second, etc. In such way, using the FILTER function, we select the number of different runs that currently exist in the triple store by detecting the number of different classes that contain the word “parcels”. %given the instance within the year that the query is executed. %In the end, the query detects the parcels that a misclassification has been detected in high confidence level using the past measurements.%%
%If for a parcel the results are higher than the season threshold, an alarm is set and the misclassification for this parcel is described as “persistent”.%
The query in Listing \ref{rule3} retrieves all the parcels that have been misclassified in past classifications more times than the defined threshold. Then the associated season for the crop type prediction of the latest available classification is compared with the associated season type of the declarations using the results of Listing \ref{enrichment5}. If there is a disagreement, the parcel is selected as an alert and candidate for OTSC. This query further narrows the smart sampling filter and is recommended for query executions early in the year (months before July), when the classification results are less trustworthy. More information about this functionality is described in Algorithm \ref{alg:alg2}. \\

\begin{lstlisting}[caption={Semantic query to calculate the smart sampling threshold according to the number of different parcel classes, that correspond to different times of the year, which exist in the triplestore},captionpos=b,label={rule_thres},basicstyle=\scriptsize]
PREFIX xsd: <http://www.w3.org/2001/XMLSchema#>
PREFIX ofn: <http://www.ontotext.com/sparql/functions/>
    
SELECT ?final_threshold WHERE { 
    {
        SELECT (count(distinct ?parcel_class) as ?count)
        WHERE { 
            ?parcel_instance a ?parcel_class.
            FILTER contains(str(?parcel_class), "parcels")
        } 
    }
    BIND(IF(ofn:floorMod(?count,2)="1"^^xsd:int,
        ((?count+"1"^^xsd:int)/2)-"1"^^xsd:int, 
        ?count/2) AS ?threshold )
    BIND(IF(?count="1"^^xsd:double, "0"^^xsd:int,
        ?threshold) AS ?threshold )
    BIND(IF(?threshold>"5"^^xsd:double, "5"^^xsd:int,
        ?threshold) AS ?final_threshold )     
}
\end{lstlisting}

%\begin{figure}
%\scriptsize{
%\begin{myverb1}
%PREFIX map: <http://example.com/#>
%PREFIX vocab: <http://example.com/ontology#>
%PREFIX xsd: <http://www.w3.org/2001/XMLSchema#>
%SELECT ?id (count(distinct ?parcel) as ?count) WHERE { %

%    ?parcel vocab:has_ID ?id.
%    ?parcel map:misclassification ?o .
%       
%    ?latest_parcel a vocab:parcels_classification_wscores_v8.
%    ?latest_parcel vocab:has_ID ?id.
%    ?latest_parcel map:same_season "false".
%} 
%GROUP BY ?id
%HAVING (?count>final_threshold)
%ORDER BY desc (?count)
%\end{myverb1}}
%\caption{Semantic query to check persistent misclassifications using the smart sampling threshold and the number of past classification instances that have been misclassified provided that in the latest classification run the period of the farmer declaration does not agree with the period of classification}
%Checking persistent misclassifications
%\label{rule3}
%\end{figure}
\begin{lstlisting}[caption={Semantic query to check persistent misclassifications using the smart sampling threshold and the number of past classification instances that have been misclassified provided that in the latest classification run the period of the farmer declaration does not agree with the period of classification},captionpos=b,label={rule3},basicstyle=\scriptsize]
PREFIX map: <http://example.com/#>
PREFIX vocab: <http://example.com/ontology#>
PREFIX xsd: <http://www.w3.org/2001/XMLSchema#>
SELECT ?id (count(distinct ?parcel) as ?count) WHERE { 

    ?parcel vocab:has_ID ?id.
    ?parcel map:misclassification ?o .
       
    ?latest_parcel a vocab:parcels_classification_wscores_v8.
    ?latest_parcel vocab:has_ID ?id.
    ?latest_parcel map:same_season "false".
} 
GROUP BY ?id
HAVING (?count>final_threshold)
ORDER BY desc (?count)
\end{lstlisting}
\subsection{Experiments and Results} 
\label{area_of_interest}
\subsubsection{Hyperparameter optimization and performance of the SVM classifier}

\noindent The crop classification model was built using the SVC function from the scikit-learn library\footnote{https://scikit-learn.org/} of Python. 
The hyperparameters were optimized using grid search over a range of values for each parameter, using 5-fold cross-validation. The hyperparameter combination that was selected was the one that produced the highest overall accuracy. For the penalty parameter of the error term, $C$, we tested values within the range $[2^{-2},\ldots,2^{9}]$. For the kernel coefficient, namely the gamma parameter, the optimal value was examined within the range $[10^{-4},\ldots,10^1]$; while for the independent term of the kernel function, coef0, we searched the range $[10^{-3},\ldots,10^2]$. The best combination consisted of $C=4$,  $\gamma = 0.001$ and $coef_0 = 10$.
%The hyperparameters were optimized  with semi-exhaustive search over a range of values. The evaluation of the optimization was made based on the overall test accuracy of the various models. For the regularization parameter, or the penalty parameter of the error term, $C$ values within the range $[2^{-2},\ldots,2^{15}]$ were tested. Low values of $C$ resulted in relatively lower accuracy, as the cost constraint is relaxed. The parameter $C$ was finally set at 128. The $\gamma$ and $coef_0$ parameters represent the kernel coefficient and the independent term of the kernel function, respectively. The optimal value of the first was examined within the range $[10^{-5},\ldots,10^2]$. Lower values of $\gamma$, required longer time for the model to be trained, and led to poor results. The best performance, in terms of both time and accuracy, was achieved with value set to 0.1. The value of $coef0$ was examined in the range $[10^{-5},\ldots,10^3]$, and each of these values resulted in satisfying scores, however, the best accuracy was observed with the value set to 1. % \\

\noindent Finally, the SVC function includes a parameter called $class\_weight$, which was set to ``balanced''. Its impact is that it sets the parameter $C$ of each class $i$ to $class\_weight[i]*C$, where $class\_weight[i]$ is inversely proportional to class frequencies of the input data. This way the negative effects of an unbalanced dataset, such as the one used for this study, are ameliorated. \\

\noindent We evaluated our core classification model~\cite{sitokonstantinou2018scalable} for three different agricultural areas, in Greece, Lithuania and Spain, for which we had independent in-situ validation data. These areas present various challenges. In the case of Greece, the agricultural landscape is significantly fragmented, resulting in small and narrow parcels, occupied by mixed pixels. In Lithuania, there is extended cloud coverage throughout most of the season, resulting in sparse time-series of cloud free Sentinel-2 imagery. Ten, eleven and fourteen crop classes have been classified in the Spanish, Greek and Lithuanian cases, respectively. \\ 

\noindent Validated results were consolidated based on OTSCs that were performed by the respective paying agencies of the three AOIs, during the 2018 subsidy applications. In the case of the Spanish AOI, out of the 107 randomly selected parcels for inspection, 105 were classified correctly. In Greece, inspectors visited only parcels classified with high confidence, namely of high posterior probability for the classification decision, to crop types other than the one declared. These instances are considered as potential breaches of compliance. It was shown that 76 out of 85 inspected parcels were indeed wrongly declared and correctly classified by our model. Finally, in Lithuania, the validated dataset acquired through the inspections resulted in an overall accuracy of 76.2\% in late June and 80\% in late August out of 3,319 parcels inspected. The results revealed the dependencies of the crop classification model performance, on the percentage of truthful declarations, the cloud coverage and the parcel shape and size. \\

\noindent In the Spanish AOI for which we focus in this work all these dependencies were optimal, i.e. more than 97\% of truthful declarations, limited cloud coverage, and an average parcel size of 2 ha. Hence more than 90\% classification accuracy was achieved.\\

\noindent Our model evaluation analysis also revealed that classification decisions for larger parcels and parcels with straighter borders tend to have higher accuracy than smaller parcels or parcels with more irregularly-shaped boundaries. The parcel area is important since accuracy depends on the number of image pixels that fall within the parcel boundaries. Sentinel-2’s 10 m pixel size equates to 50 image pixels in 0.5 ha of land. Our analysis shows that having 50 pixels of information provides accurate results, whereas for smaller parcels the decision is both less confident, namely of lower SVM score, and less accurate (Tables \ref{table:spain_acc1} and \ref{table:spain_acc2}).

\begin{table}[!ht]
\caption{Accuracy of crop type classification for different parcel area ranges}
\centering
\tabcolsep=0.05cm
 \begin{tabular}{||c c c c ||} 
 \hline
 \multicolumn{4}{|c|}{Spain} \\
 \hline
  &  & Accuracy & \% parcels \\
 \hline
Large &	\textgreater 1ha &	96.70 &	44.85 \\
Medium &	0.35ha\textless x\textless 1ha &	94.44 &	23.85 \\
Small &	\textless 0.35ha	& 87.02 &	31.30 \\

 \hline
\end{tabular}
\label{table:spain_acc1}
\end{table}

\begin{table}[!ht]
\caption{Relationship between SVM scores and overall accuracy}
\centering
\tabcolsep=0.05cm
 \begin{tabular}{||c c c ||} 
 \hline
 \multicolumn{3}{|c|}{Spain} \\
 \hline
 SVM score & Accuracy & \% parcels \\
 \hline
\textgreater 0.85 &	97.88 &	77.43 \\
0.7\textless x\textless 0.85 &	89.46 &	10.27 \\
\textless 0.7	& 66.32 &	11.30 \\

 \hline
\end{tabular}

\label{table:spain_acc2}
\end{table}

\noindent In Table \ref{table:spain_acc2} is observed that there is a strong correlation between the SVM score and accuracy. Indeed, the subset of parcels with an SVM score larger than 0.85, achieves an overall accuracy of more than 97\%. On the other hand, the subset of parcels with SVM scores less than 0.7, achieves an overall accuracy of 66\%. 

%\subsubsection{Classification results for the satellite image analysis}

\subsubsection{Scenarios for the control of the CAP}
%For the following CAP monitoring scenarios,classification maps are dynamically produced for the Spanish AOI and  the 2018 inspection  year.Crop classifications are performed iteratively for progressively larger  feature  spaces.  The  first  crop  map  is  produced  using Sentinel-2 time-series spanning January until early May; and with every new acquisition, new classifications are performed

\paragraph{Scenario 1: Smart Sampling of OTSCs} %\hfill \break

%\textbf{Description}: 
\noindent The paying agency inspectors search for parcels of farmers that have potentially falsely declared the cultivated crop type. The parcels prone to noncompliance are dynamically provided, starting from late June until the end of the cultivation period, to allow the inspectors to better target their inspections.\\

\noindent\textbf{Farmer profile}: Dan is a farmer. He has into his possession a field containing barley, while he has declared that the field’s crop type is maize. 

\noindent\textbf{Analysis results}: His field has been selected for OTSC because the crop classification has classified the cultivated crop type as barley, with high confidence. Even though the classification is not particularly trustworthy, being performed in early June, the parcel is marked as high risk by the smart sampling algorithm, as the prediction belongs to a different season class, namely winter.

\noindent\textbf{Decision}: The paying agency inspector needs to check the field, as there is a strong possibility that the farmer has wrongly declared the cultivated crop type. \\

\paragraph{Scenario 2: Greening 1: Crop Diversification} \label{scen_2} %\hfill \break

%\textbf{Description}:
\noindent As previously stated, paying agencies need an automated system to detect non-complying farmers with respect to the Greening 1 requirement.

\noindent\textbf{Farmer profile}: Bob is a farmer. He has three fields in his possession 
covering an area of 27.4066 hectares. The total area of the fields that contain 
soft wheat is 27.1058 hectares, while for the crops that contain rapeseed the total area is 0.3008 hectares.

\noindent\textbf{Analysis results}: According to our analysis there seem to be two fields with soft wheat and one with rapeseed. The farmer does not comply with the rule because even though he cultivates two different crops, the total area of the soft wheat parcels exceeds the 75\% of the total area of the fields that the farmer has into his possession. 

\noindent\textbf{Decision}: The paying agency inspectors need to check these fields
because there is a strong possibility that the farmer is not
complying with the Greening 1 requirement.\\

\paragraph{Scenario 3: detecting parcels prone to non-compliance to the SMR-1 requirement} \label{scen_3} %\hfill \break 

%\textbf{Description}: 
%The water proximity of a field is significant information for environmental planners. Using these information can lead to environmental protection and water pollution reduction. Farmers may choose to declare a wrong crop type in their fields to hide their involvement in water pollution of a specific area. %
The paying agency inspectors need to establish the compliance to the SMR 1 defined buffer zones. In order to do that the distance from the parcels to nearby surface waters is calculated. The slope and aspect of the parcels are also taken into account in order to establish if there is an actual risk for runoff. \\

\noindent\textbf{Farmer profile}: Lucy is a farmer. She has into her possession a field whose distance from surface waters is almost 3 meters. The slope of the field is $15^{\circ}$, therefore of high runoff risk, and the aspect $251^{\circ}$, namely of western orientation. The aspect of the proximity line is $267^{\circ}$. 

\noindent\textbf{Analysis results}: Since the parcel appears to be very close to surface waters, within the SMR-1 buffer, and the difference of the parcel and proximity line aspects is within the range of potential runoff, the analysis marks it as high risk.

\noindent\textbf{Decision}: The paying agency inspectors need to monitor this parcel because there is a strong risk for nitrate-rich runoff to surface waters.\\

\subsubsection{Implementation of the CAP scenarios}
\noindent \paragraph{Implementation of Scenario 1: Smart Sampling}. \label{implementation_smart_sampling}
In this scenario, the potential of better targeted, smart OTSCs is explored, exploiting accurate crop classification results, early in the year. This is achieved through the proposed semantic enrichment of classification results and the pertinent smart sampling query. The idea is to provide alarms, starting from early summer, when declarations are usually received, and dynamically update those using progressively larger feature spaces that include new Sentinel-2 acquisitions.  \\

\noindent Table \ref{table:pa_ua} shows the Producer’s Accuracy (PA) and User’s Accuracy (UA) for the predictions of all crop classes. PA is the percentage of correctly classified parcels against the total number of parcels of a given class. On the other hand, UA is the percentage of correctly classified parcels for a given class against the total number of parcels classified to that class. The metrics in Table \ref{table:pa_ua} refer to the classification results produced using the entirety of acquisitions, 24/01/2018 to 21/10/2018, and have been averaged over 20 iterations of random training dataset splits. Additionally, the PA and UA results address the lowest level of crop taxonomy, as shown in Figure \ref{fig:croptype_ontology}. \\

\begin{table}[!ht]
\caption{PA and UA for the classification of 10 crops types using the full season time-series of Sentinel imagery}
\centering
 \begin{tabular}{||c c c ||} 
 \hline
 Class & UA (\%) &	PA (\%) \\ 
 \hline
 Soft wheat &	94.00	& 95.42 \\ 
 
 Maize &	94.98 &	94.36 \\
 
 Barley &	93.09 &	93.69 \\
 
 Oats &	92.52 &	88.40 \\
 
 Sunflower &	94.87 &	92.85 \\  
 
 Rapeseed &	95.34 &	92.08 \\  
 
 Broad beans &	94.84 &	89.40 \\  
  
 Shrub grass &	84.15 &	80.70 \\  
  
 Vineyards &	85.50 &	87.76 \\  
 
 Cherry trees &	83.22 &	81.98 \\ 
 \hline
\end{tabular}
\label{table:pa_ua}
\end{table}

\noindent Table \ref{table:pa_ua} shows excellent classification performance for most of the crop types. Specifically, it is observed that the classifier achieves PA and UA values of more than 80\% for all crop classes; while reaching values as high as 95\%. It is also worth noting the excellent performance for the classes oats, soft wheat and barley. These crop types belong to both the cereals and winter superclasses, having  similar spectral and phenological characteristics. Nontheless, the classifier appears to discriminate among them easily. Finally, the weakest performances are observed for the classes of shrub grass, vineyards and cherry trees. These are all year-long crop types, with no significant phenological characteristics to assist the classifier. Additionally, the shrub grass class, which is an ambiguous crop description, is characterized by diverse spectral characteristics from parcel to parcel. \\

\noindent The metrics shown in Table \ref{table:pa_ua} theoretically allow for the effective sampling of OTSCs. However, the results presented have been produced using images until the end of October, when the notion of smart sampling becomes obsolete, as the inspections would have preceded. For this reason, the crop classification is performed at multiple instances throughout the year, starting as early as May 4. As it would be expected, the classification results of reduced feature spaces, when executed early in the year, would achieve suboptimal results of low reliability. In this regard, per class scores are computed for each sample, using class membership probability estimates ~\cite{wu2004probability}. Therefore each sample is associated with ten different membership probabilities, as many as the crop classes involved. In order to select the most reliable decisions, the difference between the two highest scores for each sample is recorded ($score$ in Algorithm \ref{alg:alg1}). This is achieved via the semantic enrichment mechanism, as described in figures \ref{enrichment1} and \ref{enrichment2}. The following figure illustrates those differences in the form of a histogram, using 100 bins, for the entire dataset. The algorithm classifies most of the data with high reliability. The ones that are classified with low score are most likely crops that belong to the same family and some of them with similar spectral signatures are expected to be misclassified at some level.\\
\begin{figure}[!ht]
    \centering
    \includegraphics[width=\linewidth]{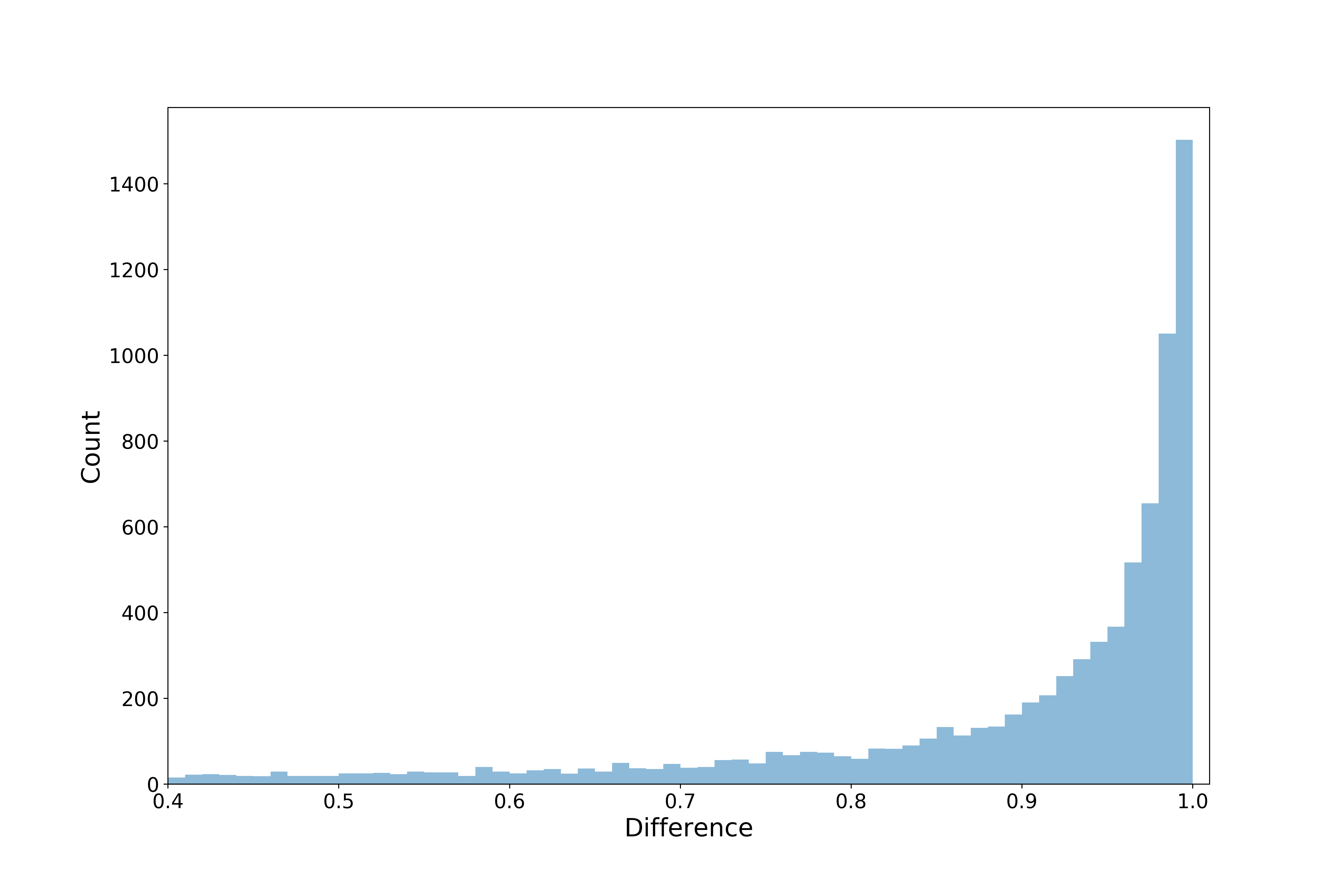}
    \caption{Histogram of differences between the two highest scores of each sample decision. The closer is the difference to 1.0 the lower the classification uncertainty.}
    \label{fig:histogram}
\end{figure}

\begin{figure}[!ht]
    \centering
    \includegraphics[width=\linewidth]{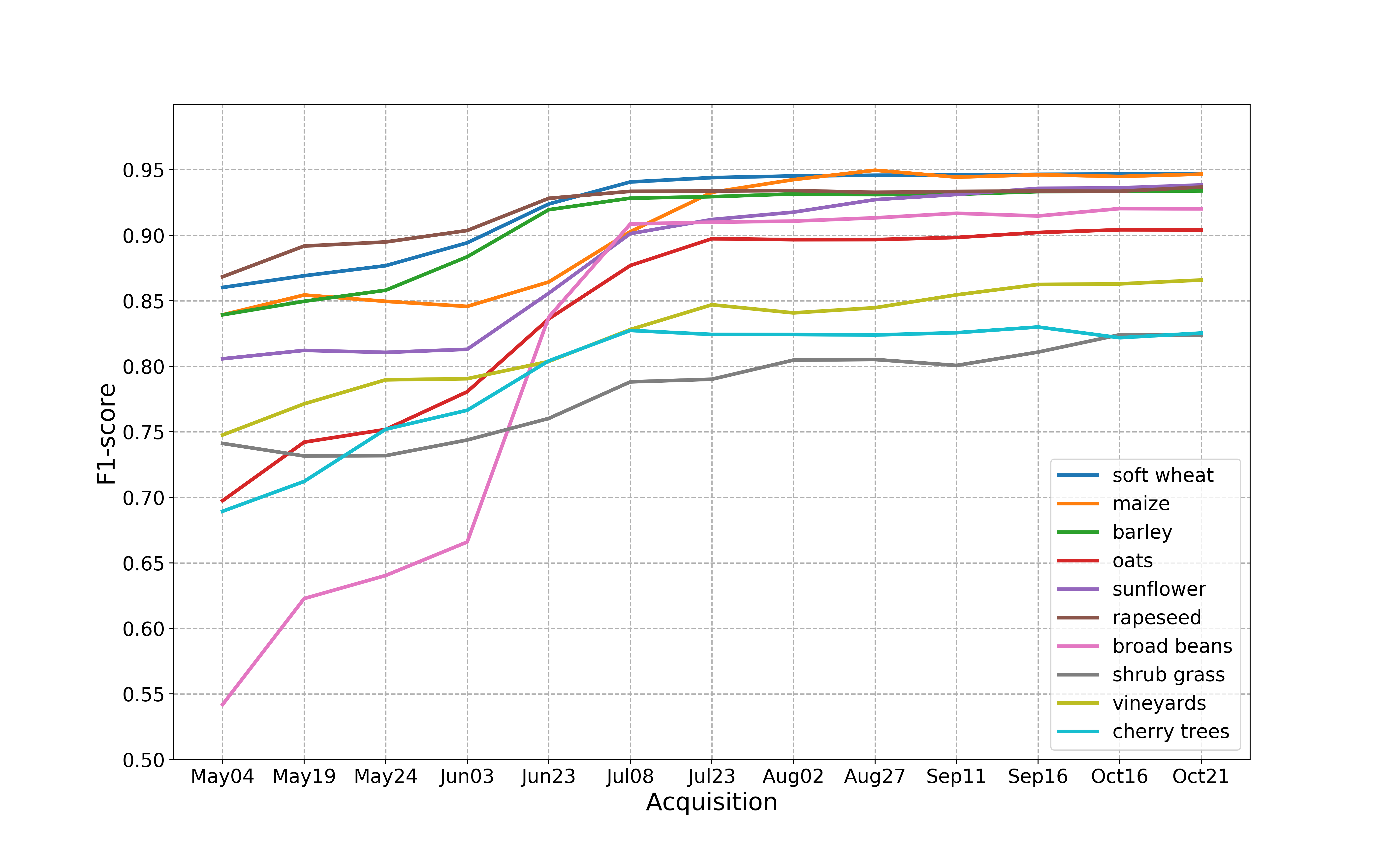}
    \caption{Evolution of F1-score for crop type classification with progressively larger feature spaces}
    \label{fig:evolution}
\end{figure}

\noindent The SVM algorithm is trained and tested for progressively larger feature spaces, whose classification results populate the knowledge base. The evolution of accuracy is shown in Figure \ref{fig:evolution}, using the F1-score metric ~\cite{sokolova2006beyond}. F1-score, defined as \(2 \times(\frac {PA \times UA}{PA + UA})\), providing an overall accuracy metric that accounts for both PA and UA. The first point on the x-axis of Figure \ref{fig:evolution} refers to the F1-score for the classification that uses images until the 4th of May. For each next run onward, features keep on populating the feature space, with every new acquisition. It is observed that the larger feature spaces result in better classification results. Specifically, feature spaces that include the July 23rd acquisition, appear to approach optimal F1-scores for most crop classes, creating a plateau on the evolution curve. The majority of these crops are harvested from late June to early July, and thus our model can separate the classes more comfortably, when features that cover this period are added to the feature space.   \\ 

% Accurate crop classification enables the monitoring of the CAP rules and allows for efficient decision making on the farmers’ compliance. Towards this direction, the inspected parcels were assorted into four reliability categories based on the traffic light system. More specifically, the categories are described as green, yellow, red and unreliable, based on varying thresholds of the difference between the two highest prediction probabilities, or SVM scores (Figure \ref{fig:evolution_misclassifications}). In order to pinpoint the parcels of the highest probability of noncompliance, the green category is investigated, for which the algorithm is assumed to predict with high reliability. 

\begin{figure}[!ht]
    \centering
    \includegraphics[width=\linewidth]{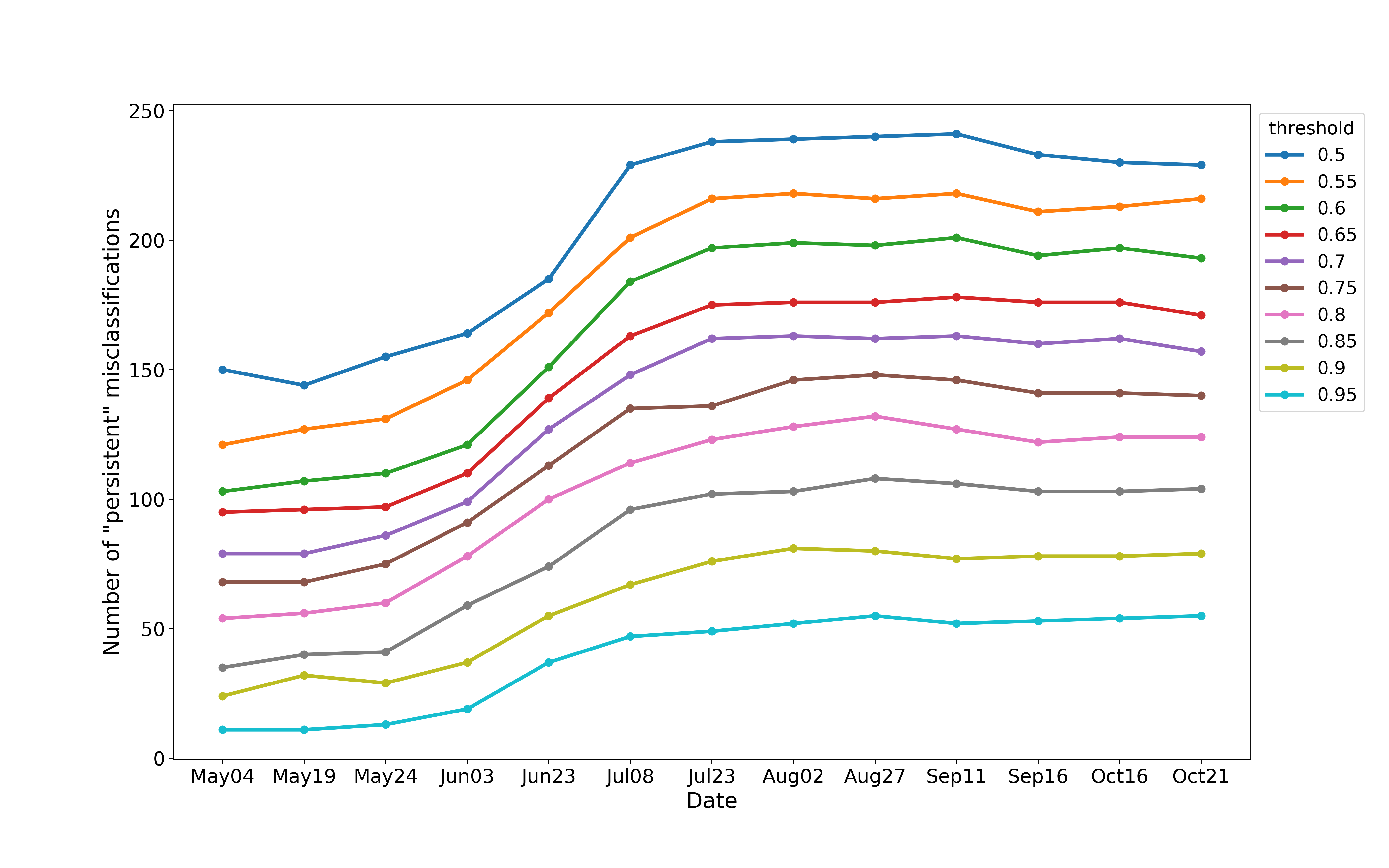}
    \caption{Evolution of the number of constant persistent misclassifications for a varying threshold of the green category}
    \label{fig:evolution_misclassifications}
\end{figure}

\noindent Accurate crop classification enables the monitoring of the CAP rules and allows for efficient decision making on the farmers’ compliance. Towards this direction, the inspected parcels were assorted based on the previously described traffic light system.Two highest prediction probabilities, or SVM scores (Figure \ref{fig:evolution_misclassifications}) are taken into consideration to pinpoint the parcels of the highest probability of noncompliance, for which the algorithm is assumed to predict with high reliability.\\

\begin{figure}[!ht]
    \centering
    \includegraphics[width=\linewidth]{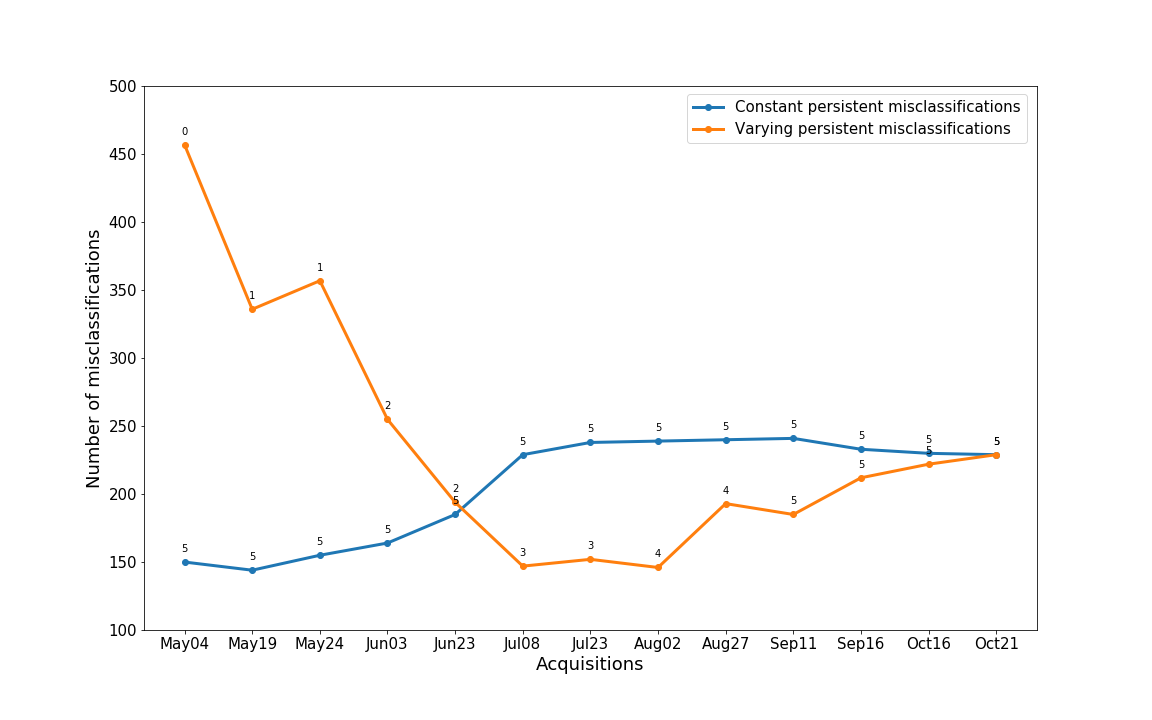}
    \caption{The evolution of the number of misclassifications under constant (blue line) and varying thresholds (orange line). The varying threshold increases by one every two image acquisitions. The thresholds define which parcels are the misclassifications for each classification instance looking at the number of times they were found misclassified in previous runs.}
    \label{fig:evolution_misclassifications2}
\end{figure}

\noindent At first, for each run with a different feature space in the knowledge base, the set of green misclassifications is recorded, namely the reliable instances for which the prediction does not match the declaration. Then, the set of persistent green misclassifications is computed using the semantic query in Figure \ref{rule3}. Persistence refers to the number of times a sample has been found misclassified in the different classification iterations. In this study, there are two different thresholds of persistence. First, if a sample has been misclassified more than five times, in all different runs, from May to late October, then it is considered to be a validated alert. This is called constant persistent misclassification, as shown in Figure \ref{fig:evolution_misclassifications2}, and is assumed to function as the validation dataset, against which performance metrics are computed.\\

\noindent Figure \ref{fig:evolution_misclassifications} presents the number of constant persistent misclassifications relative to the increasing number of features, for different threshold values of what is considered a green parcel. The latter translates to varying values of the difference between the two highest scores for each sample. As expected, lower thresholds lead to higher number of misclassifications.  It can also be observed that for each threshold that was tested, the pattern of the plot line is the same. The number of misclassifications is constant during May, then it presents an increase at the start of June, which begins to stabilize towards the start of August. It can be concluded that it is difficult for the algorithm to identify truly mislabeled data, really early in the year, but it seems to perform better with an increasing number of features. This improvement, which starts around June, is justified because most of the crops that were examined, are harvested at June, after which our model can classify the data with higher confidence.\\

\noindent However, in order to calculate the persistence of a parcel as mentioned above, a full Sentinel-2 series, for any given year of inspection, is required. Therefore, the constant persistent misclassifications are merely used as validation dataset of truly wrongly declared parcels, and are not part of the smart sampling algorithm. Thus, in order to be able to identify wrongly declared samples, in real scenarios, the total number of times that a green parcel has been misclassified, until the current run, is calculated. Then, we set a different threshold for each iteration ($P_t$ in Algorithm \ref{alg:alg1}) which is calculated using the semantic query in Figure \ref{rule_thres}. This is referred to as threshold of varying persistent misclassifications. \\

\noindent Figure \ref{fig:evolution_misclassifications2} displays the number of misclassified parcels using both constant and varying persistence thresholds and a constant green threshold equal to 0.5. The threshold value was defined based on the a priori knowledge of the percentage of false declarations, annually. INTIA has stated that usually no more than 3\% of the agricultural parcels in Navarra are falsely declared. Additionally, INTIA, as acting paying agency for the Navarra region, is obligated to conduct randomly selected inspections for at least 1\% of applications. Therefore, using 0.5 as the threshold for the green category, provides 267 constant persistent misclassifications, which amount to 3.1\% of the dataset; satisfying both the expected false declarations percentage and the minimum number of mandated OTSCs. Inspecting Figure \ref{fig:evolution_misclassifications2}, each point, of both curves, is associated with the peristence number that is used to count the misclassifications. 

\paragraph{Implementation of Scenario 2: Crop Diversification}.
\noindent Table \ref{table:results_crop} contains a list of farmers that seem to not comply with the rule described in Figure \ref{rule1} and Figure \ref{rule1b}. The first column contains the farmer identifier that corresponds to a specific farmer, the second contains the count of different crop types that the specific farmer cultivates in the fields that he has into his possession and the third the sum area of the parcels that the farmer has into his possession. The fourth column contains the area of the main crop for the cases that the farmer cultivates more than 2 crop types in an area of 10 to 30 hectares but the main crop exceeds the 75\% of the total area. It also contains the sum of the two main crops for the cases that the farmer cultivates more than 3 crop types in an area of more than 30 hectares but the sum of two main crops exceeds the 95\% of total area. The last column contains the crop types that the farmer is cultivating in their farm. \\

%\begin{table}[!ht]
%\resizebox{9cm}{!} {
%\centering
%\tabcolsep=0.05cm
% \begin{tabular}{||c c c c ||} 
% \hline
% Farmer id & crop types count &	sum area of parcels & crop types\\ 
% \hline
%2181 & 2 &	305481 & Soft Wheat, Barley \\
%
%2183 & 2 &	340440 & Soft Wheat, Barley \\
%
%1349 & 1 &	337116 & Soft Wheat \\
%
%37 & 1 &	1360397 & Soft Wheat \\
%
%1626 & 2 &	303953 & Soft Wheat, Barley \\
%
%49 & 2 &	692969 & Soft Wheat, Barley \\
%
%2255 & 1 &	346641 & Soft Wheat \\
%
%151 & 2 &	437095 & Soft Wheat, Barley \\
%
%2322 & 2 &	321067 & Barley, Oats \\
%
%2234 & 2 &	2044910 & Soft Wheat, Barley \\
% \hline
%\end{tabular}}
%\caption{Crop Diversification results}
%\label{table:results_crop}
%\end{table}
\begin{table}[!ht]
\centering
\caption{List of farmers that seem to be non-compliant to the Crop diversification requirement and information about their crop fields}
\resizebox{9cm}{!} {

\tabcolsep=0.05cm
 \begin{tabular}{||c c c c c||} 
 \hline
 Farmer id & crop types & sum area  & main crop(s) & crop types\\
 & count & of parcels & area & \\
 
 \hline
37 & 1 & 245008 & & Soft Wheat \\

1005 & 2 & 309458 & &  Soft Wheat, Barley \\

1007 & 1 & 191124 & & Soft Wheat \\

1133 & 2 & 274066 & 271058 & Soft Wheat,  Rape \\

1229 & 2 & 407609 & & Soft Wheat, Barley \\

1319 &1 & 103471 & & Barley \\

1306 & 1 & 302232 & & Soft Wheat \\

1860 & 2 & 287242 & 278455 & Soft Wheat,  Rape \\

2677 & 3 & 577902 & 576071 & Rape, Soft Wheat, \\  
 & & & & Vinification vineyard \\

3131 & 3 & 614529 & 586915 & Soft Wheat, Rape, Barley \\

\hline
\end{tabular}}
%\caption{List of farmers that seem to be non-compliant to the Crop diversification requirement and information about their crop fields}
%Crop Diversification results}
\label{table:results_crop}
\end{table}

\begin{figure}[!ht]
      %\caption{Farmers distribution under Greening 1 requirement according to their compliance and total arable farm area that have under their possession}
    \centering
    \includegraphics[width=9cm]{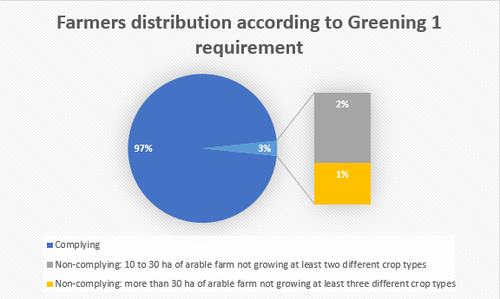}
  \caption{Farmers distribution under Greening 1 requirement according to their compliance and total arable farm area that have under their possession}
    \label{fig:distr_greening1}
\end{figure}

\noindent Figure \ref{fig:distr_greening1} presents the percentage of farmers that seem to be complying and non-complying to the Greening 1 requirement. The number of farmers that there is no need to be checked is 2610 (82\%), while the number of potentially non-complying farmers is 703 (18\%). More specifically, 506 (13\%) farmers own 10 to 30 hectares of arable farm from which 54 (1\%) grow less than two different crop types and 452 (14\%) grow two or more different crop types but the main crop covers more than 75\% of the land. On the other hand, 189 (5\%) farmers own more than 30 hectares of arable farm  from which 68 (2\%) grow less than three different crop types and 121 (3\%) grow three or more different crop types but the two main crops cover more than 95\% of the land. To perform these calculations we used the data of the classification run from early July (jul08).
%Figure \ref{fig:distr_greening1} presents the percentage of farmers that seem to be complying and non-complying to Greening 1 requirement. The number of farmers that there is no need to be checked is 3210 (97\%), while the number of non-complying is 111 (3\%). More specifically, 70 (2\%) farmers own 10 to 30 hectares of arable farm and are not growing at least two different crop types while 41 (1\%) farmers own more than 30 hectares of arable farm and are not growing at least three different crop types. To perform these calculations we used the data of the last classification run.
\paragraph{Implementation of Scenario 3: SMR 1} \label{impl_scen3}.
\noindent In the SMR 1 requirement we select the parcels that their distances from surface is under 10 meters based on the organic manure spreading buffer. From these only parcels with slope higher than $12^{\circ}$ are recorded, according to the SMR1 specifications. Additionally, in order to select only parcels with an actual risk for runoff, the aspect of the proximity line ($\alpha_{degrees}$),  namely the orientation of the line connecting the parcel to the water object, needs to fulfill the relationship:
\begin{equation}
    aspect-45<\alpha_{degrees}< aspect+45
\end{equation}
where the aspect is given per parcel and the proximity line aspect is calculated using the following formula:

\begin{align}
    \alpha_{radians}&=\atantwo(x_{1}-x_{2},y_{1}-y_{2}) \\
    \alpha_{degrees}&=\alpha_{radians}\times{180^{\circ}/\pi}
\end{align}

\begin{figure}[!ht]
    \centering
    \includegraphics[width=8cm]{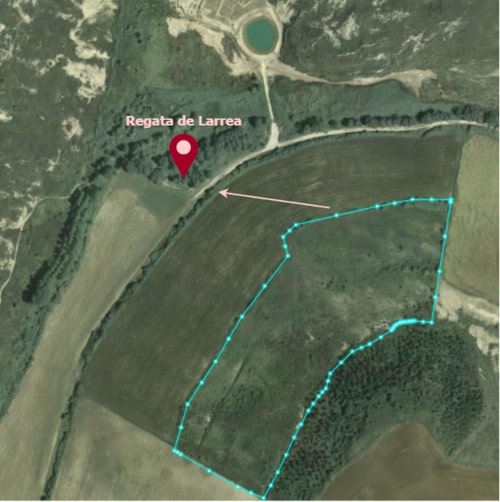}
    \caption{Example of parcel that is susceptible to runoff according to the SMR 1 requirement. The parcel slope aspect (arrow) and the location of the nearby Regata de Larrea river are also shown.}
    \label{fig:res_smr1}
\end{figure}

\noindent Figure \ref{fig:res_smr1} shows the visual representation of the parcel that was mentioned in section \ref{scen_3}. According to the aforementioned rules this parcel is of high risk. 

\begin{figure}[!ht]
    \centering
    \includegraphics[width=9cm]{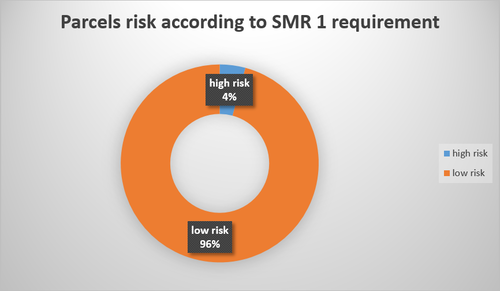}
    \caption{Distribution of low risk and high risk parcels according to the SMR1 specifications}
    \label{fig:distr_smr1}
\end{figure}

\noindent Figure \ref{fig:distr_smr1} presents the percentage of parcels that seem to be of low and high risk based on the SMR 1 requirement. The number of low risk parcels that there is no need to be checked is 14407 (96\%), while the number of high risk parcels is 630 (4\%). 

\subsubsection{Smart Sampling Accuracy}
\noindent Table \ref{table:pa_ua_evolution} presents the PA and UA of the misclassified green parcels, for varying and constant (5) persistence, with the latter functioning as the ground truth. To calculate the PA, the number of correctly misclassified green parcels of varying persistence ($P_t$) is divided by the total number of misclassified parcels in the validation dataset (267), while for the UA it is divided by the total number of misclassified green parcels of varying persistence. 

\begin{table}[!ht]
\caption{PA and UA evolution of the smart sampling algorithm}
\centering
 \begin{tabular}{||c c c ||} 
 \hline
 Date & PA (\%) &	UA (\%) \\ 
 \hline
May04 & 56.18 & 32.82 \\
May19 & 50.19 & 39.88 \\
May24 & 52.81 & 39.50 \\
Jun03 & 48.69 & 50.98 \\
Jun23 & 50.94 & 70.10 \\
Jul08 & 53.93 & 97.96 \\
Jul23 & 56.93 & 100.00 \\
Aug02 & 54.68 & 100.00 \\
Aug27 & 71.54 & 98.96 \\
Sep11 & 69.29 & 100.00 \\
Sep16 & 79.40 & 100.00 \\
Oct16 & 83.15 & 100.00 \\
Oct21 & 85.77 & 100.00 \\
\hline
\end{tabular}

\label{table:pa_ua_evolution}
\end{table}

\noindent Table \ref{table:pa_ua_evolution} shows that suboptimal PAs and UAs are achieved for classifications early in the year. This indicates that the smart sampling algorithm will select erroneously a significant percentage of the suggested OTSCs. Nevertheless, the smart sampling precision, indicated by the UA, reaches near perfect values, from early July08 onward.\\

\noindent In an attempt to further filter the OTSCs selection of the suboptimal smart sampling results, found early in the year, we exploit the season type of the crop taxonomy, as described in Figure \ref{fig:croptype_ontology}. Each of the crops that were examined belongs to one of the three season types and that is summer, winter, and year-round. Based on that, a false declaration would be more likely, if the prediction of the algorithm belongs to a different season class as compared to the corresponding season class of the declaration. Table \ref{table:percentages} presents the number of the green misclassifications that come from each of the different season types as well as the total sum of them, along with the number of green misclassifications for which the predicted crop season differs from the declared one, identified as an alarm. Finally, the last column indicates the percentage of those alarms that belong to the validation dataset of more than five persistent misclassifications.  \\

\begin{table}[!ht]
\centering
\caption{Percentages of green misclassifications for each season class and UA of persistent alarms filtered through Algorithm}
\resizebox{9cm}{!} {
 \begin{tabular}{||c c c c c c c ||} 
 \hline
Date & Total &	Winter & Summer & All Year & Alarms & Persistent Alarms UA (\%)\\
\hline
May04 & 457 & 418 & 3 & 1 & 35 & 57.14 \\
May19 & 336 & 302 & 3 & 1 & 30 & 63.33 \\
May24 & 357 & 321 & 4 & 2 & 30 & 60.00 \\
Jun03 & 255 & 226 & 4 & 1 & 24 & 70.83 \\
Jun23 & 194 & 167 & 4 & 1 & 22 & 68.18 \\
Jul08 & 147 & 125 & 3 & 1 & 18 & 94.44 \\
Jul23 & 152 & 133 & 2 & 1 & 16 & 100.00 \\
Aug02 & 146 & 128 & 1 & 1 & 16 & 100.00 \\
Aug27 & 193 & 168 & 0 & 2 & 23 & 91.30 \\
Sep11 & 185 & 164 & 0 & 2 & 19 & 100.00 \\
Sep16 & 212 & 182 & 0 & 2 & 28 & 100.00 \\
Oct16 & 222 & 186 & 1 & 2 & 33 & 100.00 \\
Oct21 & 229 & 190 & 1 & 2 & 36 & 100.00 \\
\hline
\end{tabular}
% \ref{alg:alg2}
}
\label{table:percentages}
\end{table}

\begin{figure}[!ht]
    \centering
    \includegraphics[width=14cm]{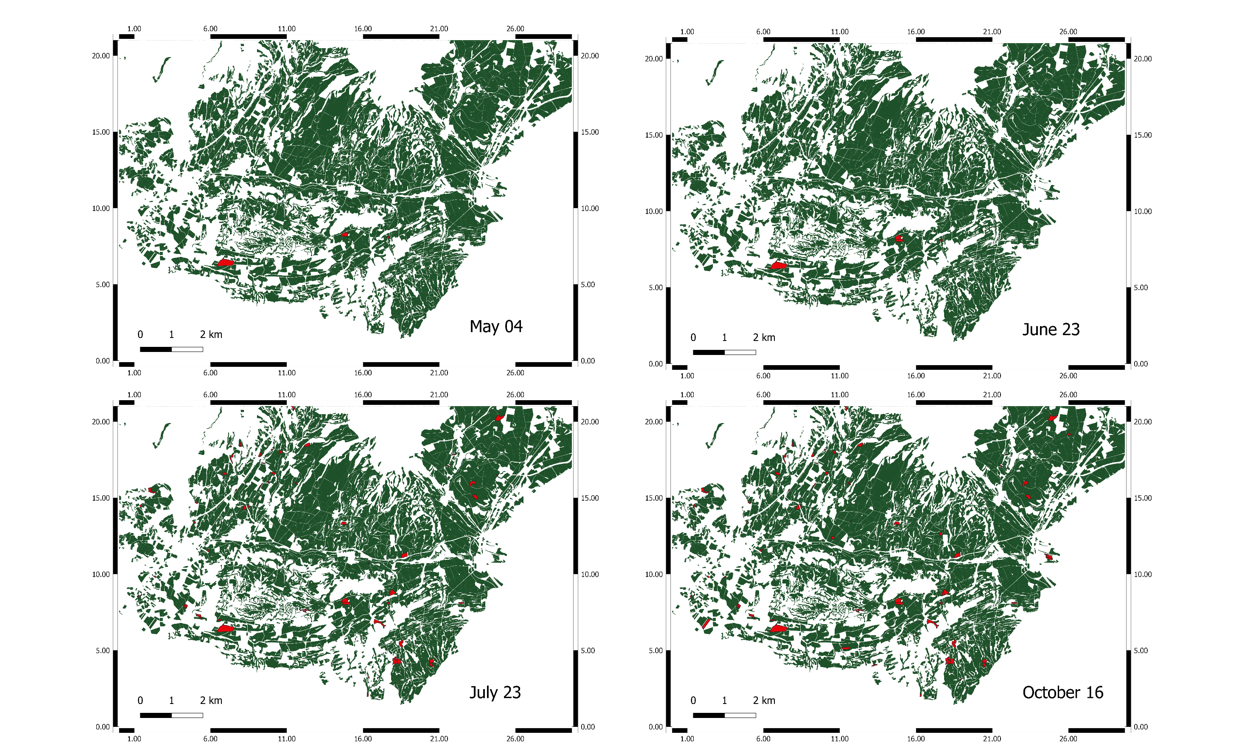}
    \caption{Smart sampling alerts at different instances throughout the year of inspection. The red parcels indicate to the paying agency inspector where to target their inspections. Early in the year the alerts are fewer, stemming from less reliable classifications. With more images as we move along in time, classifications become more reliable and thus more alerts are identified.  }
    \label{fig:ss_evolution}
\end{figure}

\noindent Comparing the “Persistent Alarms UA” column of Table \ref{table:percentages} and the “UA” column of Table \ref{table:pa_ua} reveals increase in the precision of smart sampling for runs that take place as early as May. Special attention is given to UA instead of PA, as it demonstrates the reliability of the system. Since, the inspections are not exhaustive but rather sampled based, it is more important to ensure that most alerts are indeed wrong declarations.
In Figure \ref{fig:ss_evolution} is displayed the evolution of the smart sampling alerts for different classifications throughout the year. It can be observed that for runs until June 23, the alerts are only few as they are passed through the Algorithm \ref{alg:alg2} filter. It can be seen that even though some alerts only appear early in the year, the critical mass of them can be identified from as early as July.

\subsubsection{Scalability}
%\paragraph{Scalability and Transferability of the Crop Classification scheme}
%\hl{++ Here we will reference the complexity analysis performed in Sitokonstantinou et al (2018), in support of the overall scalability of the system (add 5-7 lines max}\\
%The crop classification 

%\paragraph{Computational complexity of querying}
\noindent Table \ref{table:query_exec} presents the execution time for each one of rules that are described in this work. The execution time for each rule is calculated as the mean time of each scenario of the queries presented in section \ref{querying}. Results show that the query that takes the most time to run is the SMR 1 requirement, taking into account that this query runs per parcel, compared to the other queries which run for the whole dataset. The reason is that this query requires multiple calculations between polygons such as distance and aspect which are very time-consuming. The second column presents the execution time in seconds for the actual size of the dataset, while the last presents the execution time in a reduced dataset size in order to further understand the scalability.

\begin{table}[!ht]
\caption{Mean execution time of each scenario according to the semantic queries described in section \ref{querying}}
\centering
 \begin{tabular}{||c c c ||} 
 \hline
 Query & Execution time (sec)  & Execution time (sec)   \\ 
  &  actual size &  reduced size  \\ 
 \hline
 Smart Sampling &	0.4 & 0.4 \\ 
 
 Greening 1 requirement &	7.35 & 0.023  \\
 
 SMR 1 requirement &	1.16 & 0.016 \\
 \hline
 Number of triples & 7.369.055 & 1.260.600 \\
 \hline
\end{tabular}

%Mean execution time of each query}
\label{table:query_exec}
\end{table}

\begin{figure}[!ht]
    \centering
    \includegraphics[width=8cm]{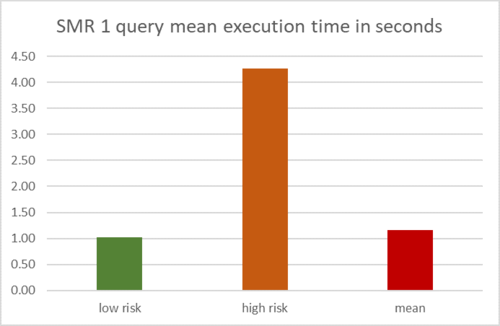}
    \caption{SMR 1 requirement mean execution time for each parcel according to the risk type (low, high) compared with total mean execution time}
    %SMR 1 requirement mean execution time}
    \label{fig:res_smr1_exectime}
\end{figure}

%the distribution of the execution time of SMR 1 requirement for each parcel%
\noindent Figure \ref{fig:res_smr1_exectime} presents the mean execution time of SMR 1 requirement for each parcel according to the risk type (low, high). The execution time is extremely low when the parcel does not satisfy the filters (aspect, distance, slope), while when the parcel satisfies the filters the execution time is significantly higher. Despite this fact, the mean execution time of all parcels is significantly low, taking into account the large number of low-risk parcels. 
\subsubsection{CAP monitoring in practice}
 \noindent The results in Section IV-C described the implementation of the practical applications of the proposed system, which have been showcased in the form of the three scenarios of Section IV-B. Scenario 2 on Greening 1, and Scenario 3 on SMR-1 requirements respectively, are performed once for every year of inspection. Therefore, these two scenarios infer a single cycle of execution, with reference to Figure \ref{fig:stages}. On the other hand, the Smart Sampling application (Scenario 1) is an iterative process. The first crop classification is performed in early May, coinciding with the CAP subsidies applications commencement, with subsequent iterations producing progressively more accurate crop type maps with every new Sentinel-2 acquisition. Hence, there is a full cycle execution (Figure \ref{fig:stages}) with every newly acquired image, from ingestion to interlinking. \\
 
\noindent According to the results in Table \ref{table:percentages}, the issued alarms are adequately trustworthy to suggest potential OTSCs from as early as July, with a user's accuracy of more than 94\%. From then on, the paying agency inspectors can have targeted OTSCs that increase in number with every new iteration, following their summer-long inspection process. Alternatively, the results from May until early July can be used in assistance of the farmer application process. The applicants and the paying agencies can have an indication of potential non-compliance even at the application stage; thus allowing the farmer to make timely changes to their application.

\subsection{Conclusions}
\label{sec:conclusion_semantics}
\noindent In this study we presented a semantic-oriented framework for knowledge discovery using supervised classification in the CAP domain. The main focus is given on the detection of possible violations according to the declaration of the farmers, the Greening 1 requirement and SMR 1 requirement. The framework can strongly assist in decision making issues by providing helpful information to paying agency inspectors and environmental consultant to detect possible breaches of compliance. The proposed solution  relies on data coming from sentinel images and open data (e.g. OpenStreetMap). Common denominator between the two datasets is the provision of georeferenced information. Data refer to a region in northeastern Spain. The SVM classification method has been applied to classify the cultivated crop types for multiple instances throughout the cultivation season. The data, which are in Shapefile format, are converted into Turtle RDF format using the GeoTriples tool. Results are saved in GraphDB triplestore. Semantic queries are executed to enrich the data with information about possible farmers non-compliance according to agricultural policies.\\

\noindent In this work we have shown how the paying agencies of the CAP can benefit from the exploitation of big Copernicus data. We showcased how with only freely available satellite data and ancillary LOD one can provide actionable information. Combining the state-of-the-art in EO-based crop classification, semantic enrichment and linking free and open data has facilitated the development of an end-to-end system, from data acquisition to CAP related decision making. The main innovations of the presented methodology include its re-usability and transferability, using predominantly open data and requiring minimal fine-tuning when applied to other regions,  and scalability, accounting for all big data considerations and choosing the computationally efficient alternative every step of the way, towards the monitoring approach of the new CAP. \\

\section{Street-level Images} \label{street}

\subsection{Literature review}
\noindent  CAP paying agencies of each Member State have to move away from the currently employed confined and sample-based checks and adopt practices that will enable them to take informed decisions for the subsidy payments of all parcels. For that, the AMS is introduced. AMS refers to the systematic monitoring and assessment of agricultural activities and practices using satellite data (e.g. Copernicus), joined with data from the LPIS, but also other ancillary sources. 
The Copernicus Sentinel satellite missions provide optical and SAR data of high spatial and temporal resolution and have been extensively used for the agricultural monitoring and the CAP purposes \cite{lopez2021sentinel, sonobe2017assessing}. \\

\noindent On the other hand, the main enabler towards the feasibility of this exhaustive monitoring, is widely considered to be AI. In particular, ML pipelines are continuously being developed and integrated into the operational framework of the  paying agencies~\cite{lopez2021sentinel, schulz2021large, navarro2021operational, rousi2020semantically, sarvia2021possible}.
However, for wall-to-wall and perfectly accurate assessment on the compliance of farmers, as required by the new CAP guidelines, the data modality and spatial resolution (10 m - 60 m) of the Sentinel imagery alone is not enough. In fact, it is required to incorporate other sources to complement their weaknesses. \\

\noindent Remote sensing systems usually produce a large volume of data due to the high spatial, spectral, radiometric and temporal resolutions needed for applications in agriculture monitoring \cite{sishodia2020applications}. 
Looking at the current, rapidly advancing, state-of-the-art in ML for EO, data availability is increasingly perceived to be of multi-dimensional nature, through the notion of data variability and data complementarity, on top of data volume. Thus, more and more heterogeneous data sources are collected and data fusion techniques are applied in an attempt to generate enhanced feature spaces from non-overlapping and complementary feature domains. In EO applications during the recent years, fusion of data in different altitude levels, such as UAV and high resolution satellite images, is very common \cite{zhao2019finer, zhou2021uav}.
However, to convert the available data into training datasets for ML/DL pipelines, we also require the relevant annotations, i.e., ground-truth labels. 
While satellite imagery is not in short supply, ground-level observations are hard to find and they lack consistency in terms of spatial and temporal availability. Apart from being difficult to acquire, such labels are also the most expensive data component to generate; not just in terms of monetary cost, but also in terms of time, workforce and the availability of expert knowledge for the annotation.\\

\noindent Going back to the new CAP monitoring requirements,  paying agencies are facing challenges in their attempt to monitor compliance. Through the implementation of the AMS, the decisions that will be made for the subsidy allocations of all parcels have to be supported by evidence, either as a result of the predictions of ML/DL pipelines, or through photo-interpretation, which is a common process that the  paying agencies undertake (in retrospect) for validation purposes and for cases of dispute resolution. Thus, the need for such a holistic approach regarding annotated data availability is becoming evident.\\

\noindent Driven by the above, we present an analysis-ready dataset, annotated with crop type labels, that includes both the space (Sentinel data) and ground (street-level images) domains. We formulate this framework of space and ground data availability using only open-access (Copernicus data and LPIS) and crowdsourced (Mapillary) data. We present a methodology and share the code for i) collecting, ii) processing to analysis-ready and iii) annotating (with crop labels) street-level images. Each street-level image is matched with a time-series of Sentinel-1  and Sentinel-2  data. We use our dataset to independently train traditional ML and state-of-the-art DL models on its space and ground data layers, and we discuss the capabilities of synergistic use by applying late fusion. All in all, we introduce the idea of space-to-ground data availability and offer an example curated dataset that the community is encouraged to work on using deep learning fusion models, on both satellite and street-level images, towards fully exploiting the complementarity of these data modalities for agriculture monitoring downstream tasks.

\subsection{Data availability for Remote Sensing applications}
\noindent Our initial approach to this multi-level (i.e., altitude, mode, angle of sensing etc.) data availability framework includes the two edges, i.e., space-level (low earth orbit) and ground-level. In each of these domains, the data outputs differ substantially in terms of a set of characteristics, such as the spatial and spectral resolution, the temporal availability, the viewing angle, the susceptibility to the weather conditions, and the nature of the conclusions that can be drawn through their use.

\subsubsection{Data sources going from space to ground}
\noindent Space-level includes acquisitions from EO satellites starting from higher ranges of low earth orbits (e.g. Sentinel 1/2), and moving down to lower altitudes (e.g. SPOT6/7 and Planet's Doves). On the other edge, the ground-level comprises imagery and acquisitions coming from the street (e.g. Google Street View, Mapillary and Kakao), as well as in-situ geotagged photos from mobile phones \cite{kenny2021empathising}, which is a source that is already utilized for agricultural monitoring \cite{wang2020mapping} as well as for validation processes and evidence collection campaigns of the  paying agencies. Between these two edges of space and ground data acquisitions, there is the potential to integrate other data sources from various domains and acquisition altitudes, with the most common being aerial photos \cite{zhang2013fusion} and images from Unmanned Aerial Vehicles (UAVs) \cite{maimaitijiang2020crop}.

\subsubsection{Space to ground data availability for CAP monitoring}
\paragraph{Space-level datasets.}
\noindent Data availability on the space-level is quite extensive, with a growing number of EO satellites providing a consistent stream of data acquisitions over the globe. The most commonly used data sources on this level for the CAP monitoring purposes are the Sentinel and Landsat satellites, with their acquisitions becoming available as open-access data and on various processing levels. Thus, there is a plethora of available datasets with annotations for a wide range of downstream applications. BigEarthNet\cite{8900532} is a collection of 590,326 pairs of Sentinel-1 and Sentinel-2 image patches over Europe, annotated with multiple land cover classes. DENETHOR \cite{Kondmann2021DENETHORTD} is a publicly available analysis-ready benchmark dataset which combines Planet Fusion data, together with Sentinel-1 radar and Sentinel-2 optical data. It includes annotations of crop type labels, covering an area in Northern Germany. ZueriCrop\cite{turkoglu2021crop} is another case of a crop classification targeted dataset, containing Sentinel-2 time series, along with ground truth labels for 116,000 fields over Zurich. Sen4AgriNet\cite{sen4agrinet} is a benchmark dataset based on Sentinel-2 imagery, annotated with crop type labels, which contains 42,5 million parcels, thus, rendering it suitable for the training of deep learning architectures. Finally, CropHarvest \cite{tseng2021cropharvest} contains Sentinel-1 and Sentinel-2 timeseries, meteorological and topographic data, together with crop type labels for 90,480 datapoints spanning all around the world.

\paragraph{Ground-level datasets.}
\noindent As previously described and, in contrast to the space-level datasets, securing availability on the ground-level is a challenging pursuit. This is both in terms of finding data relevant to the CAP monitoring domain, as well as in regards to finding or generating annotations. The main focus of street-level imagery in computer vision is commonly the identification of objects and features that are encountered in urban areas (e.g. cars, traffic signs etc.), driven by domains of application like self-driving automobiles. Thus, there is a general shortage in ground-level datasets from rural areas and, even more, a shortage in annotations of agricultural interest, with most of the available ones coming from time-consuming photo-interpretation efforts. Despite the challenges, some remarkable collections of ground-level datasets exist. iCrop\cite{Wu2021-jb} is a multiclass dataset of 34,117 road view images in China, annotated with crop type labels. CropDeep\cite{CropDeep2019} includes 31,147 in-situ images and 40,000 annotations related to plant species classification, captured in a variety of realistic conditions. However, to the authors' knowledge, the majority of the available annotated datasets on this level are of significantly smaller size, which renders the effective training of ML/DL models through their use more challenging. For example, Crop/Weed Field Image Dataset \cite{haug15}, comprises 60 images with vegetation masks and annotations for crops and weeds.

\paragraph{Combining Space and Ground.}
\noindent The combined use of datasets from different sources is a common approach in the literature for feature enhancement and augmentation of the training data volume. These combinations are applied in a variety of ways, like on the measurement level \cite{maimaitijiang2020crop}, feature level \cite{zhang2013fusion}, and decision level \cite{chen2020decision}. However, such datasets are usually limited to the space domain (e.g., Sentinel-1 and Sentinel-2 \cite{sitokonstantinou2021scalable}, Sentinel and Landsat-8 \cite{chen2020decision}, Sentinel and PlanetFusion \cite{Kondmann2021DENETHORTD} etc.).
Consequently, availability on the basis of fusing different domains (e.g. space and ground) is limited.

\subsubsection{Ground coverage with crowdsourced and open access-data}
\noindent Given the amount of ground-level data acquisitions that are required to match and fully cover a single space-level acquisition, we are driven towards the need to decentralize the effort of ground-level data collection and the exploitation of crowdsourcing platforms like Mapillary, Google Earth Pro and Kakao. However, while such platforms can offer significant volumes of ground-level data, they do come with certain particularities that need to be addressed before they can be considered analysis-ready. For instance, crowdsourced contributions may display discrepancies in the acquisition methodology. Also, there are no annotations available for most of the contributions, especially ones of agricultural interest. To counter this shortage of annotations, methodologies for their mass generation are capturing the interest of the research community. LPIS data, which contain georeferenced crop type labels coming from farmers' declarations, have already been used in DataCAP \cite{datacap} to map crop types to street-level images based on transformations to the coordinates of the image acquisition. Street2Sat\cite{street2sat} pipeline is aiming to transform geotagged street-level images to sets of georeferenced points that can be used as labels for satellite images. In \cite{dAndrimontSLI}, the authors developed a pipeline for the monitoring of crop phenology using deep learning architectures on street-level imagery, reference parcel data, and ground observations.

\subsection{The Dataset}

\noindent In this work, we introduce a multi-level, multi-sensor, multi-modal dataset annotated with grassland/not grassland labels for the monitoring of the CAP. We empower the transition towards the post-2020 CAP guidelines through the exploitation of the benefits offered by the dataset's space and ground components. Our methodology for data collection and annotation is presented, and the code is available at \url{https://github.com/Agri-Hub/Space2Ground}. We showcase the multi-faceted utility of our dataset, by applying state-of-the-art ML/DL architectures for grassland detection on data from the space and ground domains. On top of the single-domain results, we apply late-fusion (decision level) to underline the potential of harnessing our dataset's multi-modality.

\subsubsection{Data sources and data collection}
\noindent We constructed this dataset considering the two edges of data availability. On the space-level edge, we have included Sentinel-1 GRD and SLC products and Sentinel-2 multispectral imagery, acquired through the Copernicus Open Access Hub. Specifically, from Sentinel-1 we produced VV/VH backscatter and VV/VH coherence using the snappy library and from Sentinel-2, we acquired all bands except 1, 9 and 10. On the ground-level edge, we use the Mapillary crowdsourcing platform to include street-level images covering our area of interest. In order to interact with the latest version of the Mapillary API (v4), we developed a Python library. The scripts for downloading imagery from Mapillary, based on a specific time range for a specific area can be found on GitHub (\url{https://github.com/Agri-Hub/Callisto/tree/main/Mapillary}). For the annotation of these datasets with agricultural crop type labels, we use the LPIS data which, for the Netherlands, are openly available through their National Georegister website.\\

\noindent Grassland is the most dominant crop type in the Netherlands. In our AOI, there are 55,039 parcels in total and more than 80\% have been declared as Grassland. We excluded instances without clear spectral signatures or irrelevant to CAP purposes (e.g. forests, borders adjacent to arable land, etc.) and heavily impacted by cloud coverage, ending up with a total of 37,041 parcels, out of which 31,350 (84.6\%) are grasslands.

\subsubsection{Methodology of annotation}
\noindent To annotate our data we use the LPIS dataset. It consists of shapefiles with geo-referenced parcel geometries, that are assigned with crop type labels. As Sentinel data are also geo-referenced, their annotation with these labels is straightforward.
The generation of labels for the street-level images, though, is a more challenging task. Mapillary includes information about the coordinates of the sensor in each image capture and a compass angle to specify the direction towards which it is facing. To generate annotations for the street-level images, we fuse them with the LPIS data by applying transformations to the acquisition coordinates following the methodology developed in \cite{datacap}. Using this methodology we end up with up to 2 labels for each single image (one for the left and one for the right side of the field of view).

\begin{figure*}[!ht]
\centering
\includegraphics[width = \textwidth]{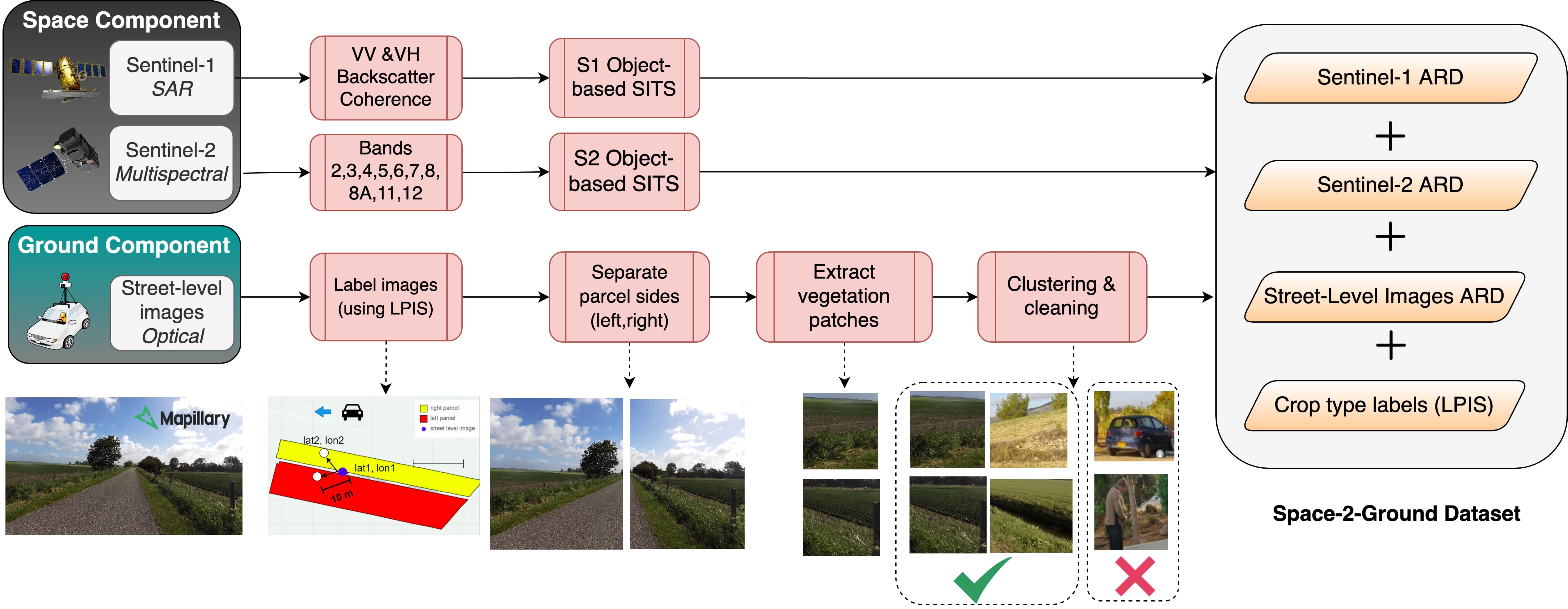}
\caption{Methodology for the dataset creation. The Space component consists of Sentinel-1 and Sentinel-2 object-based time series data. The Ground component consists of street-level imagery acquired through the Mapillary platform. Transformations to the acquisition coordinates are applied to attach LPIS crop type labels to left and right parcels of each image. Each parcel side is isolated and vegetation patches are extracted. A final cleaning step is applied through k-means clustering and photo-interpretation.}
\end{figure*}

\noindent On top of this approach, we apply additional steps to further refine the images and improve the quality of the annotations. Initially, we split each image in 2 halves, separating the labels of each side accordingly (with a front-facing acquisition, each side corresponds to a different parcel). Then, to isolate and extract vegetation, we keep the 30\% top leftmost or rightmost part (depending on the side) with regards to width, and the 20\% to 50\% part with regards to height. Through the previous steps, we end up with 36,985 image patches, which are resized to (260, 260).\\

\noindent Subsequently, we filter out ones that are irrelevant or contain noise (eg. cars, infrastructure, roads, etc.) in an unsupervised way. In particular, we use a VGG16 pretrained network to extract a representation of each image. Then, we perform Principal Component Analysis (PCA) on these representations and we keep the 100 most important features. Finally, we fit a k-means model to the aforementioned data to group them into 200 clusters, and we keep the 50 most useful ones through visual interpretation. After this filtering step, we end up with a refined set of 10,102 annotated image patches.

\subsubsection{Dataset structure}

\noindent On the space-level, we calculate mean values of Sentinel-1 and Sentinel-2 data for each parcel. Sentinel-1 data are also aggregated on a monthly basis. Consecutively, we create training and test sets (80\%-20\%), and store them in csv format. The first two columns correspond to the the crop label (Grassland/Non-Grassland) and the unique identifier of the parcel, while the rest of the columns represent the mean values of the Sentinel features.
On the ground-level, the street level images are also split into training and test data (80\%-20\%), and they are stored in different directories based on their class label. The image names are in form of $\{imageID\}\_\{direction\}$, where $\{imageID\}$ corresponds to the unique identifier of the image and $\{direction\}$ to the direction in which it has been cropped.

\subsubsection{Dataset utility \& downstream tasks}
\noindent Through the provision of an object-level mapping between its different data sources, our dataset facilitates the creation of pipelines that harness it in a multitude of ways. Each single data source can be used separately for the training of ML/DL pipelines. Various modes of data fusion are supported, like measurement fusion of same-level data, feature fusion, and late fusion of models trained separately on different data layers and levels. The object-level mapping also allows for any available annotations to be broadcast to all data sources. \\

\noindent The ground component comes with a set of additional use cases. Ground data can work as material for validation of the ML/DL predictions and for the creation of ground-truth annotations. They are also suitable for other photo-interpretation tasks. For instance, paying agencies commonly undertake photo interpretation effort, for desk inspections, known as desk OTSCs, and for resolution of disputes. Models trained on ground-level data can be mounted on ground sensors (cars, mobile phones, UAVs, etc) to support inference at the edge. Last but not least, Space-to-Ground data availability is greatly facilitating the creation of synthetic data through the use of generative architectures, like Generative Adversarial Networks (GAN), which are already being applied, particularly for the translation of data between the domains of satellite and street-level imagery \cite{sat2street,street2sat}.

\subsection{Benchmarking}
\label{sec:results2}

\noindent We evaluated the performance of a Random Forest (RF), a Support Vector Machine (SVM) model and three different deep learning models on our dataset for the task of grassland classification using Sentinel-1 and Sentinel-2 data as input. RF and SVM are commonly used for remote sensing tasks, thus we included it
as a benchmark to compare with the deep learning models. The different deep learning architectures consist of a temporal Convolutional Neural Network (CNN) architecture \cite{Pelletier2019Temporal} which has achieved state-of-the-art results in crop classification, an LSTM with 64 hidden units and the same LSTM model with an additional attention mechanism. Table \ref{tab:metrics} depicts the performance of these models on the test dataset. 

\begin{table}[!ht]
\caption{Performance metrics for grassland classification using different models.}
\label{tab:metrics}
\centering
\begin{tabular}{|c|c|c|c|c|c|}
\hline
\textbf{Method}   & \textbf{SVM} & \textbf{RF}  & \textbf{TempCNN}  & \textbf{LSTM} & \textbf{LSTM+Attention} \\ \hline
\textbf{Accuracy} & 93.69\%  & 94.68\%  & 95.22\%  & 95.14\%  & 95.20\%  \\ \hline
\textbf{F1 score} & 85.22\%  & 88.08\%  & 89.96\%  & 89.85\%  & 90.05\% \\ \hline
\end{tabular}
\end{table}

\noindent Apart from the space-level data, we also evaluated the performance of advanced CNNs (e.g. ResNet, EfficientNet, VGG, Inception) on the street level images. We experimented with pretrained (on ImageNet) models and the best performance was achieved by the Inceptionv3, with an overall accuracy of 85\%. 
Therefore, the application of fusion at the decision level is enabled, by exploiting the predictions derived from both the space and ground data components. In our case, the combination of the outputs of these different models, results to marginal improvement of the overall accuracy. However, we notice significant enhancement in terms of the confidence of the final prediction and thus the reliability of the final decision.

\subsection{Conclusions}

\noindent In this work we presented the first dataset that includes Sentinel-1, Sentinel-2 and street-level images, matched through the use of geo-referenced crop type labels (LPIS). We focused on grasslands for a Dutch AOI around Utrecht and we used openly accessible and crowdsourced data sources.
The code implementation of our methodology is shared, which renders it reproducible, transferable and extensible for the community to utilize and build on top.
The performance of the grassland classification models can be improved by using a street-level image dataset of enhance volume and quality. 
With regards to quantity, larger areas of interest and wider time-frames must be considered to increase the number of available images. To this direction,  paying agencies can also exploit the current operational framework of their daily inspections, and mount cameras on the field inspectors' vehicles to automatically capture imagery during their field visits.
From the image quality perspective, pointing cameras on the side can dramatically increase the percentage of vegetation per image and cover larger portions of the parcels. In addition, vegetation extraction can be improved through the application of semantic segmentation and creation of appropriate masks \cite{seamless_scene_segmentation}.
The potential for application of various fusion approaches will be further explored, in order to highlight the interoperability and complementarity of our dataset's layers and domains.

\section{DataCAP application} \label{datacap}

\subsection{Introduction}\label{sec:datacap_intro}

\noindent In this work, we demonstrate DataCAP, an AMS module that comprises a Sentinel datacube, ML pipelines for crop classification and grassland mowing detection, and street-level images retrieved from the Mapillary API. DataCAP offers easy and efficient searching, storing, pre-processing and analyzing of big EO data, but also visualisation tools that combine satellite and street-level imagery for verifying algorithmic decisions. DataCAP comprises two components (Figure \ref{fig:architecture1}), the data management component (Section \ref{sec:datacube}) and the visualisation component (Section \ref{sec:street}). \\

\subsection{Data management component}\label{sec:datacube}

\begin{figure}
    \includegraphics[width=\textwidth]{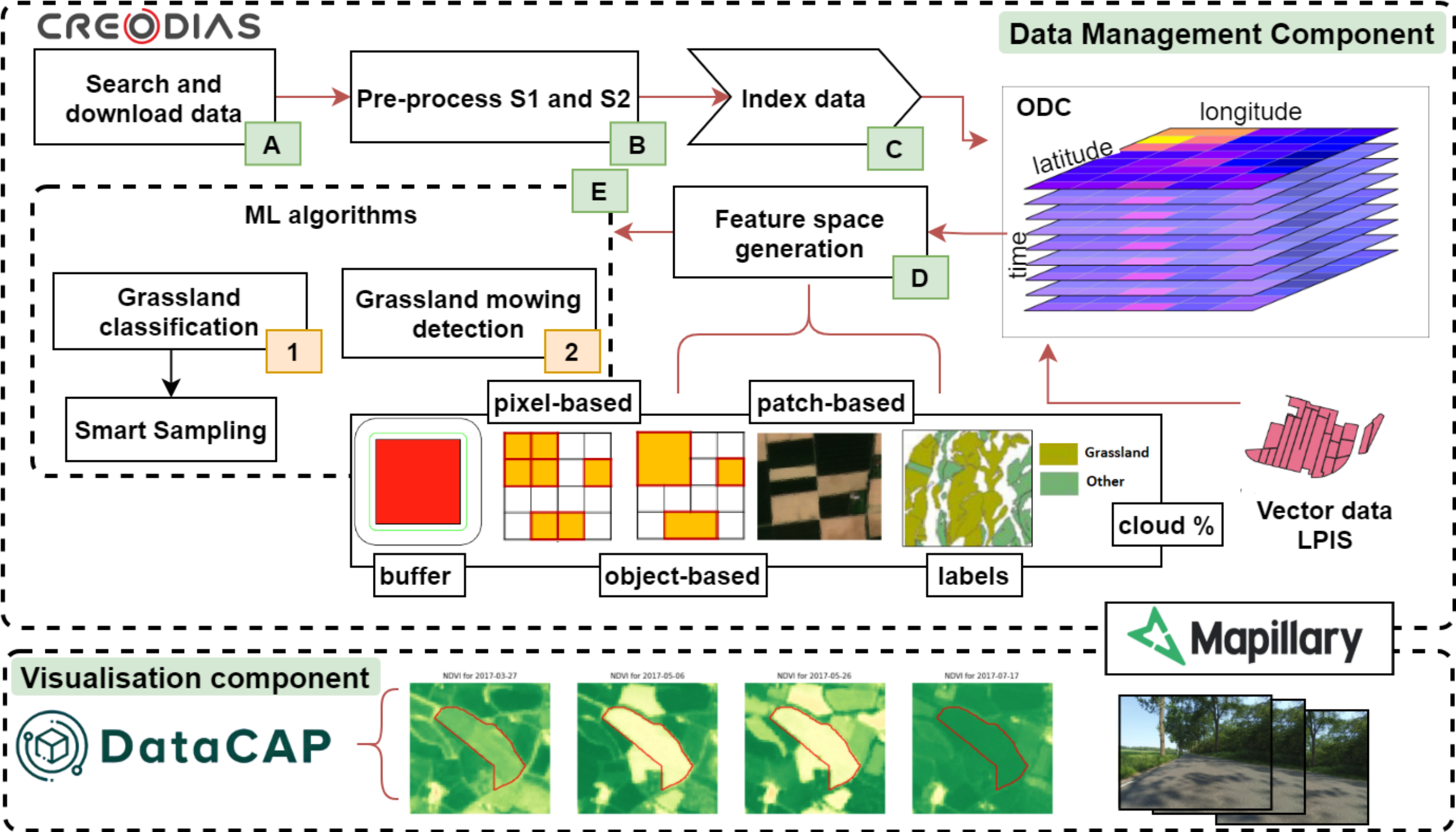}
    \caption{DataCAP architecture: data management and visualisation components} \label{fig:architecture1}
\end{figure}

DataCAP's Data Management Component (DMC) is an automated module that searches, harvests (Figure \ref{fig:architecture1} A) and pre-processes Sentinel data (Figure \ref{fig:architecture1} B), which are then indexed as ARD in its datacube (Figure \ref{fig:architecture1} C). This way, DataCAP offers easy and fast spatiotemporal data querying on SITS. The code for DataCAP DMC, including instructions on how to set up the datacube that facilitates it, can be found at \url{https://github.com/Agri-Hub/datacap}. DataCAP also enables the combination of satellite ARD with other data, including vector data for object-based image analysis, labelled data for supervised learning and crowdsourced street-level imagery for visual verification of ML outputs. The repository additionally includes two demo jupyter notebooks, showcasing the functionalities of DataCAP's DMC in the context of two CAP monitoring scenarios. Nevertheless, it should be noted that DataCAP is generic and can find applications in multiple other domains that leverage big satellite data and crowdsourced street-level images. Some indicative examples are land use/land cover change detection and burnt area mapping.\\

\noindent DataCAP's datacube has been built on the Open Data Cube (ODC) software. ODC has been developed by Geoscience Australia (GA) and is supported by the Committee of Earth Observation Satellites (CEOS) \cite{Killough2018OverviewOT}.
ODC provides a user-friendly Python API, which loads data into Xarray data structures. This structure simplifies the management of multidimensional arrays, enabling the slicing of data on each of the dimensions of the cube. Currently, our datacube contains time-series of Sentinel-1 and Sentinel-2 ARD (2017) over Netherlands (6,375 km\textsuperscript2). \\

\noindent \textbf{A - Search and download}
The Sentinel-1 and Sentinel-2 data products are automatically harvested and downloaded from the CreoDIAS API\footnote{https://finder.creodias.eu/}. The user can request Sentinel products within a specified time window over the area of interest, along with other user defined parameters such as the maximum allowed cloud coverage. The metadata of each GET response is stored in a PostgreSQL/PostGIS database in order to have full control and geo-spatial querying capabilities over the products that have been downloaded. \\

\noindent \textbf{B - Pre-processing}
The downloaded products are then processed to ARD. The output of this step returns time-series of i) Sentinel-1 backscatter coefficient and coherence products, ii) Sentinel-2 atmospherically corrected multi-spectral imagery; along with a scene classification product that includes clouds, dark pixels etc. \\

\noindent \textbf{C - Index to ODC}
The ARD are automatically loaded into the datacube, triggering a batch process whenever a new image is downloaded and pre-processed. ODC provides two methods for loading data, indexing or ingesting. The indexing method stores only the metadata of each product in the database, while having the actual data products in a file system. On the other hand, the ingestion method stores the data in both.
According to ODC documentation\footnote{https://datacube-core.readthedocs.io/en/latest/ops/ingest.html} Cloud Optimized Storage formats combined with GDAL or other software improve the performance of reading files, making indexing the preferred choice. The indexing is done using YAML files, which are blueprints for storing each data type (i.e. coherence, backscatter, multi-spectral images).  \\

\noindent \textbf{D - Data analytics and feature engineering} DataCAP assists on the fast, easy and versatile generation of SITS feature spaces (Figure \ref{fig:architecture1} D) to feed ML pipelines. Users can execute i) a number of complex spatio-temporal queries using the LPIS (vector data) and scene classification products (i.e. cloud mask); ii) create pixel-based, object-based or patch-based feature spaces; iii) apply inward buffers to avoid mixed pixels and more.  For our two downstream tasks, we generated an \textit{object-based (mean)} Sentinel-1/Sentinel-2 time-series, from \textit{01/03/2017 - 31/10/2017} over \textit{Utrecht, Netherlands}, with max cloud coverage \textit{$>$85\%} and inward buffer \textit{10 m}.\\

\noindent \textbf{E - 1 Crop classification and smart sampling} In DataCAP we use the crop classification algorithm from Sitokonstantinou et al. (2018) that we slightly modified for mapping grasslands over Utrecht, Netherlands \cite{sitokonstantinou2018scalable}. Then the classification results are passed through the smart sampling algorithm from \cite{rousi2020semantically}. The smart sampling algorithm utilizes the posterior probabilities of the classification to return to the user the most confident mismatches, between crop predictions and farmer declarations,  for further visual inspection through DataCAP's Visualisation component. \\

\noindent \textbf{E - 2 Grassland mowing event detection} DataCAP leverages the Sen4CAP\footnote{http://esa-sen4cap.org/} grassland mowing detection algorithm. The algorithm calculates VI time-series trends and identifies abrupt drops. If the drops are larger than a pre-defined threshold then they are characterized as a mowing event.  \\

\subsection{Visualisation component}\label{sec:street}

\noindent DataCAP's visualisation component is demonstrated at \url{http://62.217.82.91/}. This demo GUI is structured based on the two downstream tasks presented earlier and enables the verification of the ML outputs via means of visual inspection. The interface offers two types of visualisations, i) time-series of Sentinel-1 and Sentinel-2 images and ii) crowdsourced street-level images. \\

\noindent Using the API of the crowdsourcing platform Mapillary, we automatically download all available street-level images over the area and time window of interest. Then, each downloaded image is matched with the corresponding LPIS object(s) it illustrates. We annotate images that are taken either towards the windshield direction (Case 1) or the window direction (Case 2). \\

\noindent We use Equations \ref{eq:lat} and \ref{eq:lon} to move the initial geo-location coordinates ($\boldsymbol{lat_{1}}$, $\boldsymbol{lon_{1}}$) to new coordinates ($\boldsymbol{lat_{2}}$, $\boldsymbol{lon_{2}}$) that are $d = 10m$ away in the direction of angle $\theta$. For Case 1, we set $\theta = compass\_angle + 45^o$ for the right half of the image and $\theta = compass\_angle - 45^o$ for left half. For Case 2 we set $\theta = compass\_angle$. 

% (\boldsymbol{$lat_{1}$}, \boldsymbol{$lon_{1}$}) to new coordinates (\boldsymbol{$lat_{2}$}, 
% \boldsymbol{$lon_{2}$}) that are  $d = 10 m$ away in the direction of angle $\theta$. For Case 1, we set $\theta = compass\_angle + 45^o$ for the right half of the image and $\theta = compass\_angle - 45^o$ for left half. For Case 2 we set $\theta = compass\_angle$. 

\begin{equation}
    \label{eq:lat}
    lat_{2} = \arcsin{(\sin{lat_{1}}\cdot\cos{\frac{d}{R})} + \cos{lat_{1}}\cdot\sin{\frac{d}{R}}\cdot\cos{\theta})}
\end{equation}

\begin{equation}
    \label{eq:lon}
    lon_{2} = lon_{1} + \arctan{(\sin{\theta}\cdot\sin{\frac{d}{R}}\cdot\cos{lat_{1}},\cos{\frac{R}{d}}-\sin{lat_{1}}\cdot\sin{lat_{2}})} 
\end{equation}

\noindent where $\boldsymbol{R}$  is the radius of the Earth. If the new coordinates of an image fall within any parcel geometry, then we match the parcel(s) label with that image. The \href{https://github.com/Agri-Hub/Callisto/tree/main/Mapillary}{\color{blue}{code}} and the produced annotated street-level \href{https://github.com/Agri-Hub/Callisto-Dataset-Collection}{\color{blue}{images}} are open.\\

\noindent \textbf{Visual inspection of parcels to verify ML results.} 
One can select the flagged parcels according to the smart sampling algorithm, and can verify if the farmer declaration or ML prediction was correct. This is done by inspecting parcel-focused time-series of Sentinel-1 and Sentinel-2 images, with an adjustable buffer around the parcel.  The time-series reveal the crop growth for each parcel, thus allowing to distinguish between different crop types. In the same manner, visual inspection of the time-series of grassland parcels can reveal sudden changes in the vegetation cover, thus indicating mowing events. Finally, the street-level images offer a very high resolution complementary information to finalise the decision.\\

\subsection{Conclusion}
All in all, DataCAP is a data handling and visualisation module for the monitoring of the CAP. DataCAP consists of i) a back-end component that helps collect and prepare satellite ARD to feed pertinent ML pipelines, and ii) a front-end component that utilizes the satellite ARD and street-level images to help verify the ML outputs. The demonstrated solution is  scalable, extendable and reproducible. DataCAP's code and produced annotated datasets are open, encouraging the data science community to exploit them in similar or other pertinent domains and applications.

%changed 20/04
\ifpdf
    \graphicspath{{Chapter3/Figs/Raster/}{Chapter3/Figs/PDF/}{Figs/Chapter3/}}
\else
    \graphicspath{{Figs/Chapter3/Vector/}{Figs/Chapter3/}}
\fi

\chapter{Food Security Monitoring}
\begin{figure}
\centering\includegraphics[scale=0.14]{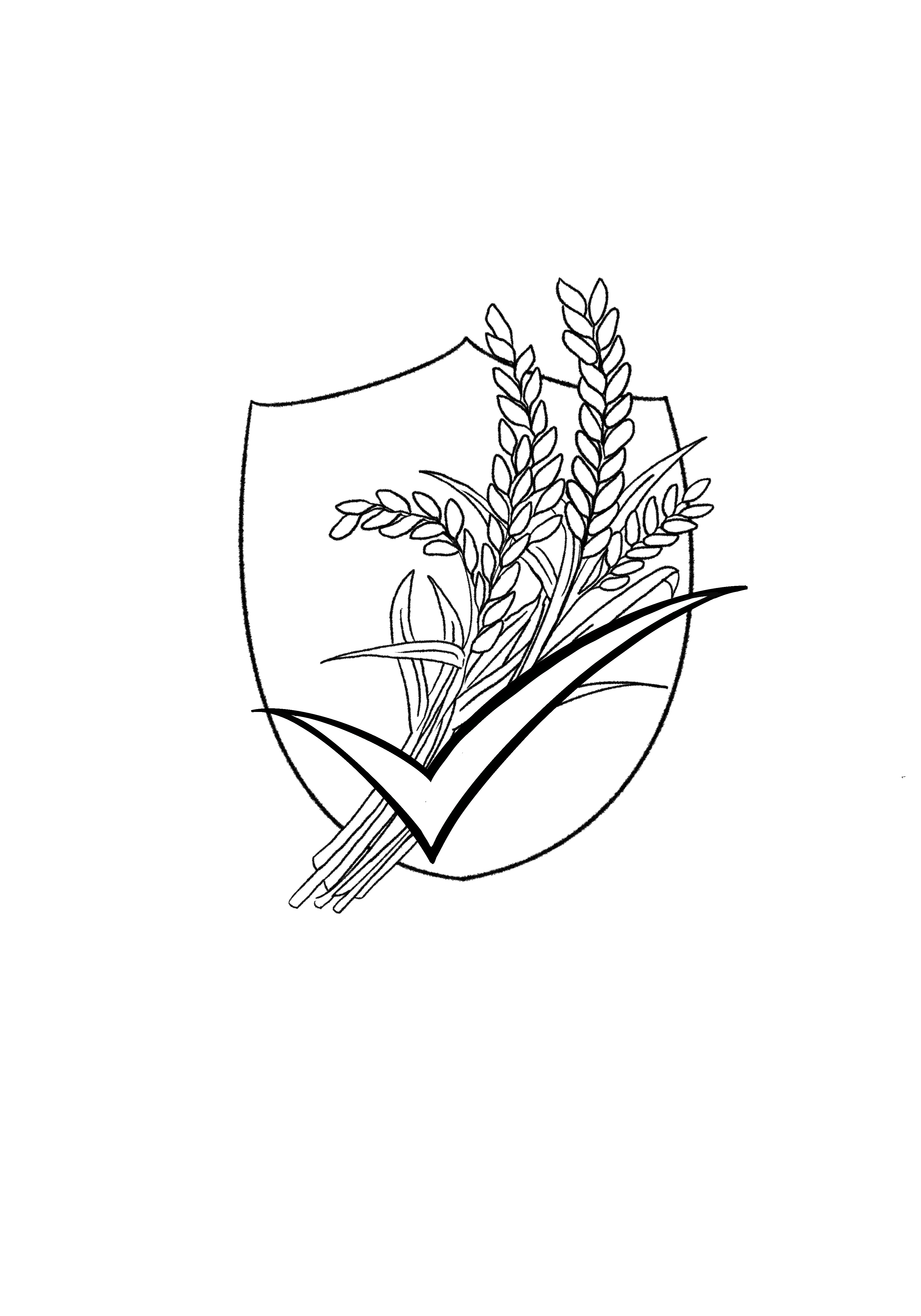}
\end{figure}

\section{Introduction}

\noindent  The demand for rice production in Asia is expected to increase by 70\% by 2050, which makes evident the need for a balanced productivity and effective food security management at national and continental level. This chapter focuses on paddy rice mapping and monitoring in South Korea. Food security monitoring demands knowledge extraction at large scales to allow for decision making at the highest level. Section \ref{ricebiomass} presents a method for monitoring the growth of rice, using the TIMESAT software on NDVI time-series, extracting phenological metrics, biomass and yield indicators. TIMESAT requires user provided parameters to define the start and the end of season to then compute the relevant metrics. In order to automate this procedure, the vegetation indices NDWI and PSRI were used to develop a data-based parameter tuning for TIMESAT.\\

\noindent  The large-scale mapping of paddy rice extent at the parcel level requires the processing of big satellite data of high spatial resolution. South Korea’s food security concerns are related to the overproduction of rice and the low self-sufficiency in the production of other major grains. For this reason, the systematic and large scale monitoring of the paddy rice extent has been identified as key knowledge for food security decision making. Section \ref{korea} reports on work that addresses the big data implications derived from a large scale and high resolution paddy rice mapping application. Towards this direction, we applied a distributed RF classifier using the cluster-computing framework Apache Spark in a high performance data analytics environment. The input data to the classifier comprise long time-series of Sentinel-1 and Sentinel-2 images, but also pertinent vegetation indices. The proposed paddy rice classification method achieves an accuracy of more than 85\%, for a study site in Northwestern South Korea.\\

\noindent Section \ref{scalablerice} builds on the work reported in Section \ref{korea} and scales the paddy rice classification to a nationwide application. We showcase the capacity of the proposed approach to cope with scarce labeled data. Specifically, we used a combination of supervised and unsupervised models. First, pseudo-labels were generated for rice classification from a single site (Seosan-Dangjin) by using a dynamic k-means clustering approach. The pseudo-labels are then used to train a RF classifier that is fine-tuned to generalize in two other sites (Haenam and Cheorwon). The optimized model was then tested against 40 labeled plots, evenly distributed across the country. The paddy rice mapping pipeline is scalable as it has been deployed in a High Performance Data Analytics (HPDA) environment using distributed implementations for both k-means and RF classifiers. When tested across the country, the proposed model provided an overall accuracy of 96.69\% and a kappa coefficient 0.87. Even more, the accurate paddy rice area mapping was returned early in the year (late July), which is key for timely decision-making. Finally, the performance of the generalized paddy rice classification model, when applied in the sites of Haenam and Cheorwon, was compared to the performance of two equivalent models that were trained with locally sampled labels. The results were comparable and highlighted the success of the model’s generalization and its applicability to other regions.

\section{Phenology, Biomass and Yield Indicators} \label{ricebiomass}

\subsection{Literature review}

\noindent The knowledge of crop’s status enables the effective decision making and planning of agricultural practices, such as the appropriate fertilization and irrigation operations, to maximize the yield \citep{porker2016improvement}. In this context, the correlation between satellite-based features, such as VIs, and biophysical parameters collected at a single date, or accumulated over some time period, is investigated. Published studies, like Lambert et al \citep{lambert2018estimating}, took advantage of Landsat and Sentinel-2 data to estimate crop yield using regression fitting of smoothed VI curves.\\

\noindent Considering the drawbacks of satellite imagery due to cloud cover, EO indices can only reveal coarse growth stages, where short-interval growth stages can be effectively identified using ancillary climate and meteorological data (e.g. air temperature). Since crops grow via the absorption of heat, this can be quantitatively measured by air temperature accumulation with the construction of accumulated growing degree days \citep{skakun2018winter}. In addition, Duchemin et al \citep{duchemin2006monitoring} tailored a first approach at coupling remote sensing vegetation indices with evapotranspiration data to investigate wheat phenology and irrigation practices in Central Morocco, revealing a strong relationship between the Leaf Area Index (LAI) and the crop coefficient. Moreover, other publications approach yield estimation through the exploitation of Crop Growth Models, such as the AquaCrop \citep{jin2016estimation,jin2017winter,silvestro2017estimating} model and the World Food Studies Simulation Model (WOFOST) \citep{ma2013assimilation,zhou2019assimilating}, in combination with satellite data, to simulate the phenology development processes and therefore predict the yield production.\\

\noindent Recent studies, employ more complex deep learning techniques in phenology extraction and yield estimation, utilizing EO data and Vegetation Indices \citep{fernandes2017sugarcane,haghverdi2018prediction,ma2016convolutional,yalcin2017plant}. Haghverdi et al \citep{haghverdi2018prediction} used Landsat 8 data, including simple ratio (SR), NDVI, green NDVI (GNDVI) and indices of greenness, wetness and soil brightness (GI, WI, SBI) as input data to an artificial neural network (ANN) in order to relate Crop Indices and field estimates and ultimately predict the yield of cotton fields. In the same context, Hulya Yalcin (2017)22 have developed a phenology recognition algorithm, using a pre-trained Convolutional Neural Network (CNN) and ground field mounted cameras to classify the phenological stages of various types of plant.\\

\noindent  This study suggests a fully automated pipeline for the continuous and timely monitoring of rice in South Korea, to produce yield estimation proxies and indicators. In this context, this study is as an extension of the transferable and scalable solution to address the Big Data issues associated with large-scale agriculture monitoring, as proposed by Sitokonstantinou, et al.\citep{sito2020}. In this publication, the authors have used a HPDA environment to enable the management of long time-series of Copernicus satellite data, required for the effective and efficient monitoring of food security. The work reported herein,builds upon the published results of distributed RF rice classification, using the cluster computing framework of Apache Spark \citep{sito2020}. The extracted rice extent for the area of interest is utilized here to subsequently extract phenological information based on the analyses of Sentinel-2 NDVI time-series. Thematic maps of phenological metrics are produced representing different stages of crop growth, such as start, peak and end of season, as well as two proxies of accumulated biomass and produced yield.

\subsection{Materials}
\subsubsection{Area of interest}
\begin{figure}[htbp!] 
\centering    
\includegraphics[width=0.85\textwidth]{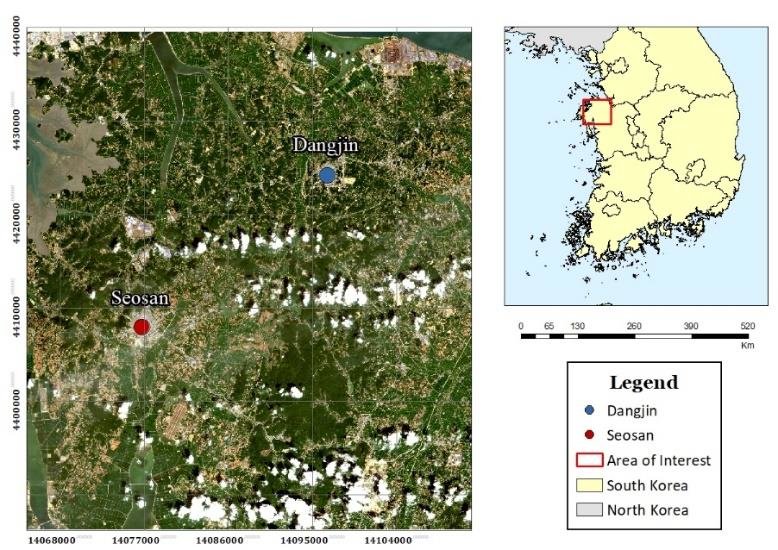}
\caption{Area of interest located in the regions of Dangjin and Seosan in South Korea.}
\label{fig:dangjin_seosan_timesat}
\end{figure}

\noindent The area of interest for the present study is located at the western part of South Korea, and that is the districts of Dangjin and Seosan. These two areas represent two of the highest rice producers of the country \citep{park2018classification} .The annual mean temperature in Dangjin and Seosan is 11.4\textsuperscript{o}C, while the annual precipitation is approximately 1158.7mm \citep{kim2017monitoring}. South Korea’s climate is highly affected by the Asian Monsoon. During the winter cold air masses dominate, while during the summer, warm and moist air masses reach the country from tropical regions. For this reason, the rainiest season in South Korea is during the summer, which results in scarcity of cloud free optical imagery. Figure \ref{fig:dangjin_seosan_timesat} presents the area of interest for which the phenology extraction
approach was implemented.

\subsubsection{Data}
\begin{figure}[htbp!] 
\centering    
\includegraphics[width=0.9\textwidth]{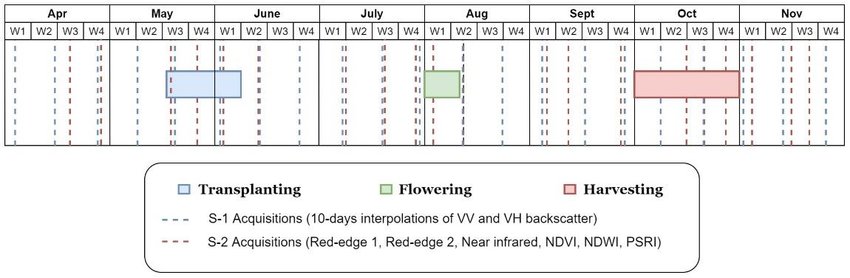}
\caption{South Korea rice paddy growth phases and the respective Sentinel-1 and Sentinel-2 acquisitions.}
\label{fig:growth_phases_sentinels}
\end{figure}

\noindent The input dataset for the rice classification, as described in Sitokonstantinou et al \citep{sito2020}, which was used to extract the rice extent for this study, comprises of a time-series of Sentinel1 VV backscatters, Sentinel 2 images (all bands except B01, B09, B10), and pertinent Vegetation Indices that have been widely used in crop monitoring and crop mapping applications \citep{lebourgeois2017combined,sitokonstantinou2018scalable}. The aforementioned feature space amounted to more than 80 GB, which justifies then need for distributed processing, as described in the first section. The phenology extraction methodology in this study, makes use of three Sentinel-2 Vegetation Indices, including the Normalized Difference Vegetation index (NDVI), Normalized Difference Water Index (NDWI) and Plant Senescence Reflectance Index (PSRI), in order to identify the beginning, as well as the end of the crop’s growth cycle. The images have been atmospherically corrected using the Sen2Cor tool and clouds have been masked out using the Sen2Cor Scene Classification product, including cloud shadows, dark pixels, high and low probability clouds, and cirrus clouds. Finally, the spatial resolution of all indices has been resampled to 10 m and amounts to several tens of gigabytes of feature space. Figure (\ref{fig:growth_phases_sentinels} illustrates how the satellite image acquisitions capture the different phenological phases in South Korea.

\subsection{Methods}
\noindent The phenology extraction methodology was implemented using TIMESAT \citep{fe7756f3b0d54bd0ad186c4aec4f1012}, a software package that smooth satellite derived time-series data, using regression fitting. It provides three different fitting approaches, which are based on the least square fits to the upper envelope of the time-series data. In this work, Normalized Difference Vegetation Index (NDVI) time-series data were analyzed in order to obtain growth information for rice. Among the three available processing methods (Savitzky Golay, assymetric Gaussian and double logistic), the adaptive Savitzky Golay filter was used, as it captures both subtle and rapid changes in the time-series, providing a better understanding for the beginning and for the ending of the growing season. Moreover, the adaptive Savitzky Golay filter, as a form of moving average, succeeds in smoothing the NDVI curves, and at the same time fitting the time-series without reshaping the raw data After the NDVI curves are smoothed, a number of phenological metrics are computed, as shown in Table 1.

\subsubsection{Data pre-processing}
\noindent In order to analyze the NDVI curves, TIMESAT needs even spacing between the points that comprise the time-series. For this reason, several interpolation methods were investigated. Ultimately, the weighted average and linear interpolation methods were utilized in order to create decadal time-series from the raw data acquisitions. The weighted average interpolation method fills the fixed timestamps (decadal time-series) that fall within a ten day range of the acquisition, and then linear interpolation follows to fill any outstanding timestamps.

\subsubsection{TIMESAT implementation}
\noindent Initially, median filtering method is employed with multiple iterations for spike removal, as outliers may seriously degrade the final function fit and interfere with the final width and weight of the annual NDVI curve. Then the Savitzky Golay function is applied to smooth the timeseries curve.\\

\begin{figure}[htbp!] 
\centering    
\includegraphics[width=0.65\textwidth]{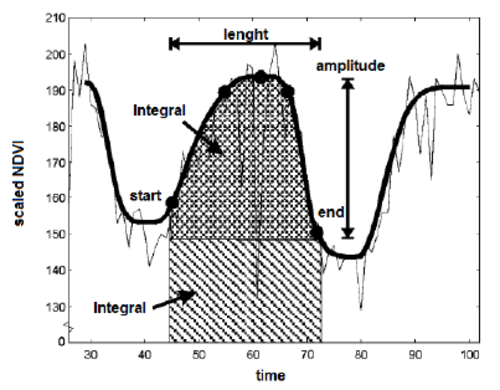}
\caption{TIMESAT \citep{penatti2012subdivision} smoothing process and the calculated phenological metrics.}
\label{fig:timesat}
\end{figure}

\noindent In order to extract the phenological metrics, TIMESAT requires a user-defined parameter, which describes the season start (emergence) and end (harvest). In more detail, TIMESAT requires a user specified threshold of i) the distance between the left minimum level and the maximum of the smoothed curve for the time of the start of the season (SOS) and ii) the distance between the right minimum level and the maximum for the time of the end of season (EOS). In order to automate this procedure, and create a generalized crop growth retrieval methodology, this study investigates two more vegetation indices (NDWI and PSRI), which have been preprocessed identically to the NDVI timeseries. Analyzing the time-series of NDWI and PSRI and considering the temporal cultivation practices relevant for the rice crop in South Korea, the SOS and EOS parameters were determined. Specifically, the transplanting (emergence) practice was identified at the intersection of NDWI and NDVI. Prior to the intersection of the curves, NDWI rises as the plots are flooded. Then NDWI drops and NDVI rises, indicating that transplanting has taken place. In the same fashion, harvesting is identified close to the intersection of PSRI and NDVI. PSRI rises, while NDVI decreases due to the yellowing of the crop. Then both NDVI and PSRI drop, indicating that harvest took place.\\

\begin{figure}[htbp!] 
\centering    
\includegraphics[width=0.8\textwidth]{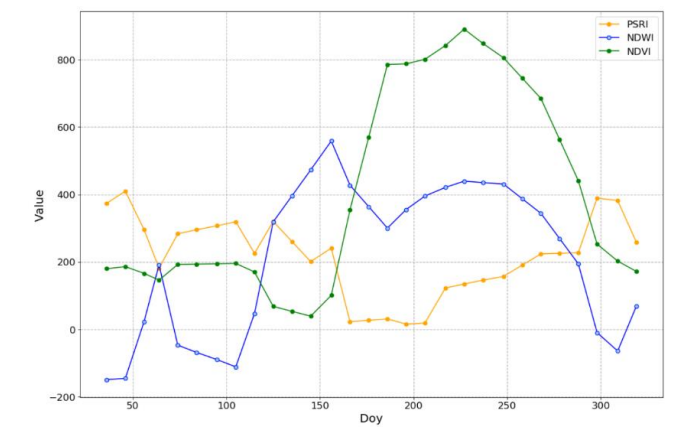}
\caption{Interpolated time-series of NDVI, NDWI and PSRI of one parcel identifying major cultivation
practices.}
\label{fig:interpolated_ts}
\end{figure}

\noindent The products of TIMESAT are eleven phenological metrics, six of which were used in this study, as seasonality parameters. The most important of them being the start, peak and end of season, measured in Days of Year (DOY), and the rate of increase and decrease and the large and small integrals (Table1 and Figure \ref{fig:timesat}). The rate of increase in NDVI during the beginning of the season can be related to the physiognomy of the vegetation, and the green up rate, as it describes the velocity at which the crop moves from the germination stage to the growth stage. On the other hand, the rate of decrease describes the rate of senescence at which the crop shifts from the flowering phase to the ripening stage, where the harvesting begins. Additionally, the large season integral (LSI) can be used as a proxy of the relative amount of vegetation biomass without regarding the minimum values, whereas the small season integral (SSI) as a proxy of the relative amount of vegetation of biomass while regarding the minimum values \citep{araya2018cropphenology}.\\

\noindent The definition of the SOS and the POS can be characterized as crucial, as the left integral defined in combination with the base level, establish a total accumulated biomass indicator of the plant. At the same time, after the finish of the phenological cycle of the crop, the POS and the EOS along with the base level, define a crop production indicator \citep{araya2018cropphenology}. In addition, high values indicate smaller reduction in the accumulated biomass after the flowering stage of the plant, which are associated with a slower and longer process of producing shoots and fruits, whereas small values represent a shorter process that has not been completed filling shoots and associated with low yield

\subsection{Results}
\noindent Using this study’s method the start, peak and end DOYs of the crop season have been identified. In addition, based on the extracted SOS and EOS, a biomass and a yield indicator are calculated from the NDVI time-series, using the TIMESAT solution. The biomass indicator can be derived at the peak of the season. Therefore, detecting the onset of the peak of season using appropriate rules, can make pertinent decision making possible, even early in the year.\\

\noindent The study of the NDVI profile can identify the three main crop growth phases. After transplanting, which in our case can specify the start of season, the rice crop is at the early vegetative phase, as germination at the seedbeds has proceeded, where the tillering stage begins until the stem elongation of the crops\citep{duchemin2006monitoring}. Subsequently, the reproductive phase follows, where the peak of season is noted with the highest value of the NDVI time-series, when the crop is at the heading stage\citep{parul2017rice}. With respect to the end of season, is referred to the DOY where the crop is mature enough to be harvested and thus at the ripening phase.\\

\begin{figure}[htbp!] 
\centering    
\includegraphics[width=1.0\textwidth]{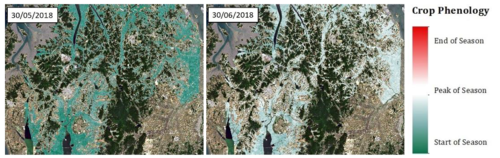}
\caption{Resulted start of season identifying the vegetative phase.}
\label{fig:start_season_veg_phase}
\end{figure}

\begin{figure}[htbp!] 
\centering    
\includegraphics[width=1.0\textwidth]{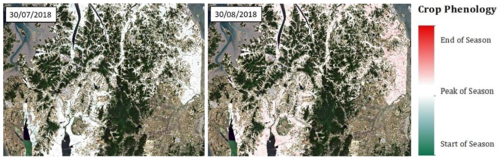}
\caption{Resulted peak of season identifying the reproductive phase}
\label{fig:peak_season_veg_phase}
\end{figure}

\begin{figure}[htbp!] 
\centering    
\includegraphics[width=1.0\textwidth]{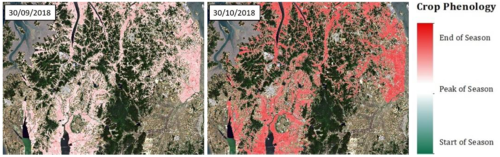}
\caption{Resulted end of season identifying the ripening phase}
\label{fig:end_season_veg_phase}
\end{figure}

\noindent The study of the NDVI time-series was implemented at the object level, by creating objects of
7x7 pixels. Figures \ref{fig:start_season_veg_phase}, \ref{fig:peak_season_veg_phase} and \ref{fig:end_season_veg_phase} illustrate the evolution of rice growth, together with the crop phase. The color pallet, in Figures \ref{fig:start_season_veg_phase}, \ref{fig:peak_season_veg_phase} and \ref{fig:end_season_veg_phase}, illustrates the differences of NDVI values with the peak (highest) NDVI value for every object.\\

\noindent The Figures \ref{fig:start_season_veg_phase}, \ref{fig:peak_season_veg_phase} and \ref{fig:end_season_veg_phase} reveal the development of the rice crop, highlighting with green color the start of season; with white color, the passing to the peak of season; and with red color, when the crop reaches the end of season. The different colors showcase that the rice fields can be at different growth phases at a specific DOY. 

\subsubsection{Phenology Metrics}
\begin{figure}[!ht] 
\centering    
\includegraphics[width=0.8\textwidth]{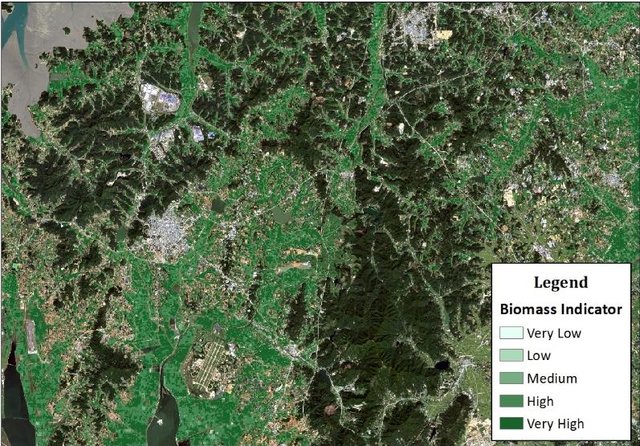}
\caption{The Biomass Indicator results for the district of Dangjin.}
\label{fig:biomass_indicator}
\end{figure}

\noindent As mentioned earlier, the two essential proxies of biomass and yield indicators are obtained from the calculation of start, peak and end of season. These two are produced from the calculation of the left and right integral, between start-peak and peak-end with the base level, respectively. A high value of the biomass indicator implies a high accumulated biomass until the peak of season, which can be used as an early indicator of the final production (Figure \ref{fig:biomass_indicator}.\\

\begin{figure}[!ht] 
\centering    
\includegraphics[width=0.8\textwidth]{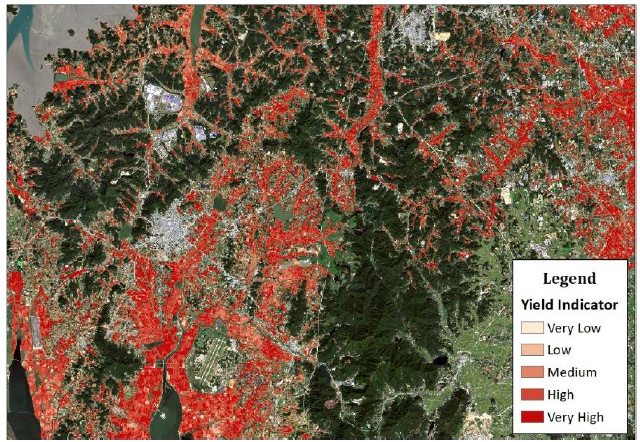}
\caption{ The Yield Indicator results for the district of Dangjin.}
\label{fig:yield_indicator}
\end{figure}

\noindent Moreover, at the end of season, the yield indicator is calculated, as a proxy of the final production. The figure \ref{fig:yield_indicator} of the yield indicator illustrates with deep red color the areas with expectedly high final production. This indicator can be utilized only as a proxy, since it is derived exclusively from the study of the NDVI curve, without considering any other parameters.

\subsubsection{Validation}
\noindent The validation of the rice status indicators was challenging, since no ground truth information was available. The absence of validation crop-calendars for the rice crop at the area of interest, led to the investigation of other means of evaluation. The analysis of the related bibliography provided a benchmark for the primary results of the crop indicators, and that is SOS and EOS. As mentioned by Park et al. \citep{li2014analysis}, based on the heading period or the growing days, the rice (Oryza sativa L.), which is cultivated widely in South Korea, can be grouped into early maturing, medium maturing and medium-late-maturing cultivars, which result in different transplanting periods, since the transplanting period is determined by the relative growth difference of rice plants at a certain time. Gutierrez et al. \citep{gutierrez2013effect} highlight that late-maturing fields are flooded twenty-four days more than the early maturing ones. Specifically, as claimed by Jang et al. \citep{jang2012mapping} , in the district of Dangjin, mainly early and medium-late maturing rice crops are cultivated. Transplanting starts at the middle of May until the early of June for the medium-late maturing and middle until end of June for the early maturing cultivars. With respect to the harvesting practices, Kim et al. \citep{kim2015sensitivity} state that it occurs in October for the medium-late maturing and starts in middle of September for the early maturing cultivars.More than 90{\%} of the objects were found to have SOS and EOS that fall within the aforementioned periods.\\

\begin{figure}[htbp!] 
\centering    
\includegraphics[width=0.75\textwidth]{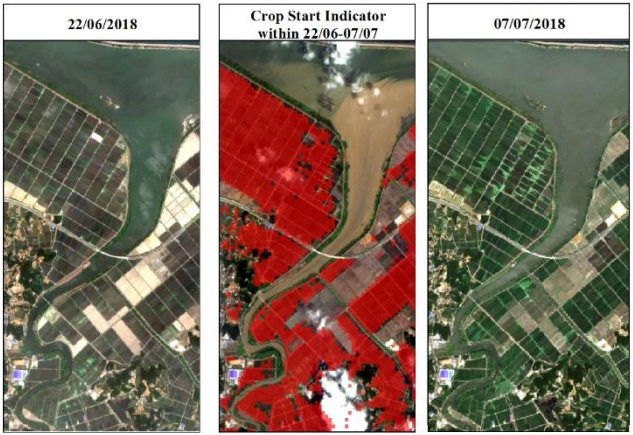}
\caption{Validation of Crop Start Indicator results through the available cloud free imagery.}
\label{fig:validation_crop_Start_indicator}
\end{figure}

\begin{figure}[htbp!] 
\centering    
\includegraphics[width=0.75\textwidth]{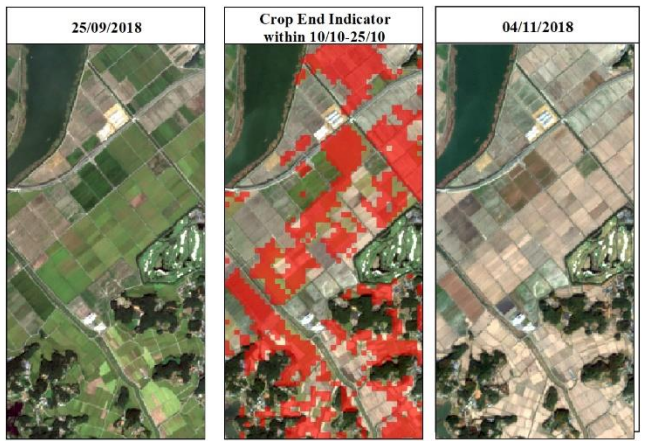}
\caption{Validation of Crop End Indicator results through the available cloud free imagery.}
\label{fig:validation_crop_End_indicator}
\end{figure}

\noindent Photointerpretation was additionally performed to visually inspect the aforementioned agreement with the periods of transplanting and harvesting mentioned in the literature. Indeed, Figures \ref{fig:validation_crop_Start_indicator} and \ref{fig:validation_crop_End_indicator} reveal that TIMESAT has correctly identified the SOS and EOS timestamps. In Figure \ref{fig:validation_crop_Start_indicator}, it is clearly shown that transplanting occurs sometime between the 22-07/06, while Figure \ref{fig:validation_crop_End_indicator} indicates that harvest takes place sometime between 10-15/10.

\subsection{Conclusions}
\noindent This work, as an extension of the results, described in Sitokonstantinou et al \citep{sito2020} , adds to
an end-to-end Sentinel based agriculture monitoring scheme for the monitoring of food security, but also the control of the new CAP. The suggested scheme describes a transferable and scalable agriculture monitoring system, accommodating for any Big Data considerations. Additionally, the proposed framework addresses the needs of food security and sustainable agriculture stakeholders, such as decision and policy makers, environmental ministries and policy implementation bodies, via providing accurate information, based on which timely actions can be undertaken. In the future, the proposed algorithms will be applied at larger scales, showcasing the scalability of the overall system. At the same time, the scale of thematic information will be increased to the intra-parcel level, incorporating climatic and meteorological data (i.e. accumulated growing degree days) that come from in-situ sensors.
%%%%
\section{Distributed Crop Classification}\label{korea}

\subsection{Literature review}

\noindent Paddy rice is a primary crop in South Korea, which is the area of interest for this study, and information about its spatial distribution and yield is of great importance for the environmental management, but also for food security related decision-making. South Korea, although of high food security index, has low food self-sufficiency that is decreasing in the long term. The country can be considered exposed to potential food security issues due to its high dependency on international supply for major crops and limited number of exporting countries for rice.\\

\noindent The monitoring of paddy rice extent and its overall production requires manual field visits for survey and is thus costly and time consuming, when compared to information gathered through EO means. Remote sensing is one of the most effective technologies to map the extent of crops. Rice area mapping at the parcel, regional and national scale has been extensively studied in the past, through several approaches, including mono-temporal and multi-temporal classification schemes that utilize both optical and microwave - SAR data \citep{qin2015mapping,Nguyen_2015}. \citep{Tian_2018} have introduced a novel multi-season paddy rice mapping method, using Sentinel-1 and Landsat-8 data under a k-means unsupervised classification methodology. Torbick et al. \citep{torbick2017monitoring} have produced an updated land cover map, including the rice class, fusing Sentinel-1, Landsat-8 OLI and PALSAR-2 data using a Random Forest classifier. Finally, Pazhanivelan et al. \citep{Pazhanivelan} have introduced a robust rule-based classification for mapping rice area with multi-temporal, X-band, HH polarized SAR imagery (COSMO Skymed and TerraSAR X), with site-specific parameters. \\

\noindent There have been multiple studies to exploit both optical and SAR data for the monitoring of agriculture, with a lot of research on the usage of Sentinel data, utilizing their unprecedented characteristics in temporal and spatial resolution \citep{inglada2016improved,immitzer2016first,sitokonstantinou2018scalable}. Studies have used multiple classification techniques including neural networks, supervised and unsupervised machine learning techniques, but also custom rule-based systems. Nevertheless, all studies had to manually collect their training and validation samples, undermining the design of a fully transferable and site independent framework of application for the described systems.\\

\noindent In this regard, and under considerations of scalability, reproducibility and transferability, which are essential for a national scale application, it was decided to design a more dynamic rice monitoring system that is largely independent of hard-to-attain non-EO information. In this work we suggest a land cover map update mechanism based on change detection to produce appropriate training and validation datasets for any given year of inspection via utilizing older ground truth information. The updated land cover map is then used to train a distributed Random Forest classifier in a HPDA environment. \\

\noindent The recent and unprecedented availability to high resolution satellite imagery, such as the Sentinels, has introduced a paradigm shift in the field of remote sensing. Increasing number of satellites and sensors along with improved spatial and temporal resolution generate big EO data that demand increased computational power for their exploitation \citep{chebbi2015big}. To accommodate for the storage and processing requirements, in this new era of EO science, decentralized and distributed environments and frameworks are utilized. High Performance Computing (HPC) combines technologies such system software, architecture and algorithms in order to increase the effectiveness and the speed of complex processing chains. Distributed processing is part of this umbrella technology. It makes use of parallel processing on multiple machines so that the data are distributed to the entire memory of the system. \\

\noindent Remote sensing research has been enhanced in recent years by high performance computing \citep{lee2011recent}. Alongside with the infrastructure, several computing frameworks have been developed, which promise more effective and efficient image processing. Apache Hadoop is one of the most famous implementations of the MapReduce model, introduced by Dean and Ghemawat \citep{dean2008mapreduce}. Apache Spark, an alternative framework, has advantages over Hadoop when it comes to speed, memory and high level operators, making it the best choice for big data tasks \citep{jianwu2016emerging} and the chosen framework for this work. \\

\subsection{Area of Interest and Data Acquisition}

\subsubsection{Area of Interest}

\begin{figure}[htbp!] 
\centering    
\includegraphics[width=1.0\textwidth]{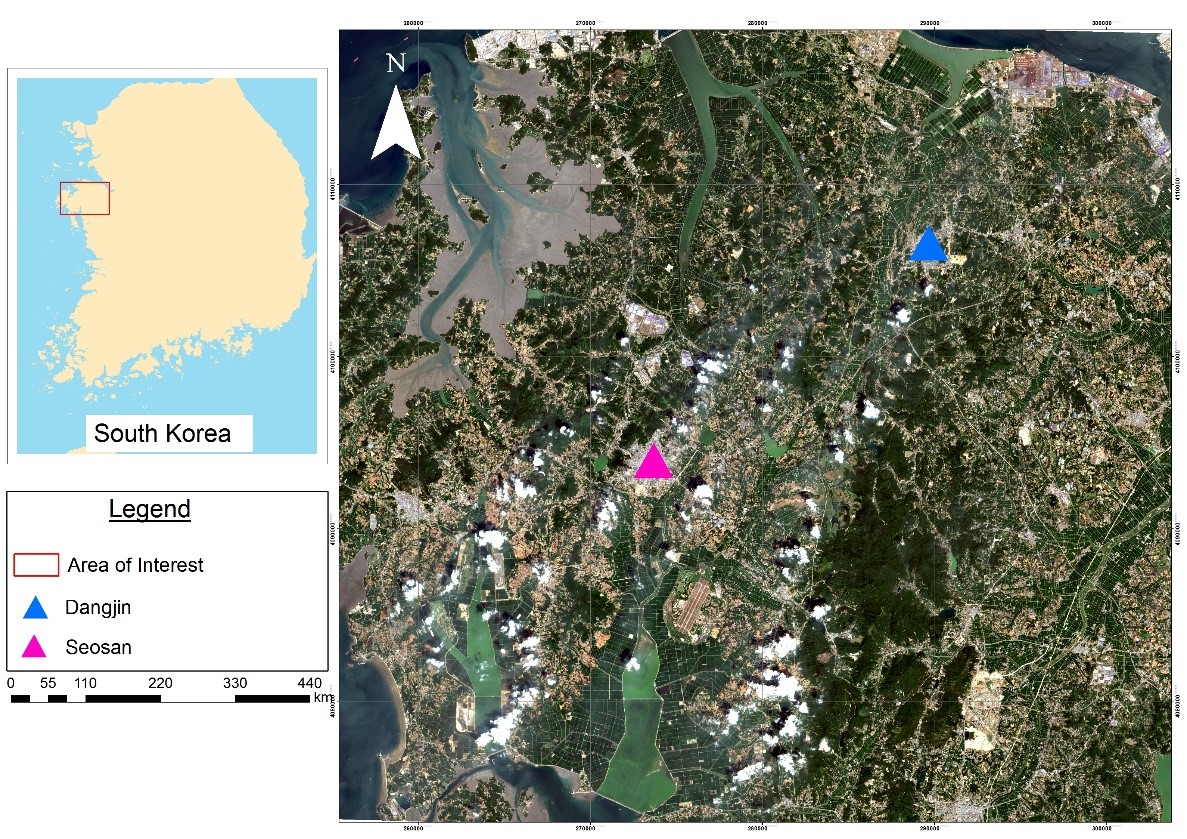}
\caption{Area of interest located in the regions of Dangjin and Seosan in South Korea.}
\label{fig:dangjin_seosan}
\end{figure}

The area of interest for this study comprises of the South Korean regions of Dangjin and Seosan. The two areas are recorded as the highest rice producing in the country \citep{area2}. The annual mean temperature and precipitation in the region are 11.4 °C and 1158.7 mm, respectively \citep{kim2017monitoring}. South Korea has a rainy season during the summer due to the Asian Monsoon, therefore cloud free optical imagery is scarce. For this reason, weather independent SAR imagery from Sentinel-1 has also been exploited in the suggested methods of this study. The figure \ref{fig:dangjin_seosan} illustrates the area of interest for which an accurate paddy rice map has been produced.\\

\subsubsection{Data acquisition}
\noindent The area of interest for this study comprises of the South Korean regions of Dangjin and Seosan. The two areas are recorded as the highest rice producing in the country \citep{area2}. The annual mean temperature and precipitation in the region are 11.4 °C and 1158.7 mm, respectively \citep{kim2017monitoring}. South Korea has a rainy season during the summer due to the Asian Monsoon, therefore cloud free optical imagery is scarce. For this reason, weather independent SAR imagery from Sentinel-1 has also been exploited in the suggested methods of this study. The figure below illustrates the area of interest for which an accurate paddy rice map has been produced.\\

\noindent Secondly, the input dataset for the paddy rice classification algorithm comprises of a time-series of Sentinel-1 and Sentinel-2 imagery to capture the entirety of the crop’s phenology. For the creation of the pixel based feature space, monthly means of VV backscatter from Sentinel-1 images, along with Red-edge, Near Infrared (NIR) bands and VIs from Sentinel-2 images, were used as features. The VIs incorporated in the feature space include the NDVI, NDWI and PSRI, which have been widely used in crop monitoring and crop mapping applications \citep{lebourgeois2017combined,sitokonstantinou2018scalable,gao1996ndwi,hatfield2010value}. The resolution of the time-series of imagery is at 10 m and amounts to several tens of gigabytes of feature space, which is comprised of a total of 167 features. The figure below illustrates how the satellite image acquisitions capture the different phenological phases of rice in South Korea. \\

\begin{figure}[htbp!] 
\centering    
\includegraphics[width=1.0\textwidth]{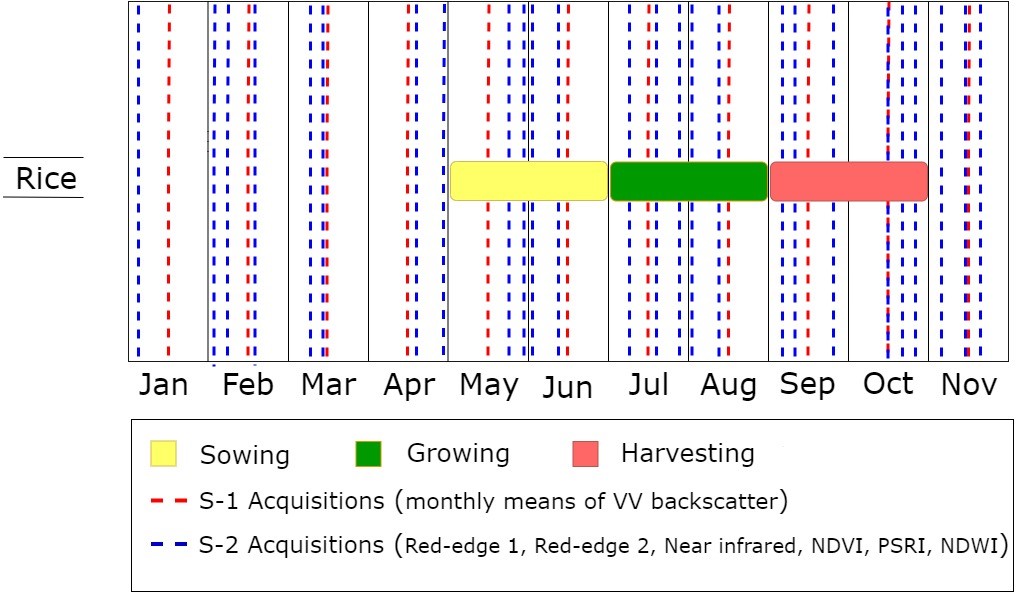}
\caption{South Korea paddy rice phenology and the respective Sentinel-1 and Sentinel-2 acquisitions that capture it}
\label{fig:rice_phenology}
\end{figure}

\noindent The recent and unprecedented availability of open access high resolution satellite imagery, such as the Sentinels, has introduced a new era in the field of remote sensing. However, the automated and timely acquisition of Sentinel imagery from the plethora of available hubs becomes challenging. The different hubs offer Sentinel data with different specifications, such as their rolling archive policy, data availability, geographic coverage and latency of acquisition. Thus, an application has been developed to connect to multiple Sentinel hubs and automatically search for the pertinent Sentinel data. This broker of Sentinel data retrieves the requested products from the most efficient hub that is decided in terms of download speed and product availability. The overall architecture of this Sentinel hubs broker is shown in the figure below. 

\begin{figure}[htbp!] 
\centering    
\includegraphics[width=0.8\textwidth]{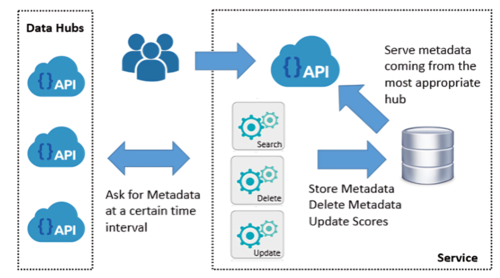}
\caption{The federated Sentinel hubs API overall architecture.}
\label{fig:umbrella}
\end{figure}

\subsection{Methods}

\subsubsection{Change Detection: Land cover map update}

\noindent The output product of the below methodology does not attempt to accurately classify rice fields for the year of inspection but merely delete changes in the land cover map of 2015 (reference map), as described in the Area of Interest and Data Acquisition section. Therefore, using only the rice pixels from the reference map we eliminate outliers to produce the updated map for the year of inspection. This way the training dataset for the distributed Random Forest classifier is refined, by removing rice pixels of changed land cover. \\

\begin{figure}[htbp!] 
\centering    
\includegraphics[width=0.75\textwidth]{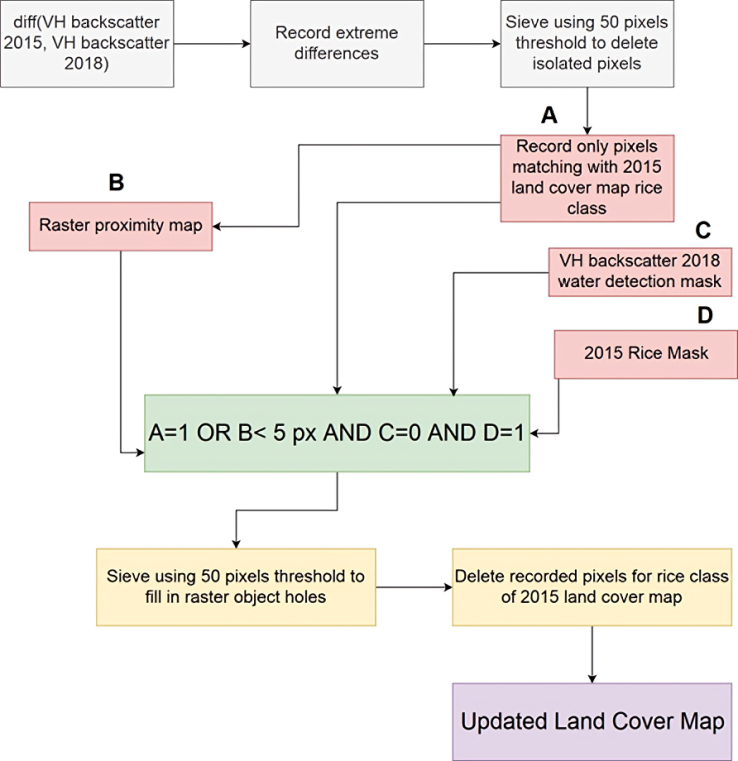}
\caption{Framework of land cover map update method.}
\label{fig:land_cover_map_framework}
\end{figure}

\noindent Figure \ref{fig:land_cover_map_framework} illustrates the workflow for the land cover map update methodology. The processing chain starts by taking the difference of the VH polarized backscatter of Sentinel-1 images in mid-June of 2015 and 2018, respectively. VH polarization is especially sensitive in the water concentration of inundated rice fields. Imagery sensed in mid-June was selected, as it is the period that rice fields are flooded and thus more easily detectable. The preprocessing of Sentinel-1 imagery involved calibration, speckle filtering and terrain correction. The two images have been additionally co-registered in order to ensure one to one pixel match.\\

\noindent From the difference product of the two VH backscatter images we record only the pixels of extreme differences, using the values of bottom 2\% and top 98\% of the cell data range as thresholds. Then raster polygons smaller than 50 pixels are replaced with the pixel value of the largest neighbor polygon, thus eliminating island pixels and filling raster polygon holes. The remaining outlier pixels are superimposed with the rice pixels of the reference map, keeping only matching records. Hence, we end up with a set of pixels identified as rice in the reference map that appear changed for the year of inspection, as illustrated in the following figure.\\

\begin{figure}[htbp!] 
\centering    
\includegraphics[width=1.0\textwidth]{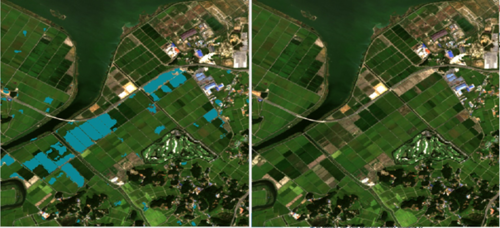}
\caption{Unrefined outlier detection product.}
\label{fig:unrefined_outlier_detection}
\end{figure}

\noindent To further refine the product illustrated above, it was attempted to fill in missing pixels within the identified non-rice parcels. A raster proximity map was generated to indicate the distance from each pixel to the nearest pixels identified as an outlier. Then pixels that haven not yet been identified as outliers and have a distance of fewer than 5 pixels from outlier pixels are recorded (B in Figure \ref{fig:land_cover_map_framework}). Product C in Figure \ref{fig:land_cover_map_framework} refers to an automated water detection map via thresholding, based on the VH backscatter, sensed in mid June of the inspection year. Since rice is inundated the algorithm also works for identifying the flooded agricultural land.\\

\begin{figure}[htbp!] 
\centering    
\includegraphics[width=1.0\textwidth]{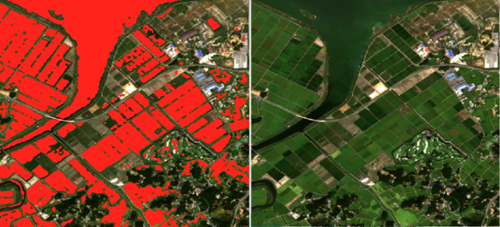}
\caption{VH backscatter 2018 water mask via thresholding.}
\label{fig:vh_2018_water_mask}
\end{figure}

\noindent Finally, product D in Figure \ref{fig:land_cover_map_framework} refers to rice pixels of the reference map. Combining products A-D under the Boolean expression in Figure \ref{fig:land_cover_map_framework}, we record all outlier pixels of the unrefined detection product (Figure \ref{fig:unrefined_outlier_detection}), but also pixels that are concurrently not included in the threshold based water mask and have a proximity value of fewer than 5 pixels (Figure \ref{fig:vh_2018_water_mask}). The final product is illustrated in Figure 8 in the Results section.

\subsubsection{Distributed Random Forest}

\noindent Food security monitoring at the national scale demands time series analysis of multiple satellite images, so as to capture the crop’s phenology. It becomes apparent that the size of this dataset, which is several tens of gigabytes, exceeds the memory limits of conventional machines. To overcome this limitation, distributed architecture and parallel data processing had to be introduced. From the storing point of view, the Hadoop Distributed File System (HDFS) is used to store large amounts of data \citep{ruzgas2016big} giving the potential of breaking down the dataset to separate blocks and distribute them to the multiple nodes of the cluster. Moving on to the processing of the dataset, Apache Spark reads the distributed pixel-based feature space from HDFS in the form of dataframes and transforms them into Resilient Distributed Datasets (RDDs), a read-only, partitioned collection of records, which allows for transformations and actions, such as filtering and sorting. Afterwards, data are prepared for the classification process by randomly splitting them into training and test subsets, with a split ratio of 30/70. \\

\begin{figure}[htbp!] 
\centering    
\includegraphics[width=1.0\textwidth]{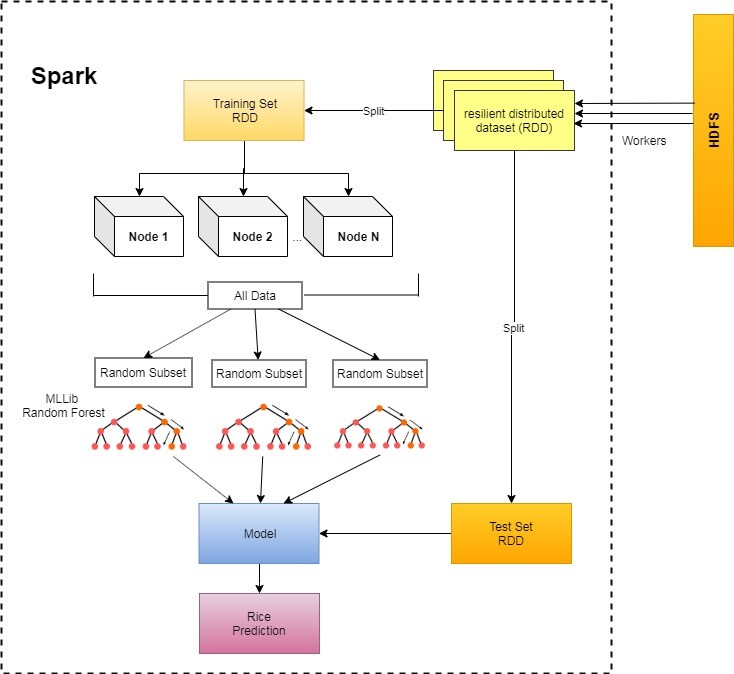}
\caption{The overall architecture of the distributed Random Forest classification.}
\label{fig:distributed_rf_architecture}
\end{figure}

\noindent Considering the need of efficient and fast data analysis for this oversized dataset, an optimized and scalable machine learning algorithm was used to exploit the power of distributed computing. The distributed Random Forest implementation comes with MLLib, a library of Apache Spark. The Random Forest algorithm is an ensemble classifier based on the decision tree model. Training data are split into k different subsets, using the bootstrap method, and each subset creates a new decision tree. Finally, a forest is constructed from the different decision trees and predictions are made through a majority voting mechanism \citep{breiman2001random}. The algorithm was parameterized to have 10 decision trees, a maximum tree depth of 30 and a maximum number of bins set to 32. The overall architecture of the system is depicted in Figure \ref{fig:distributed_rf_architecture}. 

\subsection{Results}

\begin{figure}[htbp!] 
\centering    
\includegraphics[width=1.0\textwidth]{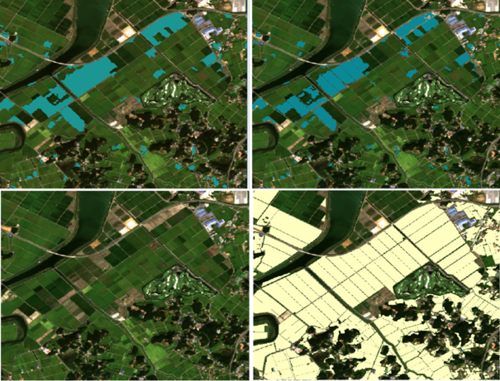}
\caption{Top left - refined outliers detection, top right – unrefined outliers detection, bottom left – true colour composite September 2018, bottom right – rice pixels based on the reference map.}
\label{fig:rice_pixels_detection}
\end{figure}

\noindent The figure below (Figure \ref{fig:rice_pixels_detection}) illustrates the final refined outliers detection output at the top left corner. Rice fields based on the reference map and the unrefined outlier detection products are also shown for visual comparison.\\

\noindent The illustrated figures throughout this section represent an indicative snapshot of the area of interest. The method was tested in a total area of 7,185 ha of paddy rice. The detected outlier pixels amounted to 396 ha, therefore resulting in a 5.5\% change. This percentage does not only refer to rice parcels that altered to other than rice cultivations in 2018, but also pixels on the parcel borders that were classified as rice due to less than optimal parcel digitization of the reference map. This further refines and removes noise from the training dataset.

% \subsubsection{Paddy rice classification}
\subsubsection{Evaluation of the Paddy rice classification}

\noindent The classification performance was assessed based on the metrics of recall, precision and F1-score, computed as shown in Equations \ref{precision},\ref{recall} and \ref{f1}. F1- score is the harmonic mean of precision and recall \citep{sokolova2009systematic}. The metrics were computed based on the below confusion matrix. Rice is considered as the positive class and non-rice as the negative class. 

\begin{table}[!ht]
\caption{Confusion matrix of paddy rice classification.}
\label{tab:confusion_matrix_rice }
\centering
\scalebox{1}{
\begin{tabular}{|c|c|c|}
\hline
\textbf{}& 
\textbf{Prediction Rice} & \textbf{Prediction Non-Rice}\\ \hline
\textbf{Truth Rice} & 4,151,744 & 616,558\\ \hline
\textbf{Truth Non-Rice} & 697,320 & 19,547,543\\ \hline
\end{tabular}}
\end{table}

\begin{equation}\label{precision}
  Precision = \frac{TP}{TP + FP}
\end{equation}

\begin{equation}\label{recall}
  Recall = \frac{TP}{TP + FN}
\end{equation}

\begin{equation}\label{f1}
  F1\textsubscript{score} = 2 \times \frac{Precision \times Recall}{Precision + Recall}
\end{equation}

\noindent where TP is the number of correctly classified rice pixels or True Positive, FP is the number of non-rice pixels classified as rice or False Positive, TN the number of correctly classified non-rice pixels or True Negative and FN is the number of rice pixels classified as non-rice or False Negative.\\

\noindent The results are very promising, with more than 85\% precision and recall for the paddy rice class. It should be noted that the above accuracy metrics are computed against the updated map, as described in the Methods section, and not a validated dataset. The validity of the updated map is thus assumed for training and validation, even though it is not perfectly representative of the truth. The dataset does not account for the new rice cultivations in the year of inspection that were described as non-rice cultivations in the reference map, as the change detection method only removes outliers from the latter. For this reason the actual precision values are expected to be slightly higher than the ones indicated.

\subsection{Conclusions}

\noindent The distributed Random Forest implementation in the HPDA environment accommodates for the big EO data and allows for the large scale pixel-based paddy rice classification. The regional application presented in this work can be linearly scaled up for the entire country, providing an accurate indication of the total paddy rice extent. The classification product can then be used for the estimation of yield, providing essential information for food security and enabling high level decisions at the national scale.\\

\noindent The exploitation of exclusively freely available data, along with the employment of big data technologies, such HDFS and Apache Spark, but also big data infrastructure, such as the HPDA, constitutes the suggested rice monitoring scheme a scalable and transferable solution.

%%%%% Added 27/04 - A scalable ML pipeline 2021

\section{Scalable Paddy Rice Mapping}\label{scalablerice}
\subsection{Literature review}\label{sec:introduction}

% \noindent Over the last decades, the continuous increase in global population and need for nutrition, in combination with the impact of climate change on food production, is expected to affect the food sector significantly \cite{fritz2013need}. The agricultural productivity needs to be strengthened in order to accommodate the needs of the growing population, while preserving environmentally-friendly and sustainable agricultural practices. In this context, there is a need for the timely, large-scale and accurate monitoring of agricultural production and the provision of the necessary knowledge for evidence-based decision making on food security matters \cite{yifang2015global,fritz2013need}.\\ 

\noindent Rice is a widely planted crop in the world and the most important food staple in Asia. Merely fifteen Asian countries, including South Korea that is the study area of this work, account for 90\% of the global rice production and the demand is expected to increase by 70\% in the next 30 years \cite{muthayya2014overview}. These figures indicate that there is pressing need for timely and accurate knowledge for rice’s spatial distribution and its expected yield at national and continental scales \cite{wang2020mapping}.\\

\noindent According to the Korean Statistical Information Service (KOSIS), South Korea cultivates approximately 730 thousand hectares and produces more than 3.5 million tonnes of rice annually, which amounts to 0.5\% of the global rice production. Furthermore, the global food security index by the Economist Intelligence Unit ranks South Korea's performance at the top 30 (29/113), with a score 72.1. However, there is systematic overproduction of rice that results in large storage costs and shortage in other major grains. This is due to the governmental direct payment scheme that focuses on rice and gives limited incentives for the cultivation of other crops \cite{lee2016crop}. Even more, South Korea scores a negative 85\%, referring to the percentage difference from the mean global score, in the crop storage facility indicator of the global food security index. This indicator is based on the assessment on the governmental investments to improve crop storage.\\

\noindent The monitoring and mapping of paddy rice extent in South Korea and the estimation of its productivity is currently performed by the Korea Rural Economic Institute (KREI). The current approach is based on costly and time-consuming field visits and the collection of field data at sampled points. This point information is then spatially interpolated through statistical techniques in order to extract the required nationwide rice production assessments. In this regard, EO derived information through the implementation of AI techniques is a key enabler for facilitating detailed assessments over large areas. This is achieved by providing spatial exhaustiveness, timeliness and high thematic precision.\\

\noindent However, AI algorithms require a significant  amount of ground truth data to train the prediction models, which in most of the cases are scarce and of poor quality. The issue of getting access to reliable ground truth data becomes even more challenging when dealing with large scale applications that cover vast areas at national or continental level. Another issue linked to the detailed, precise and exhaustive mapping of paddy rice extent relates to the spatial resolution of the EO data used.\\

\noindent Over the past decades, several studies have been conducted for paddy rice mapping, using MODIS sensor data as the main source for crop monitoring. The high revisit frequency of the MODIS missions offers significant capabilities for continuous monitoring, and therefore it has been utilized in relevant studies both at global \cite{xiao2006mapping,gumma2011mapping,pittman2010estimating} and at national scale \cite{zhang2015mapping,peng2011detection,xiao2005mapping}. However, due to the low spatial resolution of MODIS images, detailed thematic information is not feasible. This limitation has been resolved by employing higher spatial resolution satellite images, such as the Landsat TM data \cite{dong2016mapping,kontgis2015mapping,qin2015mapping}. However, the Landsat TM data are characterized by longer revisit times and hence suboptimal temporal resolution for this kind of applications. Additionally, SAR data have been used in a plethora of studies for paddy rice mapping. The SAR signal gets mirrored at the surface of calm and open water bodies, being particularly useful to detect paddy rice that is inundated during the first stages of its cultivation
\cite{nelson2014towards,shao2001rice,jo2020deep,wang2020mapping}.\\

\noindent In recent years, new satellite missions that offer imagery of improved spatial and temporal resolution have introduced a paradigm shift in the potential for new applications, but also in the ways in which data is processed and knowledge is extracted. An example of such missions are the Sentinels, which freely and systematically supply images of high spatial and temporal resolution at a global scale. However, these EO data streams, which can be described as Big Data, demand increased computational power for their effective and efficient exploitation \cite{chebbi2015big}. The Sentinel-1 and Sentinel-2 missions are ideal for the monitoring of agriculture, with the latter being primarily designed for such applications. The coverage of large areas, the short revisit times and the high spatial resolution of SAR and optical imagery has made Sentinel-1 and Sentinel-2 missions the main sources of EO data for numerous studies that address the monitoring of food security and the control of agricultural policies \cite{inglada_2016,ndvi4,sitokonstantinou2018scalable,rousi2020semantically}. The common denominator for all these studies has been the accurate and spatially detailed mapping of crops. \\

\noindent In the context of paddy rice mapping, multiple classification techniques have been utilized, including neural networks, supervised and unsupervised machine learning techniques, but also rule-based algorithms. Indicatively, Tian et al. (2018) have introduced a k-means unsupervised classifier for mapping multi-season (early, middle and late) paddy rice using Sentinel-1 and Landsat-8 data \cite{Tian_2018}. Torbick et al. (2017) have developed a rice monitoring system, combining Sentinel-1, Landsat-8 OLI and PALSAR-2 data using a RF classifier \cite{torbick2017monitoring}. Nguyen et al. (2016) have analyzed the relationship between the rice growing cycle and the temporal variation of Sentinel-1 backscatter (VV and VH) to then map rice using a decision tree on the extracted phenological parameters \cite{nguyen2016mapping}. Finally, Jo et al. (2020) have tested three kinds of deep learning approaches for paddy rice classification to overcome the scarcity of labeled data. These deep learning applications included data augmentation, semi-supervised classification and domain-adapted architecture \cite{jo2020deep}.\\ 

\noindent Nevertheless, in most studies the ground truth samples to train supervised classifiers are manually collected. In other cases, authors employ unsupervised learning techniques that are computationally costly due to their exhaustive nature. Furthermore, few studies have attempted large-scale paddy rice mapping at the field level, and the consequent management of big EO data through distributed computing or other pertinent big data technologies. These challenges undermine the development of a transferable, site independent and computationally economic framework for rice mapping. In this regard, and under the key considerations of scalability, reproducibility and transferability that are essential for large scale applications, this study implements a high spatial resolution paddy rice classification pipeline that is largely independent of the hard-to-attain ground truth information. \\

\noindent Specifically, this work implements a hybrid model that is based on a large-scale supervised paddy rice mapping architecture using pseudo-labels (weakly supervised learning). This study contributes by managing to generate useful labeled data in a fully dynamic approach, using unsupervised learning (Section \ref{sec:pseudolabeling}). Then, the extracted rice clusters are used to train an RF classifier, which is parameterized to minimize the generalization error when the model is applied to other areas (Section \ref{sec:RF}). Overall, this study implements a comprehensive pipeline for paddy rice mapping at very large scales and with high spatial resolution, using minimal ground truth information.\\

\noindent In addition, we utilize a HPDA environment for managing and efficiently processing the large feature spaces of Sentinel-1 and Sentinel-2 time-series (Section \ref{sec:RF}). We contribute with a fully distributed Apache SPARK based workflow that links the custom-made temporal interpolation on PySPARK dataframes algorithm (Section \ref{sec:pre-processing}) and the distributed implementations of the k-means and RF algorithms. This way, we fully exploit the HPDA infrastructure, minimising the computational cost. Finally, in Section \ref{sec:discussion} we discuss the robustness of the pseudo-labeling method, the generalization of the paddy rice classification model and the potential for the pipeline to scale. 
 
%%%%%%%%%%%%%%%%%%%%%%%%%%%%%%%%%%%%%%%%%%
\subsection{Materials}\label{sec:materials}

\subsubsection{Study area}\label{sec:Study Area}

\noindent The study area consists of three sites in South Korea, which are located in separate climatic and agro-climatic zones and are described by diverse paddy rice cultivation characteristics \cite{zones, kim2016classification}.The climatic and agro-climatic zones of South Korea are shown in Figure \ref{appendix_zones}. The three sites have been selected in order to first allow for and then test the generalization of the paddy rice classification model. \\

\noindent South Korea is in the temperate monsoon and continental climate zones. It has an average annual temperature that ranges from 10 \textsuperscript{o}C to 15 \textsuperscript{o}C, reaching the highest values in August (23 \textsuperscript{o}C - 26 \textsuperscript{o}C) \cite{south_korea}. The annual precipitation ranges from 1,000 mm to 1,800 mm in the southern part of the country and from 1,200 mm to 1,500 mm in the central region \cite{area1}. According to the Korea Meteorological Administration, over half of the total precipitation falls during the summer season due to the Asian Monsoon.\\

\begin{figure}[ht]
    \centering
    \includegraphics[width=12 cm]{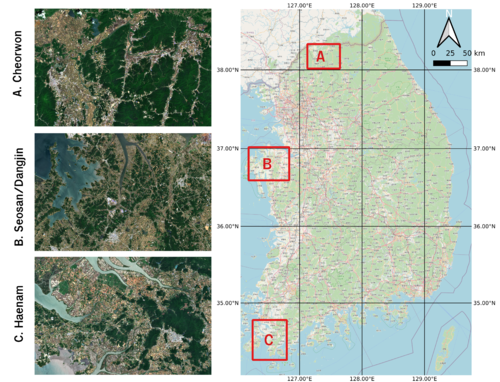}
    \caption{Study area in South Korea - a) Cheorwon, b) Seosan-Dangjin and c) Haenam}
    \label{fig:map}
\end{figure}

\noindent The reference site, based on which the generalized paddy rice classification model was trained, comprises the region around the cities of Seosan and Dangjin that are located at the northwestern end of the South Chungcheong province (Figure \ref{fig:map} b). 
\noindent Seosan and Dangjin are recorded as the highest rice producers in the country \cite{area2}. The sites shown in Figure \ref{fig:map} (a) and Figure \ref{fig:map} (c) are used for the validation and testing of the paddy rice classification model. Haenam (Figure \ref{fig:map} c) is located in South Jeolla Province, below the Taebaek Mountains, and is characterised by an oceanic and rather warm climate \cite{co2paddyrice}. Cheorwon (Figure \ref{fig:map} a) is located next to the border with North Korea, in the Gangwon province.\\

\noindent The different climatic and paddy rice cultivation characteristics for the three sites are summarized in Table \ref{tab:differences}, indicating their respective rice transplanting and harvesting periods \cite{area2, haenammeteo, cheorwon}. Table \ref{tab:differences} and Figure \ref{fig:ndvi_aois} reveal that the three sites are described by diverse conditions that is expected to first enable and then showcase the generalization of the paddy rice classification model. Specifically, the model is trained using pseudo-labels in the reference site of Seosan-Dangjin, then optimized (Section \ref{sec:RF}) and tested in the Haenam and Cheorwon sites (Section \ref{sec:Results}).

\begin{table}[!ht]
    \centering
    \caption{\label{tab:differences} Climatic and paddy rice cultivation characteristics for the three study sites. \cite{area1,area2,co2paddyrice,haenammeteo,cheorwon,zones}} \resizebox{\textwidth}{!}{%
     \begin{tabular}{c c c c}
    \toprule
    \textbf{}           & \textbf{Seosan/Dangjin} & \textbf{Haenam} & \textbf{Cheorwon} \\ 
    \midrule
 \textbf{Mean Elevation (m)} & 37 & 45 & 215 \\ 
    \textbf{Annual Mean Temperature (\textsuperscript{o}C)} & 11.9/11.4  & 13.4       &      10.2\\ 
    \textbf{Annual Mean Precipitation (mm)} & 1285.7/1158.7   & 1325.4   &  1391.2 \\
    \textbf{Annual Mean Humidity (\%)} & 74.1 &  73.4   & 70.4  \\
    \textbf{Climate zone} & Central & Southern & Northern \\ 
    \textbf{Agro-climatic zone} & {Western central plain \cite{zones}/Zone 2 \cite{kim2016classification}} & {South western coastal \cite{zones}/Zone 1 \cite{kim2016classification}} & {Northern central inland \cite{zones}/Zone 4, 5 \cite{kim2016classification}}\\
    % \textbf{Terrain Type} & - & Flat & Flat\\
    % \textbf{Slope(^$^{\circ}$)} & & 0 & 0\\
    \textbf{Rice area 2018 (ha)} & 37,728 (sum) & 18,484 & 9,429\\
    \textbf{Rice yield 2018 (tons)} & 206,475 (sum) & 89,106 & 52,653\\
    \textbf{Rice transplanting} & late May - mid June & early - mid June & mid - late May \\ 
    \textbf{Rice harvesting} & early September to mid October & late September to early October &  early to mid September\\

    \bottomrule
    \end{tabular}}
\end{table}

\subsubsection{Satellite data}\label{sec:Satellite Data}

\noindent In this study, we used both optical and SAR images. This combination of Sentinel-1 and Sentinel-2 data enhances the discrimination between the land use/land cover targets, exploiting the weather independent SAR data to create dense time-series of satellite imagery. Additionally, SAR data are ideal for rice mapping, as paddy rice is inundated during the initial stages of the cultivation. SAR backscatter coefficient values are low for calm water bodies and therefore rice is strongly differentiated from other crops \cite{sar2,sar1,sar3,sar4}. 
Sentinel-1 GRD and Sentinel-2 L1C data for the year 2018 were acquired from the Umbrella Sentinel Access Point, developed by the National Observatory of Athens.\\

\begin{figure}[!ht]
\begin{center}
		\includegraphics[width=12.5 cm]{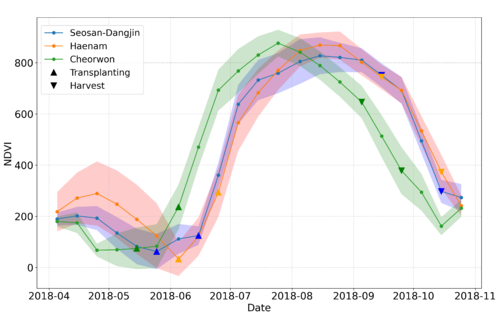}
	\caption{NDVI time-series for the three study sites; Seosan-Dangjin (Blue), Haenam (Orange), Cheorwon (Green). The up-pointing triangles enclose the transplanting period, while the down-pointing triangles enclose the harvesting period.}
	\label{fig:ndvi_aois}
\end{center}
\end{figure}

\noindent Rice paddies are inundated approximately one month before transplanting takes place. However, the start of the inundation period differs from place to place. For instance in the Haenam site, transplanting takes place a bit later in the year. Figure \ref{fig:ndvi_aois} shows the mean values of NDVI for rice paddies in each site. 

\noindent The faded colored area around each curve illustrates the standard deviation. The differences in the growth of paddy rice among the sites are evident. The high standard deviation values for the months of April and May on the Haenam curve can be explained by the fact that certain parcels go through one winter and one summer cultivation within the same year. This results in a slight delay of the transplanting period, as it is indicated by the orange curve in Figure \ref{fig:ndvi_aois}.\\

\noindent Furthermore, the period in which transplanting can take place is enclosed within the up-pointing triangles at the beginning of each NDVI curve. Respectively, the down-pointing triangles at the end of each curve correspond to the periods in which harvesting can occur. Therefore, in order to create a generalized and robust rice classification model that is not subject to the aforementioned region-based differences, the feature space is truncated to include acquisitions starting from June onward, for all sites. \\

\noindent After pre-processing all available Sentinel-2 images to Level-2A products, using the Sen2Cor software, we selected the cloud-free instances based on the percentage of cloud coverage according to the generated Scene Classification \cite{sen2cor}. The cloud mask produced from the Scene Classification product, included the classes of high probability clouds, medium probability clouds, cirrus, snow and cloud shadows. Then, only the images with cloud coverage less than 70\% over the site were included in the feature space. The relatively cloud-free Sentinel-2 datasets used for each site, covering June to October of 2018, amounted to 16 images for Seosan-Dangjin (128 GB), 13 images for Haenam (130 GB) and 19 images for Cheorwon (90 GB).\\

\begin{figure}[!ht]
    \centering
    \includegraphics[width=10.5 cm]{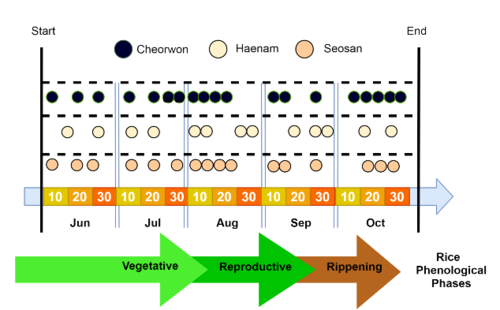}
    \caption{Timeline of the phenological stages for paddy rice in South Korea, along with the acquisition dates of the Sentinel-2 images for 2018}
    \label{fig:timeline_stages}
\end{figure}

\noindent Figure \ref{fig:timeline_stages} indicates the acquisition dates for the selected Sentinel-2 images for all three sites. These images, along with the generated VIs, construct the Sentinel-2 component of the feature space. \\

\noindent\textbf{Sentinel-1 data}

\begin{figure}[!ht]
 \centering
     \includegraphics[width=10  cm]{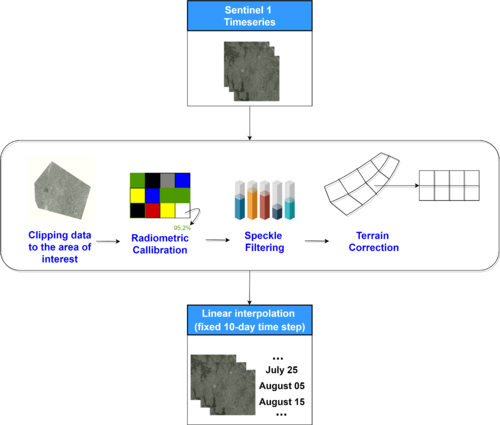}
        \caption{Pre-processing workflow for generating Sentinel-1 VV and VH backscatters}

    \label{fig:S1preprocess}
\end{figure}

\noindent A time-series of Sentinel-1 images was acquired between the dates 01/06/2018 - 31/10/2018. Specifically, we used the Level 1 Ground Range Detected (GRD) products in Interferometric Wide (IW) swath mode. The IW mode acquires data with a 250 km swath at 5 m by 20 m spatial resolution. The pre-processing of Sentinel-1 data included i) clipping to the area of interest, ii) radiometric calibration, iii) speckle filtering using the Lee filter, iv) terrain correction using Shuttle Radar Topography Mission (SRTM) 10-m and v) conversion of backscatter coefficient ($\sigma^0$) into decibels (dB). The steps are shown in Figure \ref{fig:S1preprocess}. The VV and VH bascatter coefficient time-series are sampled with a fixed 10-day step, averaging the values in each 10-day window. This is done in order to slightly reduce the dimensionality of the Sentinel-1 component of the feature space. 

\noindent\textbf{Sentinel-2 data}

\noindent For this study, a time-series of Sentinel-2 MSI scenes were acquired for the period June to October 2018. The Sentinel-2 images consist of 13 spectral bands at 10, 20 and 60 m of spatial resolution. The 60 m resolution bands, namely B01, B09 and B10, have not been utilized. Furthermore, the images have been atmospherically corrected, transforming them from Top-Of-Atmosphere (TOA) Level 1C products to Bottom-Of-Atmosphere (BOA) Level 2A products, using the Sen2Cor software \cite{muller2013sentinel}. The data were then re-sampled to 10 m spatial resolution, re-projected and clipped over the areas of interest.\\

\noindent Furthermore, three VIs were computed to enhance the feature space, namely the NDVI, the NDWI proposed by Gao et. al (1996) and the PSRI \cite{tucker1979red,gao1996ndwi,merzlyak1999non}. The NDVI and NDWI indices have been extensively used in the EO-based crop classification literature, as they are good indicators of the biomass and water content of crops. On the other hand, PSRI is particularly sensitive to the crops' senescence phase \cite{nvdi2019,lebourgeois,huang,sitokonstantinou2018scalable,ndvi3,psri}. 

% \begin{equation}\label{equ:ndvieq}
% ndvi = \frac{nir - red}{nir +red}
% \end{equation}
% \begin{equation}\label{equ:ndwieq}
% ndwi = \frac{nir - swir}{nir + swir}
% \end{equation}
% \begin{equation}\label{equ:psrieq}
% psri = \frac{red - blue}{red\_edge}
% \end{equation}

\subsubsection{Labeled data}\label{sec:validatio_data}

\noindent This study focuses on generating pseudo-labels, using unsupervised learning, in order to train a distributed RF classifier. Nevertheless, labeled sets have been both acquired and generated to a) select the optimal clustering for pseudo-labels that will form the training set (Section \ref{sec:pseudolabeling}), b) fine-tune the model for nationwide generalization (Section \ref{sec:pseudolabeling}) and c) test the model generalization (Section \ref{sec:Results}). \\

\noindent Level-3 land cover maps were used as an initial reference point for generating the labeled data, as made available from the Korean Ministry of Environment (KME). The Level-3 land cover maps are generated using VHR satellite imagery (KOMPSAT-2 and IKONOS) and aerial ortho-photos from one or two years \cite{jo2020deep}. The map objects are 3-m wide for linear elements and 100 m\textsuperscript{2} for plane elements and therefore the spatial resolution is comparable to the corresponding resolution of the Sentinel data. However, the maps are not updated every single year. Furthermore, it was deduced that they tend to be unreliable and subject to significant errors with respect to both object geometries and the land cover labeling. Nonetheless, they were an adequate starting point for our labeling process.\\

\noindent The labeled data generated in this study have been derived from a Sentinel-1 and Sentinel-2 based photo-interpretation. The labeling process refers to the identification of rice and non-rice pixels and it was applied to all three study sites, for the inspection year of 2018. The method has been based on the most recent Level-3 land cover map for each respective site, which were then appropriately corrected to be compliant to the year of inspection. To this end, the first step was to utilize a change detection method, as published in \cite{sito2020}, to update the available Level-3 land cover maps for each site. This was done by identifying large changes on rice objects between the year of the land cover map production and the year of inspection. Those instances were later removed from the rice class of the updated maps.\\

\noindent Consequently, two different experts conducted blind photo-interpretation using multi-temporal Sentinel-1 and Sentinel-2 images and the updated Level-3 land cover maps as a reference. Each of the first two interpreters independently assigned labels to approximately 5-10\% of each site. Then, a third interpreter determined the final labels by making an ultimate decision on the disagreements of the first two. The samples were evenly distributed in space and thus representative of the entire site to ensure reliable model optimization and classification assessment. Additionally, the samples were taken from diverse landscapes including agriculture, water, forest and urban classes. The locations of the photo-interpreted samples are presented in Figure \ref{fig:aois}.\\

\begin{figure}[!ht]
    \centering
    \includegraphics[width=10.5 cm]{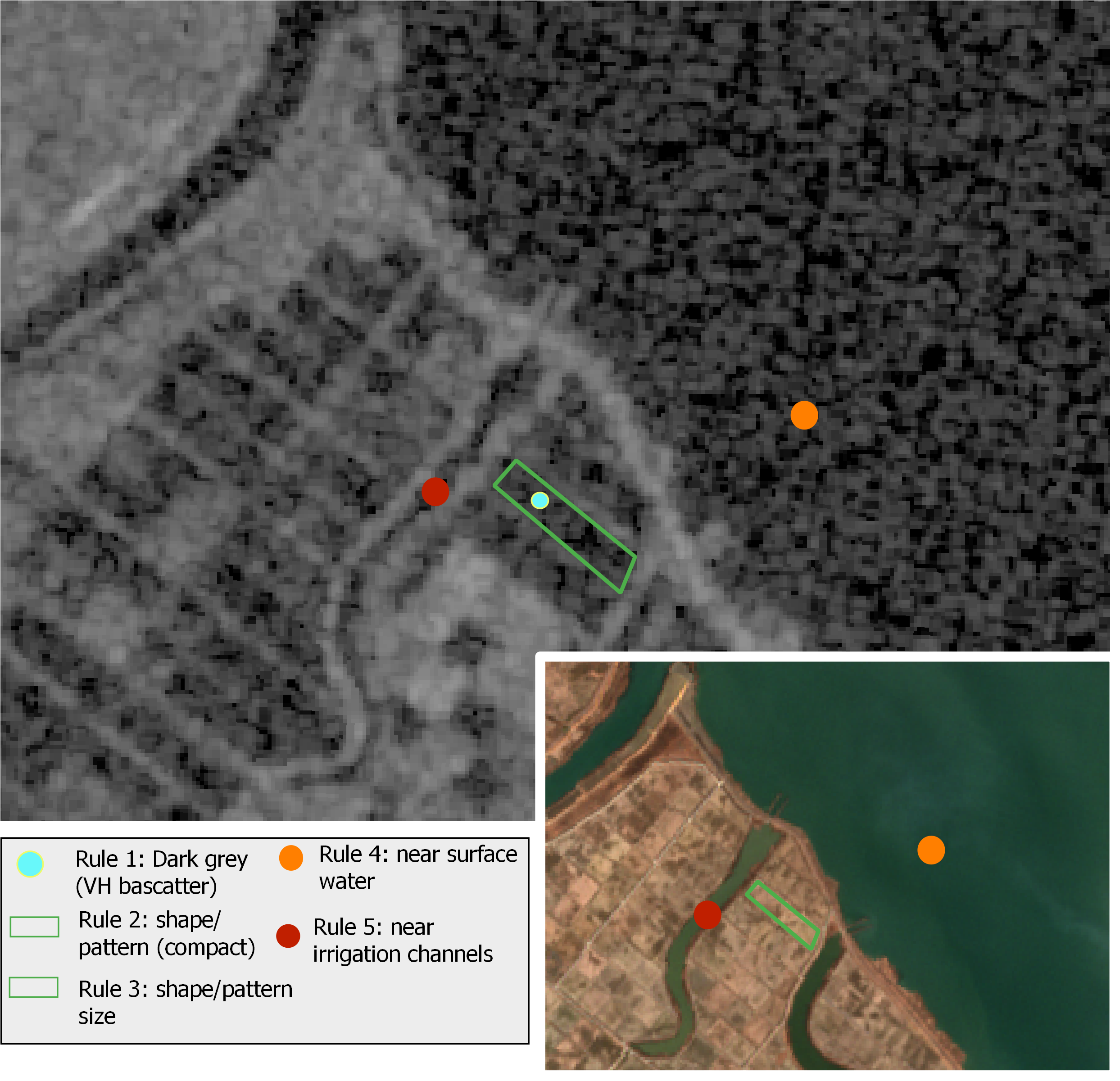}
    \caption{Example of the photo-interpretation rules used for creating the labeled dataset. The grayscale image is the VH backscatter for the 31/05/2018 Sentinel-1 acquisition and the colored image is the RGB for the 18/04/2018 Sentinel-2 acquisition. The latter is used for feature clarity.}
    \label{fig:photorules}
\end{figure}

\noindent The inundation and transplanting periods for paddy rice differ among the study areas. Specifically, in Cheorwon the inundation starts in the mid-late April, while in Seosan-Dangjin in the end of April and in Haenam in the middle of May. Therefore, the photo-interpretation analysis included Sentinel-1 and Sentinel-2 data starting from the month of April. The sensitivity of Sentinel-1 data to water has been primarily used to identify the inundated areas before the start of the transplanting phase. Low backscatter coefficient values, revealing water content, and the corresponding geometric patterns of fields have served as a first indication to separate and label the rice paddies \cite{torbick2017monitoring}. Figure \ref{fig:photorules} shows the photo-interpretation rules used in the first step of the labeling process.\\

\begin{figure}[!ht]
    \centering
    \includegraphics[width=10.5 cm]{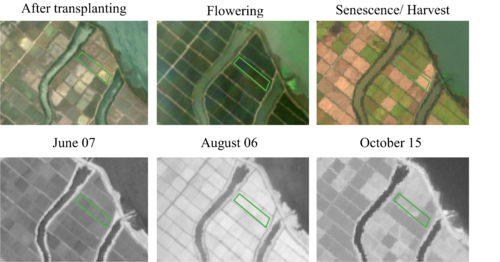}
    \caption{Examples of photo-interpretation keys based on the time-series of Sentinel-2 RGB (top) and NDVI images (bottom)}
    \label{fig:photokeys}
\end{figure}

\noindent Rules 1-3 had to be satisfied, while rules 4 \& 5 served as an additional indication for the labeling of rice paddies. To be noted that in certain cases, and especially due to the size of a paddy rice or its narrow shape (e.g. terrace paddy), it was difficult to extract reliable information on existing inundation patterns using solely Sentinel-1 data. To overcome this problem, an analysis of NDWI multi-temporal profiles was additionally carried out for candidate rice paddies that had satisfied several or all of rules 2-5, but not rule 1. Actually, in cases that NDWI was found smaller than 0.2 before transplanting, in the specific parcel was assigned the non-rice label, otherwise the decision process was moving to the next step that made use of Sentinel-2 multi-temporal photo-interpretation keys. An indicative example is illustrated in Figure \ref{fig:photokeys}. \\

\noindent The interpretation of the time-series of Sentinel-2 true color and false color (B08-B04-B03) composites has proven to be particularly useful in examining whether or not a parcel complies with the different phenological stages of rice. In the end, the examination of NDVI and NDWI multi-temporal signatures were used to guide the final decision of the photo-interpreter. Their decision was dictated by the fact that NDWI gets higher values before the transplanting period and starts lessening thereafter. On the other hand, NDVI exhibits the inverse phenomenon, as it increases after the transplanting period, reaching its peak values at the end of flowering, when it starts decreasing again as rice enters in its ripening phase (Figure \ref{fig:photokeys}). This procedure was applied in all three study sites. The generated labeled sets for the Seosan-Dangjin, Haenam and Cheorwon sites included 1,675,632 non-rice and 1,287,183 rice pixels, 2,944,147 non-rice and 892,141 rice pixels and 1,137,195 non-rice and 195,942 rice pixels, respectively. \\

\begin{figure}[!ht]
    \centering
    \includegraphics[width=10.5 cm]{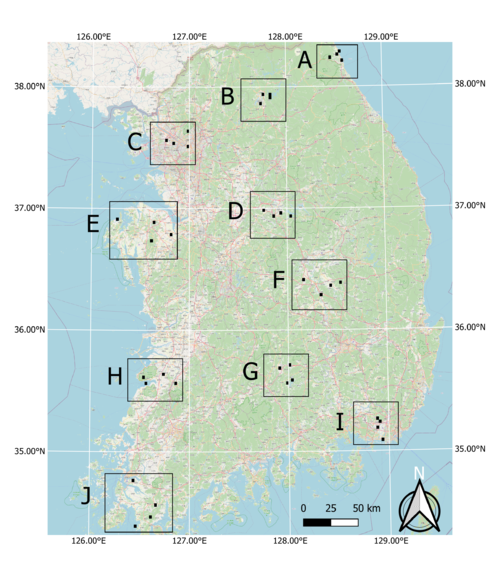}
    \caption{Location of test sites across South Korea. Each site is composed of four plots that are respectively characterized by agricultural, urban, forest and water landscapes.}
    \label{fig:test set}
\end{figure}

\noindent Finally, in order to further test the success in the generalization of the model, its transferability and robust applicability in different agro-climatic zones over the entire country, the labeled set created by the authors of \cite{jo2020deep} was additionally included and used in this study (Figure \ref{fig:test set}). For this, the authors have identified ten evenly distributed sites in South Korea that fall into ten different agro-climatic zones according to \cite{zones}. Each site includes four plots in agricultural, urban, forest and water abundant landscapes. The size of the plots ranged from 6 to 6.5 km\textsuperscript{2}. By applying a photo-interpretation study, the necessary test data were produced based on the existing level-3 land cover maps that were updated to reflect the 2018 reality. This was done using visual interpretation and cross-validation with the employment of three independent interpreters, who were assisted by Google Earth and domestic street view data (\url{https://map.kakao.com}; \url{https://map.naver.com}). 
 
% \newpage
\subsection{Methods}\label{sec:methods}

\subsubsection{Sentinel-2 time-series interpolation}\label{sec:pre-processing}

\noindent The cloud covered instances in the Sentinel-2 time-series, which were masked out, are represented as null values in the feature space. We applied linear interpolation in order to fill in these gaps. Due to the large volume of the data, this method was implemented in PySpark and processed in the HPDA. Since there is not any ready-to-use function to interpolate a PySpark Dataframe, this interpolation method was developed from scratch. This allowed for the mass processing of the time-series, efficiently filling in the missing values of large datasets.\\

\noindent Then, and in order to avoid possible temporal gaps between the available acquisitions, we transformed the Sentinel-2 feature space to a time-series of fixed step, generating values that represent the 5th, 15th and 25th day of each month. In order to do this, a workflow of two separate interpolation methods was followed. \\

\begin{figure}[ht]
    \centering
    \includegraphics[width=10.5 cm]{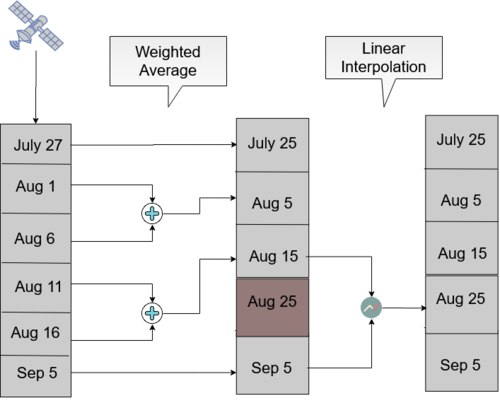}
    \caption{An example of the Sentinel-2 temporal interpolation methodology for images acquired between July 20 and September 10 2018}
    \label{fig:interpolation}
\end{figure}

\noindent Initially, the method illustrated in Figure \ref{fig:interpolation} examines the time windows between the 1st and the 10th, the 11th and the 20th and the 21st until the end of each month. A weighted average interpolation is applied to the acquisitions that fall within each window and it then constructs part of the fixed-step time-series. However, it is likely that there are not any observations within a specific window, resulting in the absence of features for the corresponding temporal instance. In order to fill these remaining temporal gaps, linear interpolation was applied between the previous and next available fixed temporal instances.
\subsubsection{Pseudo-labeling}\label{sec:pseudolabeling}

\noindent In an attempt to address the common issue of ground truth data scarcity, we examined a pseudo-labeling approach to generate labeled data for training. This can be done using only a small amount of ground truth data (see Section \ref{sec:discussion}). The pseudo-labels were generated using a distributed k-means classifier, employing as input the time-series of the vegetation indices (NDVI, NDWI, PSRI) for the months of June and July. \\

\noindent K-means is one of the simplest and most popular unsupervised machine learning algorithms \cite{lloyd1982least,steinhaus1956division,ball1965isodata,macqueen1967some}. The algorithm aims to cluster the data into k groups, where the variable k is specified by the user. At the beginning, k different clusters are created, using k randomly generated candidates. Then every entity is assigned to the nearest cluster based on their features. In the next step, the entities are reassigned to the nearest cluster that is now represented from the mean value of the entities that belonged to it from the previous step. The algorithm keeps iterating until the assignment of data points to clusters remains unchanged. The outputs of the algorithm are the means of the k clusters (centroids), as well as the labels for each sample, which are then used as training data. \\

\begin{figure}[ht]
    \includegraphics[width=15 cm]{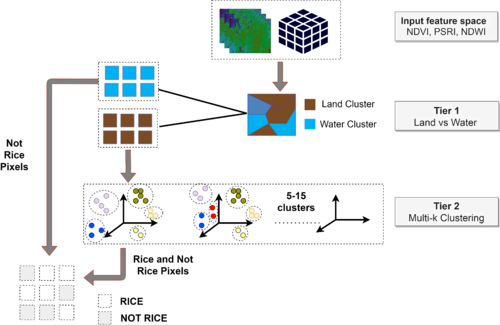}
    \caption{Workflow for generating rice pseudo-labels using k-means clustering}
    
    \label{fig:preprocess}
\end{figure}

\noindent In our approach and due to the large volume of data the distributed k-means version of the Pyspark API was used. First, an execution of k-means is performed with only two clusters (k=2), separating land from water. Having extracted the land pixels, a second-level clustering for multiple k values (5-15) takes place (Figure \ref{fig:preprocess}). \\

\noindent Each pixel is assigned to a different cluster. However, there is no further information about the land cover that each cluster represents. Thus, a method to identify the rice cluster for all individual clusterings (for k=5-15) is required. This is achieved by the Mean Squared Error (MSE) between the centroid of each cluster and the mean time-series signature of the rice class, as extracted from the labeled set (Section \ref{sec:validatio_data}). \\

\noindent Once the rice cluster has been identified for each individual run, the precision and recall for this cluster is calculated against the labeled set. This is the only instance when labeled data are utilized in the classification process. The clustering with the best precision and recall combination is selected, and these extracted rice labels are then used for training the RF (Section \ref{sec:RF}). The optimal combination of precision and recall is shown in Section \ref{sec:Results} 

\subsubsection{Model Performance Evaluation}\label{sec:metrics}
% \noindent In this study, four metrics were used to evaluate the performance of the paddy rice classification model and that is the precision, recall, kappa coefficient and F1 score. 

% \begin{equation}
%     Precision = \frac{TP}{TP + FP}
%     \label{eq:precision}
% \end{equation}

% \begin{equation}
%     Recall = \frac{TP}{TP+FN}
%     \label{eq:recall}
% \end{equation}

% \begin{equation}
%     F_{1} score = \frac{2}{2 \cdot TP+FP+FN}
%     \label{eq:f1}
% \end{equation}

% \begin{equation}
%     Kappa = \frac{p_{o}-p_{e}}{1-p_{e}}
%     \label{eq:kappa}
% \end{equation}
% where TP is the number of True Positive instances, FP the number of False Positive, FN the number of False Negative, $p_{o}$ is the relative observed agreement between the ground truth and the prediction, and $p_{e}$ is the hypothetical probability of chance agreement.
% \noindent The Random Forest (RF) is an ensemble classifier based on the decision tree model. The training data are split into multiple different subsets, using the bootstrap method, and each subset creates a new decision tree. Finally, a forest is constructed from the different decision trees and predictions are made through majority voting \cite{breiman2001random}. \\

\noindent A distributed RF implementation was employed in order to address the great complexity that comes with the large scale application of paddy rice classification. The algorithm utilized MLLib, a library of Apache SPARK, in which many of the optimizations are based upon Google’s PLANET project \cite{panda2009planet}. First of all, each tree of the RF ensemble is trained independently, thus multiple trees can be trained simultaneously on different nodes. Further, MLLib enables the parallelization on each single tree of the ensemble, allowing for the concurrent training of multiple sub-trees. The optimal number of these sub-trees is optimized on each iteration of the algorithm, depending on memory constraints \cite{databricks}. \\

\noindent MLLib optimizes the algorithm using techniques to avoid unnecessary parses on data. For example, splits for every tree node of the same level are computed simultaneously with only one pass over the data, rather than one pass for each node individually. For the best split computation, features are separated into bins, which are then used to calculate useful statistics for the splitting. These bins are precomputed for each of the instances, thereby saving computational time.\\

\noindent The distributed RF classification pipeline has been deployed on an HPDA environment, which gives the ability to train a model in significantly less time than a conventional machine. In this study, we used the Cray-Urika-GX of the High-Performance Computing Center Stuttgart (HLRS). The Cray-Urika-GX comprises 41 nodes, with 512GB of RAM and 2TB of disk capacity each. Each node is composed of 2 Intel BDW 18-Core (2.1 GHz) processors. For this work, up to 16 nodes were utilized for the experiments, making it possible to explore multiple parameterizations of the RF model in a short amount of time. \\

\begin{figure}[!ht]
\centering
		\includegraphics[width=10.5 cm]{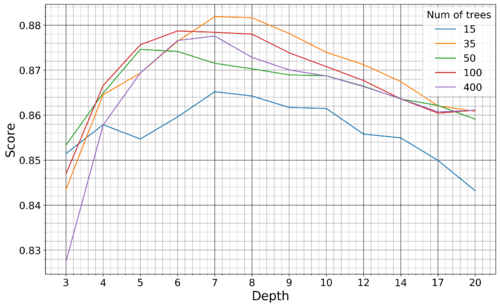}
	\caption{Kappa coefficient of the Haenam site for models trained on pseudo-labels of the Seosan-Dangjin site, with various depth sizes and number of trees}
	\label{fig:kappa_haenam}
\end{figure}

\begin{figure}[!ht]
\centering
		\includegraphics[width=10.5 cm]{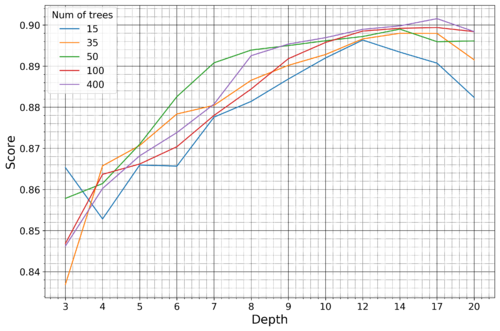}
	\caption{Kappa coefficient of the Cheorwon site for models trained on pseudo-labels of the Seosan-Dangjin site, with various depth sizes and number of trees}
	\label{fig:kappa_cheorwon}
\end{figure}

\noindent The model was trained on the pseudo-labels generated in the Seosan-Dangjin site. The hyperpameters were optimized using half of the labeled samples for the Haenam and Cheorwon sites (validation set). The other half was used for testing the performance of the optimized model, as presented in Sections \ref{sec:Results} and \ref{sec:discussion} (test set). Figures \ref{fig:kappa_haenam} and \ref{fig:kappa_cheorwon} demonstrate how the kappa coefficient responds to the depth size and number of trees of the RF model.\\
\noindent For the Haenam site, the kappa coefficient increases for larger depth sizes. This is true until the depth parameter becomes equal to 8, after which the kappa coefficient slightly decreases. For the Cheorwon site, the kappa coefficient increases almost linearly with an increasing depth, until it reaches a plateau for depth values larger than 12. \\

\noindent The response of the kappa coefficient for a varying number of trees, in combination with several depth values, was observed for tens of experiments in the range 10-400. In Figures \ref{fig:kappa_haenam} and \ref{fig:kappa_cheorwon}, we depict five indicative cases that fully describe the relevant response of the RF. For 15 trees, the kappa values for both sites are less than optimal, although satisfactory. For 35 or more trees, however, the performance stabilizes and displays marginal differences for all different depth values. \\

\begin{figure}[!ht]
    \centering
    \begin{subfigure}{\linewidth}
		\includegraphics[height = 6cm, width=8.5 cm]{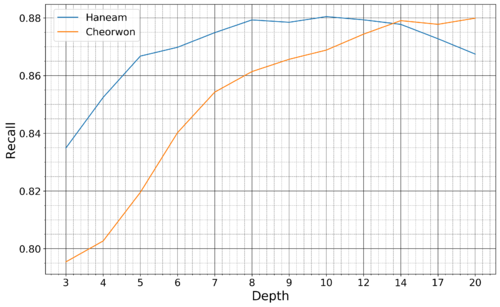}
        \includegraphics[height = 6cm, width=8.5 cm]{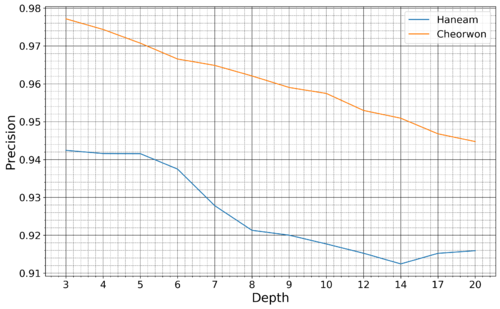}
    \end{subfigure}
	\caption{Recall and Precision of the Haenam and Cheorwon sites, for models trained with pseudo-labels of the Seosan-Dangjin site, with 50 trees, various depths}
	\label{fig:precision_recall_1}
\end{figure}

\noindent The value of choice for the number of trees parameter was 50, as a trade-off between accuracy and computational complexity. In Figure \ref{fig:precision_recall_1} is displayed the relative performance between the two sites for increasing depth size. The precision and recall evolution plots are run for a fixed number of trees equal to 50.\\

\noindent It is observed that for increasing depth values, the scores decrease for precision and increase for recall, as expected. However, the performance for the Haenam site and for depth sizes larger than 12 appears to lose not only in precision but also in recall. This indicates model overfitting. This phenomenon is evident in the Haenam site mostly because the transplanting period commences later in the year. Starting the time-series that makes up the feature space from June, and not earlier, partly resolved the issue. Nonetheless, large depth sizes capture more of the Seosan-Dangjin site particularities that has a negative impact in the generalization of the model to areas with different characteristics, such as the Haenam site. For this reason the depth parameter used was set to 12. \\

\subsection{Results}\label{sec:Results}
\subsubsection{K-means clustering}\label{sec:Clustering_reults}

\noindent As mentioned in Section \ref{sec:pseudolabeling}, we perform dynamic clustering for multiple k values to generate pseudo-labels. The RF model was then trained on the pseudo-labels that come from the Seosan-Dangjin site. Nevertheless, the k-means algorithm was executed for all three study sites. This was done in order to showcase the robustness of the pseudo-labeling method, while at the same time generate local pseudo-labels for training individual RF models for the Haenam and Cheorwon sites. The locally trained models for these two sites are then compared to the predictions of the Seosan-Dangjin site model to demonstrate the success of the generalization (Section \ref{sec:discussion}).\\

\noindent In Table \ref{tab:kmeans_rice} is shown the precision, recall, F1-score and the optimal number of clusters for the k-means clustering in each of the sites. It can be observed that for all sites the performance scores are satisfactory for both the rice and non-rice classes. 

\noindent The implemented unsupervised classification method is dynamic, as it tries out multiple k values and selects the most appropriate. It is also robust, as it returns predictions of comparable accuracy for different areas of application. The best k is selected to appropriately balance precision and recall. This trade-off combination of the two metrics was defined by observing the results of multiple experiments. \\

\noindent The experiments were evaluated using the RF accuracy metrics, therefore examining the quality of the pseudo-labels in their capacity as training samples. The experiments revealed particular sensitivity to the recall metric. Experiments of recall smaller than 85\% yielded sub-optimal RF classification results, as the training set was not adequately representative. Therefore, the first condition for selecting the best clustering was that recall must be larger than 85\%. Precision is also very important, securing the minimization of noise in the pseudo-labels. Thus the second condition was for precision to be larger than 90\%. If both conditions were satisfied for more than one clustering, we chose the one with the highest F1- score. 

\begin{table}[!ht]
\caption{\label{tab:kmeans_rice}Pseudo-labeling performance metrics and the optimal number of clusters for the rice and non-rice class per site}
    \centering
    \begin{tabular}{cccccc}
    \toprule
 \textbf{Class} & \textbf{Site} & \textbf{Precision} & \textbf{Recall}  & \textbf{F1-score} & \textbf{Clusters}\\ 
    \midrule
    \multirow{3}{*}{Rice} &
    \textbf{Seosan/Dangjin}   & 97.01\%   & 91.82\% & 94.35\% & 7  \\ 
   & \textbf{Haenam}   & 94.16\%   & 88.09\% & 91.03\% & 8 \\ 
   & \textbf{Cheorwon} & 97.63\%   & 88.59\% & 92.89\% & 7 \\ 
   \midrule
   \multirow{3}{*}{Non Rice} &
    \textbf{Seosan/Dangjin}   & 93.91\%  & 97.81\%   & 95.82\%    & 7 \\  
   & \textbf{Haenam}   & 96.45\%  & 98.34\%   & 97.93\%    & 8 \\  
   & \textbf{Cheorwon} & 98.02\%  & 99.62\%   & 98.81\%    & 7 \\ 
    \bottomrule
    \end{tabular}
\end{table}

\subsubsection{Model performance for varying training set size}\label{sec:ratio_evolution}
\begin{figure}[!ht]
  \begin{subfigure}{\linewidth}
  \includegraphics[height=4.5cm,width=.33\linewidth]{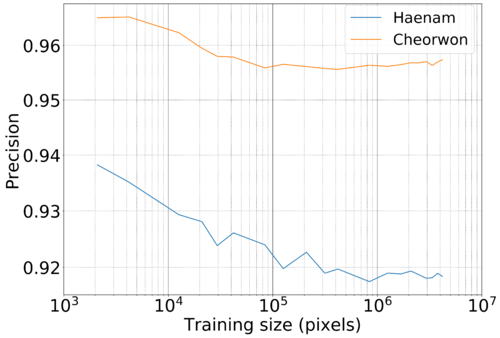}\hfill
  \includegraphics[height=4.5cm,width=.33\linewidth]{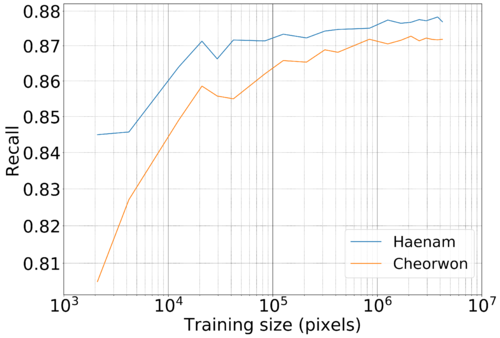}
    \includegraphics[height=4.5cm,width=.33\linewidth]{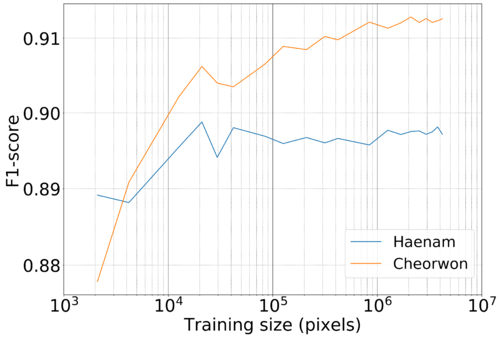}
   \end{subfigure}
  \caption{Precision, Recall and F-1 score of RF for varying training set size - for a feature space that includes acquisitions from June until the end of July - with depth = 12 and number of trees = 50}
  \label{fig:ratios}
\end{figure}

\noindent Figure \ref{fig:ratios} shows how the precision, recall and F-1 score respond to the training set size. The scores were calculated against the labeled data of the Haenam and Cheorwon sites that were not used in fine-tuning the hyperparameters of the generalized model. The training set size on the x-axis refers to the number of samples on which the model was trained. The site comprises 41,645,772 pixels and the different training sets of increasing number of samples are randomly selected. \\

\noindent It is observed that with higher training set sizes the precision slightly decreases, while recall increases. It can also be seen that the performance scores are high even with only a limited number of training data. A model that uses a training set of 0.4 M pixels, including approximately 90,000 rice pixels, reaches near-optimal recall values. The precision and the F1-score get high values even for surprisingly small training sets, e.g. 0.002 M. However, the associated recall value is substantially lower that indicates overfitting, which is expected for such small training sets. The results presented from this point on were trained with roughly 4 M training pixels.\\

\subsubsection{Feature importance}\label{sec:RF_importance}

\begin{figure}[!ht]
 
\begin{center}
		\includegraphics[height = 11cm, width=15 cm]{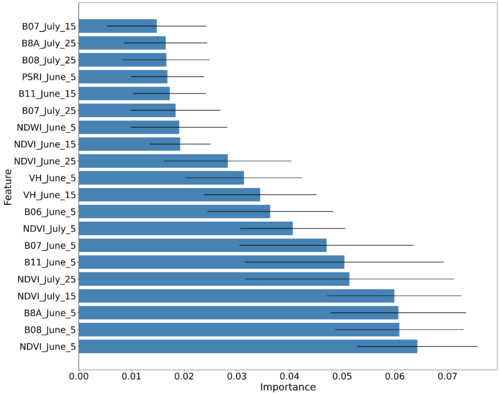}
	\caption{Average RF importance of the top ranked features, along with the respective standard deviation for 10 different runs.}
	\label{fig:import1}
\end{center}
\end{figure}

\begin{figure}[!ht]
\begin{center}
		\includegraphics[height = 7.5cm,width=10.5 cm]{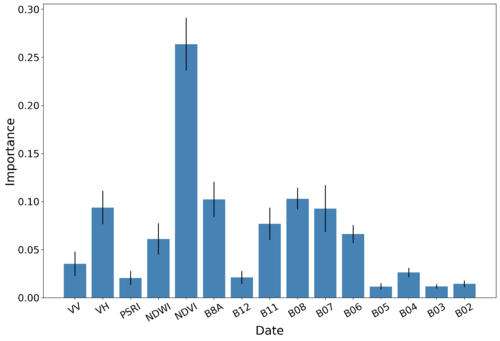}
	\caption{RF importances aggregated by feature type, along with the respective standard deviation for 10 different runs.}
	\label{fig:import2}
\end{center}
\end{figure}

\begin{figure}[!ht]
\begin{center}
		\includegraphics[height = 7.5cm, width=10.5 cm]{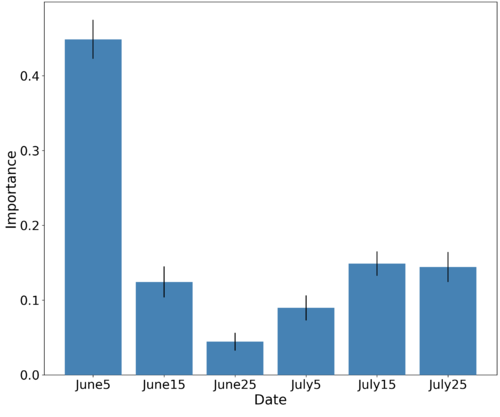}
	\caption{RF importances aggregated by time-step of the interpolated time-series, along with the respective standard deviation for 10 different runs.}
	\label{fig:import3}
\end{center}
\end{figure}

\noindent Figures \ref{fig:import1}, \ref{fig:import2} and \ref{fig:import3} show the RF importances for the optimized model of 50 trees and depth equal to 12. 
The importance values have been averaged for running the same algorithm 10 times with the same hyper-parameters using 10 different random seeds. Figure \ref{fig:import1} shows the RF importances of the top 20 ranked features, which include the Sentinel-2 bands B08, B8A, B11, B07 and B06, all vegetation indices (NDVI, NDWI, PSRI) and the Sentinel-1 backscatter coefficients VH and VV.\\

\noindent It is observed that bands B08 and B08A (NIR) and the vegetation index NDVI for June05, along with the NDVI index for July15 and July25, are the most important features for the paddy rice classification. The Sentinel-1 features (VH June15 and VH June05) appear in the tenth and eleventh position, with significant contribution.\\

\noindent Figure \ref{fig:import2} illustrates the aggregated importance values for the different feature categories. NDVI appears to be by far the most important, followed by the red-edge and NIR bands of Sentinel-2 and the VH backscatter coefficient of Sentinel-1. \\

\noindent In the same fashion, Figure \ref{fig:import3} aggregates the importance values with respect to time. The June05 instance appears to dominate the importances. This is expected, as by that date the rice fields are fully inundated, allowing for their discrimination among other crop types.\\

\subsubsection{Accuracy evolution}\label{sec:evolution_results}
\begin{figure}[!ht]
    \centering
    \begin{subfigure}{\linewidth}
		\includegraphics[width=9 cm]{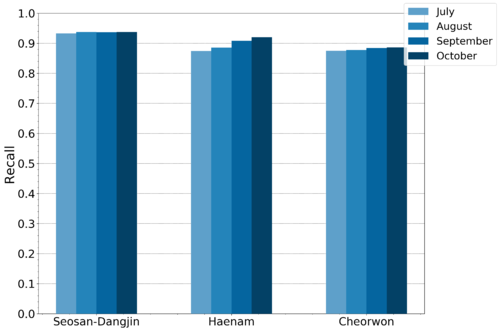}
        \includegraphics[width=9 cm]{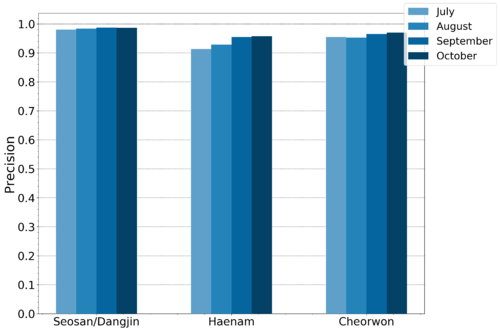}
    \end{subfigure}
	\caption{RF recall and precision for incrementally larger feature spaces. Classifications were run in July, August, September and October.}
	\label{fig:precision_recall}
\end{figure}

\noindent Figure \ref{fig:precision_recall} shows the evolution of precision and recall for incrementally larger feature spaces. The four different feature spaces, for which the classification performance is compared, are composed of features from the first acquisition of June until the last acquisitions of July, August, September and October, respectively. As expected, richer feature spaces achieve both better recall and precision. Nonetheless, it can be concluded that near-optimal performance is achieved early in the year, allowing for the timely decision making on food security matters. The model generalizes successfully even with short time-series that merely include the months of June and July.\\

\noindent Table \ref{tab:area_stats} shows the total predicted paddy rice area for each of the three sites, compared with the official KOSIS statistics for the year 2018. The last column of the table shows the deviation of the predicted area against the KOSIS equivalent, which is minimal for all sites.\\

\begin{table}[!ht]
\caption{\label{tab:area_stats}Paddy rice area for the Seosan-Dangjin, Haenam and Cheorwon sites from the official Korean Statistical Information Service (KOSIS) compared to the total paddy rice area predicted by this study's model (using images until July).}
    \centering
    \begin{tabular}{cccc}
    \toprule
  \textbf{Site} & \textbf{KOSIS area (ha)} & \textbf{Predicted area (ha)}  & \textbf{Deviation} \\ 
    \midrule
    \textbf{Seosan \& Dangjin}   &37,728  & 37,033 & -1.84\% \\ 
   \textbf{Haenam}   &18,484   & 17,969 & -2.78\%  \\ 
   \textbf{Cheorwon} & 9,429   &9,854 & +4.50\% \\ 
    \bottomrule
    \end{tabular}
\end{table}

\subsubsection{Model generalization}\label{sec:test_plots}
\noindent In order to evaluate the model performance at national scale, we have used 10 labeled sites in different agro-climatic zones across the country. Each site consists of four plots of diverse landscape characteristics, as described in Section \ref{sec:validatio_data}. Table \ref{tab:test_aois} shows the total accuracy and kappa coefficient of the optimized model for each different plot of the test set.\\

\begin{table}[!ht]
\caption{\label{tab:test_aois}Total accuracy and kappa coefficient of the model when run in i) July and ii) October. Highlighted with light blue are the plots for which the optimized model of this study performs better (>0.5\%) than the Data-Augmented Learning Material Fully Connected Recurrent Neural Network (DALM-FCRNN) in \cite{jo2020deep}. Highlighted with light orange are the plots with inferior performance (<0.5\%).}
    \centering
    \resizebox{\textwidth}{!}{%
    \begin{tabular}{ccccccccccccc}
    \toprule
\textbf{Model} & \textbf{Type}  & \textbf{A} & \textbf{B} & \textbf{C}  & \textbf{D} & \textbf{E} & \textbf{F} & \textbf{G} & \textbf{H} & \textbf{I}  & \textbf{J} & \textbf{Total}\\ 
    \midrule
    \multirow{6}{*}{July} &
\textbf{Agriculture} & \cellcolor[HTML]{FED8B1}87.47 & \cellcolor[HTML]{CBCEFB}{93.16} & \cellcolor[HTML]{CBCEFB}{93.04} & \cellcolor[HTML]{CBCEFB}{90.67} & {89.14} & \cellcolor[HTML]{FED8B1}90.10 & {84.56} & {91.71} & \cellcolor[HTML]{FED8B1}83.33 & \cellcolor[HTML]{CBCEFB}{88.10} & {89.13} \\
& \textbf{Urban}       & {99.60} & {99.97} & {99.99}  & \cellcolor[HTML]{CBCEFB}{99.66} & \cellcolor[HTML]{CBCEFB}{98.43} & {95.80}  & \cellcolor[HTML]{CBCEFB}{94.03} & \cellcolor[HTML]{CBCEFB}{99.65} & {100.00}   & \cellcolor[HTML]{CBCEFB}{99.84} & \cellcolor[HTML]{CBCEFB}{98.70} \\
& \textbf{Forestry}    & {100.00}   & {100.00}   & {100.00}   & {99.66} & {99.95} & {99.65} & {99.53} & {99.90}  & {100.00}   & {100.00}   & {99.87} \\
& \textbf{Water}       & {100.00}   & {100.00}   & {99.99}   & {99.91} & {100.00}   & \cellcolor[HTML]{FED8B1}90.67 & {100.00}   & {100.00}   & \cellcolor[HTML]{CBCEFB}{100.00}   & {100.00}   & {99.05} \\
& \textbf{Sites Acc.} & {96.77}  & {98.28} & {98.26}  & \cellcolor[HTML]{CBCEFB}{97.48} & {96.88} & \cellcolor[HTML]{FED8B1}94.05 & {94.53} & {97.81} & {95.83} & \cellcolor[HTML]{CBCEFB}{96.99} & {96.69} \\
& \textbf{Cohen's Kappa}       & \cellcolor[HTML]{FED8B1}0.80  & \cellcolor[HTML]{CBCEFB}{0.80}   & \cellcolor[HTML]{CBCEFB}{0.86}  & \cellcolor[HTML]{CBCEFB}{0.86}  & \cellcolor[HTML]{CBCEFB}{0.91}  & \cellcolor[HTML]{FED8B1}0.84  & \cellcolor[HTML]{CBCEFB}{0.78} & {0.93}  & \cellcolor[HTML]{FED8B1}0.84 & \cellcolor[HTML]{CBCEFB}{0.89}  & \cellcolor[HTML]{CBCEFB}{0.87} \\
\midrule
\midrule
\multirow{6}{*}{October} &
\textbf{Agriculture} & 87.77 & 93.37& 93.19 & 90.41 & {89.37} & {90.33} & {84.39} & {91.84 } & 82.79 & {89.64} & {89.21} \\
& \textbf{Urban} & {99.69} & {99.96}& {100.00} & {99.67} & {98.40} & {95.89}  & {94.20} & {99.60} & {100.00}  & {99.83} &{98.72} \\
& \textbf{Forestry}    & {100.00}   & {100.00}  & {100.00}   & {99.65} & {99.94}& {99.66} & {99.51} & {99.91}  & {100.00}   & {100.00}   & { 99.87} \\
& \textbf{Water} & {100.00}  & {100.00}   & {100.00}   & {100.00} & {100.00}   & {90.77} & {100.00}   & {100.00}   &{ 99.99}   & {100.00}   & {99.05} \\
& \textbf{Sites Acc.} & {96.87}  &{98.33} &{ 98.30}  & {97.43} & {96.92} & {94.16} & {94.53} & {97.15} & {96.12} & {97.37} & {96.65} \\
& \textbf{Cohen's Kappa} &{0.80}  & {0.80} & {0.87}  & {0.86}  & {0.91} & 0.84 & {0.78}  & {0.93}  & {0.84}  &{0.91}  & {0.87}\\   
    \bottomrule
    
    \end{tabular}
}
\end{table}

\noindent The authors in \cite{jo2020deep} have evaluated the performance of multiple deep learning models for paddy rice mapping against this particular test set. The values in light blue and light orange show the instances for which this study's model performs better or worse than the top performing model in \cite{jo2020deep}, respectively. The improvements when using this study's model range from 0.5\% to 10\% (kappa for D), while the decreases in accuracy range from 0.5\% to 4\% (Accuracy for I: Agriculture). This study's model achieves a total kappa coefficient of 0.87; while the overall precision and recall of the rice class is 88.91\% and 88.19\%, respectively. The results are comparable to the corresponding test scores for the Haenam and Cheorwon sites. The predicted paddy rice maps for all 40 test plots can be found in Figure \ref{fig:predicted_paddy_rice_maps} of the Supplementary Material. \\

\noindent Figure \ref{fig:good_bad} shows a good (E: Agriculture) and a bad (G: Agriculture) example of the predicted rice maps. In plot E, where the rice paddies are spatially continuous there are only few misclassifications. On the other hand, plot G is characterized by fragmented, small and narrow rice parcels, yielding some errors as a result of mixed pixels. \\

\begin{figure}[!ht]
\begin{center}
		\includegraphics[width=15.5 cm]{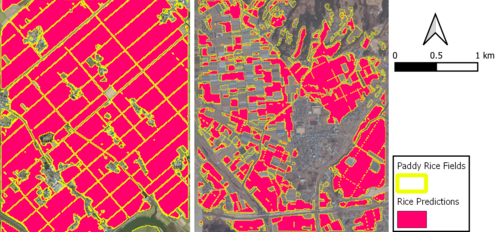}
	\caption{Paddy rice prediction maps for E (left) and G (right) test plots. }
	\label{fig:good_bad}
\end{center}
\end{figure}

\newpage
\subsection{Discussion}\label{sec:discussion}

\noindent It should be noted that using unsupervised classification for generating training data (pseudo-labels) allows for producing samples over very large areas. In the alternative case of using ground truth data from field visits, the samples would have been limited in number, fragmented in space and thus potentially less representative.\\

\noindent Despite the high accuracy of the k-means clustering, some rice instances are still mislabeled. However, these instances do not belong to a single cluster, yet they are split among other robust clusters. This suggests that there is not any specific pattern of rice paddy that is left out. Moreover, RF successfully manages to correctly classify a significant number of those mislabeled pseudo-labels. Specifically, the RF precision and recall scores for the rice class, when tested in the Seosan-Dangjin site, were  98.66\% and 93.99\%; offering an improvement over the pseudo-labels of +1.65\% and +2.17\%, respectively.\\

\noindent It is the excellent performance of unsupervised classification that allows for the generation of high quality labeled data to train an RF model. The success of clustering can be attributed to the nature of the target. Paddy rice is a distinctive crop type, mainly due to its inundation period, constituting it a particularly easy target to discriminate. This can be further supported by Figure \ref{fig:import3} that shows that features from the June05 instance are by far the most important.\\

\noindent Another important characteristic is the abundance of paddy rice in the study sites. For the year of inspection, paddy rice made up 78.2\% of the agricultural land of the Dangjin region, 71\% of the Seosan region, 54.19\% of the Haenam region and 74.6\% of the Cheorwon region. Paddy rice is the most dominant cultivation in the areas of inspection, which further enhances the performance of unsupervised classification. Additionally, South Korea is characterized by spatially continuous rice fields with minimal fragmentation. This again contributes to realizing satisfactory clustering results. \\

\noindent Therefore, it could be argued that for binary classification problems, which share similar characteristics, the implemented pseudo-labeling approach can be applicable. In fact, pseudo-labeling constitutes the overall methodology transferable. The three study sites of different agro-climatic conditions, geomorphological characteristics and land cover abundances, yielded excellent and comparable results; showcasing the robustness of the method. The idea of pseudo-labeling for crop classification could be even extended to multi-class scenarios. Multi-crop classification problems could be reduced to multiple OnevsAll binary classifiers, such as the one implemented in this work. \\

\noindent Finally, it should be noted that this approach can be employed even in cases that there is no available labeled data, using the photo-interpretation approach suggested in Section \ref{sec:validatio_data}. This way one can annotate a very small amount of rice and non-rice pixels to then use for choosing the best k value for the unsupervised classifier. Table \ref{tab:kmeans_per_size} presents the F1 scores for multiple k-means classifications in the Seosan-Dangjin site, calculated against a varying size of the photo-interpreted set of samples. Each experiment is evaluated against a spatially continuous subset of the labeled set across the Seosan-Dangjin site, which includes at least 5,000 pixels. Figure \ref{fig:kmenas_areas} depicts the locations of the seven sites (Site id 1-7) of Table \ref{tab:kmeans_per_size}. Site id 8 refers to the entire labeled set for the Seosan-Dangjin site (Figure \ref{fig:aois}).\\

\begin{table}[!ht]
\caption{\label{tab:kmeans_per_size} F1-score for the rice cluster of different k-means classifications against a varying size of the labeled data}
    \centering
    \scalebox{0.75}{
    \begin{tabular}{cc|cccccccc}
    \toprule
  &  & & & & k - clusters & & & &  \\
 \textbf{Site id} & \textbf{Labeled set size (pixels)} & \textbf{5} & \textbf{6} & \textbf{7}  & \textbf{8} & \textbf{9} & \textbf{10} & \textbf{11} & \textbf{12} \\ 
    \midrule
    \textbf{1} & \textbf{8K}   & 57.56      & \textbf{88.23} & \textbf{89.93} & \textbf{87.62} & \textbf{88.11}    & 61.99       & 60.97       & 63.34       \\
    \textbf{2} & \textbf{17K}  & 66.01      & \textbf{96.15} & \textbf{96.86} & \textbf{95.59} & \textbf{96.2}  & 80.14       & 79.69       & 80.85       \\
    \textbf{3} & \textbf{39K}  & 95.60       & \textbf{96.98} & \textbf{96.98} & \textbf{96.79} & \textbf{96.7}  & 87.65       & 87.21       & 87.99       \\
    \textbf{4} & \textbf{51K}  & 88.96      & \textbf{93.71} & \textbf{93.85} & \textbf{93.00}    & \textbf{93.07} & 85.40        & 85.21       & 85.59       \\
    \textbf{5} & \textbf{60K}    & 97.94      & \textbf{99.62} & \textbf{99.56} & \textbf{99.55} & \textbf{99.56} & 96.57       & 96.5        & 96.67       \\
    \textbf{6} & \textbf{85K}  & 93.10       & \textbf{95.47} & \textbf{95.92} & \textbf{94.78} & \textbf{95.2}  & 84.74       & 84.48       & 85.01       \\
    \textbf{7} & \textbf{260K}  & 96.46      & \textbf{96.57} & \textbf{96.68} & \textbf{95.98} & \textbf{96.13} & 90.26       & 90.12       & 90.55       \\
    \textbf{8} & \textbf{2.5M}     & 90.12      & \textbf{95.03} & \textbf{95.29} & \textbf{94.59} & \textbf{94.74} & 87.88       & 87.71       & 88.15       \\
   
    \bottomrule
    \end{tabular}}
\end{table}
% \endgroup

\noindent It is observed that for all eight experiments, the models with k $\in{[6,9]}$ exhibit the best performance, with marginal differences in their F-1 scores (<1\%). More specifically though, the model with $k=7$ achieves the best performance for all experiments, regardless of the size of the labeled set used.
This indicates the potential for easily generating confined labels through visual interpretation that is solely based on freely available earth observations (Sentinel-1 and Sentinel-2 data). \\
 
\begin{figure}[!hbt]
\begin{center}
		\includegraphics[width=11.5
		cm]{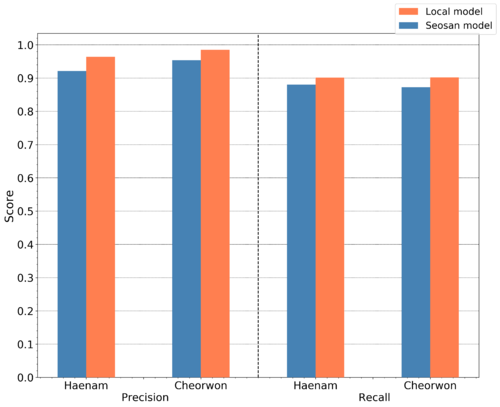}
	\caption{RF precision and recall for then Haenam and Cheorwon sites with predictions based on the i) Seosan-Dangjin model and ii) locally trained models}
	\label{fig:local}
\end{center}
\end{figure}

\noindent To further investigate the generalization of the RF model, a comparison analysis was conducted between the performance of the transferred model, from the reference site of Seosan-Dangjin, and the locally trained models in the Haenam and Cheorwon sites. The latter refer to trained models with locally sampled pseudo-labels (Table \ref{tab:kmeans_rice}). Figure \ref{fig:local} shows the precision and recall for both the locally trained models, which have been individually optimized, and the generalized model of the Seosan-Dangjin site. The differences between the local models and transferred model are minimal. Therefore, it could be argued that a single model from only a small region can be applicable to other areas. This is further supported by the results against the test set in Section \ref{sec:test_plots}.\\

\begin{figure}[!hbt]
\begin{center}
		\includegraphics[height = 6.5cm, width=9cm]{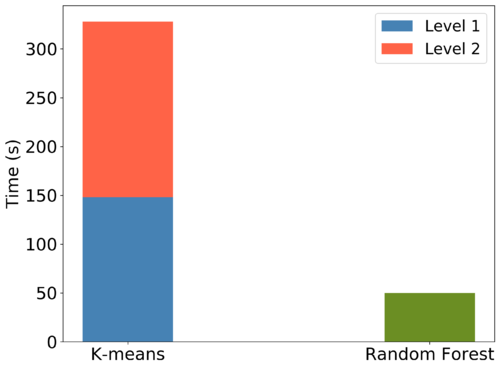}
	\caption{Computation time of k-means, showing Level 1 - land vs water clustering - and Level 2 - for k = 7 on the land cluster. Computation time of RF in Cheorwon site with depth = 50 and trees = 12}
	\label{fig:kmeansVSRFtimes}
\end{center}
\end{figure}

\noindent Furthermore, Figure \ref{fig:kmeansVSRFtimes} illustrates the execution times for the k-means clustering and RF predictions for the Seosan-Dangjin site.
The k-means execution time includes both levels, namely the binary land/water clustering (k=2) and the succeeding clustering on the land mask (k=7). Clearly, the RF model inference is significantly more efficient. The computational complexity reduction by not performing k-means clustering in each area is supported by the very good performance of the transferred model when compared to the local ones.\\

\begin{figure}[!ht]
\begin{center}
		\includegraphics[width=10.5cm]{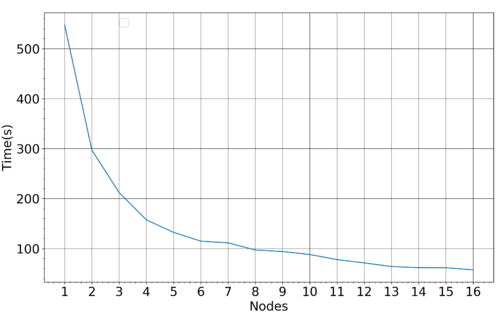}
	\caption{RF training time for a varying number of computing nodes - 4 M entities, 36 CPUs and 512 GB memory}
	\label{fig:rf_training_time}
\end{center}
\end{figure}

\noindent Figure \ref{fig:rf_training_time} illustrates the RF training time for the Seosan-Dangjin site against an increasing number of employed nodes in the HPDA. The training set used for these measurements comprise roughly 4 M pixels. \\

\noindent Each node is composed of 36 CPUs and 512 GB of memory. It is observed that the training time decreases exponentially with an increasing number of nodes, until it converges to approximately 50 seconds for more than 12 nodes. Additionally, applying the trained model to the Haenam and Cheorwon sites takes less than one minute for approximately 65 M pixels. Therefore, it would take at most one hour to produce a nationwide rice map for South Korea. \\
%%%%%%%%%%%%%%%%%%%%%%%%%%%%%%%%%%%%%%%%%%
\newpage
\subsection{Conclusions}\label{sec:conclusions}

\noindent A distributed RF classifier that was trained with pseudo-labels on a Sentinel-1 and Sentinel-2 data time-series offered an accurate, generalized and thereby transferable model for high spatial resolution paddy rice mapping. The model was trained in the reference site of Seosan-Dangjin and then fine-tuned for nationwide applicability based on two other sites of diverse characteristics. When tested in these two sites the model achieved more than 88\% recall and more than 91\% precision scores. The model was further tested on 40 test plots, evenly distributed across South Korea. The results revealed a successful model generalization, with an overall classification accuracy 96.7\% and kappa coefficient 0.87. Furthermore, the results were returned early in the year, in the first stages of the crop development that in turn allows for timely decision making. \\

\noindent The implemented pseudo-labeling method is fully dynamic and requires only a confined set of labeled data. Unsupervised classification alone proved to be effective in accurately classifying paddy rice. However, the exhaustive nature of k-means adds significant computational overhead, when framed in the context of large-scale applications. Nevertheless, it was shown that pseudo-labels can be extracted for only a small area and supervised learning models, such as RF, can generalize well to produce accurate rice maps in other areas. Finally, a distributed implementation of the entire semi-supervised pipeline was used in a HPDA environment to ensure the computational scalability of our approach.\\

\noindent All in all, it could be argued that the implemented paddy rice classification pipeline was designed to be dynamic, site independent and computationally scalable, exploiting exclusively satellite data that are freely available. Accurate, detailed and timely information on the paddy rice extent over large areas is instrumental to the national food security management and decision making. \\

\vspace{6pt}

%changed 20/04
\ifpdf
    \graphicspath{{Chapter4/Figs/Raster/}{Chapter4/Figs/PDF/}{Figs/Chapter3/}}
\else
    \graphicspath{{Figs/Chapter4/Vector/}{Figs/Chapter4/}}
\fi

\chapter{Resilient and smart farming}
\begin{figure}
\centering\includegraphics[scale=0.14]{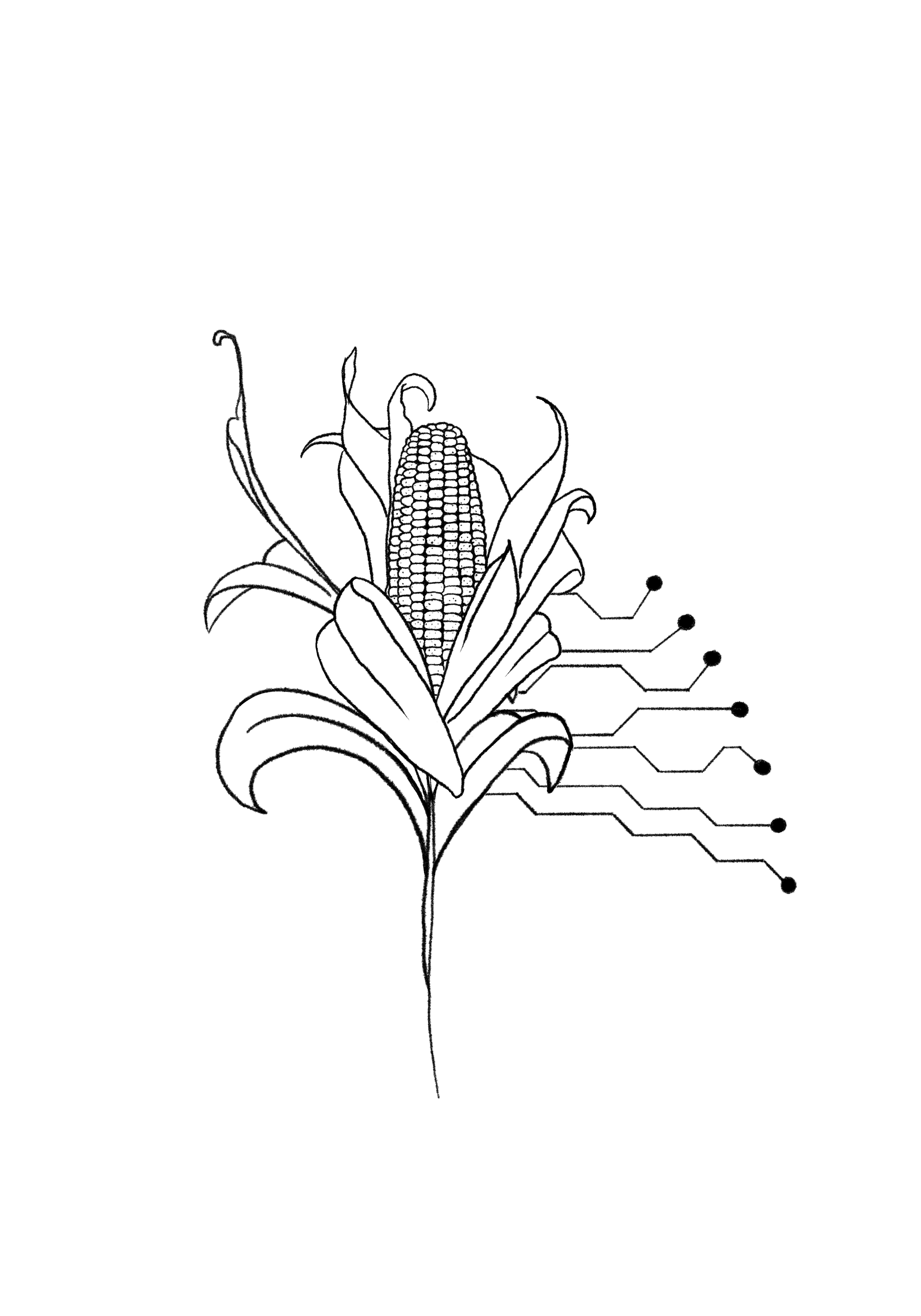}
\end{figure}

% \chapter{Resilient and smart farming}

\section{Introduction}

Crop phenology is crucial information for crop yield estimation and agricultural management. Traditionally, phenology has been observed from the ground; however
EO, weather and soil data have been used to capture the physiological
growth of crops. Section \ref{semipheno} elaborates on a dynamic phenology stage estimation
methodology for cotton towards early warning and mitigation advice against natural disasters. First, a time-series comparison algorithm, based on EO data, is used to assign pseudo-labels to approximately 1,000 parcels. For this, only a limited number of ground truth samples is employed. The pseudo-labels are then used to train RF regression models for phenology stage estimation. The pseudo-labeling process is used to augment the annotated dataset and allow for modelling the growth of cotton. The models are applied and evaluated on two different test sites in Greece; for which field campaigns were carried out to collect the labels. The results are satisfactory and showcase the successful generalization of the models to other areas. The dynamic predictions for cotton growth and extreme weather events, from numerical weather prediction models, are invaluable information for decision-making relevant to agricultural insurance schemes and farm management.\\

\noindent  In Section \ref{fuzzypheno}, a new approach for the within-season phenology estimation for cotton at the field level is proposed. For this, a variety of EO vegetation indices (Sentinel-2) and numerical simulations of atmospheric and soil parameters is exploited. The proposed method is unsupervised to address the ever-present problem of sparse and scarce ground truth data that makes most supervised alternatives impractical in real-world scenarios. A fuzzy c-means model is applied to identify the principal phenological stages of cotton and then used the cluster membership weights to further predict the transitional phases between adjacent stages. In order to evaluate the models, hundreds of crop growth ground observations were collected in Orchomenos, Greece. A new collection protocol is introduced, assigning up to two phenology labels that represent the primary and secondary growth stage in the field and thus indicate when stages are transitioning. The proposed model was tested against a baseline model that allowed to isolate the random agreement and evaluate its true competence. The results showed that the proposed model considerably outperforms the baseline one, which is promising considering the unsupervised nature of the approach. The limitations and the relevant future work are thoroughly discussed. The ground observations are formatted in an ready-to-use dataset and is available at
https://github.com/Agri-Hub/cotton-phenology-dataset.

%changed 20/04
\ifpdf
    \graphicspath{{Chapter4/Figs/Raster/}{Chapter4/Figs/PDF/}{Figs/Chapter4/}}
\else
    \graphicspath{{Figs/Chapter4/Vector/}{Figs/Chapter4/}}
\fi

% First work => semi-supervised 

\section{Semi-supervised phenology estimation} \label{semipheno}

\subsection{Literature review}
\noindent Crop yield is exposed to a number of risks that result in volatile farm profit and hence an unstable income for the farmers \citep{hardaker2004coping}. The timely knowledge of the occurrence and severity of upcoming weather perils is significant information for the development of risk management tools, but also the optimization of farm management and the control of inputs. In this context,  farmers, agricultural consultants and agricultural insurance companies could benefit greatly \citep{glauber2013growth}.\\

\noindent With respect to agricultural insurance schemes, the usually applied indemnity-based insurance is subject to asymmetries in information, and for this reason index-based insurance has been used as an alternative \citep{turvey2001weather}. However, unlike traditional compensation schemes, index-based insurance is not based on the actual farm losses but rather on exceeding an index threshold \citep{dalhaus2016can}. Given that the overall goal is the reduction of the farm losses and the increase of the transparency in the compensation process, indices should be able to describe the reality of the individual farm. This discrepancy between indices and actual farm losses is known as basis risk. The knowledge of phenology significantly decreases the basis risk and the expected utility for the farmers \citep{dalhaus2016can}.\\

\noindent The yield losses caused by adverse weather events can vary significantly, as there is high dependence on the phenological stage of the plant at the instance of the event. Therefore, instead of having fixed time windows for the index determination, phenology estimation can play a key role in the adjustment of the index results based on the expected impact of the disastrous event on yield. Big EO data that cover large areas with high frequency and at high spatial resolution have introduced new opportunities for the large-scale monitoring of phenology, without the need for costly and time-consuming manual observations \citep{jianwu2016emerging,sitokonstantinou2020sentinel}. \\

\noindent Additionally, proactive actions and mitigation measures against imminent adverse weather events, but also the management and scheduling of farm practices, can be assisted by the timely and accurate knowledge of the growth stage of the crop. For instance, soil fertilization can be applied earlier or foliar fertilization can be delayed based on the knowledge of an upcoming precipitation event. Another example is the early picking of cotton bolls, during the boll opening phase, in the case of an approaching extreme weather event. \\
In this study, we have implemented a dynamic phenology estimation methodology for cotton. To overcome the scarcity of labelled data for training the ML models, we have developed a pseudo-labeling technique that is based on a comparison analysis of EO data time-series from the Sentinel-2 mission. The pseudo-labels are used to train a RF regression model and its performance is evaluated on 27 parcels for which ground truth data was collected through field campaigns.

\subsection{Methodology}
\label{sec:methodology}
\subsubsection{Early warning for weather events}\label{early warning}
In order to predict the occurrence of an imminent extreme weather event, atmospheric parameters from numerical weather prediction (NWP) models are exploited. Currently, information both from a long-term/coarse-grid-spacing global model (15-days/0.25-degrees respectively) as well as a short-term/high-grid-spacing regional model (2.5 days/0.02-degrees) are available. The latter refers to an in-house configuration of WRF-ARW on a spatial configuration of 6-km grid spacing over Europe and 2-km over Greece. The model configuration in terms of resolution, as well as the microphysical schemes that are used, allow for an explicit resolution of complex processes such as the initiation of deep convection without the need for parameterization schemes. This also benefits the estimation of difficult to estimate processes such as hail growth, that are known to challenge the reliability of any NWP model. Forecasts from both sources are updated daily.\\

\noindent The model outputs that are used to identify approaching extreme weather phenomena include 2m temperature and soil temperature at different depths (0-10, 10-40 and 40-100cm), the wind speed at 10 m (gale risk), accumulated precipitation (flood-inducing heavy precipitation event risk) and  Convective Available Potential Energy - CAPE ( hailstorm risk).

\subsubsection{Phenology stage prediction}\label{phenology}
The phenology stage prediction methodology is comprised of two steps. Initially, a time-series comparison algorithm is used to generate predictions based on a limited amount of reference parcels, for which on-the-spot acquired growth stage timestamps are available (Section \ref{sec:heuristic}). 
The second step involves utilizing the predictions from the first step as pseudo-labels and train RF regression models (Section \ref{sec:RF}). 
The reference data refer to the collection of phenological stage timestamps for 10 parcels in Rodopi, Greece; as collected through field visits in 2018 and 2019. The growth stages that concern this study include the root establishment (germination and emergence), the leaf development, the squaring, the flowering, the boll development and the boll opening.
\begin{table}[!ht]
\centering
 \caption{\label{tab:table-name} Phenological stages with reference to this study's continuous scale, the possible DoY ranges in which a stage can occur and the expected ranges of duration}
    \begin{tabular}{|l|c|c|c|p{1.0\columnwidth}}
        \hline
        \textbf{Growth stage}                       & \textbf{Cont. scale} & \textbf{DoY range} & \textbf{Days} \\ \hline
        \textit{{Seeding}}        & 100               & 110-125                       & -              \\ \hline
        \textit{{Root Establishment}}        & 100 - 199               & 110-150                       & 15-25              \\ \hline
        \textit{{Leaf Development}} & 200 - 299               & 130-190                           & 25-40             \\ \hline
        \textit{{Squaring}}           & 300 - 399               & 160-215                         & 20-25             \\ \hline
        \textit{{Flowering}}          & 400 - 499               & 180-250                         & 35-45             \\ \hline
        \textit{{Boll   Development}} & 500 - 599               & 220-270                         & 20-25             \\ \hline
        \textit{{Boll   Opening}}     & 600 - 699               & 240-315                         & 25-45             \\ \hline
        \textit{{Harvest}}            & 700              & 265-315                               & -                 \\ \hline
    \end{tabular}
   
    \label{tab:stages}
\end{table}

\noindent Table \ref{tab:stages} lists the six growth stages, along with their corresponding values in i) a continuous growth scale that is used for the model predictions, ii) the range in DoY that stages can occur and iii) the range of their expected duration according to literature \citep{oosterhuis1990growth}. The continuous scale ranges from 100 to 700, where 100 refers to the seeding day and 700 refers to the completion of the boll opening stage. \\

\noindent \textbf{Pseudo-labeling} \label{sec:heuristic}
The methodology is based on the 1vAll comparisons of the time-series of an incoming parcel with the ones from the reference dataset. The time-series feature space comprises of Sentinel-2 images, including all spectral bands - except B01, B09 and B10 - and the vegetation indices NDVI, NDWI and Boll Opening Rate Index (BORI). Since phenology is dependent on the relative temporal progression of features and not their absolute values, the final parameter space is comprised of the slopes ($y'$) of the time-series with a step of $s$ (=5) days, as shown in Algorithm \ref{chapter4_pseudocode}. \\

% \begin{algorithm}
% \newcommand{\forcond}{i=DoP-tw \KwTo DoP}
% \SetKwInOut{Input}
% \SetAlgoLined
% \KwResult{Current phenological stage}
% \Input{data, ref\_data, DoP, tw, s}
% \ForEach{feature \textbf{in} data}{
%     y = data.subset(feature)\;

%     \For{\forcond}{
%         $y^{'}(i)$ = $[y(i)$-$y(i-s)]/s$\;
                
%     }
%     possible\_stages = get\_stages(DoP)\;
%     \ForEach{stage \textbf{in} possible\_stages }{
%         \ForEach{ref\_parcel \textbf{in} ref\_data}{
%             start = start\_day(stage,ref\_parcel)\;
%             end = end\_day(stage,ref\_parcel)\;
%             y\_ref = ref\_data.subset(feature,ref\_parcel)\;
%             \For{i=start \KwTo end}{
%                 \For{j=i-tw \KwTo i}{
%                     $y^{'}_{ref}(j)$=$[y_{ref}(j)$-$y_{ref}(j-s)]/s$\;
%                 }
%                 end\_pos = (i-start)/(end-start)\;
%                 pred\_stage = stage + 100*end\_pos\;
%                 errors.add( MAE($y^{'}$, $y^{'}_{ref}$))\;
%                 preds.add(pred\_stage)\;
%             }
%         }
%         sort(preds, by = errors) \;                             
%         results.add(preds[:3])\;
%     } 
% }
% final\_stage = median(results)\;
% \caption{Pseudo-labeling}
% \label{pseudocode}

% \end{algorithm}

% simpler edition

\begin{algorithm}
\caption{Pseudo-labeling}
\label{chapter4_pseudocode}
\textbf{Input:} $data$, $DoP$, $tw$, $s$\\
\textbf{Output:} $Current phenological stage$\\
\begin{algorithmic}
  
    \FOR{feature \textbf{in} data}
        \STATE y = data.subset(feature)\;
        \FOR{i in range(len(y))}
            \STATE $y^{'}(i)$ = $[y(i)$-$y(i-s)]/s$
        \ENDFOR
      \STATE possible\_stages = get\_stages(DoP)\;
    \FOR{stage \textbf{in} possible\_stages }
        \FOR{ref\_parcel \textbf{in} ref\_data}
            \STATE start = start\_day(stage,ref\_parcel)
            \STATE end = end\_day(stage,ref\_parcel)
            \STATE y\_ref = ref\_data.subset(feature,ref\_parcel)
            \FOR{i=start to end}
                \FOR{j=i-tw to i}
                    \STATE $y^{'}_{ref}(j)$=$[y_{ref}(j)$-$y_{ref}(j-s)]/s$
                \ENDFOR
                \STATE end\_pos = (i-start)/(end-start)
                \STATE pred\_stage = stage + 100*end\_pos
                \STATE errors.add( MAE($y^{'}$, $y^{'}_{ref}$))
                \STATE preds.add(pred\_stage)
            \ENDFOR
        \ENDFOR
        \STATE sort(preds, by = errors)                             
        \STATE results.add(preds[:3])
    \ENDFOR 
\STATE final\_stage = median(results)
    
    \ENDFOR

\end{algorithmic}
\end{algorithm}
% \label{chapter4_pseudocode}
% \begin{algorithm}[h]\label{pseudocode2}
% \newcommand{\forcond}{i=DoP-tw \KwTo DoP}
% % \SetKwInOut{Input}{Input}
% \SetAlgoLined
% \KwResult{Current phenological stage}
% \Input{data, ref\_data, DoP, tw, s}
% \ForEach{feature \textbf{in} data}{
%     y = data.subset(feature)\;

%     \For{\forcond}{
%         $y^{'}(i)$ = $[y(i)$-$y(i-s)]/s$\;
                
%     }
%     possible\_stages = get\_stages(DoP)\;
%     \ForEach{stage \textbf{in} possible\_stages }{
%         \ForEach{ref\_parcel \textbf{in} ref\_data}{
%             start = start\_day(stage,ref\_parcel)\;
%             end = end\_day(stage,ref\_parcel)\;
%             y\_ref = ref\_data.subset(feature,ref\_parcel)\;
%             \For{i=start \KwTo end}{
%                 \For{j=i-tw \KwTo i}{
%                     $y^{'}_{ref}(j)$=$[y_{ref}(j)$-$y_{ref}(j-s)]/s$\;
%                 }
%                 end\_pos = (i-start)/(end-start)\;
%                 pred\_stage = stage + 100*end\_pos\;
%                 errors.add( MAE($y^{'}$, $y^{'}_{ref}$))\;
%                 preds.add(pred\_stage)\;
%             }
%         }
%         sort(preds, by = errors) \;                             
%         results.add(preds[:3])\;
%     } 
% }
% final\_stage = median(results)\;
% \caption{Pseudo-labeling}
% \label{pseudocode}
% \end{algorithm}

\noindent In order to predict the phenological stage for a parcel, a time window ($tw$) that refers to the number of examined days prior to the Day of Prediction (DoP) must be defined. This way, we generate the feature subspace for the given DoP. For this study, $tw$ was set to 75 days. In other words, each time-series segment represents the last 75 days prior to the DoP. Then, based on the literature derived DoY ranges, as given in Table \ref{tab:stages}, we record all possible phenological stages for any given DoP. The $tw$-long feature subspace of the examined parcel is compared with multiple equivalent segments of the reference parcels using the Mean Absolute Error (MAE). These segments stem from sliding in time, from the start ($start$) to the end of each possible stage ($end$), for a given DoP. \\

\noindent The comparisons are made for each feature individually, and the three smallest errors for each are recorded. The prediction targets refer to the continuous phenology scale, as defined in Table \ref{tab:stages}.\\

\noindent Each error corresponds to a particular segment from the reference parcels and in turn to a specific value in the continuous scale. For instance, the prediction 510 refers to the 5th phenological stage (Boll development), with a 10\% completion percentage. The three segments with the smallest MAE are recorded for each parameter. The median value of the respective predictions for those recorded segments is the final growth stage prediction at a given DoP. \\

\noindent \textbf{RF regression model using pseudo-labels} \label{sec:RF}
The reference parcels are few in numbers and the extracted knowledge from those can only be fully representative for parcels of the same region and of similar agro-climatic conditions. Therefore, the time-series comparison prediction was applied to 994 cotton parcels in close proximity to the reference parcels. Due to the lack of quality available ground truth information and the need to generalize the phenology estimation methodology to be applicable in other areas, we examined a pseudo-labeling approach, using those first predictions as training data for an RF regression model. \\

\noindent Since the phenology prediction methodology needs to be dynamic and capable to execute at any time instance, multiple RF models are trained for every 5 days throughout the growing cycle of cotton. Therefore, predictions are made every 5 days, using different RF models of increasingly larger feature spaces. The feature spaces comprise of images from seeding to the DoP. Each model has been fine-tuned individually. 

% To reduce the computational complexity the time-series are temporally interpolated with a time-step of 5 days. 

\subsection{Experimental Results}
\label{sec:results4}

\noindent The validation data, based on which the performance of the two phenology estimation techniques were evaluated (Sections \ref{sec:heuristic} and \ref{sec:RF}), has been collected through a field campaign on 16 parcels in Rodopi, Greece and 11 parcels in Thessally, Greece. Two experts visited each of the fields 3 or 4 times from the beginning of August 2020 until their harvest. On each visit, the experts recorded the prevailing phenological stage of the parcel in the BBCH scale, which was then translated to the continuous scale of this study~\citep{meier1997growth}. The BBCH scale makes use of a two digit representation, with the first digit referring to the principal growth stage and the second digit describing the secondary growth stage that corresponds to an ordinal number of percentage value. The BBCH scale ranges from 00 to 99. \\
% Figure \ref{fig:RFerrors} shows the train and test MAE errors in the continuous scale for 15 RF models at different DoY. The 994 parcels in Rodopi, for which pseudo-labels have been assigned based on the time-series comparison method, have been randomly split to 70\% training and 30\% test sets. The errors in Figure \ref{fig:RFerrors} have been calculated against the pseudo-labels. It is worth-noting that the temporal progression of errors appears to follow the growth cycle of cotton. A faster rate of change in biomass translates to larger errors, which is indeed reasonable.
% Table \ref{tab:AccError} shows the percentage of predictions that were found to 

\noindent Table \ref{tab:AccError} shows the percentiles of predictions with respect to error ranges in the continuous scale. The results are given for the two validations datasets, namely of Thessally and Rodopi, separately. It should be noted that the Thessaly region is situated far from both the reference data and the pseudo-labels. For both regions, but particularly for Thessaly, the RF performs better than the time-series comparison method (pseudo-labeling). Even though the validated samples are limited, it could be argued that the model generalizes well, providing satisfactory results when transferred to different regions.

\begin{table}[!ht]
\centering
\caption{Percentiles of predictions of RF and pseudo-labeling (PL) for different ranges of error in the continuous scale}

\begin{tabular}{|c|c|c|c|c|}
\hline & \multicolumn{2}{c|}{\textbf{Thessaly}} & \multicolumn{2}{c|}{\textbf{Rodopi}} \\ \hline
\textbf{Error}   & \textbf{RF}     & \textbf{PL}     & \textbf{RF}      & \textbf{PL}     \\ \hline
\textbf{\textless 10} & 29\%   & 31\%       & 24\%       & 31\%   \\ \hline
\textbf{\textless 20} & 76\%   & 51\%       & 54\%       & 51\%   \\ \hline
\textbf{\textless 30} & 100\%  & 74\%       & 63\%       & 64\%   \\ \hline
\textbf{\textless 40} & -      & 86\%       & 80\%       & 82\%   \\ \hline
\textbf{\textless 50} & -      & 96\%       & 92\%       & 91\%   \\ \hline
\end{tabular}
 \label{tab:AccError}
\end{table}
\noindent The MAE of the two methods has been averaged over all predictions for the 97 field visits and was computed in both the continuous scale and in days. The pseudo-labeling method resulted in a MAE of 23.98 in the continuous scale and 6.88 in days, while the RF regression gave a MAE of 20.33 and 5.82 days. The results are satisfactory, particularly given the inherent ambiguity of the target. 
The different plants in a parcel do not all grow exactly at the same pace. Therefore, both the reference and validation data, which were given with a single BBCH description for a single DoY timestamp, are subject to observation errors. The experts have quoted that even though they are confident of their aggregated decision, they have witnessed an intra-parcel deviation for the growth stage of up to 4 days. The results are within the limits of this aggregation error and thus estimation errors should not be interpreted based on their absolute value but rather their relative importance. Having said that, both approaches appear to perform well.

\subsection{Discussion}
\label{sec:Discussion}
\noindent The prediction of extreme weather events combined with the timely and large-scale knowledge of the current growth stage is invaluable information for pertinent decision making. An indicative application would be an alerting mechanism that results in i) prevention and mitigation actions and ii) evidence-based insurance processes. Below follow some examples in order to showcase this argument.\\
% in the case of natural disaster induced damages.

\noindent \textbf{Ideal conditions for seeding based on temperature predictions}. Temperatures exceeding 16 \degree C at the top 20 cm of soil for 10 consecutive days are required for the germination of cotton \citep{Bradow2010}. Soil moisture should also be adequate, with the value being dependent to the soil type of each parcel. For non irrigated parcels an indicative average rainfall of 50mm should be recorded prior to seeding \citep{Bradow2010}.\\

\noindent \textbf{Predicted heatwaves and intensification of irrigation.} This is relevant particularly during the flowering and end of flowering stages.Temperatures larger than 32 \degree C during flowering, but also during boll development can be thought as threshold temperatures \citep{ashraf1994tolerance}. Nevertheless, the knowledge of an imminent heatwave is important for every stage of the cultivation. \\

\noindent \textbf{Interruption of irrigation based on the estimation of the current phenological stage and expected rainfall.} Irrigation can be interrupted on the onset of boll opening \citep{Cotman}, in order to stop the continuous growth of cotton and for the photosynthetic carbohydrates to start contributing to the development of bolls and not the development of leaves and flowers. On the other hand, the interruption of irrigation at the physiological cutout could be compromised due to an upcoming rainfall, with a significant cost to the yield. \\

\noindent \textbf{Application of fertilizers depending on rainfall predictions.} There is a multitude of fertilization methods, i.e. application of standard fertilization prior and during seeding and application of superficial fertilization at the first stages of the plant (plant height 10-15 cm - prior to squaring) \citep{mcconnell2005yield}. The knowledge of an imminent rainfall can assist in rushing the fertilization prior to the event in order to better integrate the fertilizer. Also, knowledge of the volume of rainfall is important in order to avoid leaching at the lower soil layers. Furthermore, foliar fertilization and fungicide/insecticide application could be postponed based on the knowledge of an imminent rainfall, avoiding rinsing prior to their absorption.\\

\noindent \textbf{Prediction of adverse weather event and early harvest}. The knowledge of imminent gale, frost, hail and flooding risk near the end of the cultivation can trigger an earlier harvest of cotton. If the hazard and the consequent damage are expected to be severe, then the early harvesting is justified even in cases of merely 30\% of bolls are open.\\

\subsection{Conclusions}
\label{sec:typestyle}
\noindent In this study we implemented a dynamic phenology prediction pipeline that is based on generating pseudo-labels to then train RF regression models. The available ground truth data is scarce and limited to a single region. Therefore the proposed semi-supervised methodology can potentially provide a scalable and geographically transferable solution. The time-series comparison and the RF regression methods provided satisfactory and comparable results. However, within the context of large scale applications, the RF solution is more computationally efficient than the exhaustive time-series comparison equivalent and with greater potential for generalization. \\

\noindent The results presented showcase the performance of the methods only for the last stages of cotton growth. In the next section, the methodology is tested for the full growth cycle in order to inspect and understand the potential performance differences among the various growth stages. Nonetheless, the results clearly illustrate the potential of dynamically identifying the growth stages over large areas with only minimal ground truth information. This in turn can have great impact towards a more resilient agricultural sector, both from the farm management and agricultural insurance perspective.

\section{Unsupervised phenology estimation}\label{fuzzypheno}

\subsection{Literature review} \label{sec:related}

\noindent Crop phenology is key information for crop yield estimation and agricultural management and thereby actionable knowledge for the farmer, the agricultural consultant, the insurance company and the policy maker. Crop phenology is the physiological development of the plant from sowing to harvest. The precise and timely knowledge of the growth status of crops is crucial for estimating the yield early in the season, but also for taking prompt action on controlling the growth to i) maximize the production and ii) reduce the farming costs \cite{gao2021mapping}.\\

\noindent Crops' water needs are a function of the phenological stage. Using the example of cotton, which is the crop of interest in this study, there is higher water usage between the flowering and early boll opening stages than in the emergence and late boll opening stages \cite{vellidis2016development}. Irrigation can be interrupted on the onset of boll opening to stop the continuous growth of cotton and allow the photosynthetic carbohydrates to start contributing to the development of bolls and not the development of leaves and flowers \cite{Cotman}.  Therefore, we need crop phenology information in order to make irrigation recommendations towards fully utilizing the expensive and often scarce water, and at the same time reduce water stress and its potential adverse effects on the yield \cite{anderson2016relationships}. Irrigation is one of the many examples of how phenology can benefit the agricultural practice management. Other examples include the application of plant growth regulators,, pest management and harvesting \cite{gao2021mapping}. For instance, pix (mepiquat chloride), which is the most widely used cotton growth regulator, when applied on the early flowering stage can reduce the  excessive cotton vegetation growth and therefore  reduce the probability of diseases and also improve lint yield and quality \citep{reddy1996mepiquat}. In the same manner, cotton picking could be rushed prior to an anticipated extreme weather event  (e.g.,  hail), if phenology estimations show a near complete boll.\\

\noindent For many years, phenology has been observed from the ground, through field visits and in-situ sensors. These approaches however are expensive, time-consuming and lack spatial variability. To this end, space-borne and aerial remote sensing VI time-series have been used to systematically monitor crop phenology over large geographic regions; often termed land surface phenology \cite{gao2021mapping, duarte2018qphenometrics}. The freely available Sentinel-2 data offer optical imagery of high temporal and spatial resolution that introduced new opportunities for the large-scale and within-season monitoring of phenology \cite{jianwu2016emerging,sitokonstantinou2020sentinel}.\\ 

% \noindent However, there are still a number of challenges to be tackled. First of all, cloud coverage disrupts the continuity of the optical Satellite Image Time-Series (SITS), which is particularly damaging when monitoring fast progressing targets like phenology. Additionally, the sensitivity of SITS on crop growth varies for different crop types and different growth stages. To this end, there have been multiple studies that additionally incorporate atmospheric and soil parameters that unlike VIs do not monitor the change in vegetation but the drivers of these changes.\\

\noindent The recently published positioning papers \cite{gao2021mapping,lacueva2020multifactorial} and \cite{potgieter2021evolution}, identify the problem of remote phenology estimation as a fundamental one for the future of agriculture monitoring. Particularly, the authors underline the importance and the expected impact of within-season estimations at high spatial resolutions. Many of the related studies focus on the prediction of few principal phenological stages, failing to truly exploit the frequency of remote sensing data. This is mostly true because the required ground observations for training and/or evaluation are usually infrequent and lack spatial variability. This kind of ground observations are usually achieved with the use of networks of phenology stations or phenocams, which are always sparse and limited in number. \\

\noindent In the past two decades, there has been a number of related studies that focus on the estimation of vegetation phenology using both EO and weather data, under a wide variety of methodological frameworks. Initial approaches to the problem, many of which continue to develop to this day, offered after-season phenology estimations and were usually applied at large geographic scales using medium resolution imagery \cite{liang2011validating,xin2020evaluations, almeida2012remote, verbesselt2010phenological, tian2020development, bolton2020continental, dineshkumar2019phenological,chen2016simple, sitokonstantinou2020sentinel}. The term after-season indicates that phenology is estimated after the crop is harvested and thus leverages the entire data time-series. This large-scale monitoring of the dynamics of phenology has been very popular in the scientific domains of ecology and climate change monitoring. Nevertheless, detailed information that is offered within the cultivation period is very important from the perspective of the farmer. Using timely and high spatial resolution phenology predictions, farmers can protect their yield and maximize their profit. Towards this direction, there has been a number of studies that provide within-season phenology predictions at the field level \cite{lopez2013estimating, yang2017improved, nieto2021integrated,zheng2016detection,czernecki2018machine,vicente2013crop,de2016particle}. Phenology estimation can be found in literature both as a classification and as a regression problem. For the first, the phenological cycle is divided into stages or classes that last for a given period of time \cite{lopez2013estimating,yang2017improved, nieto2021integrated}. Usually the crop growth period is broken down into i) the sprouting, ii) the vegetative, iii) the budding, iv) the  flowering and v) the ripening phases. One common generic distinction of such stages is i) the sprouting, ii) the vegetative, iii) the budding, iv) the flowering and v) the ripening stages. On the other hand, phenology as a regression problem translates to predicting the day of the onset of these key phenological phases, such the ones mentioned previously \cite{zheng2016detection,sakamoto2010two,czernecki2018machine,de2016particle}.\\
% Additionally, land surface phenology methods can be categorized as curve-based or trend-based \cite{gao2021mapping}. The curve-based methods rely on historical data from previous years in order to learn the phenological stages \cite{sakamoto2010two, zeng2016hybrid}. On the other hand, trend-based methods attempt to identify significant changes in the trend of EO time-series that would indicate the transition to a different growth phase \cite{gao2020within, gao2020detecting}. 

% \noindent The rest of this section is structured as follows. Section \ref{sec:related} presents related publications. Section \ref{datasets} describes the field campaign that was performed to collect the ground observations. This section also describes the predictor variables - optical, atmospheric and soil parameters - and how they were processed. Section \ref{methods} describes the phenology prediction methodology and the evaluation metrics. Section \ref{sec:experiments} demonstrates the experiments performed and their results, whereas Section \ref{discussion} discusses the results, underlines the limitations and suggests future work. Section \ref{conclusion} concludes this work.

\noindent Each crop type has its own growth cycle and hence unique characteristics with respect to i) how phenology is affected by agro-climatic conditions and ii) how responsive are the space-borne or aerial EO data with respect to the various growth stages. 
% These particularities would determine which are the appropriate data and methods to estimate the phenology of each crop, and how well the methods will ultimately perform. 
There are crop types for which phenology and yield are highly correlated to the vegetation canopy we see from EO platforms, e.g., tobacco. This is not the case for other crops, especially those that bear fruits or bolls, where phenology and vegetation canopy are not closely coupled. In literature, one can find many publications related to the estimation of phenology for rice, barley, soybean and maize \cite{yang2017improved, lausch2015deriving, zeng2016hybrid, sakamoto2010two, nieto2021integrated}. In this study, however, we focus on the rather underrepresented crop type of cotton (Gossypium hirsutum L.), which is a unique case with non-linear relationships between growth and the VIs that we have in our capacity to monitor it \cite{toulios1998spectral, gutierrez2012association, jiang2018quantitative}. In literature, one can find many publications related to the estimation of phenology for rice, barley, soybean and maize \citep{yang2017improved, nieto2021integrated,sakamoto2010two,lausch2015deriving,zeng2016hybrid}. However, cotton appears to be underrepresented. When searching for phenology estimation studies on cotton, one can find only few publications that date back more than a decade \citep{palacios2012derivation,tsiros2009assessment}, and some more recent ones that deal with multiple crop types and do not focus explicitely on cotton \citep{sakamoto2018refined,vijaya2021algorithms} . There are alsoa handful of papers that evaluate the process-based model CSM-CROPGRO-Cotton, but with small-scale experiments (few fields) \citep{ur2019application,li2019simulation,mishra2021evaluation}. On the other hand, there are dozens of recent papers that focus on the large-scale predictionof phenology for other major crops \citep{misra2020status,zeng2020review}. Indicatively, there have been interesting recent studies on maize \citep{zeng2016hybrid,zeng2020review,gao2020within,niu202230,huang2019optimal,diao2020remote}, rice \citep{huang2019optimal,luo2020chinacropphen1km,onojeghuo2018rice}, wheat \citep{huang2019optimal,nasrallah2019sentinel,mercier2020evaluation} and soybean \citep{zeng2016hybrid,gao2020within,diao2020remote}.\\

\noindent Phenology is affected by the temperature \cite{chuine2010does, cleland2007shifting}, the photoperiod and the effective solar radiation that enables photosynthesis \cite{flynn2018temperature, korner2010phenology}, the soil properties \cite{menzel2002phenology}, and many other agro-meteorological parameters \cite{piao2019plant}. Indeed, in cotton phenology literature we can find older studies that use exclusively meteorological data, such as soil and/or air temperatures \cite{tsiros2009assessment,reddy1993temperature, reddy1994modeling} and other more recent ones that combine them with optical images \cite{nieto2021integrated,czernecki2018machine,cai2019integrating}. SAR data, usually in combination with optical images, have been mostly used for estimating rice phenology \cite{yang2017improved}, but also other crop types \cite{meroni2021comparing}. The combination of Sentinel-2 and MODIS has been one of the most popular in the field. This is true because the Sentinel-2 missions offer data of high spatial resolution that enable information extraction at the field level, whereas MODIS data and their daily acquisitions, in contrast with the 5-days of Sentinel-2, allow for the generation of dense SITS \cite{sakamoto2010two,de2016particle,zheng2016detection,arun2021deep,vicente2013crop,zhang2003monitoring,zeng2016hybrid,vina2004monitoring}. Other data sources found in literature include Unmanned Aerial Vehicles (UAV) \cite{yang2020near,selvaraj2020machine} and in-field RGB sensors \cite{wang2021deepphenology}. \\

% Most studies in order to address the problem of phenology estimation use supervised learning approaches, either traditional ML \cite{nieto2021integrated, czernecki2018machine, 9553456} or DL algorthims \cite{arun2021deep}. There also techqniches 

\noindent There are several published studies that employ supervised Machine Learning (ML) methods for land surface phenology. For instance, the authors in \cite{nieto2021integrated} have used SVM and RF to integrate field, weather and satellite data for maize phenology monitoring. Furthermore, the authors in \cite{czernecki2018machine} and \cite{9553456} have used traditional ML regressors to model plant phenology based on both satellite EO and gridded meteorological data. There have also been a few DL based approaches. One example is \cite{arun2021deep}, where the authors explore the use of capsules, i.e., a group of neurons to address the issues of translation invariance prevalent in conventional (Convolutional Neural Networks) CNN, to learn the characteristic features of the phenological curves. There is also a number of methods that do not use ML. A few examples of such methods include dynamic multi-temporal modeling and Kalman filtering in \cite{vicente2013crop}, particle filtering in \cite{de2016particle}, sigmoid modeling in \cite{zhang2003monitoring}, first-derivative
analysis in \cite{meroni2021comparing} and \cite{zheng2016detection}, and wavelet-based filtering and shape model fitting in \cite{sakamoto2005crop}.\\

\noindent In this work, we exploit EO data (Sentinel-2), together with numerical simulations of atmospheric and soil parameters  (\ref{tab:variables}), to address the within-season phenology estimation for cotton at the field level. Even more, since ground truth data are scarce and expensive to collect, we predict phenological stages using clustering, in order to be truly useful in real world scenarios. We go beyond the estimation of principal phenological stages and additionally identify the fuzzy transitions between stages as individual metaclasses (two ranked labels). We focus on cotton that is a vital crop for the Greek economy and agricultural ecosystem, and even more has been underrepresented in the phenology estimation literature. 
% Indicatively the search "TITLE (cotton phenology)" returns 36 results, while the same search for wheat, rice, maize (or corn) and soybeans returns 228, 168, 153 and 62 publications respectively. 
Finally, we developed and made publicly available a unique dataset of cotton growth ground observations, collected by an expert who performed hundreds of field visits in Orchomenos, Greece.\\

\subsection{Materials}
\label{datasets}

\subsubsection{Ethics Statement}

\noindent The field campaigns were conducted in cotton fields in Orchomenos, Greece, which are privately owned by the members of the agricultural cooperative of Orchomenos.A memorandum of understanding was signed with the cooperative that explicitly allowed to perform visits in selected fields. During the experiments, no other specific permission was required, as only observational activities were carried out and no endangered or protected species were involved.

\subsubsection{Study Area and Field Campaigns}
\label{campaigns}

\noindent Greece has the fourth largest production of cotton per person (approx. 29kg) and is the number one producer in the European Union (EU), with 304,000 tons per year \citep{faostat}. Cotton is extensively cultivated and is very important for the national economy, with Greece being the fifth largest exporter in the world. Unfortunately, however, there is no organized effort to record practice calendars and phenological observations. In Greece, cotton needs between 150 to 200 days in order to complete its phenological cycle. The duration depends on the cotton variety and the agro-climatic conditions  \cite{tsiros2009assessment}.\\

\begin{figure}[!ht]
    \centering
    \includegraphics[width=10cm]{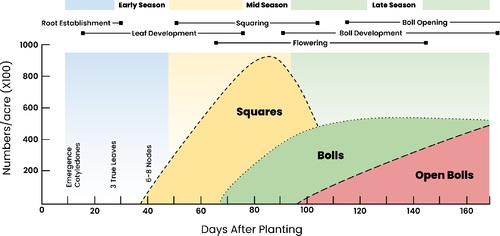}
    \caption{The phenological cycle of cotton in Greece. The principal phenological stages of cotton are Root Establishment (RE), Leaf Development (LD), Squaring (S), Flowering (F), Boll Development (BD) and Boll Opening (BO). The temporal overlaps between adjacent phenological stages are illustrated. Retrieved from \citep{oosterhuis1990growth} and modified}
    %SMR 1 requirement mean execution time}
    \label{fig: phenologicalcyclecotton}
\end{figure}

\noindent For this study, cotton growth consists of six principal phenological stages. These refer to higher level groupings of the cotton growth micro-stages defined in the official manual for damage assessment of the Greek Agricultural Insurance Organization (ELGA) \cite{elga}. These groupings have been made after consulting experts in cotton growth and the relevant literature \cite{oosterhuis1990growth}. Fig \ref{fig: phenologicalcyclecotton} illustrates the phenology of cotton in Greece. The first stage is Root Establishment (RE), referring to the period from sowing to the development of three leaves. This stage lasts between 15 to 30 days, but this can be greatly affected by weather conditions and particularly low temperatures that can slow down the process \cite{tsiros2009assessment}. The second stage is Leaf Development (LD) and encompasses the period from the development of the fourth leaf to the appearance of the first squares. This usually takes between 35 to 45 days, but once again it is subject to weather conditions \cite{danalatos2007introduction}. The third growth stage is Squaring (S) that includes the period between the formation of the first squares to the appearance of the first flowers. This stage takes between 15 to 30 days. Then the first flowers open with the onset of the Flowering (F) stage that lasts for 20 to 40 days. Then follows the Boll Development (BD) stage that takes 25 to 45 days until the start of leaf discoloration and the onset of Boll Opening (BO) that lasts roughly 10 to 20 days until harvest.\\

\noindent It should be noted that phenology is a dynamic variable and in principle can be described in great detail, going beyond these six principal crop growth stages. At any given instance, a cotton plant can be characterized by a combination of adjacent stages. For example, during the late flowering stage, a plant would have both flowers and cotton bolls, i.e., it would be transitioning to the BD stage. One of the most popular crop growth identification scales is the BBCH  \cite{meier1997growth}. The BBCH scale makes use of a two digit representation, with the first digit referring to the principal growth stage and the second digit describing the secondary growth stage that corresponds to an ordinal number of percentage value. The BBCH scale ranges from 00 to 99. However, collecting ground observations of this detail is a challenging task because i) a large number of samples cannot be observed in near-daily frequency and ii) it is difficult, even for experts, to assign precise growth stages, especially when this decision needs to be aggregated at the field level.\\

\begin{figure}[!ht]
    \centering
    \includegraphics[width=14cm]{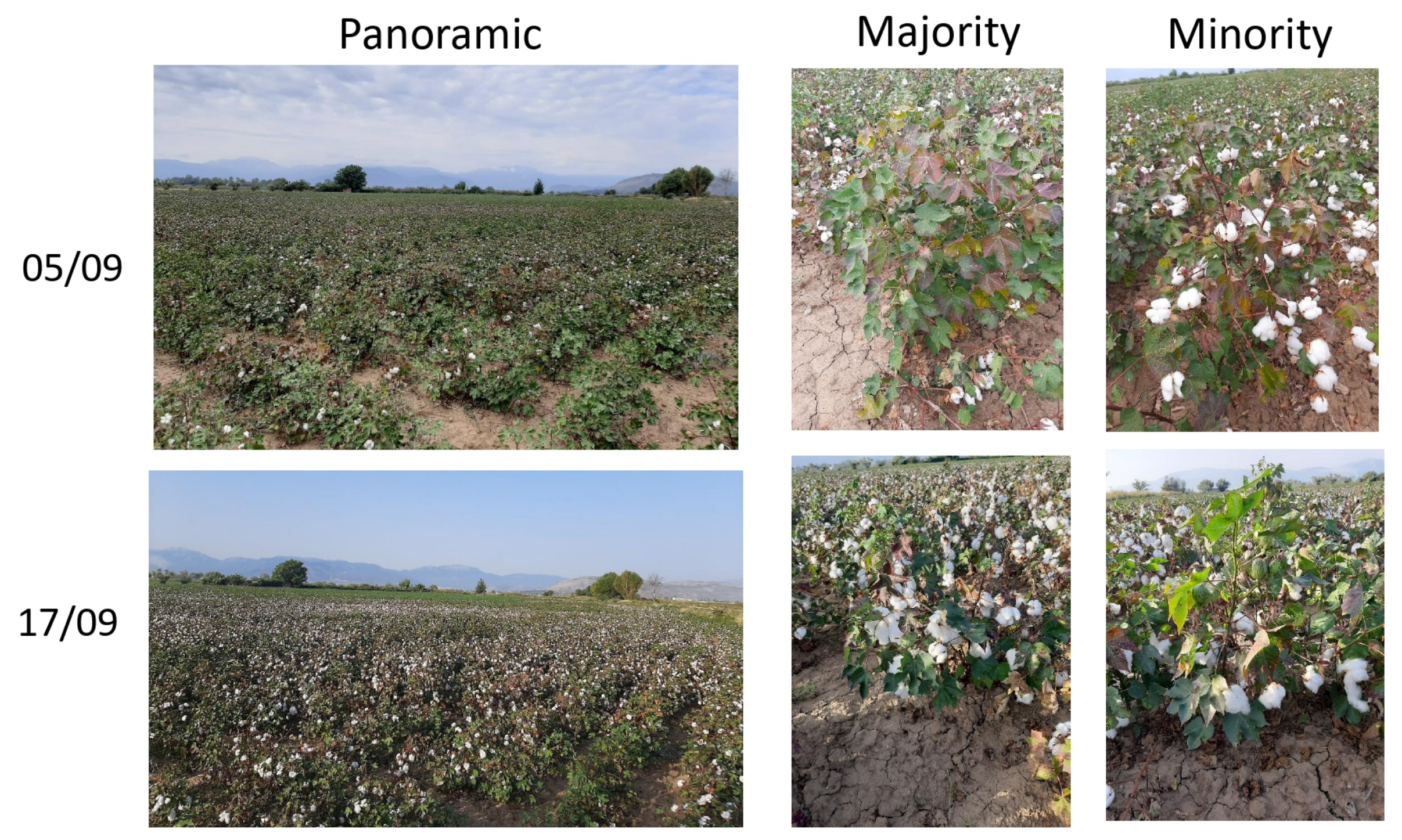}
     \caption{Examples of field photos.
    Field photos that were captured in two consecutive visits of the same field. For each visit there is a panoramic photo of the field and two close-up photos of a plant that represent the majority and minority of the field, respectively. \textbf{(a)} Visit at 05/09, when the primary stage was BD and the secondary stage was BO. \textbf{(b)} Visit for 17/09, when the primary stage was BO and secondary was BD.}
    \label{fig:bo_bd}
\end{figure}

\noindent In order to collect ground truth data that would allow us to evaluate the models of this study, an agronomist, who is a cotton grower and seasoned field scouter, performed an extensive and intensive field campaigns in Greece. The campaigns took place during the growing season of 2021, from root establishment to boll opening, which extends between late April and early October. The expert followed the instructions that are summarized below: 
\begin{itemize}
    \item  At least 15 visits per field (approx. 3 per month) during the growing period, including at least one visit per phenological stage. 
    \item  Ideally, visit the fields in the days that Sentinel-2 passes over. If this is logistically impossible, visit the fields maximum one day prior or after the Sentinel-2 pass. 
    \item  If it is cloudy, check the next Sentinel-2 pass, consult weather forecasts and decide if the inspection could be delayed for a few days or should happen irrespective of the cloud coverage.
    \item  Walk with a zig-zag pattern for typical scouting through the field and inspect the growth status and how it varies in space.
    \item  Decide on the phenological stage, choosing among the six principal stages that were defined earlier, which best describes the majority of the plants in the field. If the field is in a transitioning phase between two phenological stages, mention both and decide which is the prevailing one, i.e., the primary stage. It should be noted that the phenological stages must be among the six principal phenological stages defined earlier in this section.
    \item  Decide on the percentage that is explained by the primary and the secondary stage
    \item  Take a panoramic photo of the entire field. Take two close-up photos of plants. The first one should be representative of the majority of the plants in the field. The second one should be representative of a minority of plants in the field. The latter close-up photo should be captured only when the percentage of the minority class, in terms of area, is deemed significant (Fig. \ref{fig:bo_bd}).

\end{itemize}

\noindent Fig. \ref{fig:bo_bd} helps to further illuminate what is meant by the terms primary, secondary and percentage of prevalence. \\

\noindent The close-up photo labelled "majority" shows a representative plant of this aggregate status of the the field, also conveyed by the panoramic photo, which can be described with BD as primary and BO as secondary. On the other hand, the close up photo labelled "minority" refers to a plant that is less common in the field and is more representative of the secondary stage. This becomes even more clear by inspecting the "minority" close-up photo for the 05/09 visit, which shows a plant that is well into the BO phase. The expert would identify the percentage of prevalence based on the area that the plants represented by the "majority" and "minority" close-up photos cover in the field, but also the average density of the two stages in each plant. In this example and for the 05/09 visit, BD was 100\% and BO was 80\% prevalent, which means that BD was found in every plant of the field while BO on 80\% of the plants. Respectively, for the 17/09 visit BO was 100\% and BD was 30\% prevalent.\\

\noindent The close-up photo labelled "majority" shows a representative plant of this aggregate status of the the field, also conveyed by the panoramic photo, which can be described with BD as primary and BO as secondary. On the other hand, the close up photo labelled "minority" refers to a plant that is less common in the field and is more representative of the secondary stage. This becomes even more clear by inspecting the "minority" close-up photo for the 05/09 visit, which shows a plant that is well into the BO phase. The expert would identify the percentage of prevalence based on the area that the plants represented by the "majority" and "minority" close-up photos cover in the field, but also the average density of the two stages in each plant. In this example and for the 05/09 visit, BD was 100\% and BO was 80\% prevalent, which means that BD was found in every plant of the field while BO on 80\% of the plants. Respectively, for the 17/09 visit BO was 100\% and BD was 30\% prevalent. \\

\noindent During the growing season of 2021, our expert made 1285 visits to 80 cotton fields in Orchomenos. Orchomenos is an agrarian municipality in Viotia district of central Greece. The fields that participated in the ground observation campaigns are part of the agriculture cooperative of Orchomenos cooperative that has the highest selling price for cotton in Greece. It should be noted that among the 80 fields, 10 different cotton varieties were cultivated. This variability is important, as one can evaluate the performance of the phenology estimation models and draw conclusions on their generalization. The field visits were appropriately scheduled in order to have minimal differences between ground and satellite observations. In total, we acquired 67 different Sentinel-2 images, from mid March until the end of October. The mean difference between the ground and the cloud-free Sentinel-2 observations was 0.86 days and the standard deviation was 0.89 days. Table \ref{tab:day_diff} depicts the distribution of the difference in days between the ground and the satellite observation pairs. \\

\begin{table}[!ht]
\centering
\caption{
{Difference in days  between ground and satellite observations pairs.}}
\begin{tabular}{|c|c|c|}
\hline
\textbf{Distance in days} & \textbf{\#Cloud-free S2 captures} & \textbf{Cum. Frequency(\%)} \\ \hline
\textbf{0}                & 475     & 37 \\ \hline
\textbf{1}                & 594     & 83 \\ \hline
\textbf{2}                & 173     & 97 \\ \hline
\textbf{\textgreater{}3}  & 43      & 100  \\ \hline
\textbf{Total}            & 1285    & -   \\ \hline
\end{tabular}

\label{tab:day_diff}
\end{table}

\noindent Table \ref{table:observations} shows the number of primary and secondary ground observations for each principal phenological stage. The expert was instructed to assign a primary stage label to any visit, thus the number of primary observations equals to the number of field visits. On the other hand, a secondary stage is not necessarily present, since it is observed only in a transitioning phase between two principal phenological stages (e.g. from flowering to boll development). Specifically, a secondary stage was observed only in 669 out of the 1285 visits (52\%).\\

\begin{table}[!ht]
\centering

\caption{The number of ground observations for each principal phenological stage of cotton that have been classified as primary and secondary stage labels.}
\begin{tabular}{c|cc|}
\cline{2-3}
\multicolumn{1}{l|}{}                & \multicolumn{2}{c|}{\textbf{Ground observations}} \\ \hline
\multicolumn{1}{|c|}{\textbf{Stage}} & \multicolumn{1}{c|}{\textbf{Primary stage}} & \textbf{Secodary stage} \\ \hline
\multicolumn{1}{|c|}{\textbf{RE}}    & \multicolumn{1}{c|}{75}            & 4            \\ \hline
\multicolumn{1}{|c|}{\textbf{LD}}    & \multicolumn{1}{c|}{421}           & 20           \\ \hline
\multicolumn{1}{|c|}{\textbf{S}}     & \multicolumn{1}{c|}{212}           & 5            \\ \hline
\multicolumn{1}{|c|}{\textbf{F}}     & \multicolumn{1}{c|}{229}           & 148          \\ \hline
\multicolumn{1}{|c|}{\textbf{BD}}    & \multicolumn{1}{c|}{252}           & 315          \\ \hline
\multicolumn{1}{|c|}{\textbf{BO}}    & \multicolumn{1}{c|}{96}            & 177          \\ \hline
\multicolumn{1}{|c|}{\textbf{Total}} & \multicolumn{1}{c|}{\textbf{1285}}          & \textbf{669}       \\ \hline
\end{tabular}
\label{table:observations}
\end{table}

\begin{figure*}[!ht]
    \centering
    \includegraphics[width=14cm]{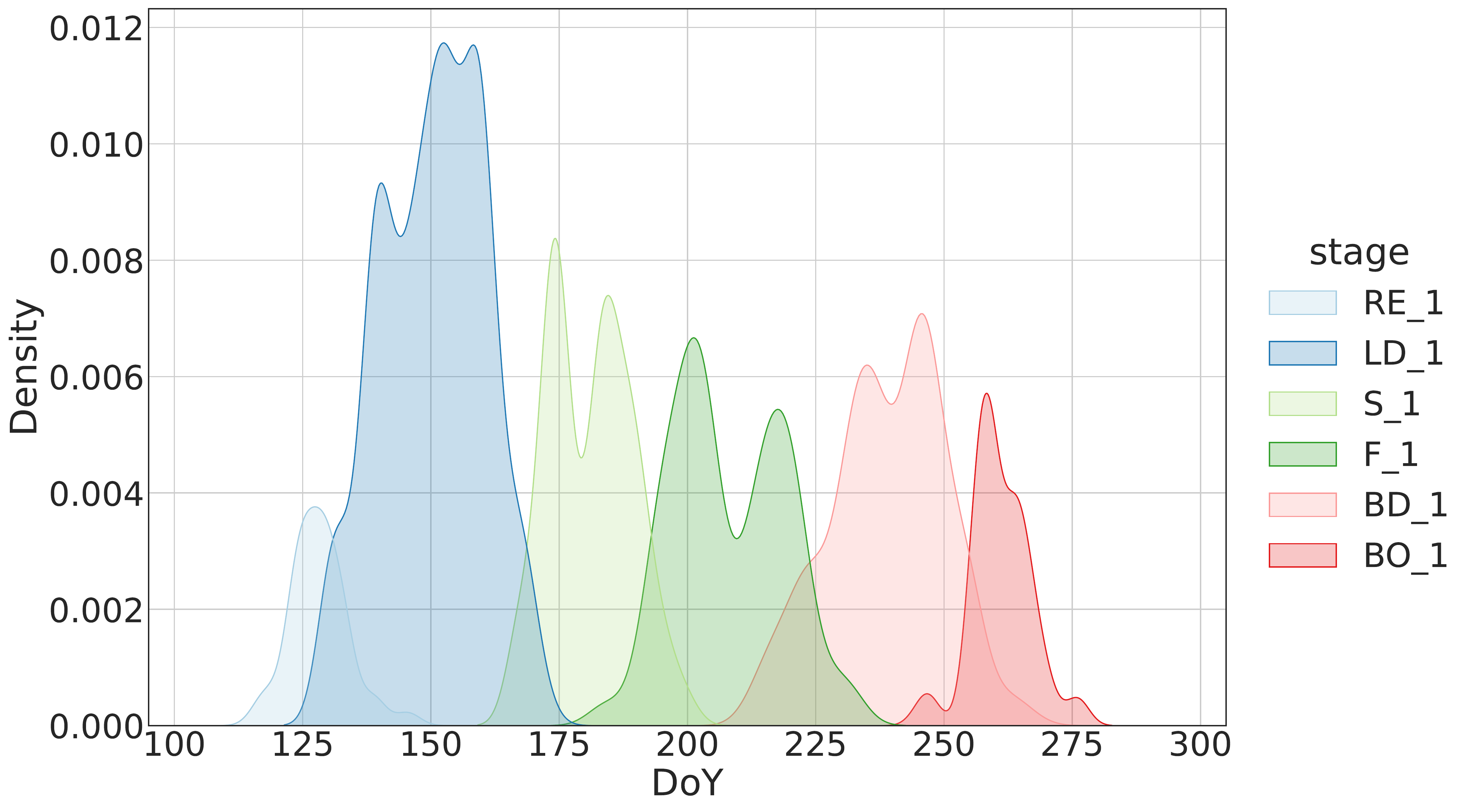}
    \caption{Kernel density estimation of DoY for which the inspector observed the different phenological stages of cotton as primary.\label{fig:thirdfigure}}
\end{figure*}

\begin{figure*}[!ht]
    \centering
    \includegraphics[width=14cm]{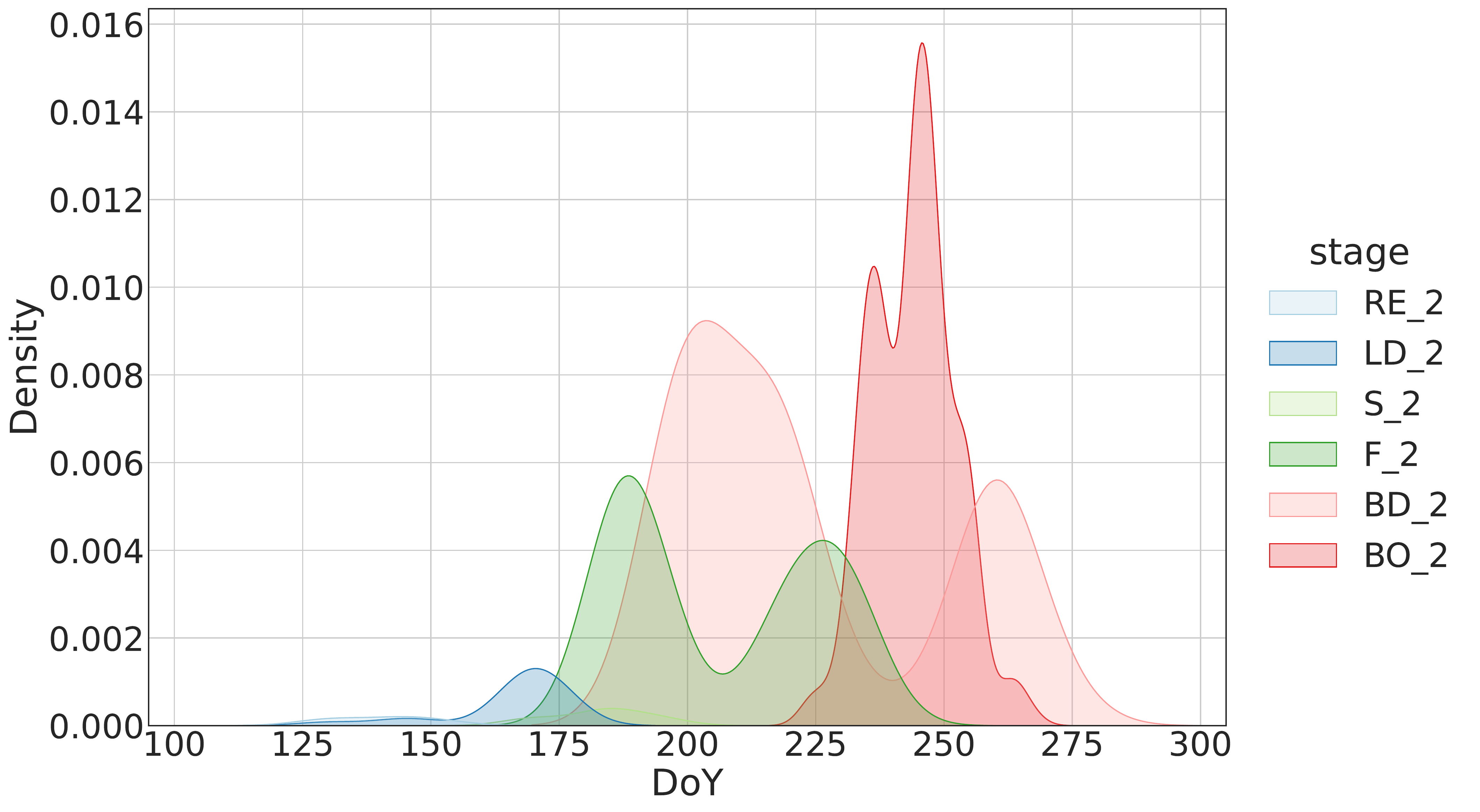}
    \caption{Kernel density estimation of DoY for which the inspector observed the different phenological stages of cotton as secondary.\label{fig:fourthfigure}}
\end{figure*}

\noindent Fig. \ref{fig:thirdfigure} and Fig. \ref{fig:fourthfigure} show the Kernel Density Estimation (KDE) of the Days of Year (DoY) for which the expert observed the different principal phenological stages as primary and secondary, respectively. It becomes clear that there are many chronological overlaps among the stages, for both the primary and secondary annotations. Inspecting Fig. \ref{fig:thirdfigure}, we see that the overlaps get progressively larger as we move towards the end of the growing cycle. This is expected as differences in growth accumulate with time and thus get more pronounced. The two figures also highlight how different the rate of growth can be even for fields that cultivate the same crop type, have similar sowing date and are in close proximity. Finally, it should be noted that for the secondary stage observations, the KDEs appear to have two modes (Fig. \ref{fig:fourthfigure}). The first and second mode refer to observations that were made prior and after the onset of the "primary" phase of a phenological category. This is the reason why there are extensive overlaps among the KDEs for the secondary stage observations.\\

% \begin{figure}[!ht]
%     \centering
%     \includegraphics[width=8.8cm]{Figs/Chapter4/sitok2.pdf}
%     \DeclareGraphicsExtensions{.png,.pdf}
%     \caption{Kernel density estimation of DoY for which the inspector observed the different phenological stages as primary.}
%     \label{fig:thirdfigure}
% \end{figure}

% \begin{figure}[!ht]
%     \centering
%     \includegraphics[width=8.8cm]{Figs/Chapter4/sitok3.pdf}
%     \DeclareGraphicsExtensions{.png,.pdf}
%     \caption{Kernel density estimation of DoY for which the inspector observed the different phenological stages as secondary.}
%     \label{fig:fourthfigure}
% \end{figure}

\noindent The ground observation dataset is public (\url{https://github.com/Agri-Hub/cotton-phenology-dataset}), encouraging the community to use it for training, testing and evaluating the performance of cotton phenology estimation or yield estimation models. The dataset  includes a) the geographic location and geometry of fields (EPSG:4326 - WGS 84),  b) the days of inspection, c) the primary phenological stage and the percentage it describes, d) the secondary phenological stage and the percentage it describes, e) the sowing and harvest dates, f) a panoramic photo of the field, g) close-up photos of representative plants for the "majority" and "minority" phenological stages. \\

\noindent The quality of the dataset has been evaluated by another expert, Expert 2, who reviewed a randomly selected subset of 145 ground observations (11.28\%) using the available panoramic and close-up photos. Expert 2 was not aware of the ground observations and was asked to decide on the primary stage and secondary stage, if there was one. Then a third expert reviewed the disagreements between the decisions of Expert 1 and Expert 2 using once more the photos captured during the visits. Table \ref{table:intra-rater_metrics} shows a number of metrics for the interrater agreement. \\

\noindent The analysis from Expert 3 yielded the percentages of direct agreement and disagreement, N/A observations and undecided observations. The agreement score referes to the cases for which Expert 2 provided the same label with Expert 1. N/A observations are the observations that were considered unfair to include in the evaluation. In particular, we do not take into account 3 cases:
\begin{enumerate}
    \item When Expert 1, i.e., the one that visited the field, provided only a primary stage observation and Expert 2 provides the same observation as a secondary stage. Such mismatch should be penalized only for the primary stage and it is considered N/A for the secondary stage.  
    \item When Expert 1 provided a primary stage observation with 100{\%} prevalence and secondary stage observation
    with prevalence less than or equal to 60{\%}, whereas Expert 2 gave the same primary stage observation but no secondary stage observation. Such stages should not be penalized for the secondary stage observation since the photos may not show it clearly. 
    \item When the primary observation of Expert 1 agrees with the secondary observation of Expert 2 and vice versa, and the prevalence percentage provided by Expert 1 is above 50{\%}. For example, Expert 1 observes F as primary with 100\% prevalence and stage BD as secondary with 60\% prevalence, whereas Expert 2 observes BD as primary and F as secondary. In fact, such cases can be very similar because both describe the transitional phase from one main stage to the other. Based on that, and since Expert 2 judges only from 2-3 photos, we do not think that it is fair to penalize these cases as wrong annotations. 
    %  \item When the primary stage observation of Expert 1 agrees with the secondary Expert 2 and the primary stage observation of Expert 2 agrees with the secondary Expert 1 . 
\end{enumerate}

\noindent The undecided category refers to instances that Expert 2 claimed and Expert 3 confirmed that the photos were not good enough to make a fair assessment. Finally, the rest of the cases are considered direct disagreements. \\

\begin{table}[!ht]
\centering
\caption{Interrater agreement metrics for the primary and secondary growth stage annotations (Expert 1 v. Expert 2).}
\begin{tabular}{l|c|c}
\cline{2-3}
 & \multicolumn{1}{l|}{\textbf{Primary stage}} & \multicolumn{1}{l|}{\textbf{Secondary stage}} \\ \hline
\multicolumn{1}{|l|}{\textbf{Direct agreement}}               & 0.72 & \multicolumn{1}{c|}{0.60} \\ \hline
\multicolumn{1}{|l|}{\textbf{Indirect agreement}}             & 0.17 & \multicolumn{1}{c|}{0.11} \\ \hline
\multicolumn{1}{|l|}{\textbf{Disagreement}}                   & 0.10 & \multicolumn{1}{c|}{0.09} \\ \hline
\multicolumn{1}{|l|}{\textbf{N/A}}                            & -    & \multicolumn{1}{c|}{0.08} \\ \hline
\multicolumn{1}{|l|}{\textbf{Undecided}}                      & 0.01 & \multicolumn{1}{c|}{0.11} \\ \hline
\multicolumn{1}{|l|}{\textbf{Krippendorff”s alpha (ordinal)}} & 0.95 &                           \\ \cline{1-2} 
\end{tabular}
\label{table:intra-rater_metrics}
\end{table}

\noindent Crop growth labeling is not straightforward because one attempts to derive ordinal categories to what is actually a continuous scale problem. Thus, the results will be heavily dependent on the choice of the category limits, i.e. the instance a principal phenological stage transitions to the next. These limits, although pre-defined (e.g.onset of LD is the appearance of three fully formed leaves) and explained in detail to the various experts, are subject to different interpretations. The annotations collected through the field visits are ordinal and thus we need to select an appropriate measure to reveal information on the reliability. Krippendorff's alpha is a versatile interrater agreement metric that is applicable to any level of measurement, e.g. nominal, ordinal or interval.\cite{krippendorff2011computing} The Krippendorff's alpha between Expert 2 and Experts 1 for ordinal level of measurement was 0.95. This indicates a strong agreement on the primary stage observations, which combined with the analysis performed by Expert 3, constitutes the annotation method reliable. From this point on, we only use Expert 1's ground observations. The aforementioned analysis was merely performed to assure the quality of the ground observations. \\

\subsubsection{Predictor variables}

\noindent Table \ref{tab:variables} lists the predictor variable candidates with which we experimented in this study: i) the Sentinel-2 derived products and ii) the atmospheric and soil numerical simulations. In this section we focus on the acquisition and pre-processing the various prediction variables, whereas in the next sections we elaborate on how these variables are incorporated in the feature space and feed the phenology estimation models.\\

% \begin{table*}

\begin{table}[!ht]
\centering
\caption{Summary of the predictor variables. B is the spectral reflectance value of the band number of the Sentinel-2 band. The variables with resolution of 2km are our NWP, whereas those of 10 or 20m resolution are Sentinel-2 derived products.}
\label{tab:variables}
% \scalebox{0.7}{
% \begin{tabular}{p{0.45\linewidth}|p{0.35\linewidth}|p{0.15\linewidth}}
\resizebox{\textwidth}{!}{%
     \begin{tabular}{c c c }
\hline
\textbf{Variable}& \textbf{Formula} & \textbf{Resolution} \\ \hline
Day of Year (DoY)                          & sine, cosine                                        & -                        \\
Temperature at surface                          & min, max                                         & 2 km                        \\
Growing Degree Days (GDD) 2m                                  & (T\textsubscript{max}-T\textsubscript{min})/2 - T\textsubscript{base}                                   & 2 km                         \\
Accumulated Precipitation                       & max                                         & 2 km                     \\
Downwards Shortwave Radiation                   & max                                         & 2 km                     \\
Soil temperature 0-10 cm depth                   & min, max                                         & 2 km                     \\
Soil moisture 0-10 cm depth                      & min, max                                         & 2 km                       \\
%  RGB and NIR  & B02, B03, B04, B08                       & 10 m                  \\
% Red-edge, NIR and SWIR   & B05, B06, B07, B08A, B11, B12                      & 20 m                 \\
Normalized Difference Vegetation Index (NDVI) \citep{pettorelli2013normalized}   & (B08-B04)/(B08+B04)                       & 10 m                  \\
Normalized Difference Water Index (NDWI)      \citep{mcfeeters1996use}   & (B03-B08)/(B08+B03)                       & 10 m                  \\
Normalized Difference Moisture Index (NDMI) \citep{gao1996ndwi}       & (B08-B11)/(B08+B11)                    & 20 m                 \\
Plant Senescence Reflectance Index (PSRI) \citep{merzlyak1999non}         & (B04-B02)/ B06                     & 20 m                 \\
Soil-Adjusted Vegetation Index (SAVI)  \citep{huete1988soil}                        & ((B08 - B04)/(B08 + B04 + 0.428))*(1.0 + 0.428) & 10 m \\
Enhanced Vegetation Index (EVI)  \citep{huete2002overview}                & 2.5*(B08-B04)/((B08+6*B04-7.5*B02) + 1.0) & 10 m                     \\

% Renormalized Difference Vegetation Index (RDVI) & (B08-B04)/(B08+B04)**0.5                  & 10 m                      \\
% Ratio Vegetation Index (RVI)                    & B08/B04                                 & 10 m                        \\
% Greenness Index (GI)    & B03/B04                                   & 10 m                        \\
Visible Atmospherically Resistant Indices Green (VARIgreen) \citep{gitelson2001non}    & (B03-B04)/(B03+B04-B02)                         & 10 m \\
Green Atmospherically Resistant Index (GARI) \citep{gitelson1996use}                  & (B08-(B03-(B02-B04)))/(B08-(B03+(B02-B04)))     & 10 m \\

Structure Insensitive Pigment Index (SIPI) \citep{penuelas1995semi}       & (B08-B02)/(B08-B04)                       & 10 m                  \\
Wide Dynamic Range Vegetation Index (WDRVI) \citep{gitelson2004wide}      & (0.2*B08-B04)/(0.2*B08+B04)               & 10 m                        \\ 
Global Vegetation Moisture Index (GVMI) \citep{ceccato2002designing}                        & ((B08+0.1)-(B12 + 0.02))/((B08+0.1)+(B12+0.02)) & 20 m \\ \hline
% Green Chlorophyll Vegetation Index (GCVI)       & (B08/B03)-1                               & 10 m                       \\ 
\end{tabular}}
\end{table}

\noindent\textbf{Optical images}
\noindent The optical spectrum variables used in this study were derived from Sentinel-2 images. As mentioned earlier, optical SITS have been popular in related research studies, with special attention given to Sentinel-2, but also MODIS data. The method of this work exploits only Sentinel-2 images in order to provide crop-specific phenology predictions at the field level. There are cases where the agricultural landscape is dominated by a single crop cultivation, e.g. U.S. Corn belt, and MODIS can be a tremendous help in crop-specific phenology predictions. But in Greece the landscape is fragmented and the medium spatial resolution of MODIS would yield mixed optical signatures of multiple crop types.\\

\noindent The optical component of this study's variable space comprises the RGB, NIR and SWIR spectral bands of Sentinel-2, but also several VIs. VIs are combinations of the spectral bands that can highlight particular vegetation properties \cite{segarra2020remote}. Table \ref{tab:variables} lists the various VIs that have been used in this work. The VIs investigated are some of the most common in the relevant literature. As the crop grows, its spectral signature changes with time. It starts with bare soil and then there are stems and leaves, and then flowers and bolls. These different phases have different biophysical and biochemical properties and thus different light reflectance profiles. Therefore we investigate multiple VIs so as to capture the maximum possible information at every stage of the growth. With regards to pre-processing, the Sentinel-2 images have been atmospherically corrected using the Sen2Cor software \cite{louis2016sentinel}. Additionally, clouds have been removed using the Sen2Cor scene classification product. Then, the null-valued pixels have been filled using linear interpolation on the SITS. \\

\noindent\textbf{Atmospheric and soil parameters}

\noindent A dense, long-term and efficient monitoring of the atmospheric state is rarely met in real-life conditions except for experimental campaigns. Automatic weather stations can contribute, but insufficient spatial coverage or bad distribution of them are typical problems, let alone temporal gaps and discontinuities due to sensor failures or human aspects (poor maintenance). Given the absence of a dense in-situ network or weather radar scans over our area of study that could provide a necessary insight, we relied upon high resolution Numerical Weather Predictions (NWP) (2-km) from our convection-permitting operational configuration of WRF-ARW model \cite{skamarock2019description}. The model is initialized daily with the latest available analysis and after the exclusion of the first few hours (spin-up time) in order to reach a statistical equilibrium, the following 24-hour estimates are utilized. \\

\noindent While this grid spacing may appear quite coarse when compared to the resolution of the EO-derived products, we should consider that this is an outcome of NWP simulations. We are able to provide estimates of atmospheric parameters every 2 km over regions that are heavily  under-monitored (in-situ weather station can be available every 100 km over croplands). This scale is considered high resolution in NWP terms and the resolving of cloud microphysical processes, such as convection, starts to happen explicitly under this spatial threshold which is particularly important to resolve fine atmospheric processes on a local scale without having to rely on parameterization schemes. A 2 km forecast fulfils our needs given that the physiographic characteristics of the crop regions we focus upon are not areas of high topographical complexity, so great gradients are not expected. \\

\noindent The specific parameters that were used in our pipeline were an outcome of consultation with agronomist experts and systematic literature review upon their correlation with the evolution of cotton \cite{piao2019plant}. They include Air Temperature, Surface (skin) Temperature, 0-10cm Soil Temperature and Moisture, Precipitation and Incoming Shortwave Radiation. Growing Degree Days (GDD) is additionally computed, as it is one of the most essential indicators of phenology. Inspecting the GDD equation in Table \ref{tab:variables}, T\textsubscript{max} and T\textsubscript{min} are the maximum and minimum daily air temperatures at 2m (from surface) and T\textsubscript{base} is the crop's base temperature (15.6 °C). The latter is defined as the temperature below which cotton does not develop. The GDD variable is also known as thermal time and is an indicator of the effective growing days of the crop \cite{sharma2021use}. \\

\subsection{Methods}
\label{methods}
\subsubsection{Fuzzy clustering}
\begin{figure*}[!ht]
    \centering
    \includegraphics[width=14cm]{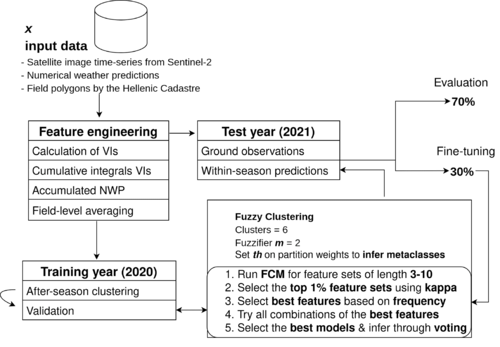}
    \caption{The proposed methodology for cotton phenology estimation depicted as a wokflow.}
    %SMR 1 requirement mean execution time}
    \label{fig:architecture}
\end{figure*}

\noindent We propose a fuzzy clustering method for the within-season estimation of cotton phenology. The workflow of the proposed approach is depicted in Fig. \ref{fig:architecture}. We use clustering to circumvent the ever-present problem of sparse, scarce and difficult-to- acquire ground observations, which constitute the supervised alternatives of limited applicability in operational scenarios. Nevertheless, we visited tens of fields and collected hundreds of ground observations in order to test the performance of our models. The aim of our newly introduced ground observation collection protocol was to extract more information than the principal growth stages, as it is usually the case in related works. This happens at the labelling level via taking advantage of the primary and secondary stage ground observations, as described in Section \ref{datasets}.\\

\noindent\(\mathcal{L} = \{\lambda_1, \lambda_2,..., \lambda_k\}\) is the finite ordered scale of the principal cotton growth class labels, where \(\lambda_1\) is RE and \(\lambda_6\) is BO. The ground observation protocol allowed \(k=6\) phenological stages to choose from and a maximum of \(n =2\) labels to assign to each field, i.e., the primary and secondary stage categorization. This restricted the set of allowable labels \(L_r\) to all possible permutations of \(n =2\) elements of \(\mathcal{L}\) and all unit sets. Specifically, there are \(k^2 = 36\) allowable labels. This multi-label problem can be reduced to a single-label one by considering each subset as a distinct metaclass \citep{Cheng2010GradedMC}. In reality, however, there are only 16 possible metaclasses, as the primary and secondary stage for a field can differ only a single position in the ordinal scale. Equation (\ref{eq1}) shows this set of 16 metaclasses in ordered scale. 

\begin{multline}
% \begin{split}
\begin{aligned}
\label{eq1}
L_r =  \{ & \lambda_1, (\lambda_1, \lambda_2), (\lambda_2, \lambda_1), \lambda_2,(\lambda_2, \lambda_3), (\lambda_3, \lambda_2), \\ &   
          \lambda_3, (\lambda_3,\lambda_4),  (\lambda_4,\lambda_3), \lambda_4, (\lambda_4,\lambda_5), (\lambda_5,\lambda_4), \\ & \lambda_5, (\lambda_5,\lambda_6), (\lambda_6,\lambda_5), \lambda_6\}
% \end{split}
\end{aligned}
\end{multline}

\noindent In order to estimate the phenological metaclass for each field at any given instance, we used the Fuzzy C-Means (FCM) clustering algorithm. \(X \in R^{K \times E}\) denotes the element space that was used as input to the FCM, where K is equal to the number of fields (M) multiplied by the number of Sentinel-2 acquisitions (J), and E is equal to the number of features. Each row of the two-dimensional space in Eq \ref{eqe} represents field \(i\) at the time instance \(j\) (DoY of Sentinel-2 acquisition). For the Sentinel-2 variables, each element \(x_{(i,j),d}\) gets the mean value of variable \(d\) for all pixels of field \(i\) at the instance \(j\). In the same fashion, for the NWP variables, each element \(x_{(i,j),d}\) gets the value of the nearest grid cell to field \(i\) for the instance \(j\). To calculate the mean value of the pixels and grid cells that fall within each field we used the parcel boundaries retrieved from the Hellenic Cadastre (scale 1:5000).\\

\begin{equation}\label{eqe}
\centering
X_{(i,j),d} = 
\begin{pmatrix}
x_{(1,1),1} & x_{(1,1),2} & \cdots & x_{(1,1),E} \\
x_{(1,2),1} & x_{(1,2),2} & \cdots & x_{(1,2),E} \\
\vdots  & \vdots  & \ddots & \vdots  \\
x_{(1,J),1} & x_{(1,J),2} & \cdots & x_{(1,J),E} \\
% z_{(2,1),1} & z_{(2,1),2} & \cdots & z_{(2,1),E} \\
\vdots  & \vdots  & \ddots & \vdots  \\
x_{(M,J), 1} & x_{(M,J), 2} & \cdots & x_{(M,J), E} 
\end{pmatrix}
\end{equation}

\noindent Since phenology is dependent on the relative temporal progression of variables, we use the accumulated NWP parameters and the cumulative integrals of the VIs. The starting point for the accumulation is set around the earliest sowing DoY for the fields in the area of interest. In our case, this starting point was the 10th of April (DoY 100). Therefore, the feature space includes the Sentinel-2 VIs and their cumulative integrals, the accumulated NWP parameters, and the cosine and sine of the Sentinel-2 acquisition DoY (Table \ref{tab:variables}).\\

\noindent During the learning phase, the FCM algorithm attempts to partition K elements \(X = \{\boldsymbol{x}_1, ..., \boldsymbol{x}_K\}\) that capture the entirety of the season into c = 6 clusters that is assumed they represent the six principal growth stages of cotton (after-season clustering in Fig. \ref{fig:architecture}) \cite{bezdek1984fcm}. This is considered to be a valid assumption given that each element is described by the EO and NWP variables, their time-accumulated variants, and the associated DoY. The assumption is also supported by the results in the next section. The algorithm returns a list of \(C = \{\boldsymbol{c}_1, \boldsymbol{c}_2, ..., \boldsymbol{c}_6\}\) cluster centers and a partition matrix \(W = (w_{k,l}) \in R^{K \times c}\), where \(w_{k,l}\) is the degree to which the element \(\boldsymbol{x}_{k}\) belongs to cluster \(\boldsymbol{c}_l\).  We applied FCM on the 2020 variable space (training year) in Orchomenos and then used the clusters C to produce within-season predictions, in dynamic fashion, during the 2021 season (test year). 
The training cotton fields of 2020 were 194 in total, and were extracted from a pre-trained crop classification model, based on \cite{sitokonstantinou2021scalable}.\\

% \noindent In order to detect cotton fields and apply the FCM in the training year (2020), we used the binary crop classifier (RF) proposed in \citep{sitokonstantinou2021scalable}, which was slightly modified to fit our case. The input data used for the cotton classification model comprised the field-based time-series of the RGB and NIR bands of Sentinel-2, the NDVI and NDWI VIs. The RF model was trained on the known cotton fields that were visited in 2021 and was then applied to the 2020 feature space. The classification result was then filtered based on the posterior probability to keep only the most confident classification decisions. The idea was to extract a considerable amount (194 in this case) of fields for which we were certain that cotton was cultivated in 2020. The field boundaries used for the object-based classification were retrieved from the Hellenic Cadastre. \\

\noindent After the clustering, the phenological stages are assigned to the different clusters via exploiting the time order. For this, the most common order of clusters is recorded and then matched to the ordered scale of labels in \(\mathcal{L}\). It is common to address a multi-label problem in an indirect way using a scoring function \(f : X \times \mathcal{L} \rightarrow R\) that assigns a real number to each element-label pair \cite{Cheng2010GradedMC}. The assumption here is that this scoring function corresponds to the probability of each label being relevant to an element. % In this work we address the multi-label problem of within-season phenology estimation in a weakly supervised and purely unsupervised way.
In our case, the scoring function \(f\) is the FCM and the scores are the membership grades \(w_{k,j}\) of each element \(\boldsymbol{x}\). In other words, the FCM attempts to find the labels in \(\mathcal{L}\) and then the partition scores or weights are used for multi-label prediction via thresholing. Even more, sorting the labels according to their score provides label ranking, enabling the identification of the primary and secondary stages, as given through the field inspections (Eq \ref{eq2}). 

\begin{align}
\label{eq2}
\lambda_i \leq_x \lambda_j \Leftrightarrow f(\boldsymbol{x}, \lambda_i) \leq f(\boldsymbol{x}, \lambda_j), i,j = 1..6
\end{align}

where $\lambda$ refers to the 6 principal phenological stages from RE to BO,x is an  element from the element space in \ref{eqe} and f is the scoring function of the FCM algorithm, i.e., the partition score. 

\subsubsection{Evaluation metrics}

\noindent We considered the ML task of phenology estimation as a multi-label classification problem, given the potential duality of phenological stages at a given instance. The two labels are ranked as primary and secondary phenological stages according to their relevance, or prevalence, in the field. Therefore, the metrics for assessing our model should capture these properties.\\

\noindent First, we categorize the predictions to error classes according to the difference or displacement between the prediction and the ground observation in the ordinal scale of metaclasses. These error classes are labelled as diff-\(o\), with \(o \in \{0, 1, 2, 3\}\). For instance, if our model predicted \(\lambda_2\) and the ground observation was \((\lambda_3, \lambda_2)\), then according to Eq \ref{eq1} the prediction is categorized as diff-\(2\). Similarly to the top-N accuracy, we devised the maxdiff-\(o\) accuracy, with \(o \in \{0, 1, 2, 3\}\), measuring the percentage of predictions with a displacement no greater than \(o\). For instance, maxdiff-\(2\) is the percentage of predictions that have at most an error of two displacement units. We also use the well known kappa coefficient, together with its linear and quadratic weighted variants. The weighted kappa metrics allow for disagreements to be weighted differently and are commonly used when labels are ordered.\\

% \begin{align}\label{eqX}
% % \begin{aligned}
% \text{maxdiff-\(o\)} = \frac{1}{K}\sum_{i=0}^{K} \begin{cases} 
% 1 & y(i) - \hat y(i) <= o \\
% 0 & otherwise
% \end{cases} 
% % \end{aligned}
% \end{align}

\noindent We additionally incorporate the Normalized Discounted Cumulative Gain (NDCG). It is a popular metric in the world of information retrieval and specifically in tasks such as top-N ranking and item recommendations. In our case, we use the various partition weights or membership probabilities (\(w_{k,j}\)) as relevance values and rank the phenological clusters accordingly. We use NDCG\(@2\), as we take into account only the top 2 ranked stages/clusters that we assume represent the primary and secondary annotations. The highly relevant phenology stage should be ranked higher than the less relevant stage, which is in turn expected to be ranked higher than non-relevant stages. NDCG\(@2\) captures and evaluates exactly this capability of the model.\\

\noindent NDCG is based on the cumulative gain that simply sums the relevance scores for top\(@p\) (\(p=2\)). This is mathematically expressed by (\ref{eq4}). 
\begin{align}\label{eq4}
% \begin{aligned}
{CG_{p}} = \sum_{i=1}^{p} rel_{i}
% \end{aligned}
\end{align}
\noindent In this case, we set \(rel = 2\) for the primary stage and \(rel = 1\) for the secondary stage. The cumulative gain however does not take into account the position of the phenological stage in the rank. This is done by the discounted cumulative gain, as in (\ref{eq5}), which makes use of a log-based penalty function and reduces the relevance score that is normalized by a penalty equivalent to each position. 
\begin{align}\label{eq5}
% \begin{aligned}
{DCG_{p}} = \sum_{i=1}^{p} \frac{rel_{i}}{\log_{2}(i+1)} = rel_1 + \sum_{i=2}^{p} \frac{rel_{i}}{\log_{2}(i+1)}
% \end{aligned}
\end{align}
\noindent Finally, the discounted cumulative gain is simply normalized by the ideal order of the relevant items and we end up with NDCG \cite{jarvelin2002cumulated}.\\

\subsection{Results}
\label{sec:experiments}

\noindent The FCM clustering was performed on the EO and NWP variables for the Orchomenos region in 2020 (training year) and was then applied on the equivalent feature space of 2021 (test year) for the within-season prediction of phenological metaclasses. Our FCM-based approach has three parameters, i) the number of clusters, ii) the fuzzifier \(m \in R\), with \(m \geq 1\) and iii) the partition score threshold, above which a cluster is considered as a valid phenological stage label. The fuzzifier m was set to 2, which is commonly preferred when using the FCM algorithm \cite{de2007advances, SINGH2016114, VERMA2016543, lei2018significantly}. According to \cite{pal1995cluster} the best choice for m is in the interval [1.5, 2.5], with $m=2$ being the most common choice. Finally, the partition threshold (th\textsubscript{w}) depends on the distribution of the partition scores during the learning phase of the FCM algorithm.\\

% First, the fuzzifier value was estimated using (\ref{eq3}), which exploits the dataset properties and specifically the number of unique elements \(K\) and the number of features \(E\) \citep{schwammle2010simple}. 
% \begin{multline}\label{eq3}
% f(E,K) = 1 + \left(\frac{1418}{K} + 22.05\right)E^{-2} \\ + \left(\frac{12.33}{K} + 0.243\right)E^{-0.0406\ln(K)-0.1134}
% \end{multline}
% As the elements of the dataset are constant (i.e., equal to the number of Sentinel-2 acquisitions), the fuzzifier value ranged depending on the number of features, as we tested thousands of combinations among the different variables in Table \ref{tab:variables}. Indicatively, we used \(m=1.99\) for \(E=5\),  \(m=1.29\) for \(E=10\), \(m=1.16\) for \(E=15\) and values close to the minimum \(m=1\) for \(E>20\). Larger values of \(m\) result in greater fuzziness and thus smaller partition weights for the top clusters. Therefore, we set different partition weight thresholds, \(th_w\), for different intervals of \(m\) values. 

\begin{figure}[!ht]
    \centering
    \includegraphics[width=10cm]{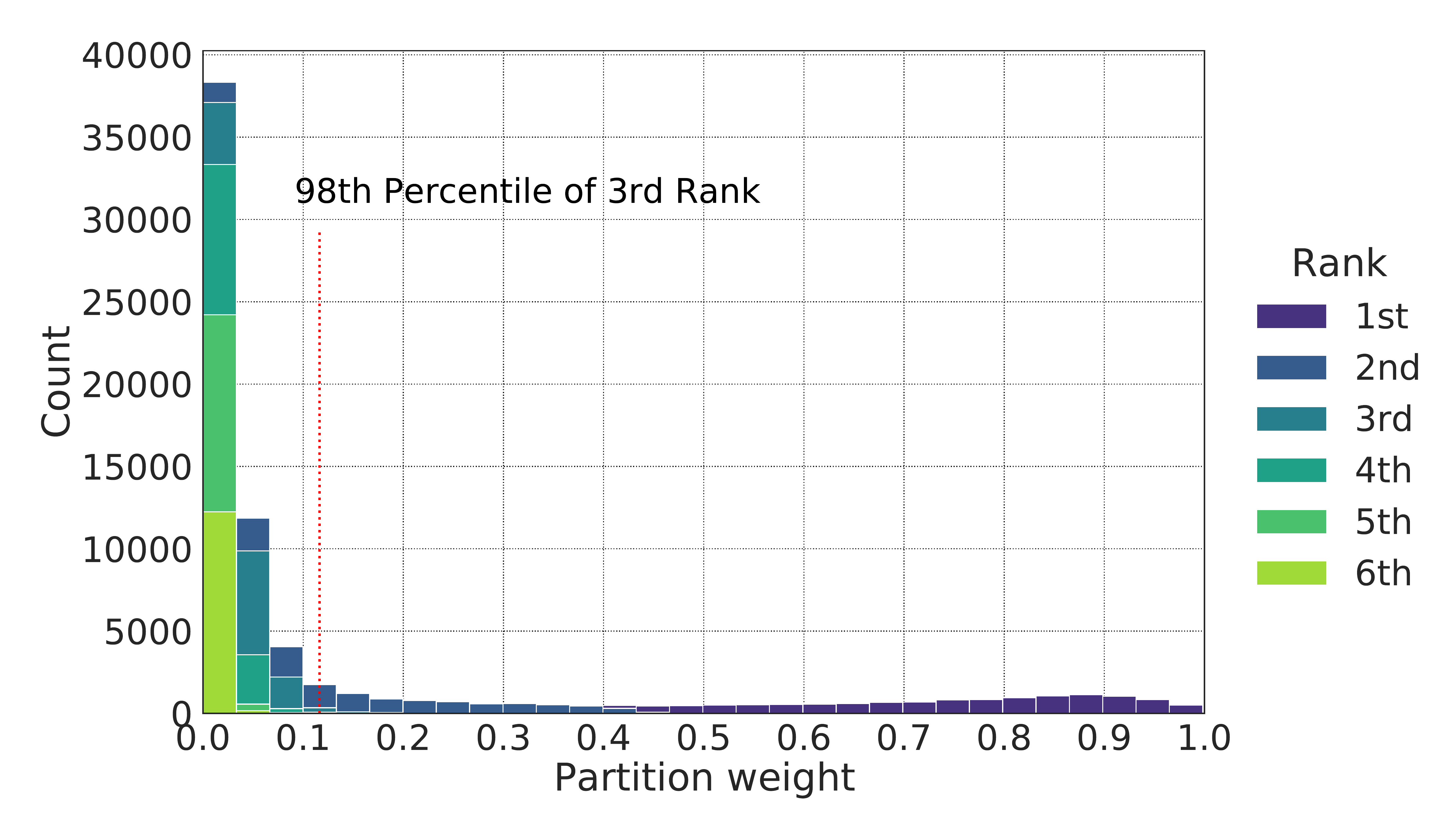}
    \caption{Distribution of FCM partition weights for \(m=2\). The hue indicates the rank in which the weights were given for each prediction. The dotted line shows the threshold above which 2nd ranked clusters are considered as valid secondary phenological stages.}
    %SMR 1 requirement mean execution time}
    \label{fig:fcmscores}
\end{figure}

\noindent For each element \(\boldsymbol{x}\), FCM gives a partition weight w\textsubscript{k,j} that refers to the degree the element belongs to each of the clusters. The weights are then sorted for each element. The two-labelled nature of our target (primary and secondary stages) implies that the values of partition weights ranked 3rd or lower, should not be considered as valid phenological stage labels. Thus, we set the threshold th\textsubscript{w} equal to the value of the 98th percentile of the partition weights ranked in the third place (Fig. \ref{fig:fcmscores}). We used the 98th percentile to eliminate the influence of potential outliers. Indicatively, 
Fig. \ref{fig:fcmscores} illustrates the distribution of the partition weights (for $m=2$) for a representative model. The different colors indicate the rank in which the weights were given for each prediction, i.e., from 1st to 6th. In this case, the threshold is set to 0.11 based on the aforementioned rule. A different threshold was computed for each different model that we tested.\\

\noindent Having configured the three parameters, we trained FCM models (after-season prediction) on the 2020 cotton fields (training year), with multiple combinations of features (~80K). The FCM algorithm works well in low dimensions. Specifically, \cite{winkler2011fuzzy} suggested that for dimensions larger than ten, the FCM starts to present ill behaviour; and for this reason we ran our experiments on hyperspaces of up to ten features. We had 32 features to choose from, i.e., the variables described in Table \ref{tab:variables} and their cumulative variants, and generated feature sets of length from 3 to 10 features. However, an exhaustive analysis would have required more than 200 million clusterings. To avoid running so many experiments, which is translated in months of experimentation, we fitted the FCM for feature sets of length \(E \in \{3, ..., 10\}\) using ten thousand random feature combinations for each length. It should be noted that all feature sets include the features cos(DoY) and sin(DoY) that set the time frame.\\

\noindent In order to evaluate the performance of the algorithm on the different feature spaces, we applied each individual FCM model (i.e., the cluster centers, C, and thresholds, th\textsubscript{w}, from the training year) on the 2021 cotton fields, for which we have ground truth labels. We split the data into test and validation (70-30), and then used the kappa coefficient values of the validation data to find the best 1\% (mean kappa = 0.45) of feature combinations. Based on these top feature combinations, we selected the 15 most appeared features. These were i) the VIs SIPI, GVMI, EVI, NDMI and SAVI, ii) the cumulative integrals of WRDVI, PSRI, NDWI and SIPI, iii) the accumulated maximum soil temperature, maximum surface temperature, maximum solar radiation and Accumulated GDD (AGDD) and iv) the always present time features of sin(DoY) and cos(DoY).  \\

% The mean kappa coefficient of the best models for the validation set was 0.45, whereas the mean kappa value for the test set was 0.43. 

\noindent Having concluded to the aforementioned set of predictors, we ran the FCM algorithm again for the training year (2020), for every possible combination of those features in spaces of length 6 to 15. This time, apart from the kappa coefficient, we also considered the maxdiff-$1$ score. Specifically, based on the performance of the best 1{\%} of feature combinations, as mentioned above, we kept solutions with kappa coefficient larger than $0.46$ and maxdiff-$1$ larger than $0.86$, which resulted to 604 cases out of 7,814. Moreover, by setting these values as such, we ensure that we acquire better solutions compared to a baseline  model. The baseline model refers to an FCM with only DoYs as input. Since phenology is closely related to the DoY, the baseline is used to capture this chance agreement and showcase by comparison the real competence of our model. Detailed comparisons with the baseline model follow later in this section. \\ 

\noindent The analysis revealed that for most cases feature sets of length $E > 10$ yielded sub-optimal results. Specifically, from the top 604 models, 84\% contained no more than 10 features. Besides the DoY features, the AGDD is by far the most common feature since it appeared in more than 86\% of the best solutions. Another important observation here is that the cumulative integrals of VIs and the accumulated NWP features are more important than than the single-date VIs. The importance of these features is great since they also capture the dimension of time, which is essential for an unsupervised phenology prediction model. Nevertheless, the results showed that the majority of well-performing feature sets contained at least one feature from each category, namely VIs, cumulative integrals of VIs and accumulated NWPs.  \\

% In order to interpret why larger feature spaces do not enhance the model's performance, we examined what happens to the centers of the clusters, paying particular attention to the top three features, i.e., cos(DoY), sin(DoY) and AGDD. We observed that in general the centers of these features remained relatively stable, in each of the best model. On the other hand, when extending the feature sets these centers start to diverge, which thus leads to sub-optimal solutions. These three features gravitate the centers towards the expected DoYs for each of the six principal phenological stages. It is the rest of the EO and NWP variables, which slightly reposition the centers based on the observed reality. This is the reason why longer feature sets, which tend to alter these stable centers of the time-reflecting features, do not perform well. 

\noindent From the top 15 features, the cumulative integral of SIPI, the accumulated maximum solar radiation, NDMI and SIPI did not appear as frequently in the best 604 models. For this reason, we discarded models that included these features. The final set of models, which was used for our predictions, comprised models with 8 and 9 features that included at least one feature from each category. This resulted to a total of 82 models. In order to ensure the generalization and robustness of our methodology, by not depending on a single feature set, we generate the final predictions through majority voting on those best models. The 82 feature sets are listed in Table S1.  \\

\noindent Table \ref{tab:unsupmetrics} shows the performance of our model and the baseline model on the test set. Indeed, it is shown that our model provides a significantly larger number of diff-$0$ predictions, namely perfect agreements between predicted and ground truth metaclasses. This is also evident via observing Cohen's kappa that is notably higher for our model. In terms of absolute values though, the model shows moderate performance in these two metrics. However, given the unsupervised nature of the FCM algorithm as well as the fact that it works very wellin the other metrics and avoids outlier errors, we claim that the overall performance is satisfactory and the proposed approach shows potential. It is also worth mentioning how our model significantly outperforms the baseline in terms of NDCG, denoting a better ranking capacity. \\

\begin{table}[!ht]
\centering
\caption{Metrics of performance for our phenology prediction model and the baseline model.}
\begin{tabular}{c|c|c|}
\cline{2-3}
\multicolumn{1}{|c|}{}                                     & \textbf{Ours} & \textbf{Baseline} \\ \hline
\multicolumn{1}{|c|}{\textbf{maxdiff-0}}                  & 0.53         & 0.38              \\ \hline
\multicolumn{1}{|c|}{\textbf{maxdiff-1}}                  & 0.88       & 0.86             \\ \hline
\multicolumn{1}{|c|}{\textbf{maxdiff-2}}                  & 1.00         & 0.97              \\ \hline
\multicolumn{1}{|c|}{\textbf{maxdiff-3}}                  & 1.00          & 1.00             \\ \hline
\multicolumn{1}{|c|}{\textbf{Cohen’s kappa}}              & 0.48        & 0.33              \\ \hline
\multicolumn{1}{|c|}{\textbf{Weighted kappa (Linear)}}    & 0.88          & 0.84           \\ \hline
\multicolumn{1}{|c|}{\textbf{Weighted kappa (Quadratic)}} & 0.98         & 0.97            \\ \hline
\multicolumn{1}{|c|}{\textbf{NDCG}}                       & 0.93          & 0.88              \\ \hline
\end{tabular}
\label{tab:unsupmetrics}
\end{table}

\noindent Table \ref{tab:displacement} shows the prediction errors in metaclass displacement units, for phenology metaclasses that had at least 10 ground observations. The displacement units are computed via multiplying the normalized confusion matrix with a weight matrix, for which the cells one off the diagonal of the confusion matrix are weighted 1, those two off are weighted 2, etc. Then the weighted displacement units are summed for each ground observation metaclass. Our model offers a smaller average displacement for six out of the eight metaclasses that account for the  majority of ground observations.\\ 

\begin{table}[!ht]
\centering
\caption{Prediction errors in metaclass displacement units. Our model is compared with the baseline model for phenology metaclasses that had more than 10 ground observations.}
\begin{tabular}{clc|cc|}
\cline{4-5}
\multicolumn{3}{l|}{}                                           & \multicolumn{2}{c|}{Displacement}                  \\ \hline
\multicolumn{2}{|c|}{\textbf{Metaclass}}                 & \textbf{Support} & \multicolumn{1}{c|}{\textbf{Ours}} & \textbf{Baseline} \\ \hline
\multicolumn{1}{|c|}{1}  & \multicolumn{1}{l|}{(RE, -)}  & 72   & \multicolumn{1}{c|}{\textbf{0.41}}          & 1.33         \\ \hline
\multicolumn{1}{|c|}{4}  & \multicolumn{1}{l|}{(LD, -)}  & 415  & \multicolumn{1}{c|}{\textbf{0.62}} & 0.89          \\ \hline
\multicolumn{1}{|c|}{6}  & \multicolumn{1}{l|}{(S, LD)}  & 17   & \multicolumn{1}{c|}{\textbf{0.41}} & 1.00          \\ \hline
\multicolumn{1}{|c|}{7}  & \multicolumn{1}{l|}{(S, -)}   & 122  & \multicolumn{1}{c|}{0.58}          & \textbf{0.17} \\ \hline
\multicolumn{1}{|c|}{8}  & \multicolumn{1}{l|}{(S, F)}   & 73   & \multicolumn{1}{c|}{\textbf{0.30}} & 0.56          \\ \hline
\multicolumn{1}{|c|}{11} & \multicolumn{1}{l|}{(F, BD)}  & 225  & \multicolumn{1}{c|}{1.04}          & \textbf{0.92} \\ \hline
\multicolumn{1}{|c|}{12} & \multicolumn{1}{l|}{(BD, F)}  & 75   & \multicolumn{1}{c|}{\textbf{0.37}} & 0.75          \\ \hline
\multicolumn{1}{|c|}{14} & \multicolumn{1}{l|}{(BD, BO)} & 177              & \multicolumn{1}{c|}{\textbf{0.54}} & 0.72              \\ \hline
\multicolumn{1}{|c|}{15} & \multicolumn{1}{l|}{(BO, BD)} & 90   & \multicolumn{1}{c|}{\textbf{0.03}} & 0.68          \\ \hline\hline
\multicolumn{2}{|c|}{\textbf{Average}}                & 1266 & \multicolumn{1}{c|}{\textbf{0.48}} & 0.78          \\ \hline

\end{tabular}
\label{tab:displacement}
\end{table}

\noindent It is observed that the metaclass (BO, BD) offers the smallest error in displacement units, whereas the (F, BD) metaclass gives by far the largest. This can be explained by the fact that that metaclass (BO, BD) is at the edge of the growing cycle and can only be confused with the metaclass that precedes it. On the other hand, metaclass (F, BD) is at the vegetation peak, where plants are well into the flowering phase and some bolls have started to develop. The consecutive metaclasses (F, -), (F, BD) and (BD, F) are situated near the plateau that is formed around the peak of the VI time-series curve (or valley, given the VI). Therefore, there are not significant differences in VI values among the three metaclasses, which explains the less than optimal performance for metaclass (F, BD). Overall, the average displacement shows significant difference between the two models. Our model achieves a respectable average error of less than half a metaclass.  \\

\begin{figure}[!ht]
    \centering
    \includegraphics[width=10cm]{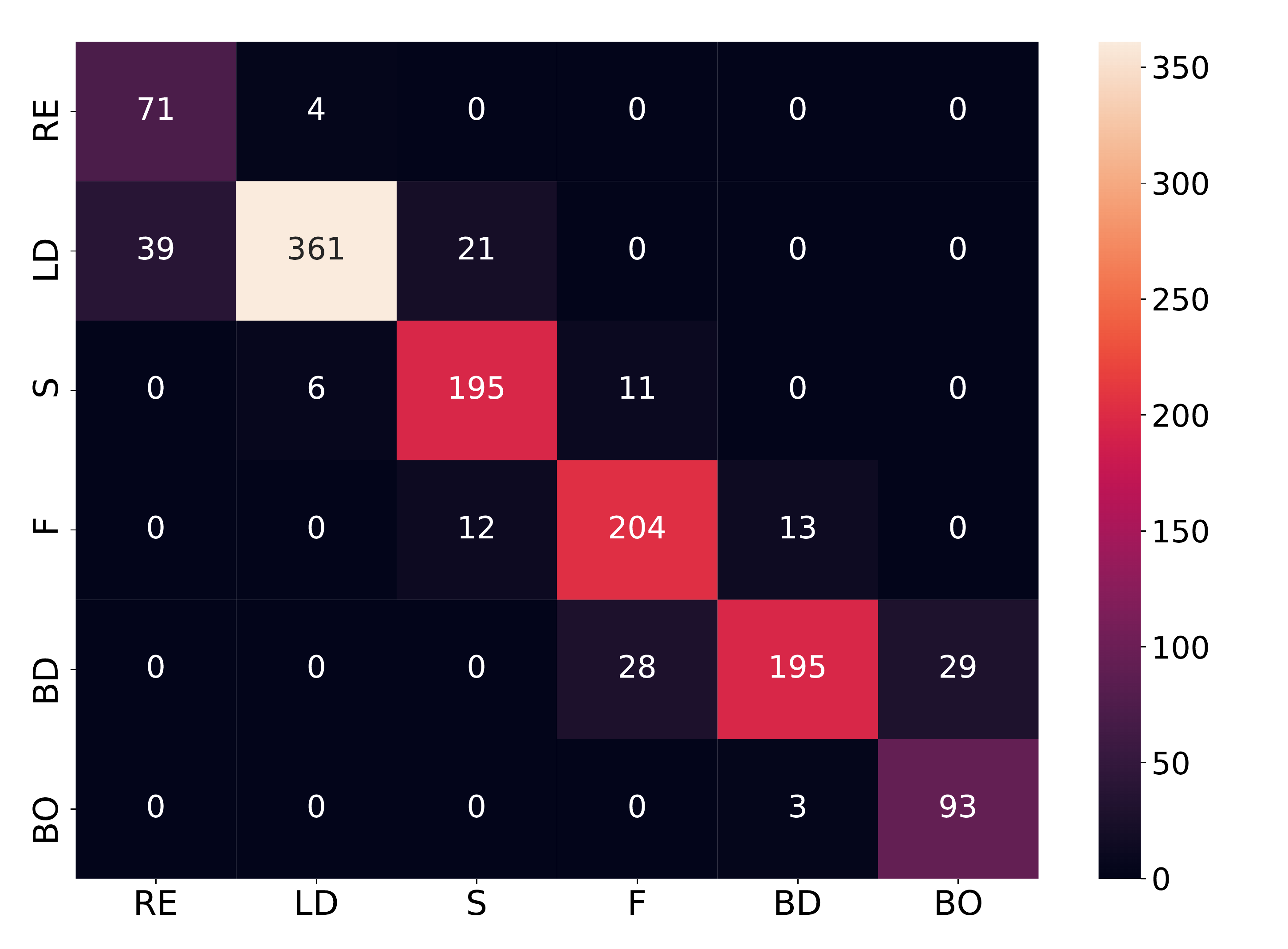}
    \caption{Confusion matrix of our phenology estimation model for the six principal phenological stages of cotton.}
    \label{fig:cmOurs_PC}
\end{figure}

\noindent Table \ref{fig:cmOurs_PC} shows the confusion matrix for the hard clustering predictions. It can be observed that for the principal phenological stages the model performs rather well, with an overall accuracy 87\%. Most misclassifications are observed for the BD stage. This is actually expected given the number of observations for which BD is observed in one of the transitional metaclasses (Table \ref{tab:displacement}). As a matter of fact, BD is never observed as unit set metaclass. 

\noindent As mentioned in Section \ref{datasets}, the expert would visit each field three or four times a month. Thus, it was common to observe a field in a particular crop growth stage for multiple consecutive visits (3 to 6). The chronological order of observation, i.e., the relative position of the ground observation in the range enclosed between the first and last time a stage was observed as primary for a particular field, is useful since it indicates if it is observed in its early, middle or late phases. \\

\begin{figure}[!ht]
    \centering
    \includegraphics[width=9.8cm]{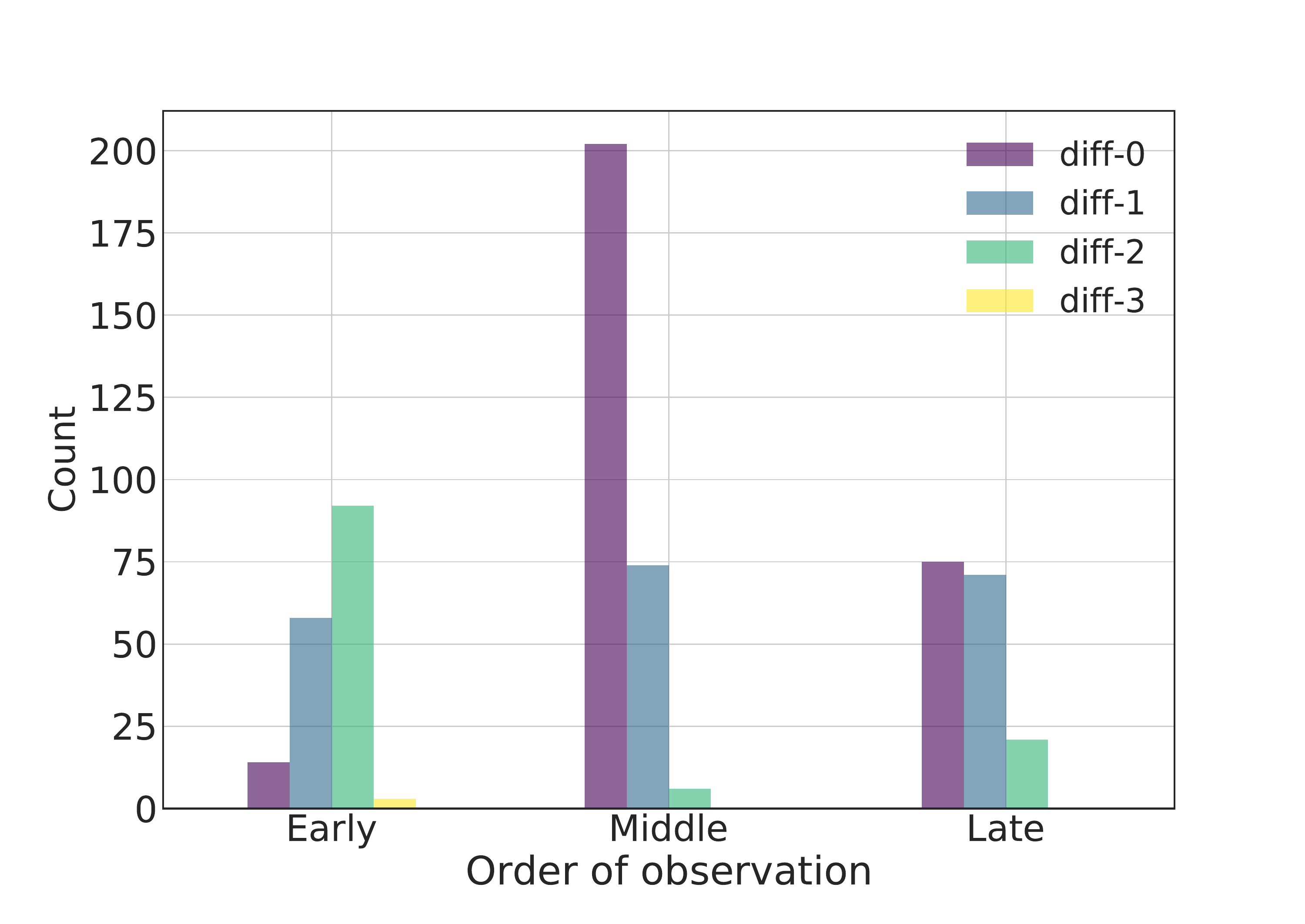}
    \caption{Distribution of the chronological order of observation, i.e., the relative position of the ground observation in the range enclosed between the first and last time a stage was observed as primary for a field. The hues show the predictions according to their diff-\(o\) categorization of prediction error.}
    \label{fig:stage4}
\end{figure}

\noindent Fig. \ref{fig:stage4} shows the distribution in the order of observation for the different disagreement categories, diff-$o$. The order of observation is categorized into early, middle and late visits. For a field that was observed in a particular phenological stage for three to five consecutive visits, early visit was the first visit and late visit was the last visit. In the seldom cases that a phenological stage was observed in six consecutive visits, then the first two and last two would be characterized as early and late visits, respectively. \\

\noindent Figure \ref{fig:stage4} shows that the majority of the perfect agreements, diff-$0$, were mostly middle and late observations. On the other hand the vast majority of big disagreements, i.e., diff-$1$, diff-$2$ and diff-$3$, were for early stage observations or, for fewer cases, late stage observations. This is expected, as middle observations would indicate that the field is well into a particular stage, whereas early or late observations denote transitional phases.

\subsection{Discussion}\label{discussion}

 \subsubsection{Crop phenology}
\noindent The results indicated that using clustering for the within-season phenology estimation is promising but challenging. EO and NWP data are competent predictor variables for this type of problems, as they represent both the land cover changes and the crop growth drivers, but cannot fully capture the physiological growth stages of crops. Furthermore, for operational applications in agricultural management, phenology predictions should be made at a within-season basis. However, there are still several challenges in this undertaking. First, within-season predictions are possible using only a part of the data time-series. Additionally, temporally dense EO time-series are crucial to ensure that critical phenological changes are detected as soon as possible \cite{gao2021mapping}. The density of the time-series however is subject to certain trade-offs. Considering the freely available satellite images, higher temporal resolution implies lower spatial resolution. Furthermore, optical SITS is significantly affected by cloud coverage. This might have not proved a problem for this study, having its area of interest in Greece, but is certainly an issue for other parts of the world. Finally, crop growth stages might not be directly related to EO and NWP atmospheric and soil variables. For this, there is a clear gap in the literature for sophisticated modelling that would be able to capture these complex relationships.\\

\subsubsection{Ground observations}
\noindent Ground observations are required to assess how well the EO and NWP based land surface phenology relates with the actual phenological stages. Most validation datasets are focused only on few stages using aggregated statistics over large areas \cite{gao2021mapping}. The National Agriculture Statistics Service (NASS) crop progress reports is such an example. Field-level ground observations are limited and rarely systematic. Recently, phenocams have been used to evaluate land surface phenology approaches \cite{zhang2018evaluation}. However, the network of phenocams is still sparse and confined. Furthermore, labeling phenological stages is a complex process and cannot be fully solved by observing photos from the field. Therefore, ground observations are still a necessity. \\

\noindent In this regard, this study offers a unique dataset of ground observations at the field level. The ground observations are accompanied by a panoramic and a couple of close-up photos of representative plants. We introduce a new field campaign protocol for collecting ground observations for crop growth that allows up to two label assignments. If two labels are assigned then the inspector should specify which one is the primary growth stage and which one is the secondary growth stage that describes the field. This allows for the detailed description of crop growth stages through the metaclasses that result from combining the primary and secondary stage annotations. Additionally, not having to decide on a single label makes the ground observation easier and can potentially increase the number of people who can perform them. This is true since the choice of the limits that define the principal growth stages differs among studies and ground observation protocols. Even more, closer to the start and end of those limits it gets tricky to decide on a single label. Having two labels can enable the reliable and large-scale crowdsourcing of ground observations.\\

\noindent Furthermore, the reliability of the ground observation collection method has been thoroughly evaluated. For this, we used the blinded interpretation of the field photos by an expert. Then the decisions of the field inspector and the photo inspector were evaluated by a third expert that decided on the percentage of agreement between the two. The quality assurance process yielded satisfactory results, deeming the ground observations reliable. The community is thus encouraged to use the openly available dataset and test their own models. The dataset is accompanied by the photos captured during the visits, which can be used for further interpretation but also computer vision tasks, such as crop classification and phenology classification.\\

\subsubsection{Clustering for phenology estimation}

\noindent It was shown that the introduced clustering method managed to learn from the complete time-series of 2020 and successfully infer the phenological metaclasses in a within-season fashion for 2021. Our model significantly outperforms the baseline, making the proposed approach very promising. Furthermore, the model predicts 16 different metaclasses and goes beyond the 6 principal phenological stages, extracting more information on their fuzzy transitions. This is particularly important since more intricate and precise agricultural management is now possible at the field level.  \\

\noindent In many studies, phenology estimation is addressed as a regression problem, aiming to predict the DoYs of growth stage onsets, which is essential information for operational applications. This study's metaclass approach can also be viewed as a classification-based alternative of onset detection. The fuzzy metaclasses (\(\lambda_a, \lambda_b\)) and  (\(\lambda_b, \lambda_a\)) denote this transitional phase between principal growth stages. In other words, the end of stage \(a\) and the onset of stage \(b\), respectively. In future work, further testing will be conducted for the evaluation of the spatial and temporal generalization of the proposed methodology. This will require additional ground observations at different areas and years of inspection.\\

\noindent We showed that formulating phenology estimation as a clustering problem, via incorporating the time in our features, is valid. The authors suggest that there is great potential and encourage the community to test more models on the proposed premise. During experiments, it was observed that the FCM was sensitive on the DoY of the first and last element included in the learning phase. This is expected since the clustering is largely dependent on time component of our features. It is therefore important to set the "start" and "end" instances with the aim to enclose the average length of the cotton season. This is easy when the sowing and harvest dates are known, but could also be approximated via observing the mean and standard deviation of the VI time-series. \\

\noindent This work relied on i) feature engineering, incorporating the time in the form of cumulative EO and NWP variables, and ii) feature selection to decide on a set of optimal feature spaces. Tens of thousands of experiments were performed for feature selection, yielding robust results. The robustness lies in the fact that the top 15 features systematically appeared in the best performing models. Furthermore, phenology predictions are based on the majority vote of the best-performing combinations of the top features, making sure our approach can generalize.\\

\noindent Having said that, there is a number of recent studies that look into DL based unsupervised change detection on SITS \cite{kalinicheva2020unsupervised, kalinicheva2018neural, kondmann2021spatial, andresini2021leveraging}. We see great potential in such approaches and we believe could be applicable in the proposed unsupervised premise for phenology estimation. Common denominator of these methods is the learning of a smaller latent or embedding space, in which entities that bear resemblance are located closer to each other. This is particularly important for clustering techniques that aim to group similar samples in the hyperspace. Usually, clustering algorithms, such as FCM, measure this similarity among entities using pair-wise distances. It is known that high dimensional spaces are not ideal for distance based techniques, as they usually fail to capture meaningful clusters. In addition, a latent manifold representation is not greatly dependent on feature engineering and can generalize well.\\

\subsection{Conclusion}\label{conclusion}
\noindent In this work we proposed a fuzzy clustering method for the within-season phenology estimation for cotton in Greece. Our method is unsupervised to tackle the problem of sparse, scarce and hard to acquire ground observations. It provides predictions within-season and thus enables its usage in operational agricultural management scenarios. It focuses on cotton which is important for three reasons -  i) it is an underrepresented crop type in the related literature, ii) the relationship between remote sensing phenology and the physiological growth of cotton is complex and iii) cotton is a very important crop for the economy and agricultural ecosystem of Greece, which is the study area. \\

\noindent We conducted field visits to collect ground observations that are offered to the community as a ready-to-use label dataset. For this, we used a new protocol that leverages two ranked labels. This makes the observations easier and at the same time provides enhanced information on the growth status. Therefore, we approach the problem as a multi-label one, introducing the notion of metaclasses. We go beyond the principal phenological stages of cotton by providing prediction for 16 metaclasses using the membership probabilities of the FCM classifier.  \\

\noindent Finally, we experimented with numerous combinations of features, including accumulated numerical simulations of atmospheric and soil paramaters, Sentinel-2 based VIs and their cumulative integral variants. Based on these experiments, we provided a list of optimal feature sets that can be used for cotton phenology estimation through majority voting. \\

\chapter{From Knowledge to Wisdom}

\section{Introduction}

\noindent The achievements of this thesis are indeed of great value towards the evidence-based decision making for a more sustainable and resilient agriculture. Big data technologies, multi-source datasets and advanced ML methods were employed to extract knowledge via remote detection; with crop classification and crop phenology estimation being the main detection outputs of this work. Nevertheless, bridging the gap between extracted knowledge and actionable advice is not straightforward, especially in the domain of agriculture. This thesis already provided relevant work in this direction with the semantic enrichment of crop type maps. However, significant work can be done in the future by employing interpretable and causal ML that will allow us to move from knowledge to wisdom. In the last couple of years, these ML domains enjoy significant attention and I believe their impact will be great in applications such as the ones introduced in this thesis. Causal ML allows for explicitly assessing the impact of cultivation practices, which is of great importance in policy making, especially in view of the fast changing climate. On the other hand, interpretable ML can assist in the rapid adoption of smart farming advice by the farmers. People tend to trust what they understand. Even more, if the results are interpretable they can be combined with the expert knowledge of the farmer to truly achieve wisdom. I believe this is the way to increase the value of outcomes, i.e., when the data meets domain knowledge. \\

\noindent To illustrate the importance for the aforementioned future perspectives, a causal ML method was developed to assess the impact of crop rotation and landscape crop diversity on climate regulation. In the same spirit, an interpretable ML method was implemented for estimating the onset of harmfulness of pests in cotton fields. This work is presented in the following sections.

%changed 20/04
\ifpdf
    \graphicspath{{Chapter5/Figs/Raster/}{Chapter5/Figs/PDF/}{Figs/Chapter5/}}
\else
    \graphicspath{{Figs/Chapter5/Vector/}{Figs/Chapter5/}}
\fi

\section{Causal ML: Land suitability}

%%%%%%%%% BODY TEXT
\subsection{Literature review}
\label{sec:intro}

\noindent One of the greatest challenges faced by humankind is producing and supplying food to a rapidly growing population in a world with changing climate~\cite{niedertscheider2016mapping}. Given the demand for natural resources, the expansion of croplands exerts substantial pressure on natural ecosystems. Ecosystem prosperity can be compromised by complex human-nature dynamics, risking human well-being itself~\cite{fu2019unravelling}. \\

\noindent The inclusion of complex causal relationships in environmental decision and policy making processes is key for  policy implementation under sustainable management regimes ~\cite{hunermund2019innovation, zheng2020causal}. Towards this direction, recent research utilizes environmental causal analyses to evaluate potential causes driving an observed or hypothesized change in specific target metrics~\cite{kluger2021combining}. In this work, we discuss causal inference and ML in the context of agricultural policy making that aims at increasing productivity to meet the global food requirements, while ensuring ecosystem resilience. \\

\noindent Predictive ML models can be inefficient for agricultural policy making since they are based on correlations and aren't inherently capable of addressing causal questions that a policy maker is after~\cite{hunermund2019causal}. Even when causal analyses are employed, they are usually based on localized experiments on few samples and cannot account for the spatial variability caused by changes in climatic conditions, management practices and other factors ~\cite{deines_satellites_2019}. Finally, policies are often horizontally implemented, leading to the lack of spatial targeting among areas with different ecological characteristics ~\cite{Cullen2018-mc}. To mitigate these issues, the CAP proposes the introduction of agri-environmental measures and eco-schemes that are tailored and optimized to the specificities of the different regions ~\cite{ec2018post}. \\

\noindent Every policy measure aims at implementing an agricultural management practice to achieve a certain goal. In order to assess the efficiency of policy measures to agroecological resilience, policy makers need to know whether particular management practices can achieve a given set of objectives \cite{lampkin2021policies}. Specifically, it is required to know if and how practices are effective at influencing key target metrics, such as the supply of ecosystem services or climate change mitigation and adaptation \cite{scarano2017ecosystem}. There are several studies that approach policy evaluation in experimental settings\cite{colen2016economic}. Nevertheless, large-scale experiments on agricultural policies are hardly feasible and thus the only way to assess their impact is through the use of observational data ~\cite{del2019causal}.\\

\noindent Land and crop information derived from satellite images, and environmental data derived from numerical simulations can both capture large areas with high frequency and at high spatial resolutions. ML based causal inference on observed and simulated data is a promising complementary approach to field experiments, as it enables the detailed assessment of the heterogeneous effects of agricultural practices over millions of fields. The exploitation of this kind of data ensures high spatial representation, accounting for regional differences in climatic, soil and crop conditions. Nevertheless, it should be noted that causal inference on observational data is subject to biases from measurement, selection and confounding ~\cite{rubin2005causal, pearl2009causality}. \\

\noindent The impact evaluation of the agricultural practices of Crop Rotation (i.e., growing different crops across a sequence of growing seasons, abbreviated as CR) and Landscape Crop Diversity (i.e., increasing the number and evenness of crops grown in a landscape, abbreviated as LCD) can be used to allocate them in space. If this evaluation happens on the basis of agroecosystem productivity, this would in turn enable the optimization of climate regulation, while sustaining productivity ~\cite{folberth2020global}.\\

\begin{figure*}[!ht]
  \centering
    \includegraphics[width=1\linewidth]{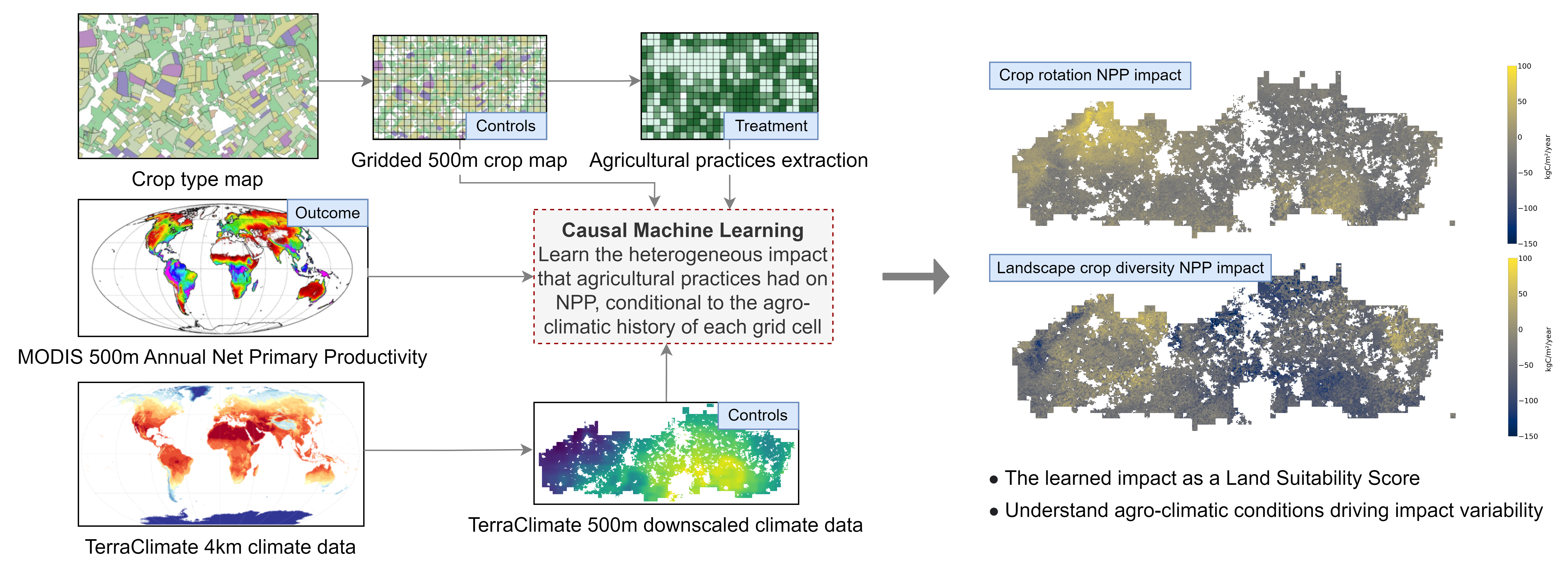}
    \caption{An overview of our process for learning the heterogeneous impact of CR and LCD on  NPP. The study period extends from 2010 to 2020. First, the parcel-level crop type maps are transformed to gridded maps of 500 m cells to match the pixel size of MODIS. For any given year and grid cell, by considering the percentage area of the cell that a crop type occupies, a crop abundance value for all major crop types is generated. Based on this, for each grid cell we compute i) the LCD using the Shannon diversity index and ii) CR via quantifying the change between consecutive years in crop abundances within each cell, and summing for the whole study period. Our input data also include environmental variables from the TerraClimate dataset and MODIS NPP values for each cell. We run two separate analyses, one for each practice. Considering the value of each practice as the treatment, NPP as the outcome, and controlling for relevant crop abundances and environmental factors, the dataset is modeled with DML. The heterogeneous impact of a practice per grid cell is then studied and used as a land suitability score.}
    \label{fig:method}
\end{figure*}

\noindent \textbf{Contributions.} We demonstrate the applicability of causal ML, specifically Double Machine Learning (DML), to estimate the impact that the LCD and CR agricultural practices had on Net Primary Productivity (NPP), that we use as an ecosystem service proxy of climate regulation. By deriving treatment effects and analyzing their heterogeneity, we infer a data-driven and context-aware agricultural land suitability score for both practices. Specifically, we propose a flexible observation-based framework that can be extended to include any management practice and agro-environmental metric that practitioners might have data on. Our approach, illustrated in Figure \ref{fig:method}, was tested in Flanders for the period 2010-2020, where we studied the practices' effect heterogeneity and discussed results in the context of local characteristics. 

\subsection{Related Work}

\noindent \textbf{Agricultural land suitability.} 
Understanding agricultural land suitability is critical for policy applications aiming to foster sustainable agricultural practices and ensure food security. According to Akpoti ~\cite{akpoti2019agricultural}, the analysis of land suitability depends on many factors, including the purpose of the assessment and data availability, and therefore there is no one-size-fits-all approach. Studied techniques are commonly categorized into traditional and modern approaches. Traditional approaches include quantitative and qualitative methods accounting for biophysical characteristics of land ~\cite{danvi2016spatially,el2016mapping}. On the other hand, modern methods exploit a plethora of variables and, according to Mugiyo et al.~\cite{mugiyo2021evaluation}, are categorized into i) computer-assisted overlay mapping, ii) geo-computation or ML and iii) Multi-Criteria Decision Making (MCDM). Jayasinghe et al. ~\cite{layomi2019assessment} used the Analytical Hierarchy Process (AHP), which is the most common MCDM method in literature, and the Decision-Making Trail and Evaluation Laboratory (DEMATEL) model to assess the suitability of land for tea crops. Using DEMATEL they were able to visualize interrelationships between factors in a causal diagram. Here, we explicitly frame agricultural land suitability as a causal ML task, attempting to spatially optimize a metric of interest via intervening on the land use; and to the best of our knowledge this is the first work that does so.\\

\noindent \textbf{Heterogeneous Treatment Effects.} The Average Treatment Effect (ATE) has traditionally been the main concern of causal inference. However, it does not necessarily convey all the needed information, particularly in cases where treatment effects vary systematically with sample characteristics. This is frequently the case in applications, and treatment effect heterogeneity has been studied in diverse fields such as medicine, social sciences, and economics~\cite{kravitz2004evidence, gabler2009dealing, kent2010assessing, heckman1997making, heckman2001, RePEc:nbr:nberwo:12145, imai2013estimating, davis2017using, lechner2002program}. Such methods have also been researched in the context of earth and environmental sciences, where heterogeneity of effects is present. The magnitude of effects is determined by various factors, including geographical, climatic and land cover. For instance, Serra-Burriel et al.~\cite{serra-burriel_estimating_2021} estimated the heterogeneous effects of wildfires and Deines et al. \cite{deines_satellites_2019} found significant variability in the effects of conservation tillage on yield. Both studies utilized EO to uncover the heterogeneity of treatment effects.\\

\noindent \textbf{Causal Machine Learning.} While statistical methods can be used for the estimation of heterogeneous treatment effects~\cite{crump2008nonparametric, lee2009non, willke2012concepts}, there has been a recent surge of ML methods tailored for the same task~\cite{chernozhukov2018double, wager2018estimation, kunzel_metalearners_2019, dudik2014doubly, jesson_quantifying_2021}. Such methods are suitable for high-dimensional data and are able to learn general functional forms, thus taking advantage of the ever-increasing volume of data. Causal ML was recently used in environmental studies utilizing both satellite observations and reanalysis data. The interaction of aerosols and clouds was studied in Jesson et al.~\cite{jesson_using_2021} with causal neural networks~\cite{jesson_quantifying_2021}. In the context of agriculture, Deines et al. ~\cite{deines_satellites_2019} investigated the effect of conservation agriculture on yield using causal forests~\cite{wager_estimation_2018, athey_generalized_2019} and studied its heterogeneity across space. \\

\noindent By employing remote sensing data and numerical predictions, one can achieve large spatio-temporal coverage that is much needed for effect heterogeneity to manifest. We also took advantage of the scale of EO and climate data and combined them with data that reflect historical agricultural use. We were then able to derive the LCD and CR practices, and estimate their heterogeneous impact on the climate regulation for each land sample.

\subsection{Data \& Methods}\label{3}

\noindent We first provide a brief overview of the theory of Conditional Average Treatment Effects in Sec. \ref{3.1}. We then provide details on the data that were used in our approach, as illustrated in Figure \ref{fig:method}. In Sec. \ref{3.2}, we describe the data derived from crop type maps, including i) the LCD and CR agricultural practices that are used as treatments in our analysis and ii) the crop type abundance features that are used as controls. In Sec. \ref{3.3}, we provide information on the NPP, which is the outcome of our heterogeneous treatment effects analysis, and in Sec. \ref{3.4} we describe the environmental factors that are also used as controls. The section concludes with Sec. \ref{3.5} and the methodological setup. 

\subsubsection{Conditional Average Treatment Effects}\label{3.1}

\noindent \textbf{Terminology and Assumptions.} Using the potential outcomes framework~\cite{rubin2005causal}, let $Y(T)$ denote the value (outcome) of a random variable $Y$ if we were to treat a unit with a binary treatment $T \in \{0,1\}$. Given a vector of features $X$ describing the units, we want to estimate:

\begin{equation}
    \theta(x) = \mathbb{E}[Y(1) - Y(0) | X = x]
    \label{cate}
\end{equation}

\noindent This heterogeneous treatment effect is referred to as the Conditional Average Treatment Effect (CATE)~\cite{abrevaya2015estimating}. There are three important assumptions: Overlap (i.e. the probability of receiving treatment $t$ given features $x$ is bounded away from $0$ and $1$), Unconfoudedness (i.e. all confounders for the relation under study are observed), and Consistency (i.e. observed outcomes of units that were assigned treatment $t$ are identical to their potential outcomes if they are assigned treatment $t$). 
\begin{table}[!ht]
  \centering
  \begin{tabular}{@{}lc@{}}
    \midrule
    Overlap & $0 < \mathbb{P}(T=t|X=x) < 1 \quad \forall t$ \\
    Unconfoundedness & $(Y(1), Y(0))\indep  T | X$ \\
    Consistency & $T=t \implies Y = Y(t)$\\
    \bottomrule
  \end{tabular}
  %\caption{Results.   Ours is better.}
  \label{tab:causalassumps}
\end{table}

\noindent Provided these assumptions hold, CATE is identifiable from observational data $\{(y_i, t_i, x_i)\}_{i=1}^N$, and the following is an unbiased CATE estimator:
\begin{equation}
    \hat{\theta}(x) = \mathbb{E}[Y | T =1, X = x] - \mathbb{E}[Y | T = 0, X = x]
\end{equation}

\noindent CATE can be similarly defined for continuous treatments, and by averaging over $X$ it reduces to the standard ATE $\mathbb{E}\big[Y(1) - Y(0) \big]$. Variables included in the feature vector $X$ are known as controls. Depending on the causal structure of the phenomenon under study, controlling for a variable might reduce or increase bias of effect estimates~\cite{cinelli_crash_2020}.\\

\noindent\textbf{Double Machine Learning.} Assuming unconfoudedness, Double Machine Learning~\cite{chernozhukov2018double} formulates the data generating process in terms of the Partially Linear Model~\cite{robinson1988root}:

\begin{align}
    Y &= \theta(X)\cdot T + g(X) + \varepsilon \label{doublemllinearity}\\
    T &= f(X) + \eta \label{confounding}
\end{align}

\noindent where $\theta(X)$ is the CATE, and $g,f$ are arbitrary functions (nuisance parameters). Notably, \eqref{confounding} keeps track of confounding as features $X$ drive both the treatment $T$ and outcome $Y$. CATE $\theta(X)$ is then estimated using a two-stage estimation procedure. During the first stage, both the outcome $Y$ and treatment $T$ are separately predicted from features $X$ using arbitrary ML models. Then, in the final stage, the CATE $\theta(X)$ is estimated by solving (for the linear case)

\begin{equation}
    \hat{\theta} = \argmin_{\theta \in \Theta}\mathbb{E}\big[(\Tilde{Y} - \theta(X)\cdot \Tilde{T})^2 \big]
    \label{doublemlfinalstage}
\end{equation}

\noindent over a model class $\Theta$, where $\Tilde{Y}$ are the residuals of the $Y \sim X$ regression, and $\Tilde{T}$ are the residuals of the $T \sim X$ regression. In \eqref{doublemllinearity} and \eqref{doublemlfinalstage} the linearity assumption can be dropped to allow for fully non-parametric CATE estimation.

\subsubsection{Data derived from crop type maps}\label{3.2} \label{subsection:croppractices}

\noindent Co-designing practical and effective solutions to major challenges, such as climate change, requires understanding the pattern of interventions carried out by agents of change in agricultural ecosystems, i.e farmers \cite{bohan2022designing}. LCD and CR are considered to be important management practices since they support synergistic improvements in crop productivity, environmental health, and ecological sustainability \cite{bowles2020long, tamburini2020agricultural, egli2021more}. Hence, such practices should be considered in strategies that promote sustainable agricultural production. \\

\noindent We exploited a series of yearly Land Parcel Identification System (LPIS) data to produce variables representing agricultural management practices \cite{mikolajczyk2021species}. LPIS is a geo-spatial database that contains the geometries of the parcels and the declared crop type by the farmers, as part of their  application for CAP subsidies \cite{rousi2020semantically}. \\

\noindent For each year, we extracted the proportion of the total area occupied by each crop type (crop abundance) per grid cell of the MODIS 500m grid. We then calculated the Shannon diversity index (H')~\cite{shannon1948mathematical, morris2014choosing} as a metric of LCD per grid cell. We finally took the mean of all years to end up with an average LCD value per cell for 2010-2020. For CR, we summed the absolute difference per crop type abundance per grid cell for two consecutive years; this procedure was repeated for all pairs of adjacent years with which we calculated the total (sum) rotations for the studied period. Both LCD and CR were used as treatments to test their effect on NPP, while crop type abundances for the major crop types in Flanders (grassland, maize, potato, wheat) were used as controls.

\subsubsection{Net Primary Productivity}\label{3.3}

\noindent Net primary productivity (NPP) is the uptake flux, in which carbon from the atmosphere is sequestrated by plants through the balance between photosynthesis and plant respiration. It is a fundamental ecological variable in biosphere functioning, the quantification of which is needed for assessing the carbon balance at regional and global scales\cite{yuan2021effects}. As NPP has been widely used for measuring vegetation dynamics, this index is highly suitable for capturing environmental changes due to natural and anthropogenic factors \cite{gang2017modeling,zhou2021identifying}. The MODIS NPP is produced by the US National Aeronautics and Space Administration (NASA) Earth Observing System. It is based on an energy  budget approach, utilizing EO on the  fraction of photosynthetically active solar radiation absorbed by the vegetation surface \cite{pan2006improved,running2004continuous}. The results have been validated as being able to  capture spatio-temporal patterns across various biomes and climate regimes \cite{zhao2005improvements}. We used annual time-series MODIS NPP (MOD17A3) gridded at 500 m for 2010-2020 as the target metric (outcome) representing the ecosystem function of climate regulation over Flanders (North Belgium). The MOD17A3 annual product is derived by summing the corresponding 8-day product, allowing it to capture seasonal variability.

\subsubsection{Environmental factors}\label{3.4}

\noindent Variability in agricultural NPP could arise due to environmental effects, with specific conditions favoring specific crop types. To uncover the diverse environmental conditions that support or inhibit climate regulation we used a series of parameters provided by Terraclimate\cite{abatzoglou_terraclimate_2018}. The gridded meteorological data  were produced through a climatically aided spatiotemporal interpolation of the WorldClim datasets to estimate monthly time-series. The confounding environmental factors used within this study included maximum and minimum Temperature, actual evapotranspiration, climate water deficit, precipitation, soil moisture, downward surface shortwave radiation and vapor pressure. The idea behind the selection of the aforementioned variables relies on the wide recognition that temperature and precipitation directly affect NPP \cite{chu2016does, gholkar2014influence, yuan2021effects}, the influence of water availability on soil productivity and vegetation growth \cite{hailu2015reconstructing}, and the processes themselves (such as evapotranspiration) that are closely related to the ecosystem function of climate regulation \cite{yang2020emergy,sun2017ecohydrological}.

%In particular, the key determinants for changes in NPP include precipitation, temperature, solar radiation, atmospheric $CO_2$, soil productivity, irrigation water, fertilizer and pesticides.      

\subsubsection{Methodological setup}\label{3.5}

\noindent We have thus created a population of $N$ units indexed by $i$, where $Y_i(T_i)$ is the observed outcome of unit $i$ treated with $T_i$ and $X_i$ are the features of $i$ that generate systematic variation on the treatment effect. Specifically, the population consists of $N$ grid cells exhaustively covering agricultural land within Flanders, $Y_i$ is the observed NPP value of each cell, $T_i$ is (the value of) the agricultural practice that was applied to the cell, and $X_i$ refers to important characteristics of the specific grid cell, comprising environmental data and agricultural use (i.e., crop types cultivated within the grid cell). The CATE \eqref{cate} is then the average impact that practice $T_i$ had on NPP $Y_i$ for the $i$-th grid cell, conditional to its characteristics $X_i$. Obtaining such local insights on where and why the impact magnitude of agricultural practices on important agro-ecosystem metrics differed, is of primary interest to policy makers as it allows for targeted agricultural policy making. \\

\noindent All data mentioned in the previous subsections are retrieved as time-series, with different spatio-temporal resolutions. Data were temporally aggregated over the period of study (2010-2020), since treating every cell-year combination as different units would potentially introduce interference effects~\cite{tchetgen2012causal}, where treatments (agricultural practices) during any year might also influence the outcome (climate regulation) of the next year. For the case of CRs, temporal aggregation over the 2010-2020 period happens by summing all the rotations that happened in grid cell $i$ from 2010 to 2020, while for LCD we consider the average Shannon diversity index over all years. This is the treatment $T$ we are using for the analyses.\\

\noindent By adding the crop abundances to the feature vector $X$, we control for the four most dominant crop types mentioned in Sec. \ref{3.2}; the median abundance value of all other crop types is less than $2\%$. We also control for important environmental variables, listed in Sec. \ref{3.4}. Crop types are confounders for the causal relation under study and thus good controls as they drive the magnitude of both the agricultural practices and the NPP. As such, they should be included in the model. On the other hand, environmental data are not driving any practice, but they are driving the NPP values, thus making them strong candidates for good controls \cite{cinelli_crash_2020}. We finally binarize the treatment by letting the median CR value and median LCD value to be thresholds, over which we designate cells as treated units and the rest as control.\\

\noindent The assumption of unconfoudedness can't be tested, and within any observational study it is likely to be invalid. Even if the variables we controlled for will help with bias reduction, some bias from unobserved confounders might still be present. Nevertheless, we note that by being cautious with the selection of controls we try to avoid selection bias, and reported results provide a large-scale complementary approach to localized field experiments.

\subsection{Experimental Results \& Discussion}

\noindent\textbf{Filtering.} For the experiments we used the EconML Python package implementing the DML method \cite{econml}. We derived CATE estimates for both LCD and CR at the 500m native NPP resolution. \\
\noindent We ensured that fitting takes place over actual croplands by restricting the dataset to grid cells where the sum of crop abundances, given by the crop type maps, exceeded a threshold of $80\%$. To aid the overlap assumption, propensity scores were first estimated using a Gradient Boosting Propensity Model from the CausalML Python package~\cite{chen2020causalml}, and units with extreme propensity scores ($\leq 0.2$ or $\geq 0.8$) were filtered out. Within DML, it is crucial to avoid overfitting first stage models; otherwise a portion of the outcome and treatment variability explained will be due to factors other than the controls. During the first stage of DML only, we therefore split the dataset to train and test (80-20) to evaluate the predictive performance and assess overfit.\\

\noindent\textbf{Double Machine Learning.} In the first stage, for both the CR and LCD analyses, Random Forest Regressors were selected to predict NPP from controls, outperforming Lasso and Gradient Boosting Regression. To predict the binary treatments themselves, Logistic Regression was used, outperforming Random Forest and Gradient Boosting classifiers.\\

\noindent All model selection procedures happened by performing 3-fold cross validation and a grid search for hyperparameter optimization. During the first stage only, the maximum absolute scaler was applied , as it retains the sparsity structure that is prevalent with crop abundance data. For the final stage regression, for both analyses, a Causal Forest \cite{wager_estimation_2018} with $1000$ trees was used with heterogeneity score as the splitting criterion. The final stage causal forest was fine-tuned based on the out of sample performance. Minimization of \eqref{doublemlfinalstage} happened with unscaled features to maintain interpretation in the original units of measurement. Fitting results for CR and LCD  can be found in Table~\ref{tab:dmlfits}. First stage models captured a significant part of the variability of both the outcome and treatment variables in both cases. The difference between the training and test performance was marginal, which indicates that models avoided overfitting.

\begin{table}[!ht]
  \centering
    \caption{Performance ($R^2$ for outcome modeling, F-1 score for treatment modeling) of ML models internally used by DML. Outcome $Y$ is the NPP, binary treatment $T$ is CR and LCD, and $X$ is a vector of features}
  \begin{tabular}{@{}lcr@{}}
    \toprule
   \textbf{Crop Rotation} & Train & Test\\
    \midrule
    $Y \sim X$ (Outcome Modeling) & $0.75$ & $0.74$\\
    $T \sim X$ (Treatment Modeling) &  $0.63$ & $0.62$ \\
    \toprule
    \textbf{Landscape crop diversity}  & Train & Test\\
    \midrule
    $Y \sim X$ (Outcome Modeling) & $0.73$ & $0.71$\\
    $T \sim X$ (Treatment Modeling) &  $0.59$ & $0.62$ \\
    \bottomrule
  \end{tabular}

  \label{tab:dmlfits}
\end{table}

\begin{figure*}[!b]
  \centering
  \begin{subfigure}{0.8\linewidth}
    \includegraphics[width=1\linewidth]{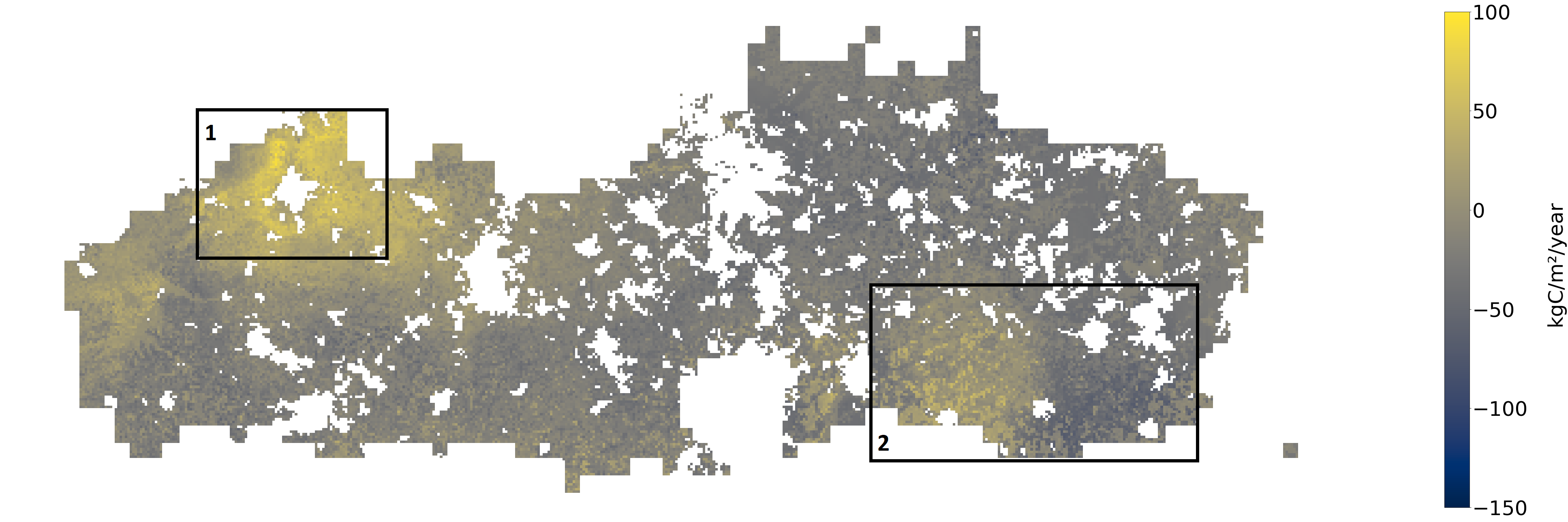}
    \caption{Impact of crop rotation on climate regulation (NPP) in Flanders at 500 m resolution.}
    \label{fig:map-a}
  \end{subfigure}
  \hfill
  \begin{subfigure}{0.8\linewidth}
    \includegraphics[width=1\linewidth]{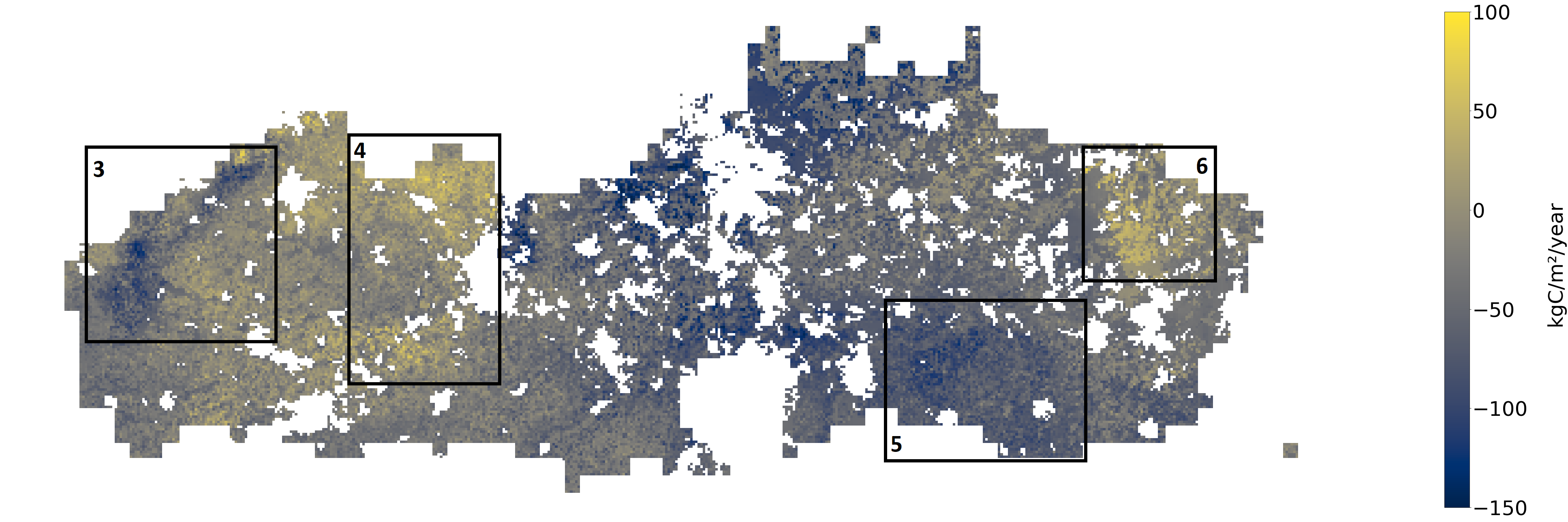}
    \caption{Impact of landscape crop diversity  on climate regulation (NPP) in Flanders at 500 m resolution.}
    \label{fig:map-b}
  \end{subfigure}
  \caption{Estimated impact of crop rotation (a) and landscape crop diversity (b) on climate regulation (NPP) based on the heterogeneous treatment effect analysis for the Flanders region from 2010 to 2020. The annual NPP outcome variable is measured in kgC/m²/year.}
  \label{fig:map2}
\end{figure*}

\subsubsection{CATE Results}\label{4.1}

\noindent Figure~\ref{fig:map2} illustrates the DML CATE estimates over the region of Flanders for LCD and CR. Intervening by applying a practice in a cell with a CATE estimate of e.g. $20$, we would expect an increase of $20$ kilograms of carbon for each squared meter of land within a year. From such results , a spatially explicit policy for sustainable agriculture can be extracted, by prioritizing for each cell the application of practices whose effect estimate (land suitability score) is high. Such analysis can be repeated for any agricultural practice and outcome metric of interest, while appropriate controls can be added using expert knowledge.\\

\noindent Our results indicated that in areas in the north-west of the study area, which are also characterised by high abundance of grasslands, CR increased NPP by approximately 100 kgC/m²/year (Figure \ref{fig:map-a}; square 1). This suggests that the contribution of grasslands in crop rotations is effective in positively driving climate regulation \cite{zarei2021evaluating}. A positive effect was also found on the north eastern regions where the dominating crop type is winter wheat followed by maize and grasslands (Figure \ref{fig:map-a}; square 2). \\

\noindent For LCD, negatively affected areas on the western coastal region (Figure \ref{fig:map-b}; square 3) and the south eastern region (Figure \ref{fig:map-b}; square 5) that are dominated by grasslands and winter wheat presented a decrease in NPP of over 100 kgC/m²/year. Due to the importance of the previously mentioned crop types to the carbon cycle, one would expect a positive impact on climate regulation. However, as they appear to be clustered in space, implementing LCD in such cropland regions led to a negatively affected NPP.
On the other hand, where multiple crop types exist (hence, higher diversity), maize showed a significant role in carbon uptake as its existence in a diversified agricultural landscape indicated an increase of approximately 50 kgC/m²/year (Figure \ref{fig:map-b}; square 4 and 6). In fact, maize has a high capacity in capturing large amounts of carbon from the atmosphere \cite{lal2010managing}.\\

\begin{figure}[!ht]
  \centering
  \includegraphics[width=1\linewidth]{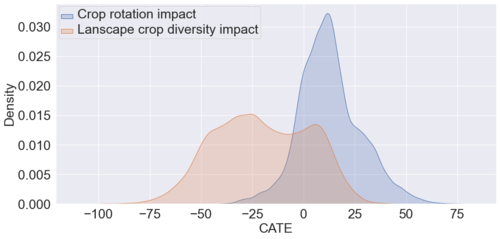}
    \caption{Distribution of CATE estimates for crop rotation and landscape crop diversity.}
  \label{fig:rota500distr}
\end{figure}
\begin{figure*}[b!]
  \centering
  \begin{subfigure}{1\linewidth}
    \includegraphics[width=1.2\linewidth]{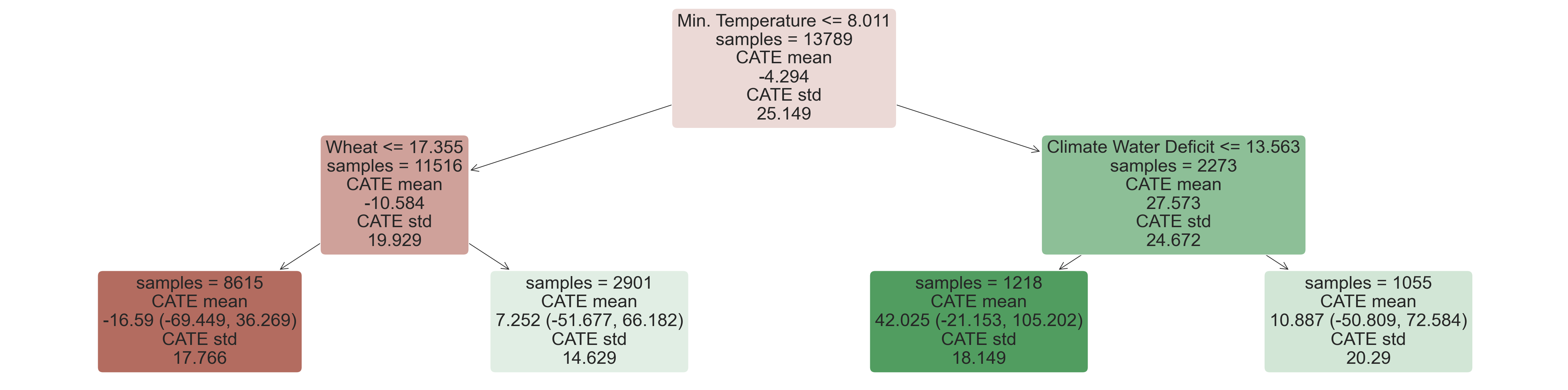}
    \caption{Heterogeneity tree for crop rotation.}
    \label{fig:tree-a}
  \end{subfigure}
  \hfill
  \begin{subfigure}{1\linewidth}
    \includegraphics[width=1\linewidth]{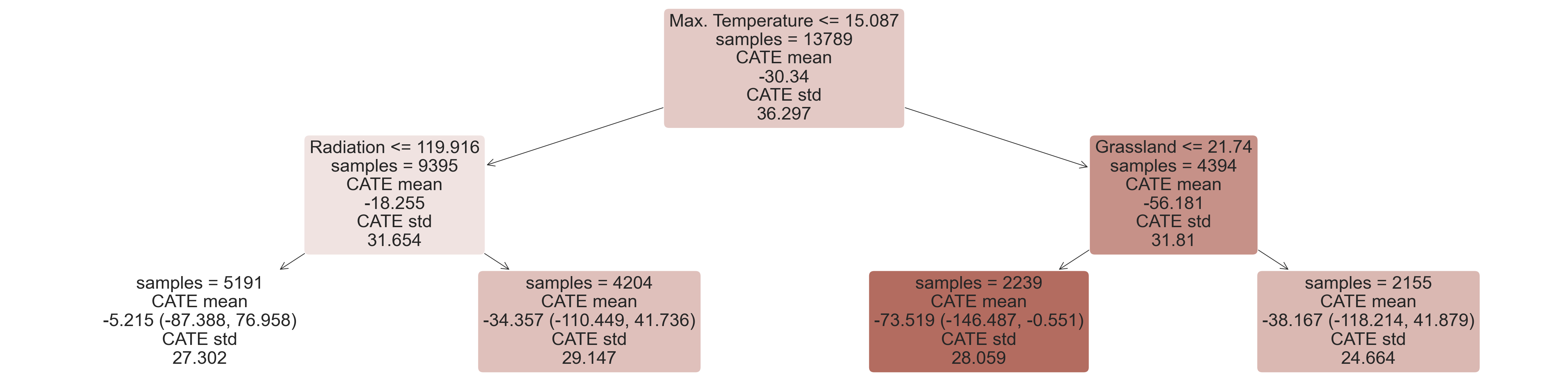}
    \caption{Heterogeneity tree for landscape crop diversity.}
    \label{fig:tree-b}
  \end{subfigure}
  \caption{Trees interpreting effect heterogeneity for both practices studied. To be read from top to bottom, going left if the Boolean condition at the top of each box is true, and right if it is false. Sample size for each leaf is reported, alongside the CATE mean, standard deviation and confidence intervals. Wheat and Grassland abundance are in \%, Min. and Max. Temperature are in °C, Climate Water Deficit is in mm and Radiation is in W/m².}
  \label{fig:tree}
\end{figure*}

\noindent The ATE estimates for CR were $1.08$ $(95\% \text{ CI } [-20.35, 22.51])$ and for LCD were $-35.73$ $(95\% \text{ CI } [-58.73, -12.73])$. Additionally, the CATE distribution of both practices over the Flanders cropland can be seen in Figure \ref{fig:rota500distr}.
Besides the more granular CATE insights we are reporting, we also note that a LCD practice implemented over the entire cropland of Flanders would have a weak yet significant negative impact on NPP. On the other hand, a CR practice implemented across the entire terrain will be able to bring multiple agro-environmental benefits without impacting the average NPP \cite{karlen2006crop, havlin1990crop}.

\subsubsection{Heterogeneity Analysis}

\noindent We used causal trees to analyze the heterogeneity of the model, accounting for all features in $X$ ~\cite{athey_recursive_2016, econml}. The tree successively splits on feature values that maximize the treatment effect difference across leaves. We constrained the tree depth to $2$ to retain explainability. In Figure~\ref{fig:tree-a} we see that minimum temperature was found to be the most important driver of effect heterogeneity for CR. Particularly, high minimum temperatures combined with low climate water deficit seems to benefit CR performance in Flanders. The abundance of wheat was also detected as a major contributor to CR heterogeneity, spatially coinciding with high CATE estimates as reported in Sec. \ref{4.1}.\\

\noindent The heterogeneity in the impact of LCD (Figure~\ref{fig:tree-b}) was found to be affected by maximum temperature, solar radiation and grassland abundance. LCD negatively affected NPP the least when maximum temperature was lower than 15.1\textcelsius, and radiation was lower than 119.9 W/$m^2$. Additionally, in high maximum temperatures, diversifying crops in grassland-abundant cells appears more preferable than diversifying cells lacking grasslands.\\

\noindent While trees are useful to detect the most important features in the sense described above, a complementary understanding of effect heterogeneity can be obtained by plotting CATE estimates as a function of selected covariates. In Figures \ref{fig:rota500tmax} and \ref{fig:div500tmax}, we see clear patterns that show maximum temperature driving the effect heterogeneity for both CR and LCD. For the rest of the heterogeneity features, we report Spearman correlation coefficients between them and the estimated CATE. The results showed moderate correlations for almost all factors, which indicates that the effect under study is indeed heterogeneous. The correlation values for environmental features are larger but comparable to the ones of crop features, highlighting that both control categories contribute to the heterogeneity.

\subsubsection{Agricultural land suitability \& climate change}

\noindent So far, we estimated the impact of agricultural practices on climate regulation from historical data (2010-2020). 
Nevertheless, important climatic variables are expected to change over the next decades as a result of climate change. The estimation of their trajectory has been the subject of numerous scientific studies \cite{stocker2014climate}. In order to devise impactful agricultural policies, we need to evaluate the performance of practices in future climatic conditions. \\

\noindent In the context of causal ML, we learned the function $\theta(x) = \mathbb{E}[Y(1) - Y(0) | X = x]$ where feature vector $X$ contains multiple climate variables. In theory, we are able to derive the impact of a practice on an outcome metric under future climatic conditions by changing the relevant variables of feature vector $X$ and re-calculating. However, $\theta(x)$ was learned from historical data, and by definition future climatic conditions were not observed. By changing climate variables to match future projections, we are essentially extrapolating the function to points outside the observed feature space. \\

\begin{figure}[!ht]
  \centering
  \includegraphics[width=1\linewidth]{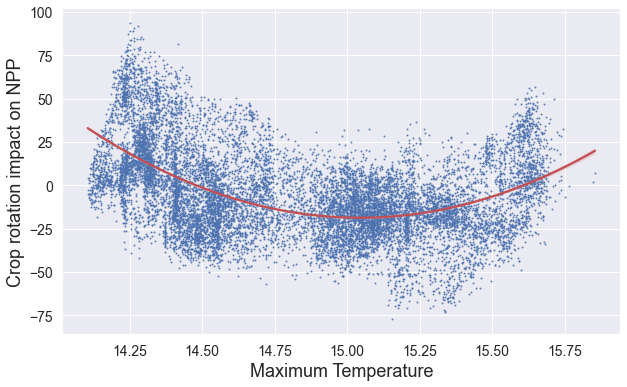}
    \caption{CR impact as a function of Max. Temperature (°C).}
  \label{fig:rota500tmax}
\end{figure}

\begin{figure}[!ht]
  \centering
  \includegraphics[width=1\linewidth]{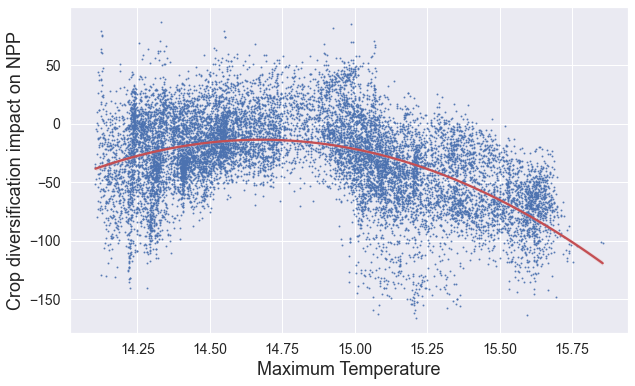}
    \caption{LCD impact as a function of Max. Temperature (°C).}
  \label{fig:div500tmax}
\end{figure}

\noindent In Figure \ref{fig:rota500tmax} for example, a quadratic trend between the CR CATE and maximum temperature is seen, allowing us to hypothesize that in slightly warmer conditions the benefit of CR would increase. We thus note that these results provide preliminary insights on an ex-ante impact assessment for the agricultural practices studied in the context of climate change and enable the formulation of scientific hypotheses for further study. \\

\begin{figure}[!ht]
  \centering
  \includegraphics[width=1\linewidth]{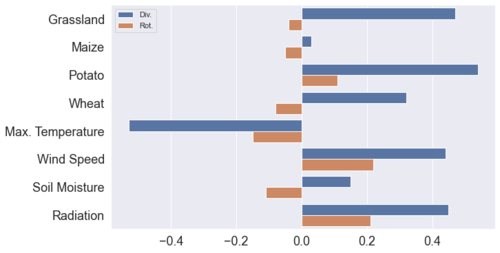}
    \caption{Barplot showing Spearman correlation coefficients of estimated CATEs with all features for both practices. }
  \label{fig:barplot}
\end{figure}

\subsection{Conclusions}

\noindent We presented the first approach for assessing agricultural land suitability using causal ML.  We estimated the heterogeneous treatment effects of CR and LCD on climate regulation (NPP), accounting for historical crop and environmental data. We used EO (MODIS), climate data (TerraClimate) and openly available yearly crop type maps that cover North Belgium (Flanders) from 2010 to 2020. CATE results show significant heterogeneity in space, highlighting the usefulness of extended spatio-temporal data coverage, the importance of spatially targeted measures and the relevance of CATE estimates as land suitability scores.
The ATE for CR is insignificant, while for LCD it is significant but relatively small compared to the influence environmental drivers have on NPP \cite{zhou2021identifying}. It would be difficult to extract such subtle effects from localized experiments of confined samples.\\

\noindent Significant challenges remain towards the application of the proposed approach in real-life land planning scenarios. Observational studies enable the manifestation of heterogeneity, which is vital for spatially explicit recommendations, but suffer from unobserved confounding. For this reason, our results should be used together with insights from local field experiments. Furthermore, we offered insights on how our two treatments affect climate regulation but we did not provide detailed information on the contribution of specific crop types. For the CR treatment, we do not take into account the crop types involved in the transition but simply the transition itself. Similarly, for the LCD all crop type mixtures are treated equally. Additional experiments with crop-aware treatments can offer enhanced understanding for evidence-based decision making.

%%%%%% Second work %%%%%%
\section{Interpretable ML: Pest presence prediction}

\subsection{Literature review}

\noindent Greece is the primary cotton producer for the European Union (EU) and the fifth-largest cotton exporter in the world; which makes cotton one of the most important crops for the national economy \cite{A1, tsiros2009assessment}. It is therefore imperative to protect cotton fields from pests and diseases with timely and effective actions to avoid damages. Helicoverpa armigera, also called cotton bollworm, is a serious threat to the crop, resulting in yield losses and suboptimal lint quality. This pest is widely abundant, especially in countries with warm or temperate climate like Greece \cite{A2}. In this work, we implement an interpretable ML approach to detect pest presence using satellite observations and meteorological data. By doing so, we identify the onset of harmfulness. In other words, consecutive presence estimations of the pest act as information about upcoming population peaks. This enables the farmers to optimize their pest management process. \\

\subsection{Related work}

\noindent In order to successfully protect crops from pests, we need to understand their life cycle and thereby intervene before the abundance of insects becomes detrimental. Cotton bollworm's life cycle is relatively fast, and it takes place in four stages. First, the female moths lay the eggs. If the weather conditions are favourable, these eggs can hatch in less than three days. Then the eggs hatch and larvae emerge. This is the destructive phase of the pest. Larvae then turn into pupae, which are usually buried at a depth of $4$ to $10$ centimetres in soil and take $10$ to $15$ days to develop in a cocoon. An adult moth hatches out of the cocoon, ready to start the reproductive cycle all over again. Generally, it takes the pest approximately one month to complete its life cycle \cite{A3}.\\

\noindent Deep learning models, and mostly Convolutional Neural Networks (CNNs), have been widely used to recognise plant pests and diseases on plant images \cite{A12}. The aforementioned techniques work on close-up photos from the field that cannot be available frequently over large areas. As an alternative, one can use coarser spatial resolution data that can provide better coverage and temporal resolution, such as weather predictions and satellite images. For this, however, we need pest traps to know the occurrence and population of insects. There are different types of pest traps, i.e., pheromone, light or sticky traps etc. \cite{B2}. Using such traps renders the problem quantitative, i.e., the interest no longer lies on the pest or damage recognition, but on the estimation of its population.\\

\noindent Several studies have worked on estimating pest population (regression) \cite{A9} or detecting significant presence (binary classification) \cite{A7,A8}. With reference to the latter, knowing the exact population of pests is not always crucial. Instead, action thresholds are used to classify the amount of pests as harmful or not \cite{A5}. One can find approaches that make use of i) physical models \cite{A10} or ii) data driven models that employ remote sensing and in-situ measurements \cite{A7,A11}. It is worth noting that most related studies use weather data, whereas in fewer cases remote EO are used to capture the changes caused by the pests, but also the favourable vegetation conditions for their occurrence \cite{A11}.  Recurrent Neural Networks (RNN) have been applied to weather data time-series, accounting for the temporal evolution of features and capturing the cyclical nature of pest abundance \cite{A4}. In other words, when trying to predict pest occurrence at a given time instance, one cannot ignore the weather conditions and the vegetation status of previous days \cite{A13}.\\

\noindent Traditional ML is also widely used in the literature, with particular focus on regression models. In \cite{A20}, the authors perform multivariate regression analyses, and they find that temperature and  rainfall have significant correlations with the cotton bollworm population. Similarly, in \cite{A14} they find temperature, wind speed and sunshine hours to be important. Furthermore, in \cite{A6} they perform binary classification for pest occurrence detection using relative humidity and temperature. Complementary to weather factors, some studies also use host plant phenology represented by NDVI or other vegetation indices \cite{A7}. This is done in order to study the association between the crop's growth and the pest population. Satellite derived vegetation indices are also used in \cite{A16}, where they introduce an ecology oriented model for population dynamics.\\ 

\noindent Motivated by real life requirements, we designed and implemented a predictive method to detect harmful cotton bollworm presence in cotton fields. Our model is interpretable, which means that one can understand how it arrived to a specific decision and which are the drivers that mattered the most. This allows the user of the model's outputs (e.g., farmers) to understand and trust them and therefore adhere to the recommendations, and even fine-tune them with their expert knowledge. Our approach makes use of both vegetation indices and weather data, and to the best of our knowledge there is no similar work for cotton bollworm presence estimation. We perform two experiments, i.e., one that makes use of past trap catches and one that does not, yielding satisfactory results for both. The latter experiment is of great significance as it allows for the method to be applicable anywhere in space, irrespective of the presence of pest traps in the vicinity.\\

\subsection{Methodology}
\begin{figure}[!ht]
  \centering
  \includegraphics[scale=0.7]{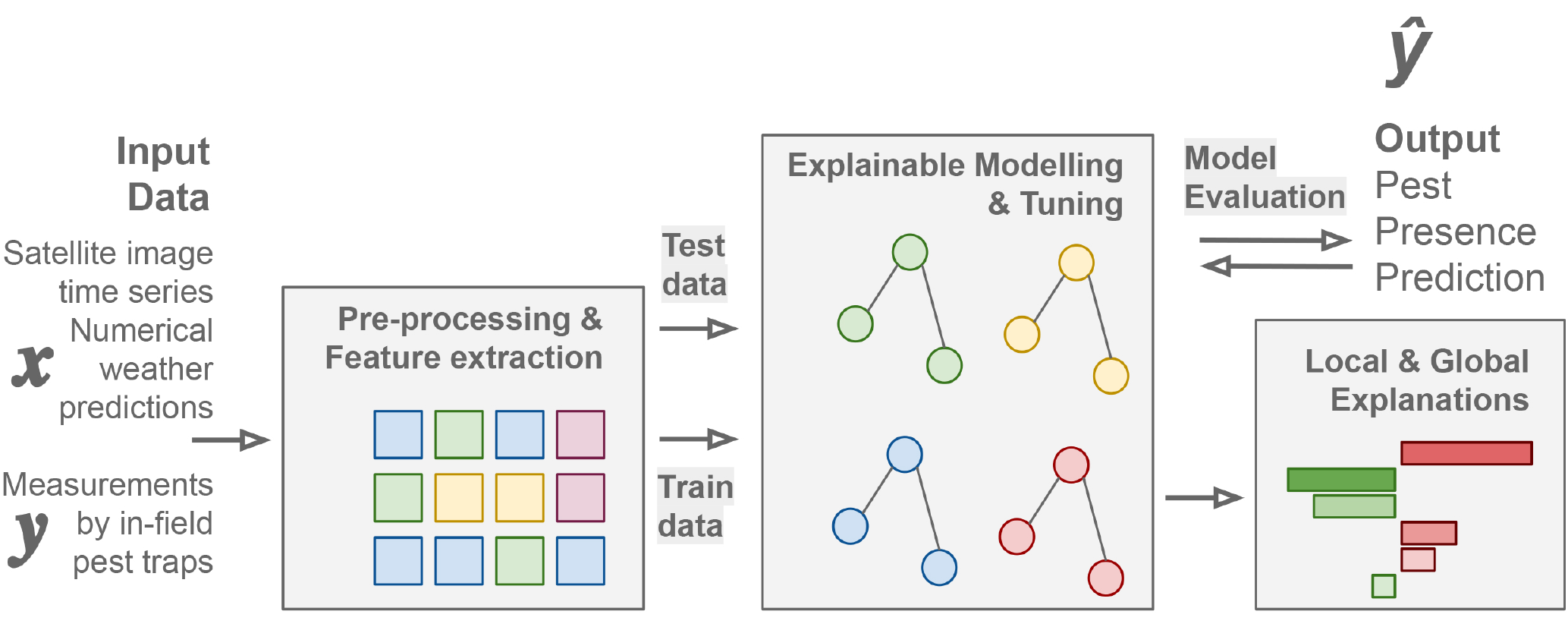}
  \caption{Workflow of the method for bollworm presence prediction. \label{fig:firstfigure5}}
  \vspace{-2em}
\end{figure}

\noindent Our work focuses on the prediction of cotton bollworm presence based on meteorological data, satellite EO and past trap catches. This section elaborates on the formulation of the problem, and the acquisition, pre-processing and engineering of the data to analysis-ready features. Fig. \ref{fig:firstfigure5} shows an overview of the steps followed. 
\vspace{-1em}
\\

\subsubsection{Problem formulation}

\noindent Let $T$ be the set of traps we have at our disposal. We note as $t_{ij}\in T$ the entity describing trap $i$ on date $j$. Every $t_{ij}$ can be expressed as
$t_{ij} = (x_{ij},y_{ij})$, where $x_{ij}\in{\rm I\!R^{d}}$ is a $d$ dimensional vector containing the corresponding weather and satellite features, and $y_{ij}\in\{0, 1\}$, with $0$ referring to pest absence and $1$ referring to pest presence. Our aim is to find a model $M:{\rm I\!R^{d}} \rightarrow \{0, 1\} $, so that for any trap $t_{ij}$, to predict the onset of harmful presence of bollworm population that will in turn signal the time for intervention. \\

% so that for any new $t_{ij}$, to be able to predict the lacking $y_{ij}$, given $x_{ij}$. 

\noindent \textbf{Explainable classifier approach:} There exist many sophisticated models that could be appropriate for our problem. However, we should bear in mind that in the agricultural sector, whoever be the user, it is common to have substantial empirical knowledge. For example, someone who cultivates cotton is likely to know when to sow, the crop's enemies, the favourable conditions etc. Therefore, it is crucial to make these users trust the model and its outputs \cite{inproceedings}. Furthermore, if the users have some insight on the main drivers and find the result unjustifiable, they can disregard the prediction or frame it in context and then take appropriate action. The above examples point out the need for model transparency, which can be achieved using explainable ML \cite{A17}. We therefore chose to proceed using an interpretable classifier. \\

%The model we used is the Explainable Boosting Machine (EBM), as implemented in the InterpretML framework. EBM is an augmented version of generalized additive models (GAMs) which are expressed by Eq.\ref{eq1}. In practise, EBM is a generalized additive model with interactions based on the (GA2M) method \cite{preprint,pairwise,accinter}. Eq.\ref{eq:2} formalizes this relationship. Every feature function $f_{i}$ is learnt using many shallow decision trees (up to 3 leaves). The learning is achieved using one predictive feature at a time over the train data, while performing gradient updates with a slow learning rate. The slow learning rate is advantageous in two ways. Firstly, it allows the features to be used independently of their order. Secondly, the one feature per time training reduces the impact of collinearity. \cite{preprint,pairwise,accinter} Interactions, being nothing but pairs of predictor variables, are learnt through the $f_{ij}$ functions. The FAST method \cite{pairwise} accounts for the selection of the subset of the interactions among all the available ones. One can consider the initial input features together with the interactions as the final feature space. Each of these features' logit score is calculated. It is then summed and passed to the link function $g$, which is chosen depending on the context. In binary classification, which was our case, EBM uses the logarithmic loss.

\noindent The model we used is the Explainable Boosting Machine (EBM), as implemented in the InterpretML framework. In ML, there is often a trade-off between accuracy and intelligibility. This is not true for EBM that achieves a performance comparable to powerful black-box models, while offering global and local explanations on the predictions \cite{preprint}. EBM is an augmented version of the Generalized Additive Models (GAMs), which are expressed by (\ref{eq1_5}).

% EBM is a tree-based, cyclic gradient boosting Generalized Additive Model (GAM, expressed by \ref{eq1}) with automatic interaction detection (GA2M, expressed by \ref{eq:2}). 

\begin{equation}
\label{eq1_5}
g(E(y)) = \beta_{0} + \sum{f_j(x_j)}
\end{equation}

\noindent In practice, EBM is a GAM model with interactions based on the GA2M method, expressed by (\ref{eq:2}) \cite{pairwise}.

\begin{equation} \label{eq:2}
g(E(y)) = \beta_{0} + \sum{f_j(x_j)} + \sum{f_{ij}(x_i,x_j)}  
\end{equation}

\noindent Every feature function $f_{j}$ is learnt using many shallow decision trees. The learning is achieved using one predictive feature at a time in round-robin pass over the train data, while performing gradient updates. In more detail, we train a tree on the first feature, then in boosting fashion we update the residual and move to the second feature, then we train another tree and so on. The learning rate is low, and thus the feature order does not matter. Interaction functions $f_{ij}$, are learnt via the FAST method \cite{pairwise}.\\

\noindent All features' contributions, together with the interactions, yield a logit score. Scores are then summed and passed to the link function $g$, which is chosen depending on the context. In binary classification, as in our case, EBM uses the logarithmic loss. Since the model is additive, the contributions of the terms can be sorted and visualized. This way, we can understand the main drivers of the model and produce new domain knowledge, but also enhance the trust on the outputs. 

\subsubsection{Data collection}

\noindent \textbf{Trap data:} To obtain insect population measurements, we collaborated with the company Corteva Agriscience Hellas that has a trap network for helicoverpa armigera. The network consists of pheromone traps in 26 different locations in the wider region of Central Macedonia, Greece. The traps are located in such a distance so as to not interact with each other and are examined (trapped insect counting) by a specialist every 3-5 days. The observations take place from June 2020 until early September 2020 and June 2021 until early September 2021. The trap data include i) the trap locations, ii) the number of pests on each trap and iii) the Day of Year (DoY) they were recorded. In total, there are 10 locations for 2021 and 16 locations for 2020. \\

\noindent \textbf{Meteorological data.}
In real life scenarios, exact weather conditions are very hard to obtain. This is true due to the absence of a dense network of weather stations. To overcome this, we use numerical weather predictions using our own configuration of WRF-ARW \cite{skamarock2019description, sitokonstantinou2021semi}. The model is of high spatial resolution (2 km), and predictions are made hourly. For each trap location, we obtained daily values for air (2 m) and soil temperature (0 m), relative humidity (RH), accumulated precipitation (AP), dew point (DP), and wind speed (WS). These parameters have been extensively used in related work. \\

\noindent \textbf{Satellite data.} 
We used Sentinel-2 images to capture the vegetation status (at the location of the traps) and its evolution through time \cite{A21}. Using atmospherically corrected Sentinel-2 images, we calculated a number of vegetation indices that highlight particular characteristics of the crops (e.g., moisture, physiological stress etc.) that are known to i) drive the occurrence of pests, but also ii) capture the vegetation cover changes caused by the insects \cite{segarra2020remote}. In detail, we used the normalized difference vegetation (NDVI), water (NDWI), moisture (NDMI), greenness (GI) and greenness chlorophyll (GCVI) indices. 

\subsubsection{Feature engineering} 
% In this section, we describe the pre-processing of the data and the feature engineering conducted. We also explain the parameters chosen for the model. All the data at our disposal, be it the trap, meteorological or satellite data are multi-temporal. In order to capture their time-dependent structure some pre-processing and feature engineering was conducted.
\noindent We provide pest presence predictions for each trap and for each date that we have catches. Each of those instances in time and space (trap/inspection day, 526 instances) is described by a series of features that are engineered using the data variables described earlier. Regarding the trap data, we use the coordinates of the trap locations, the corresponding dates they were visited and the number of insects observed per visit. In the feature space, we included the catches of the three latest visits prior to the day of prediction. The day of visit was encoded using the sine and cosine of the DoY (expressed in degrees), which is common for cyclical features. The number of catches for the day of prediction function as the labels for training and evaluation. The labels are binarized to pest presence or absence according to a certain threshold ($t=10$), which represents the number of insects above which the pest is considered harmful to the crop, also known as the action threshold. \\

\noindent The meteorological and vegetation index variables were engineered into accumulated features to capture the near-past (7 days) information that drives pest occurrence. Specifically, the accumulated vegetation indices were calculated using the cumulative integral of the time-series curve, whereas the accumulated weather features were calculated by summing the daily values over the last 7 days. Growing degree days, which capture the effective growth time of the plant,  were also calculated according to (\ref{eq:3}) \cite{A19}. $T_{base}$ refers to the temperature under which the cotton does not develop and is equal to 15.6$^{\circ}$
C, and $T_{max}$, $T_{min}$ are the maximum {\&} minimum air temperatures (2 m), respectively.
\begin{equation} \label{eq:3} GDD = \max\left({\frac{T_{max}+T_{min}}{2}-T_{base},0}\right) \end{equation}
The data was standardized using a standard scaler. The EBM was run using 100 inner bags, 100 outer bags, and a learning rate of 0.01 that according to \cite{preprint} is an appropriate tuning to achieve both high accuracy and interpretability. 
%\newpage

\subsection{Experiments \& Results} 

\noindent We used cumulative integrals of vegetation indices, accumulated and current weather data, and past trap catches to predict the occurrence of bollworms. We also ran experiments without including the insect catches in order to explore the capability of our model to predict pest presence anywhere in space, and not just for regions for which we have traps. Table \ref{tab:ms} depicts the accuracy and F1 score for both cases, averaged over 10 random train/test splits (70/30).\\
\vspace{-1em}
\begin{table}[!ht]
\caption{Performance of the pest prediction model for Case A with all features, and Case B without trap related features.}
\label{tab:ms}
\centering
\scalebox{1}{
\begin{tabular}{c|c|c|c|}
\cline{2-4}         & \textbf{Class} & \textbf{Case A} & \textbf{Case B} \\ \hline
\multicolumn{1}{|c|}{\textbf{Accuracy}} & -              & 0.75 ± 0.03     & 0.69 ± 0.02     \\ \hline
\multicolumn{1}{|c|}{\multirow{2}{*}{\textbf{F1 score}}} & Pest absence & 0.75 ± 0.04 & 0.68 ± 0.03 \\ \cline{2-4} 
\multicolumn{1}{|c|}{}                  & Pest presence  & 0.74 ± 0.04     & 0.69 ± 0.02     \\ \hline
\end{tabular}}
\end{table}
\vspace{-1em}

\noindent As expected, Case A considerably outperforms Case B. It should be noted, however, that Case B showcases promising performance given the model was able to predict pest presence using only EO and weather data. In the context of an operational pest management service, the Case B approach has great value, as it can be applicable to any cotton field without the need of being near a pest trap. \\

\noindent The pest presence threshold, set at 10 catches, amounts to approximately 3 insects a day, as traps are visited every 3-5 days. According to local agronomists that we consulted, a couple of catches a day are not considered harmful. Nevertheless, the threshold is only an approximation. Fig. \ref{fig:fourthfigure5} illustrates the seriousness of the errors of our model (Case A) by visualizing how close to the action threshold they occurred. The left histogram depicts the distribution of catches for the model errors. As expected, most errors are situated near the action threshold.\\ 

\begin{figure}[!ht]
\centering
\includegraphics[scale=0.7]{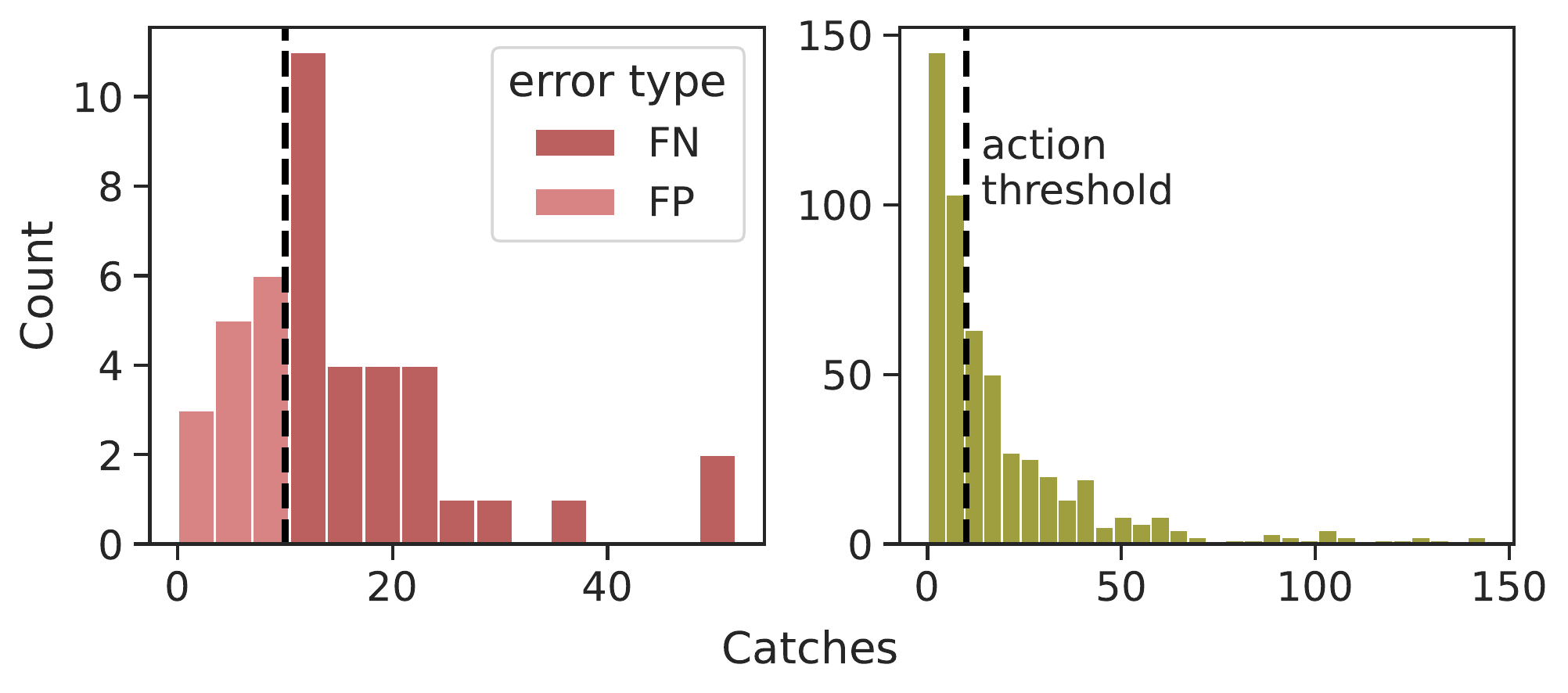} 
\caption{Distribution of the number of catches for all trap measurements (right) and for the model errors (left).\label{fig:secondfigure}}
\vspace{-1.4em}
\end{figure}

\begin{figure}[!ht]
\centering
\includegraphics[scale=0.75]{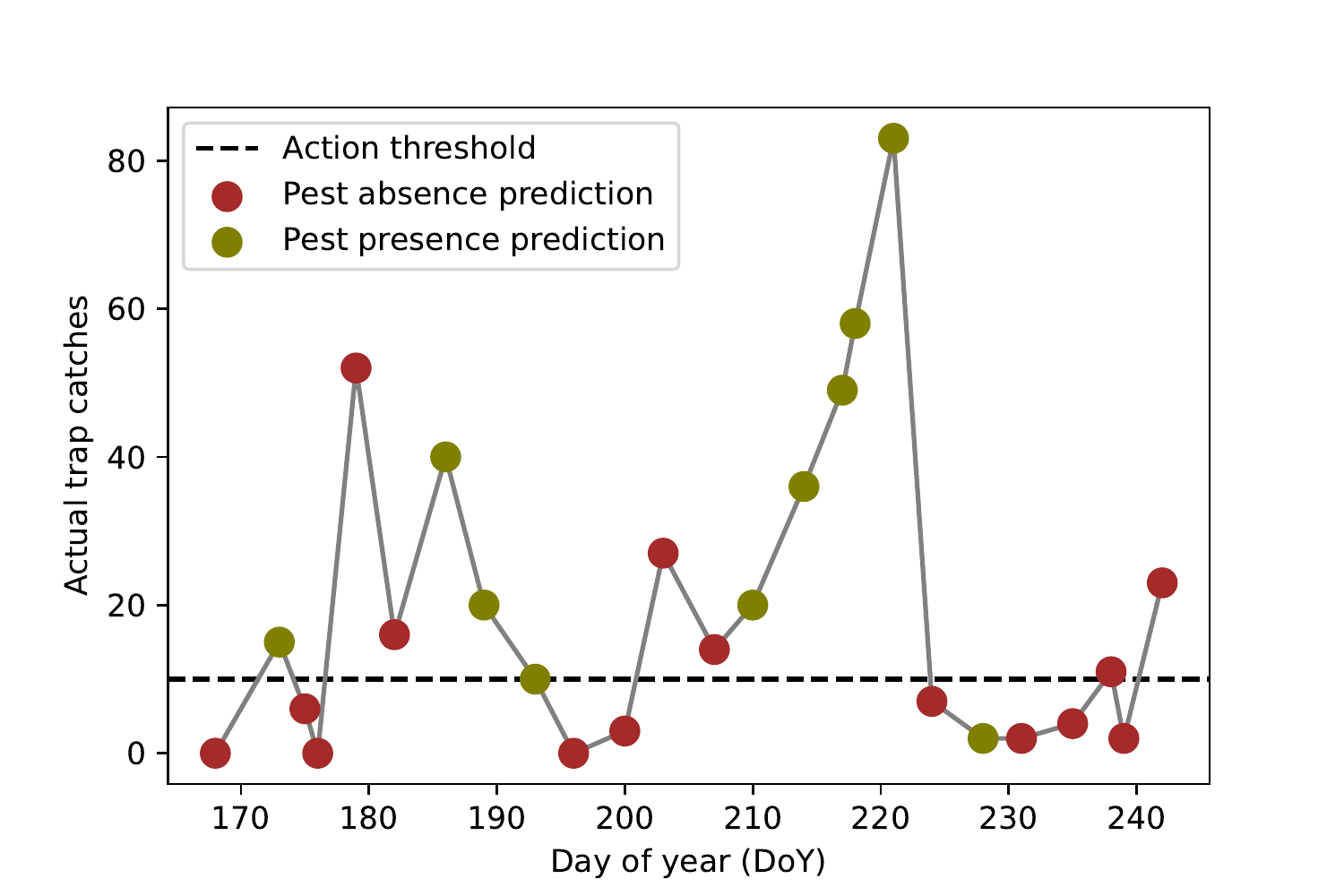} 
\caption{An indicative time-series of insect catches from a single trap in Macedonia, Greece. The green dots are the pest presence predictions and the red dots are the pest absence predictions. The dashed line represents the action threshold set at $t=10$. }\label{fig:thirdfigure5}
\end{figure}

\begin{figure}[!ht]
\centering
\includegraphics[scale=0.6]{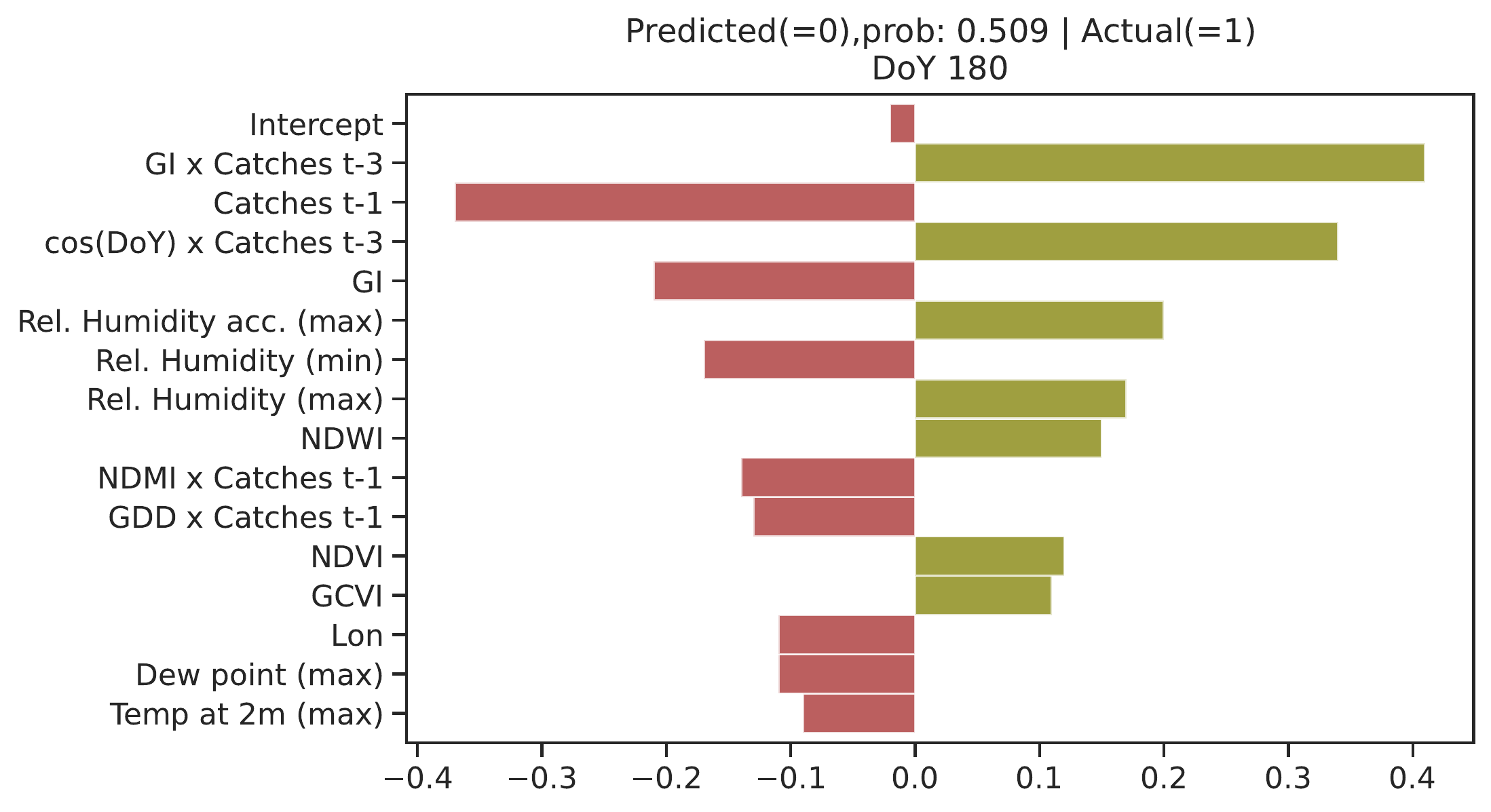} 
\caption{Local explainability plot for an uncertain prediction.}\label{fig:fourthfigure5}
\vspace{-1em}
\end{figure}

\noindent Using the InterpretML framework, we extracted local and global explainability reports. Global explanations account for all model predictions, whereas local explanations are provided for each individual prediction. Table \ref{tab:global_exp_table} shows the global explanations for Cases A and B. The scores indicate how much each feature affected the result. For Case A, the features with the strongest impact were the past trap catches. This is expected since pest presence depends on the population of the previous days. For Case B, it is mostly the weather features, both current and accumulated, as well as their interactions (acting as a new parameter) that compensate for the lack of trap information. Wind speed, relative humidity, precipitation, air, and soil temperature appear to be very important features for both cases. This is also supported by the literature \cite{A20,A14,A16}. Time features, such as cos(DoY) and sin(DoY), are also important since there is clear seasonality in bollworm population (see Fig. \ref{fig:thirdfigure5}). Finally, the vegetation indices, and interactions of them with weather features for Case B, are also found among best features. This supports the choice of combining satellite EO with weather data, which has not been done before for bollworm presence estimation.\\

\vspace{-0.9em}
\begin{table}[!ht]
\caption{Feature importance according to EBM's global explainability report for Case A with all features and Case B without trap related data.}
\label{tab:global_exp_table}
\centering
\scalebox{1}{
\begin{tabular}{|cc|cc|}
\hline
\multicolumn{2}{|c|}{\textbf{Case A}}      & \multicolumn{2}{c|}{\textbf{Case B}}                     \\ \hline
\multicolumn{1}{|c|}{\textbf{Features}} & \textbf{Scores} & \multicolumn{1}{c|}{\textbf{Features}}              & \textbf{Scores} \\ \hline
\multicolumn{1}{|c|}{Catches t-1}   & 0.29 & \multicolumn{1}{c|}{WS (min) - AP acc.}           & 0.13 \\ \hline
\multicolumn{1}{|c|}{Catches t-2}       & 0.16            & \multicolumn{1}{c|}{WS (max) - T at 0 m (max) acc.} & 0.12            \\ \hline
\multicolumn{1}{|c|}{T at 2 m (min)}    & 0.15            & \multicolumn{1}{c|}{RH (min) acc. - GDD}            & 0.12            \\ \hline
\multicolumn{1}{|c|}{cos(DoY)}      & 0.13 & \multicolumn{1}{c|}{cos(DoY)}                     & 0.11 \\ \hline
\multicolumn{1}{|c|}{GI}            & 0.13 & \multicolumn{1}{c|}{T at 0 m (max) acc. - cos(DoY)} & 0.11 \\ \hline
\multicolumn{1}{|c|}{RH (max)}      & 0.13 & \multicolumn{1}{c|}{WS (max)}                     & 0.10 \\ \hline
\multicolumn{1}{|c|}{RH acc. (max)} & 0.12 & \multicolumn{1}{c|}{NDMI - WS (min)}              & 0.09 \\ \hline
\multicolumn{1}{|c|}{NDWI}          & 0.11 & \multicolumn{1}{c|}{GCVI}                         & 0.09 \\ \hline
\multicolumn{1}{|c|}{Lon}           & 0.11 & \multicolumn{1}{c|}{NDVI - NDMI}                  & 0.09 \\ \hline
\multicolumn{1}{|c|}{GCVI}          & 0.10 & \multicolumn{1}{c|}{GCVI - sin(DoY)}              & 0.09 \\ \hline
\end{tabular}}
\end{table}
%\vspace{-1em}

\noindent The aforementioned experiments, used train and test data randomly selected from all the available traps. Nevertheless, it is also important to observe the temporal pattern of our predictions. The idea is to identify when the presence is significant, indicating the population peak over the next days and thus the need to intervene. To do so, we needed to see how the model would perform when tested on the entire time-series of a trap. Specifically, instead of training our model (Case A) using samples from  all the traps, we would leave out one trap at a time and use it as a test set. Fig.\ref{fig:thirdfigure5} shows the actual trap catches against time, together with the action threshold that indicates whether a point should have been classified as pest presence or not. This is an indicative example for a single trap located in Central Macedonia, Greece. The green dots represent pest presence predictions, the red dots represent pest absence predictions, and the dashed line represents the action threshold ($t=10$). We can conclude that the model performs rather well for this particular trap. The principal troughs and peaks are clearly identified, indicating that the model is able to predict the onset and termination of pest harmfulness. It is also worth noting how the model was able to identify the end of the bollworm population cycle after DoY 220. \\

\vspace{-1em}

\noindent To further investigate this test trap, we used the local explanations of the InterpretML framework. We indicatively examine the prediction for DoY=180. This one, despite being a peak, was not detected as such. Looking at the local explainability graph in Fig. \ref{fig:fourthfigure5}, we can gain insight on this decision. The coloured bars indicate which features drove the decision towards pest presence (green) and which features towards pest absence (red). The predicted class is assigned with a probability of only 0.509, indicating uncertainty. The feature which had the strongest impact towards the true label was the interaction of GI with Catches t-3. This is very logical since the catches at the instance t-3 indicated pest presence. On the other hand, the catches at the instance t-1 strongly suggest pest absence. By examining Fig. \ref{fig:thirdfigure5}, we can see that the pest catches in the previous visit were zero and hence unlikely to have strong pest presence only four days later. \\

\subsection{Conclusions}

\noindent In this work, we used an interpretable classifier to detect significant cotton bollworm presence to indicate the onset of pest harmfulness. This is the first study that uses both weather data and EO to model the occurrence of this pest. We ran two experiments: i) one that incorporates past trap catches in the feature space and ii) one that uses only meteorological parameters and vegetation indices. Both approaches yielded satisfactory results, with model (i) outperforming model (ii). Model (ii), however, is of particular interest in the context of pest management, as it can be applicable anywhere in space without being dependent on the presence of pheromone (or other) traps. Moreover, by using the InterpretML framework, we can understand and trust the model's decisions. Not only we predict the onset of the rise in pest population, but we also provide significant insight on the main drivers of the model's outcomes. This makes our approach transparent and easy to understand. \\

% conference works do not normally have an appendix

\section{Conclusions and Future Work}

In this chapter, I conclude by summarizing the main achievements and contributions of this thesis and how they addressed the objectives. Furthermore, I perform a critical evaluation of the outputs in the grand scheme of things and provide future perspectives that are deemed as natural continuation of the presented work. \\

\noindent Chapter 2 presented methods for handling and extracting knowledge from big Earth data and ancillary remote sensing datasets (i.e., street-level images) towards simplifying and improving the monitoring of the CAP. Specifically, a data cube-based framework was developed for the production of harmonized, analysis-ready, Sentinel-1 and Sentinel-2 time-series over large areas. Additionally,  a multi-crop classification ML pipeline that exploits the farmers declarations for training was developed. This allowed to use labels that are widely available and thus resolve the ground-truth scarcity issue, while managing the  noise stemming from self-declaration mistakes by the farmers. \\

\noindent Furthermore, a semantic enrichment method was developed to enhance the knowledge extracted for the crop type maps towards addressing the operational needs of CAP inspectors. Using the crop type maps and the semantic enrichment method,  a framework for the smart sampling of the on-the-spot checks was developed. Therefore, the inspectors do not visit randomly sampled parcels for checks, but rather the ones that are most probable wrong declarations. Finally, satellite data cannot provide 100\% certainty for decision-making. For this reason, ancillary crowdsourced data, such as street-level images were exploited as high resolution complementary source. In this context, a Sentinel and street-level analysis-ready dataset was developed and the value of fusing data from heterogeneous sources was discussed. Additionally, preliminary results for a late fusion technique were provided. \\

\noindent Future work in crop classification includes fusion of Sentinel-1 and Sentinel-2 data (e.g., RNNs or GANs to generate Sentinel-2 images from Sentinel-1 images) for the generation of dense image time-series for cases that suffer from extended cloud coverage. Additionally, pixel-based approach or patch-based approaches should be considered in cases where parcels are small. Recently, DL models have proved to enhance the performance for such cases significantly; particularly transformers and Conv. LSTMs. The proposed semantic enrichment framework can also be extended to similar problems, beyond the control of the CAP, using semantic enrichment and reasoning to support farmers in monitoring their crops, insurance companies to assess the risk in a specific area of interest, and public agencies that monitor the sustainability of rural areas as a consequence of climate change. \\

\noindent Regarding the space to ground data availability,  a proof of concept was provided on how to combine satellite and street-level images. In the future, UAV and VHR satellite data will be included and fusion techniques among the different modalities will be explored towards providing near perfect crop classification. Currently, the dataset of street-level images is confined and thus the crop classification results suboptimal. In the near future, extensive campaigns of collecting street-level images in Cyprus will be concluded, allowing to generate a huge benchmark dataset of more than 1 million street-level images and that will in turn enable the experimentation with various DL models.\\

\noindent Chapter 3 dealt with the satellite-based monitoring of food security, with particular focus on the challenges imposed by large-scale applications, i.e., geographic generalization, transferability and efficient computing. In this context, a weakly-supervised rice classification method was developed, which by using only a confined number of labels from a single region can scale to an entire country. Furthermore, a distributed data storage and processing framework was developed for the large-scale classification of rice in a high performance computer.\\

\noindent Based on the conclusions of Section \ref{scalablerice}, it could be argued that for binary classification problems, which share similar characteristics (i.e., distinct land cover class, spatially continuous and spatially abundant), the proposed pseudo-labeling approach can be applicable. In fact, pseudo-labeling constitutes the overall methodology transferable. The three study sites of different agro-climatic conditions, geomorphological characteristics and land cover abundances, yielded excellent and comparable results; showcasing the robustness of the method. The idea of pseudo-labeling for crop classification could be even extended to multi-class scenarios. Multi-crop classification problems could be reduced to multiple OnevsAll binary classifiers, such as the one implemented in this work.\\

\noindent Chapter 4 discussed issues related with the resilient farming of cotton and presented developments for the monitoring of phenology, which if provided at high temporal and spatial resolution can assist the within-season farm intervention towards reduced costs and increased yields. In this direction, a semi-supervised crop phenology estimation method was developed that uses merely a handful of labels. The method introduced a continuous phenology scale and addressed phenology estimation as a regression problem. This provided enhanced specification on the growth status of the crop, unlike the more common classification methods that are able to detect merely the principal phenological stages. Additionally, an unsupervised fuzzy clustering method for phenology estimation was implemented. The method does not use ground truth that makes it truly useful in real-world scenarios. Additionally, using the fuzziness scores one can identify the transitional periods between two phenological stages, increasing the specificity of the phenological stage prediction.\\

\noindent In future work, further testing will be conducted for the evaluation of the spatial and temporal generalization of the clustering method. This will require additional ground observations at different areas and years of inspection. There is a number of recent studies that look into DL based unsupervised change detection on SITS \citep{kalinicheva2020unsupervised, kalinicheva2018neural, kondmann2021spatial, andresini2021leveraging}. There is great potential in such approaches and it is argued that they could be applicable in the proposed unsupervised premise for phenology estimation. Common denominator of these methods is the learning of a smaller latent or embedding space, in which entities that bear resemblance are located closer to each other. This is particularly important for clustering techniques that aim to group similar samples in the hyperspace. Usually, clustering algorithms, such as FCM, measure this similarity among entities using pair-wise distances. It is known that high dimensional spaces are not ideal for distance based techniques, as they usually fail to capture meaningful clusters. In addition, a latent manifold representation is not greatly dependent on feature engineering and can generalize well. \\

\noindent Chapter 5 discussed how we shift from knowledge to wisdom. How do we move from detection towards decision making and actionable advice? The literature clearly highlights the importance of causal inference and interpretable or explainable machine learning. The chapter includes two works with preliminary results that showcase the potential of such approaches towards: i) policy making using causal machine learning for assessing the impact of agricultural practices in climate regulation and ii) rapid adoption of recommendations by the farmers through robust and trustworthy predictions, using interpretable machine learning to indicate upcoming pest abundance that indicates the need for spraying. \\

\noindent Future work for (i) includes incorporating climatic projections in the pipeline and comparing the impact learned from them to the impact learned from historical data. Further work also comprises the study of more agricultural practices, e.g., conservation tillage and minimum soil cover, but also the study of more target outcomes, such as erosion prevention and pollination potential. This way, we can move towards a synthesis of ecosystem service trade-offs and the formulation of a comprehensive suite of policies in response to climate change. Finally, the correlations between agricultural practices and NPP, and consequently the treatment impact, depend on the spatial scale of analysis \citep{zhou2021spatial}. Therefore, agricultural practices should be evaluated at different scales to gain insights towards optimizing agriculture land suitability from farm to landscape level and achieve optimum value.\\

\noindent Future work for (ii) includes forecasting pest presence for the next two to five days using numerical weather predictions and projections of vegetation indices. Moreover, using a larger trap dataset we can enhance the performance of the model and thus expand inference spatially, in areas where no traps are set, resulting in pest presence-absence maps.

% ********************************** Back Matter *******************************
% Backmatter should be commented out, if you are using appendices after References
%\backmatter

% ********************************** Bibliography ******************************
\begin{spacing}{0.9}

% To use the conventional natbib style referencing
% Bibliography style previews: http://nodonn.tipido.net/bibstyle.php
% Reference styles: http://sites.stat.psu.edu/~surajit/present/bib.htm

% \bibliographystyle{apalike}
\bibliographystyle{unsrt} % Use for unsorted references  
\cleardoublepage
\bibliography{References/references} % Path to your References.bib file

% If you would like to use BibLaTeX for your references, pass `custombib' as
% an option in the document class. The location of 'reference.bib' should be
% specified in the preamble.tex file in the custombib section.
% Comment out the lines related to natbib above and uncomment the following line.

%\printbibliography[heading=bibintoc, title={References}]

\end{spacing}

% ********************************** Appendices ********************************

\begin{appendices} % Using appendices environment for more functionality

\appendix

\chapter{Monitoring of the CAP}\label{appendix1}

\section{Crop Type Mapping}

\subsection{Crop type classification using a dense Sentinel-2 image time-series
}

\begin{figure}[!ht]
    \centering
    \includegraphics[width=\linewidth]{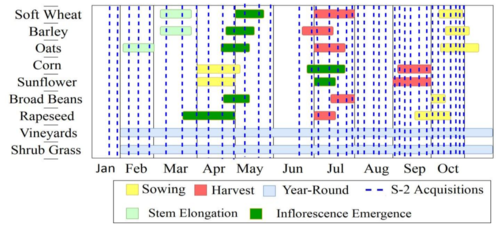}
    \caption{Timeline of phenology stages for key crops in Navarra, overlaid with the acquisition dates of Sentinel-2 images for 2017.}
    \label{fig:phenology_stages_Navarra}
\end{figure}

\noindent In order to assess the importance and impact of a very dense time-series of Sentinel-2 imagery in crop type classification accuracy, we have performed an SVM (quadratic kernel) classification scheme that incorporates almost all available imagery within the year of inspection, irrespective of the cloud percentage. Thus, for the study site in Navarra, Spain we have acquired, then pre-processed and used 39 Sentinel-2 images (January to October – Figure \ref{fig:phenology_stages_Navarra}) to classify crop types based on the 2017 farmer declarations, as part of their Common Agricultural Policy (CAP) subsidy applications

\subsubsection{Methodology}

\noindent The dataset comprises all available Sentinel-2 imagery (including acquisitions from both S2A and S2B) and thus substantial cloud coverage is evident throughout the time-series. Therefore, cloud covered pixels within the time-series are first masked out and then missing values are filled in.\\

\begin{figure}[!ht]
    \centering
    \includegraphics[width=\linewidth]{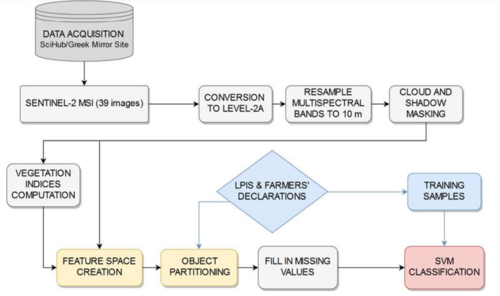}
    \caption{Flowchart of overall methodology: imagery acquisition, data pre-processing, cloud masking, feature space creation, fill in missing values, crop type classification}
    \label{fig:methodology_navarra}
\end{figure}

\noindent Figure \ref{fig:methodology_navarra} displays the processing workflow of the crop type identification scheme, from the acquisition of Sentinel-2 data to the SVM classification. The feature space, on which the supervised classification algorithm is applied to, comprises the time-series of Sentinel-2 imagery and the vegetation indices Normalized Difference Vegetation Index (NDVI), Normalized Difference Water Index (NDWI) and Plant Senescence Reflectance Index (PSRI), computed for each image acquisition (Table \ref{table:features_navarra}).\\

\begin{table}[!ht]
\caption{Sentinel-2 MSI feature space of multispectral band reflectances and vegetation indices}
\centering
\begin{tabular}{p{0.35\linewidth}|p{0.35\linewidth}}
 \hline
 Type & Sentinel-2 MSI \\ 
 \hline
Reflectances & 10 Bands [all except B09 and B10] x
39 image acquisition (390 features) \\
Vegetation Indices  & NDVI, PSRI, NDWI x 39 image acquisitions
(117 features) \\

\hline
\end{tabular}

\label{table:features_navarra}
\end{table}

\noindent In order to mask out the shadowed and clouded pixels of each image, we used the Level-2A scene classification product (SC)1 that is offered as a byproduct of the Sen2Cor atmospheric and terrain correction process. Each mask accounts for shadows, clouds of medium and high probability, cirrus clouds and snow (classes 3, 8, 9, 10, 11 of the SC product). Then for each image in the time-series, the pixels classified to the aforementioned categories are given “no data” values.\\

\begin{figure}[!ht]
    \centering
    \includegraphics[width=\linewidth]{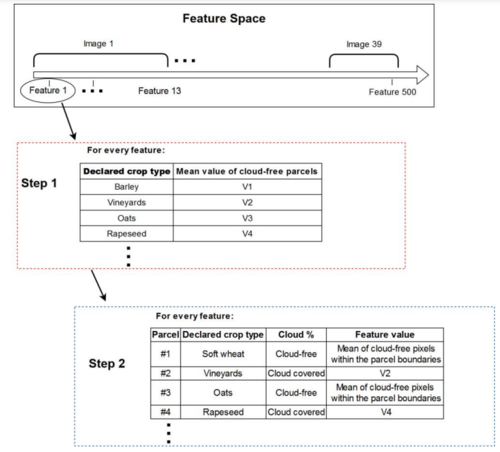}
    \caption{Filling missing values of “cloud covered” parcels. The average value for each crop type class is calculated from the corresponding “cloud-free” parcels for each feature. Then “cloud-covered” parcel values, for each feature, are substituted with the respective average class value, based on their declared crop type.}
    \label{fig:missing_cloud_values_navarra}
\end{figure}

\noindent Prior to the application of the classifier, the feature space is transformed from the pixel level to the object level by averaging the pixel values within the boundaries of each Land Parcel Identification System (LPIS) polygon. During this object partitioning process, the parcels that have more than 50\% of their pixels masked out are marked as “cloud covered”, while all else are marked as “cloud-free”. The feature values of the “cloud-free” parcels are computed using only the unmasked pixels of the parcel. On the other end, utilizing the a priori knowledge of the farmers’ declarations, and by assuming their validity, the “no data” feature values of the “cloud covered” parcels are filled in, as described in the following two steps (Figure \ref{fig:missing_cloud_values_navarra}).\\

\begin{enumerate}
    \item For each Sentinel-2 acquisition used in the time-series stack, calculate the average value of the 9 crop type classes for each feature (spectral bands and vegetation indices). Only “cloud-free”
    \item For each feature and for each “cloud covered” parcel, the “no data” values are replaced with the average remote sensing value, as calculated in step 1, considering their declared crop type class parcels partake in the computation.
\end{enumerate}

\noindent Images of more than 80\% cloud coverage, over the area of interest, bypass the process of
masking and missing value filling and are directly incorporated into the feature space after
resampling (see Figure \ref{fig:methodology_navarra}). This is done to avoid having features, constituted by parcel objects of forced values that have been computed based on a limited number of “cloud-free” instances.\\

\noindent In order to evaluate the influence of cloud coverage in crop type classification, the 39 Sentinel-2 images were then split into “clouded” and “cloudless” classes, with 19 and 20 images respectively. The division into the two classes was performed by calculating the number of all masked pixels falling within area occupied by the parcels. Images with cloud coverage less than 3\% are marked as
“cloudless”, while the “cloudy” class contains all remaining imagery. Prior to the crop type
classification, features are ranked based on their importance that is computed using the ReliefF
algorithm in MATLAB

\subsubsection{Results}
Figure \ref{fig:cloud_and_cloudless} below displays the contribution of the “clouded” and “cloudless” image classes, in
number of features, for different quintiles of the ranked feature space.

\noindent The analysis of feature importance clearly shows that “cloudless” images are the most
important in the classification process. Sorting the feature space according to the individual feature
importance weights revealed that the 85 top ranked features came exclusively from the “cloudless”
class. Utilizing solely these top 85 features resulted in a crop type classification of an overall Cohen’s
kappa 90.85\%. The features come only from images that were sensed in March (30\%), April (30\%),
June (18\%) and July (22\%). Imagery that was sensed in May is absent, as all three acquisitions within
the month had cloud coverage larger than 62\%.

\begin{figure}[!ht]
    \centering
    \includegraphics[width=\linewidth]{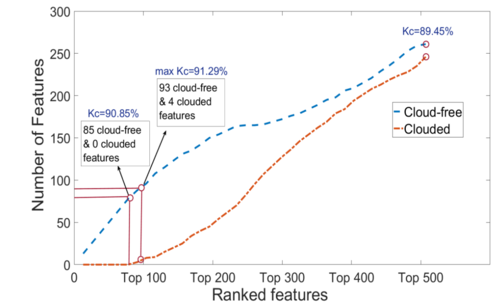}
    \caption{Contribution of “clouded” and “cloudless” images in the top ranked features of the variable space.}
    \label{fig:cloud_and_cloudless}
\end{figure}

\noindent The highest accuracy, Cohen’s kappa of 91.29\%, is achieved when classifying with the first 97
features of the ranked feature space; utilizing 93 cloud-free and 4 clouded features. On the other
hand, using the entirety of the feature space (507 features) resulted in 89.45\% accuracy. It becomes
evident that images coming exclusively from the “cloudless” class are responsible for a near
maximum overall accuracy, and therefore the incorporation “clouded” imagery is considered
redundant.\\

\noindent It can be argued that this approach does not justify the extra processing effort that is demanded,
for merely a marginal increase in thematic accuracy; particularly for operational applications, where
computational efficiency is an important consideration. It can be also observed that the top ranked
features stem from imagery that was sensed over several different months, thus covering the better
part of the phenology timeline, as shown in Figure S1. All in all, cloud-free images that cover critical
phenological stages make up the optimal feature space, for both accurate and computationally
efficient crop type classification.
\chapter{Food Security Monitoring}\label{appendix3} 

\section{Scalable Paddy Rice Mapping}

\begin{figure}[ht]
\begin{center}
		\includegraphics[width=15.5 cm]{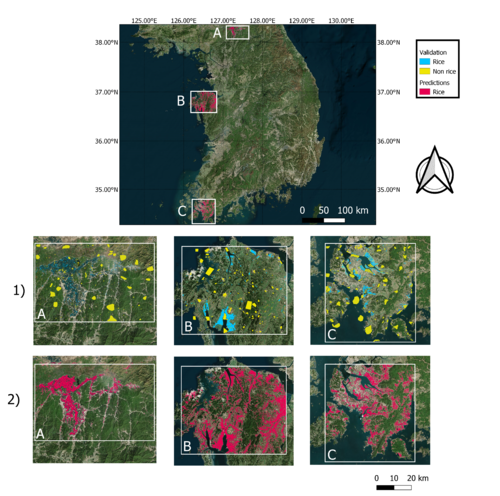}
	\caption{1) Labeled maps of Cheorwon (A), Seosan-Dangjin (B) and Haenam (C) produced with photo-interpretation (Section 3.2). Blue color represents the rice class, while yellow the non rice class. 2) Rice maps produced by applying the optimized Random Forest model trained on the Seosan-Dangjin site}
	\label{fig:aois}
\end{center}
\end{figure}

\begin{figure}[ht]
\begin{center}
		\includegraphics[width=15.5 cm]{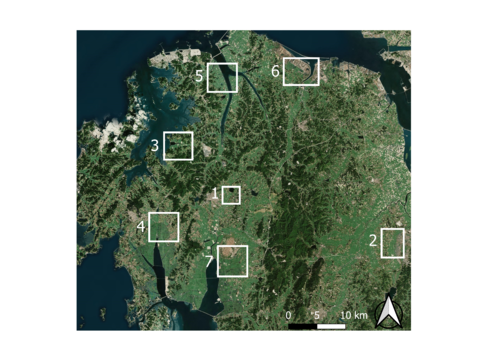}
	\caption{Locations of the Seosan-Dangjin labeled subsets used to evaluate k-means clustering experiments in Table 4 - Site IDs.}
	\label{fig:kmenas_areas}
\end{center}
\end{figure}

\begin{figure}[ht]
\begin{center}
		\includegraphics[width=15.5 cm]{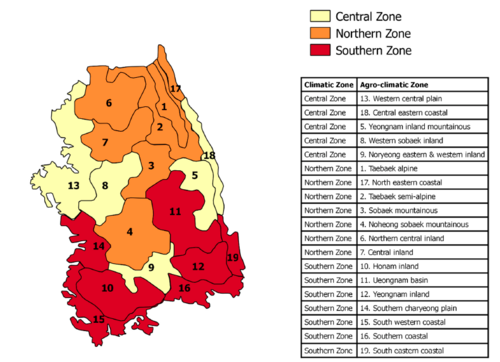}
	\caption{Agroclimatic zones of South Korea }\label{appendix_zones}
\end{center}
\end{figure}

\begin{figure}[ht]
\begin{center}
		\includegraphics[width=8 cm]{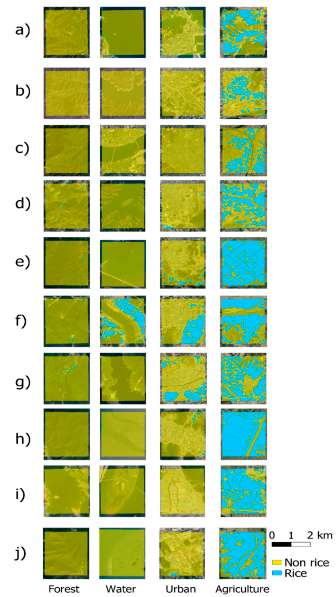}
	\caption{Paddy rice maps produced by applying the optimized RF model on 40 test plots across
the country. The locations of the sites a)–j) can be seen in Figure 7.}\label{fig:predicted_paddy_rice_maps}
\end{center}
\end{figure}

\chapter{Resilient and smart farming}\label{appendix2}

\section{Unsupervised phenology estimation}

\begin{table*}[!ht]
\resizebox{\textwidth}{!}{\begin{tabular}{|c|c|c|c|c|c|c|c|c|c|c|c|c|c|}
\hline
\textbf{feature1} & \textbf{feature2} & \textbf{feature3} & \textbf{feature4} & \textbf{feature5} & \textbf{feature6} & \textbf{feature7} & \textbf{feature8} & \textbf{feature9} & \textbf{maxdiff-0} & \textbf{maxdiff-1} & \textbf{maxdiff-2} & \textbf{kappa} & \textbf{wkappa} \\ \hline
cos\_doy          & sin\_doy          & savi              & wdrvi(I)          & ndwi(I)           & gdd               & max\_soil         & max\_surf         & -                 & 0.5813             & 0.864              & 0.9947             & 0.5323         & 0.8884          \\ \hline
cos\_doy          & sin\_doy          & gvmi              & wdrvi(I)          & ndwi(I)           & gdd               & max\_soil         & max\_surf         & -                 & 0.5653             & 0.8773             & 0.9947             & 0.5148         & 0.8883          \\ \hline
cos\_doy          & sin\_doy          & evi               & savi              & wdrvi(I)          & psri(I)           & gdd               & max\_surf         & -                 & 0.552              & 0.8693             & 0.9973             & 0.504          & 0.8846          \\ \hline
cos\_doy          & sin\_doy          & gvmi              & savi              & wdrvi(I)          & psri(I)           & ndwi(I)           & gdd               & max\_soil         & 0.5547             & 0.888              & 0.9893             & 0.5031         & 0.8875          \\ \hline
cos\_doy          & sin\_doy          & gvmi              & evi               & wdrvi(I)          & gdd               & max\_soil         & max\_surf         & -                 & 0.5493             & 0.8747             & 0.9973             & 0.4994         & 0.8843          \\ \hline
cos\_doy          & sin\_doy          & evi               & savi              & wdrvi(I)          & gdd               & max\_soil         & max\_surf         & -                 & 0.5493             & 0.8667             & 0.9973             & 0.4975         & 0.8836          \\ \hline
cos\_doy          & sin\_doy          & evi               & savi              & wdrvi(I)          & psri(I)           & ndwi(I)           & gdd               & max\_surf         & 0.5467             & 0.872              & 0.9787             & 0.4934         & 0.8799          \\ \hline
cos\_doy          & sin\_doy          & gvmi              & savi              & wdrvi(I)          & psri(I)           & gdd               & max\_surf         & -                 & 0.5413             & 0.88               & 0.984              & 0.4929         & 0.8825          \\ \hline
cos\_doy          & sin\_doy          & gvmi              & wdrvi(I)          & psri(I)           & gdd               & max\_soil         & max\_surf         & -                 & 0.544              & 0.8667             & 0.9973             & 0.4912         & 0.8816          \\ \hline
cos\_doy          & sin\_doy          & evi               & savi              & psri(I)           & ndwi(I)           & gdd               & max\_soil         & -                 & 0.544              & 0.856              & 0.992              & 0.491          & 0.8782          \\ \hline
cos\_doy          & sin\_doy          & evi               & savi              & ndwi(I)           & gdd               & max\_soil         & max\_surf         & -                 & 0.5387             & 0.8907             & 0.9973             & 0.4887         & 0.8861          \\ \hline
cos\_doy          & sin\_doy          & gvmi              & evi               & savi              & wdrvi(I)          & gdd               & max\_soil         & -                 & 0.5413             & 0.88               & 0.984              & 0.4883         & 0.8822          \\ \hline
cos\_doy          & sin\_doy          & gvmi              & wdrvi(I)          & psri(I)           & ndwi(I)           & gdd               & max\_soil         & max\_surf         & 0.5413             & 0.864              & 0.984              & 0.4879         & 0.8777          \\ \hline
cos\_doy          & sin\_doy          & evi               & savi              & psri(I)           & ndwi(I)           & gdd               & max\_surf         & -                 & 0.5413             & 0.8747             & 0.9867             & 0.4872         & 0.8807          \\ \hline
cos\_doy          & sin\_doy          & evi               & wdrvi(I)          & ndwi(I)           & gdd               & max\_soil         & max\_surf         & -                 & 0.5387             & 0.8293             & 0.9787             & 0.4867         & 0.8699          \\ \hline
cos\_doy          & sin\_doy          & gvmi              & evi               & wdrvi(I)          & ndwi(I)           & gdd               & max\_soil         & max\_surf         & 0.536              & 0.88               & 0.984              & 0.4863         & 0.8805          \\ \hline
cos\_doy          & sin\_doy          & gvmi              & evi               & wdrvi(I)          & psri(I)           & gdd               & max\_soil         & -                 & 0.536              & 0.856              & 0.9867             & 0.4858         & 0.8758          \\ \hline
cos\_doy          & sin\_doy          & evi               & savi              & wdrvi(I)          & ndwi(I)           & gdd               & max\_soil         & -                 & 0.536              & 0.8427             & 0.9813             & 0.4848         & 0.8722          \\ \hline
cos\_doy          & sin\_doy          & evi               & savi              & wdrvi(I)          & psri(I)           & gdd               & max\_soil         & -                 & 0.536              & 0.8667             & 0.9947             & 0.4844         & 0.8803          \\ \hline
cos\_doy          & sin\_doy          & evi               & savi              & wdrvi(I)          & ndwi(I)           & gdd               & max\_soil         & max\_surf         & 0.5333             & 0.904              & 1.0                & 0.4832         & 0.8884          \\ \hline
cos\_doy          & sin\_doy          & evi               & savi              & wdrvi(I)          & psri(I)           & ndwi(I)           & gdd               & max\_soil         & 0.5333             & 0.8613             & 0.9787             & 0.4827         & 0.8753          \\ \hline
cos\_doy          & sin\_doy          & gvmi              & evi               & ndwi(I)           & gdd               & max\_soil         & max\_surf         & -                 & 0.536              & 0.88               & 0.9973             & 0.482          & 0.8831          \\ \hline
cos\_doy          & sin\_doy          & gvmi              & evi               & wdrvi(I)          & ndwi(I)           & gdd               & max\_surf         & -                 & 0.5333             & 0.872              & 0.9947             & 0.4815         & 0.88            \\ \hline
cos\_doy          & sin\_doy          & gvmi              & evi               & savi              & wdrvi(I)          & psri(I)           & gdd               & max\_soil         & 0.5307             & 0.9173             & 0.9893             & 0.4784         & 0.889           \\ \hline
cos\_doy          & sin\_doy          & evi               & savi              & psri(I)           & ndwi(I)           & gdd               & max\_soil         & max\_surf         & 0.528              & 0.896              & 0.9947             & 0.477          & 0.8844          \\ \hline
cos\_doy          & sin\_doy          & savi              & wdrvi(I)          & psri(I)           & ndwi(I)           & gdd               & max\_soil         & max\_surf         & 0.528              & 0.8693             & 0.992              & 0.4755         & 0.8785          \\ \hline
cos\_doy          & sin\_doy          & gvmi              & evi               & psri(I)           & ndwi(I)           & gdd               & max\_soil         & -                 & 0.5253             & 0.8613             & 1.0                & 0.4753         & 0.8777          \\ \hline
cos\_doy          & sin\_doy          & savi              & wdrvi(I)          & psri(I)           & ndwi(I)           & gdd               & max\_soil         & -                 & 0.5253             & 0.872              & 0.9867             & 0.4752         & 0.8774          \\ \hline
cos\_doy          & sin\_doy          & gvmi              & evi               & wdrvi(I)          & psri(I)           & gdd               & max\_soil         & max\_surf         & 0.528              & 0.8827             & 0.9947             & 0.475          & 0.8806          \\ \hline
cos\_doy          & sin\_doy          & gvmi              & savi              & wdrvi(I)          & ndwi(I)           & gdd               & max\_surf         & -                 & 0.528              & 0.8747             & 0.9627             & 0.4737         & 0.8748          \\ \hline
cos\_doy          & sin\_doy          & gvmi              & wdrvi(I)          & psri(I)           & ndwi(I)           & gdd               & max\_surf         & -                 & 0.5227             & 0.8453             & 0.984              & 0.4715         & 0.8706          \\ \hline
cos\_doy          & sin\_doy          & gvmi              & savi              & psri(I)           & ndwi(I)           & gdd               & max\_soil         & -                 & 0.52               & 0.8853             & 0.992              & 0.471          & 0.882           \\ \hline
cos\_doy          & sin\_doy          & gvmi              & evi               & wdrvi(I)          & psri(I)           & gdd               & max\_surf         & -                 & 0.5227             & 0.848              & 0.9947             & 0.4709         & 0.8731          \\ \hline
cos\_doy          & sin\_doy          & gvmi              & evi               & savi              & psri(I)           & ndwi(I)           & gdd               & max\_soil         & 0.5253             & 0.8987             & 1.0                & 0.4701         & 0.8856          \\ \hline
cos\_doy          & sin\_doy          & gvmi              & savi              & wdrvi(I)          & ndwi(I)           & gdd               & max\_soil         & -                 & 0.52               & 0.8773             & 0.9787             & 0.4695         & 0.8771          \\ \hline
cos\_doy          & sin\_doy          & evi               & savi              & wdrvi(I)          & psri(I)           & gdd               & max\_soil         & max\_surf         & 0.52               & 0.8667             & 0.9973             & 0.4687         & 0.8775          \\ \hline
cos\_doy          & sin\_doy          & gvmi              & savi              & psri(I)           & ndwi(I)           & gdd               & max\_soil         & max\_surf         & 0.5173             & 0.904              & 1.0                & 0.468          & 0.8852          \\ \hline
cos\_doy          & sin\_doy          & gvmi              & savi              & psri(I)           & ndwi(I)           & gdd               & max\_surf         & -                 & 0.52               & 0.8773             & 0.9813             & 0.4675         & 0.8764          \\ \hline
cos\_doy          & sin\_doy          & gvmi              & savi              & ndwi(I)           & gdd               & max\_soil         & max\_surf         & -                 & 0.52               & 0.896              & 0.9947             & 0.4671         & 0.8836          \\ \hline
cos\_doy          & sin\_doy          & gvmi              & evi               & savi              & wdrvi(I)          & psri(I)           & gdd               & max\_surf         & 0.5173             & 0.88               & 0.9893             & 0.4653         & 0.8781          \\ \hline
cos\_doy          & sin\_doy          & gvmi              & evi               & savi              & wdrvi(I)          & psri(I)           & gdd               & -                 & 0.5173             & 0.8667             & 0.9893             & 0.4649         & 0.8742          \\ \hline
cos\_doy          & sin\_doy          & gvmi              & evi               & psri(I)           & ndwi(I)           & gdd               & max\_surf         & -                 & 0.5173             & 0.856              & 0.992              & 0.4647         & 0.8727          \\ \hline
cos\_doy          & sin\_doy          & evi               & psri(I)           & ndwi(I)           & gdd               & max\_soil         & max\_surf         & -                 & 0.5147             & 0.8453             & 0.984              & 0.4638         & 0.8693          \\ \hline
cos\_doy          & sin\_doy          & savi              & wdrvi(I)          & psri(I)           & ndwi(I)           & gdd               & max\_surf         & -                 & 0.512              & 0.8507             & 0.976              & 0.4622         & 0.8691          \\ \hline
cos\_doy          & sin\_doy          & gvmi              & evi               & savi              & wdrvi(I)          & ndwi(I)           & gdd               & -                 & 0.5173             & 0.8453             & 0.9573             & 0.462          & 0.8657          \\ \hline
cos\_doy          & sin\_doy          & evi               & wdrvi(I)          & psri(I)           & gdd               & max\_soil         & max\_surf         & -                 & 0.512              & 0.8107             & 0.9787             & 0.4614         & 0.8601          \\ \hline
cos\_doy          & sin\_doy          & gvmi              & savi              & wdrvi(I)          & psri(I)           & ndwi(I)           & gdd               & -                 & 0.5147             & 0.848              & 0.96               & 0.4608         & 0.8654          \\ \hline
cos\_doy          & sin\_doy          & gvmi              & evi               & savi              & wdrvi(I)          & ndwi(I)           & gdd               & max\_soil         & 0.512              & 0.872              & 0.976              & 0.4603         & 0.8737          \\ \hline
cos\_doy          & sin\_doy          & evi               & savi              & wdrvi(I)          & ndwi(I)           & gdd               & max\_surf         & -                 & 0.512              & 0.8613             & 0.9867             & 0.4593         & 0.8731          \\ \hline
cos\_doy          & sin\_doy          & savi              & psri(I)           & ndwi(I)           & gdd               & max\_soil         & max\_surf         & -                 & 0.5093             & 0.864              & 1.0                & 0.4583         & 0.8745          \\ \hline
cos\_doy          & sin\_doy          & gvmi              & wdrvi(I)          & psri(I)           & ndwi(I)           & gdd               & max\_soil         & -                 & 0.512              & 0.872              & 0.984              & 0.4569         & 0.8739          \\ \hline
cos\_doy          & sin\_doy          & gvmi              & evi               & wdrvi(I)          & psri(I)           & ndwi(I)           & gdd               & -                 & 0.5093             & 0.8347             & 0.9707             & 0.4556         & 0.8622          \\ \hline
cos\_doy          & sin\_doy          & gvmi              & evi               & savi              & wdrvi(I)          & gdd               & max\_surf         & -                 & 0.5093             & 0.8827             & 0.9893             & 0.4551         & 0.8775          \\ \hline
cos\_doy          & sin\_doy          & gvmi              & evi               & psri(I)           & gdd               & max\_soil         & max\_surf         & -                 & 0.504              & 0.8587             & 0.9973             & 0.4547         & 0.8719          \\ \hline
cos\_doy          & sin\_doy          & gvmi              & savi              & wdrvi(I)          & psri(I)           & gdd               & max\_soil         & -                 & 0.504              & 0.8747             & 0.9893             & 0.454          & 0.8749          \\ \hline
cos\_doy          & sin\_doy          & gvmi              & evi               & savi              & wdrvi(I)          & gdd               & max\_soil         & max\_surf         & 0.504              & 0.88               & 0.9947             & 0.4535         & 0.8771          \\ \hline
cos\_doy          & sin\_doy          & gvmi              & evi               & savi              & ndwi(I)           & gdd               & max\_soil         & max\_surf         & 0.504              & 0.8773             & 0.9947             & 0.4531         & 0.8759          \\ \hline
cos\_doy          & sin\_doy          & gvmi              & evi               & savi              & wdrvi(I)          & psri(I)           & ndwi(I)           & gdd               & 0.5067             & 0.8587             & 0.96               & 0.4512         & 0.8666          \\ \hline
cos\_doy          & sin\_doy          & gvmi              & evi               & wdrvi(I)          & psri(I)           & ndwi(I)           & gdd               & max\_soil         & 0.504              & 0.8427             & 0.9813             & 0.4506         & 0.866           \\ \hline
cos\_doy          & sin\_doy          & evi               & wdrvi(I)          & psri(I)           & ndwi(I)           & gdd               & max\_soil         & max\_surf         & 0.504              & 0.8187             & 0.9867             & 0.4498         & 0.8621          \\ \hline
cos\_doy          & sin\_doy          & evi               & savi              & psri(I)           & gdd               & max\_soil         & max\_surf         & -                 & 0.5013             & 0.8693             & 1.0                & 0.448          & 0.8756          \\ \hline
cos\_doy          & sin\_doy          & gvmi              & evi               & savi              & psri(I)           & gdd               & max\_soil         & -                 & 0.496              & 0.8693             & 1.0                & 0.4469         & 0.8743          \\ \hline
cos\_doy          & sin\_doy          & gvmi              & evi               & savi              & ndwi(I)           & gdd               & max\_soil         & -                 & 0.5013             & 0.9067             & 0.992              & 0.4468         & 0.8806          \\ \hline
cos\_doy          & sin\_doy          & gvmi              & evi               & wdrvi(I)          & psri(I)           & ndwi(I)           & gdd               & max\_surf         & 0.5013             & 0.8533             & 0.9893             & 0.4468         & 0.8691          \\ \hline
cos\_doy          & sin\_doy          & gvmi              & evi               & wdrvi(I)          & ndwi(I)           & gdd               & max\_soil         & -                 & 0.4987             & 0.8667             & 0.9893             & 0.4464         & 0.8729          \\ \hline
cos\_doy          & sin\_doy          & gvmi              & evi               & savi              & psri(I)           & ndwi(I)           & gdd               & -                 & 0.4987             & 0.8693             & 0.9813             & 0.4454         & 0.8709          \\ \hline
cos\_doy          & sin\_doy          & gvmi              & evi               & psri(I)           & ndwi(I)           & gdd               & max\_soil         & max\_surf         & 0.496              & 0.88               & 0.9973             & 0.4445         & 0.8757          \\ \hline
cos\_doy          & sin\_doy          & savi              & wdrvi(I)          & psri(I)           & gdd               & max\_soil         & max\_surf         & -                 & 0.4933             & 0.8667             & 0.9973             & 0.4444         & 0.8725          \\ \hline
cos\_doy          & sin\_doy          & gvmi              & savi              & wdrvi(I)          & ndwi(I)           & gdd               & max\_soil         & max\_surf         & 0.4933             & 0.8907             & 0.9867             & 0.4442         & 0.8757          \\ \hline
cos\_doy          & sin\_doy          & gvmi              & evi               & savi              & psri(I)           & ndwi(I)           & gdd               & max\_surf         & 0.488              & 0.8853             & 0.9893             & 0.435          & 0.8736          \\ \hline
cos\_doy          & sin\_doy          & evi               & savi              & wdrvi(I)          & psri(I)           & ndwi(I)           & gdd               & -                 & 0.488              & 0.856              & 0.9627             & 0.4328         & 0.862           \\ \hline
cos\_doy          & sin\_doy          & gvmi              & savi              & wdrvi(I)          & psri(I)           & ndwi(I)           & gdd               & max\_surf         & 0.4853             & 0.888              & 0.9813             & 0.4325         & 0.872           \\ \hline
cos\_doy          & sin\_doy          & evi               & wdrvi(I)          & psri(I)           & ndwi(I)           & gdd               & max\_soil         & -                 & 0.4853             & 0.8133             & 0.9787             & 0.4319         & 0.8549          \\ \hline
cos\_doy          & sin\_doy          & gvmi              & evi               & savi              & psri(I)           & gdd               & max\_soil         & max\_surf         & 0.4827             & 0.888              & 0.9947             & 0.4314         & 0.8747          \\ \hline
cos\_doy          & sin\_doy          & gvmi              & evi               & savi              & wdrvi(I)          & ndwi(I)           & gdd               & max\_surf         & 0.4827             & 0.8667             & 0.9733             & 0.4301         & 0.8666          \\ \hline
cos\_doy          & sin\_doy          & gvmi              & psri(I)           & ndwi(I)           & gdd               & max\_soil         & max\_surf         & -                 & 0.48               & 0.8587             & 1.0                & 0.4276         & 0.8685          \\ \hline
cos\_doy          & sin\_doy          & gvmi              & evi               & savi              & psri(I)           & gdd               & max\_surf         & -                 & 0.4773             & 0.888              & 0.9947             & 0.4274         & 0.8735          \\ \hline
cos\_doy          & sin\_doy          & evi               & wdrvi(I)          & psri(I)           & ndwi(I)           & gdd               & max\_surf         & -                 & 0.4827             & 0.84               & 0.976              & 0.4259         & 0.8598          \\ \hline
cos\_doy          & sin\_doy          & gvmi              & savi              & psri(I)           & gdd               & max\_soil         & max\_surf         & -                 & 0.4773             & 0.9013             & 1.0                & 0.4258         & 0.8768          \\ \hline
cos\_doy          & sin\_doy          & gvmi              & savi              & wdrvi(I)          & psri(I)           & gdd               & max\_soil         & max\_surf         & 0.4747             & 0.88               & 0.9893             & 0.4233         & 0.8711          \\ \hline
cos\_doy          & sin\_doy          & gvmi              & evi               & savi              & ndwi(I)           & gdd               & max\_surf         & -                 & 0.4773             & 0.8933             & 0.9893             & 0.4225         & 0.8729          \\ \hline
cos\_doy          & sin\_doy          & gvmi              & savi              & wdrvi(I)          & gdd               & max\_soil         & max\_surf         & -                 & 0.472              & 0.8827             & 0.9973             & 0.4196         & 0.8714          \\ \hline
\end{tabular}
}
\caption{\small{The feature sets of the top 82 FCM models (with size 8 or 9 features) for cotton phenology estimation and their corresponding performance. With (I) we show the cumulative integrals of the VIs. max\_soil refers to the cumulative maximum soil temperature, max\_surf to the cumulative maximum surface temperature and wkappa to the linear weighted kappa coefficient.}}
\end{table*}

\end{appendices}

% *************************************** Index ********************************
\printthesisindex % If index is present

\end{document}